Department of Electrical Engineering

National Taipei University

Master Thesis

# Reinforcement Learning based control of Cancer Chemotherapy treatments

Student：Jayakody Kankanamalage Chamani Shiranthika

Advisor：Chan Yun Yang

Co-Advisors: Chung Yi Wang, Chen Guowei

# DECLARATION

I declare that this dissertation does not incorporate, without acknowledgment, any material previously submitted for a Degree or a Diploma in any University and to the best of my knowledge and belief, it does not contain any material previously published or written by another person or myself except where due reference is made in the text. I also hereby give consent for my dissertation, if accepted, to be made available for photocopying and for interlibrary loans, and for the title and summary to be made available to outside organization.

| | |
|---|---|
| Name of Student | Signature of Student |
| J.K.C. Shiranthika | Date: 30/05/2021 |

Supervised by

| | |
|---|---|
| Name of Supervisor | Signature of Supervisor |
| Dr. Chan Yun Yang | Date: 30/05/2021 |
| Name of Supervisor | Signature of Supervisor |
| Dr. Chung Yi Wang | Date: 30/05/2021 |
| Name of Supervisor | Signature of Supervisor |
| Dr. Chen Guowei | Date: 30/05/2021 |

**DEDICATION**

*I dedicate this thesis to my parents and my twin sister who are the pioneers and great pillars that reinforced my education and always support and inspire me.*

*I dedicate this thesis to my motherland, Sri Lanka and my mother university, University of Moratuwa, Sri Lanka.*

*I dedicate this thesis to all the professors of the College of Electrical Engineering and Computer Science, National Taipei University, Taiwan.*

*A warm dedication of my findings for all the researchers in the world who are trying find better solutions to eradicate the deadly disease of Cancer.!*

# ACKNOWLEDGMENT

Firstly. I would like to express my sincere gratitude to my supervisor Dr. Chan Yun Yang who guided and profoundly nurtured me throughout the project. He extended his fullest corporation to me whenever I sought for advice. From the very first day, when proposed my research idea, Professor Yang was actively collaborating in each and every step I was keeping and his insights, suggestions and guidance cleared a troublesome path of mathematical obstacles with new concepts.

Secondly, I'm grateful to all the lecturers of the Electrical Engineering, National Taipei University. During the course period of two years, these valuable lectures helped me to think differently. Their enthusiasm and active teaching style greatly improved my curiosity and motivation in deciding a valuable research topic and in carrying out the research successfully with my highest potentials.

Thirdly, I would like to express my heartfelt bundle of thanks for my mother university in Sri Lanka, University of Moratuwa and my Faculty, Faculty of Information Technology. Because of her foundation I was blessed with, I could enter to National Taipei University with a funded scholarship for my masters study period. And, I am thankful for Dr. Sagara Sumathipala, Mr. B.H Sudantha, Dr. Upeksha Ganegoda, Dr. Surangani Bandara, who were inspiring me always to finding my optimal success with ultimate motivation.

Fourthly, I am thankful for the Ministry of Education, Taiwan in assisting me with the funded scholarship for the 2 years. Many thanks on considering me as a potential candidate in receiving the scholarship and I am so much grateful for the opportunity on behalf of my country.

Although not explicit on this thesis, there was a tremendous amount of hours, I spent learning about reinforcement learning, cancer research area, real clinical chemotherapy practices and finally offline reinforcement learning algorithms which is still an emerging research area in this paradigm. Therefore, for this research, I had to refer many books and research papers as reference. I would like to thank all the authors of those publications. I'm especially thankful to *"Commonwealth of Learning-Canada",* online professional development program who assisted the fully access to online Specialization on Reinforcement Learning offered by *University of Alberta, Canada*, where I could grasp unparalleled knowledge on Reinforcement Learning.

And, I express my heartfelt thanksgiving for my friends around me here at Taiwan and also my friends at Sri Lanka, who gave me tremendous support by words and motivations. My friends at the Lab No.607 and Lab No. 601, my roommates, many thank you for all. And, among them, I would like to specially thank my best friend, Nilantha Premakumara, who was there helping me for the whole 2 years from the day I wished to apply to higher studies at Taiwan. As the very first students who entered the NTPU from Sri Lanka and also from our mother university, University of Moratuwa, We are proud to say we worked together, on our best to achieve our goals.

Finally, an honest and special thank you to my little family, who are waiting for me at home, with hopes, in my country, my mother, my father and my twin sister. They were a quintessential support during my studies, research and the development of this thesis, by always encouraging me and believing in me, even on the darkest times.

**ABSTRACT**

Reinforcement Learning based control of Cancer Chemotherapy treatment

by

Jayakody Kankanamalage Chamani Shiranthika

July 2021

ADVISOR: DR. CHAN YUN YANG

DEPARTMENT: DEPARTMENT OF ELECTRICAL ENGINEERING

DEGREE: MASTER OF SCIENCE IN ELECTRICAL ENGINEERING

Reinforcement Learning has achieved tremendous success in recent years. Extending its scope from the traditional complex games such as Atari, Go, Pokeman, Chess, Reinforcement Learning have caught researcher's attention on modeling real life scenarios. Personalized medicine and the Dynamic Treatment Regimen is one such preliminary research field, researchers are focusing their mindsets currently. Estimating the optimal dynamic treatment regimens from the observational real clinical data is being popularized among the researchers widening its scope towards more advanced versions of Reinforcement Learning algorithms.

This research is about a research study that focused on Reinforcement Learning techniques that have been approached in the context of cancer chemotherapy treatments. Primarily this thesis addresses the result of the study by means of the state-of-the art on Reinforcement Learning techniques used in generating the optimal policy for sequential treatments including the existing Online Reinforcement Learning approaches. Thereby a simulation for Cancer Chemotherapy via traditional Q-Learning would be presented.

Secondly, study will extends towards using the Offline Reinforcement Learning in the optimal policy generation. Thus, the proposed solution follows a data driven approach, which works on the real human colon cancer clinical data.

The novelty of the proposed solution resides on the implementation of an Offline Reinforcement Learning approach by means of a model-based architecture using the Conservative Q-Learning and the Model Based Offline Policy Optimization algorithm. As per the experimental results and the evaluations it was revealed that the proposed Chemo RL model caters well the problem of cancer chemotherapy treatment scheduling and as a supportive model for the junior oncologists in deciding their treatment decisions.

## Table of Contents

## List of Figures

## List of Tables

## Abbreviations

DL - Deep Learning

RL - Reinforcement Learning

DRL - Deep Reinforcement Learning

ORL – Offline Reinforcement Learning

QL – Q Learning

DQL - Deep Q-Learning

DQN - Deep Q-Network

DP – Dynamic Programming

MDP - Markov Decision Process

GPU - Graphics Processing Unit

FQI – Fitted Q-Iteration

BCQ - Batch Constrained deep Q Learning

BEAR - Bootstrapping Error Accumulation Reduction

REM - Random Ensemble Mixture

IRIS - Implicit Reinforcement without Interaction at Scale

AWAC - Advantage Weighted Actor Critic

BRAC - Behavior Regularized Actor Critic

CQL - Conservative Q-Learning

QR-DQN - Quantile Regression DQN

BEAR-QL - Bootstrapping Error Accumulation Reduction Q Learning

SPIBB-DQN - Safe Policy Improvement with Baseline Bootstrapping DQN

MOPO - Model Based Offline Policy Optimization

BAIR - Berkeley Artificial Intelligence Research

ECOG PS – Eastern Cooperative Oncology Group Performance Scale

CCI - Charleson's Comorbidity index

CHAPTER 1

# 1. INTRODUCTION

## 1.1. Prolegomena

Cancer, rated as the second leading cause of death all around the world, as stated by the death statistics reported by the World Health Organization, is the most dangerous disease human beings have to suffer. Cancers occur as a result of changing the nature of cells in body, and leads the abnormal cells to invade or spread out to other tissues of the body whenever they grow out of control. In the field of oncology, several treatments for cancer are being carrying out such as surgery, chemotherapy, radiation therapy, bone marrow transplant, immunotherapy, hormone therapy, Cryoablation, etc., where surgery, chemotherapy and radiation are the most common cancer treatments. Chemotherapy, as one of major categories of therapy, focus on pharmacotherapy, which drugs are given through mouth or inject to body. New applications for mathematical modeling have been generated by strong interest in cancer treatment strategies for chemotherapy treatment planning and optimal drug administration. In this study, we develop an agent-based model, based on Reinforcement Learning (RL) for cancer chemotherapy treatment planning that is capable of facilitating oncologist's decision in selecting the optimal treatment regimen according to individual patient's conditions. Initializing the study from the mathematical modeling for cancer chemotherapy modeling on the online setting of reinforcement learning, our study extends up to the offline setting of reinforcement learning and build an innovative framework which can accurately predict the optimal treatment plans for Colon cancer Chemotherapy treatments.

## 1.2. Background and motivation

In the field of medical research, the term Dynamic Treatment Regime (DTR) or Dynamic Treatment Planning is the strategy of selecting appropriate therapy, dose or treatment schedules for individual patients. It is also termed as adaptive strategies, adaptive interventions or adaptive treatment strategy. A DTR comprises of a set of sequential decision rules, in which each rule corresponds to several key decision points in a disease or a disorder. These rules will take the patient diagnosis information as input and returns the probable treatment option he or she should receive. This set of rules will be generated based upon the patient's past medical history records, covariate history, and probable evolving diagnosis measures the patient is experiencing. Thus, a DTR will dictate how to individualize the treatments to patients successfully in a sequential process. (S. A. Murphy, 2003), (Lavori & Dawson, 2004), (Chakraborty, 2011), (Chakraborty & Murphy, 2014), (Laber et al., 2014), (Haraldsson et al., 2017).

Historically, medical science and medical practice have continued to focus on an acute care model for the management of all medical conditions. Despite of that, treatment decisions made in a dynamic treatment regime for a single patient are based on the observations and experiences of that person, with the intention of improving their long-term health results. DTR specifically integrate the heterogeneity in need of treatment between individuals and the heterogeneity in need of treatment over time within an individual, as opposed to classical treatments in which all individuals are given the same level and type of treatment. In DTR, the experimental or observational data will be used to explore the decision regimes that result in a maximal mean response. Theoretically, statistical methods for identifying optimal DTRs from observational data are complicated and not readily implementable by researchers, especially when survival time is the result of interest. However, in the course of this study, it was found that a considerable number of researchers have laid their eyes towards this practical issue that could be a great leap forward in medicine and also the whole human kind.

In applying dosages to the patients, this DTR is a cyclic treatment advanced with time horizons. A state evaluation consecutive to the treatment, and given by appropriate dosages in next horizon based on the maximized wellness of the patient. The applied dosages might be irregular. It depends on personal drug response and presents as varieties of side effects. The idea of introducing adaptive control arises therefore to react the problem, and gives rise to a more precise solution to adapt the irregular dynamics of the patient care.

Reinforcement learning, deep learning and their combinations, with their potential to find particularly specific solution for manipulate a complex system, are promising in managing this kind of adaptive control of diseases by given personalized medicine during the treatment schedule. Accompanied with the kinds of artificial intelligence this dreadful cancer disease would somewhat be controlled such that it would minimize the overall death rates and successfully enhance the recovery rates. The reinforcement learning is a machine learning technique capable of a problem target adaption where an agent in an environment learns with experience and tries to achieve a particular target in an optimal manner by improving the overall reward it gets.

The reinforcement learning adaption to real-life scenarios is still under emerging area which bears many research barriers, *e.g.*, the cancer therapy which we will focus further in this study. Until today, the adapted reinforcement learning principles in the arena of dynamic patient regimes and personalized medicine has been still a challenge (Chakraborty & Moodie, 2013), (Laber et al., 2014), (Zhang & Bareinboim, 2020) especially, applying to the real-life therapy. In this study, a comprehensive literature survey in the fields of Dynamic Treatment Regime will be carried out at the onset and then a data driven approach of the reinforcement learning focused on the chemotherapy treatments for cancer in deciding the optimal drug schedule for a cancer treatment will be setup. Using a reinforcement-learning scheme, we created a route to reach the goal of the optimal chemotherapy.

This study mainly focus on modeling the cancer chemotherapy problem as an agent-based environment, which could decide the patients' behavioral characteristics during the chemotherapy cycles. The chemotherapy doses with the reward scheme that satisfies how effective the treatment is, and thereby finds the optimal drug schedule for a successful recovery for cancer patients.

## 1.3. Cancer and Chemotherapy

### *1.3.1. Cancerous cells and their growth*

Human body is made up of billions of cells which has formed our organs and tissues. Genes in the nucleus of the cells tell when to grow, work, divide and die. In a healthy body, this process happens accordingly. But, in a situation where there is a change in the DNA or damage to DNA, a gene can mutate. Mutated genes won't work properly since the instructions in their DNA gets mixed up. This cause to divide the cells out of control and grow without a limit, which can cause cancer. Cancer cells also do not mature as normal cells and thus they stay immature in the body ("Cancer," 2021), ("The Hallmarks of Cancer," 2020),(*Prostate Cancer: Image Details - NCI Visuals Online*, 2001).

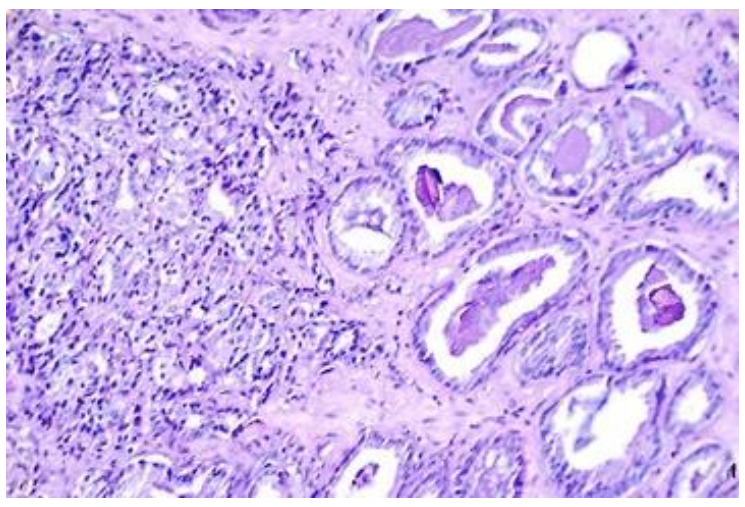

Figure 1: Prostate cancer under microscope (x300) (*Prostate Cancer: Image Details - NCI Visuals Online*, 2001).

As cancer cells divide without a control, a tumor will develop and grow. These cancer cells too have the same needs as the normal cells as they need blood supply to bring oxygen and nutrients to grow. Likewise, a small tumor will grow when it gets oxygen and nutrients and continuously the tumor will get bigger. And, cells always need to be close to the neighbor cells, known as the Allee effect in biology("Allee Effect," 2021) . Because cells stay very closer, cancer cells will spread to surrounding tissues by pushing on normal tissue beside the tumor. Therefore, cells around cancer cells either can proliferate by growing as a cancer cell or can migrate by growing as a non-cancer cell ("Cancer," 2021), (*Cancer Research UK*, 2010) (Figure 1).

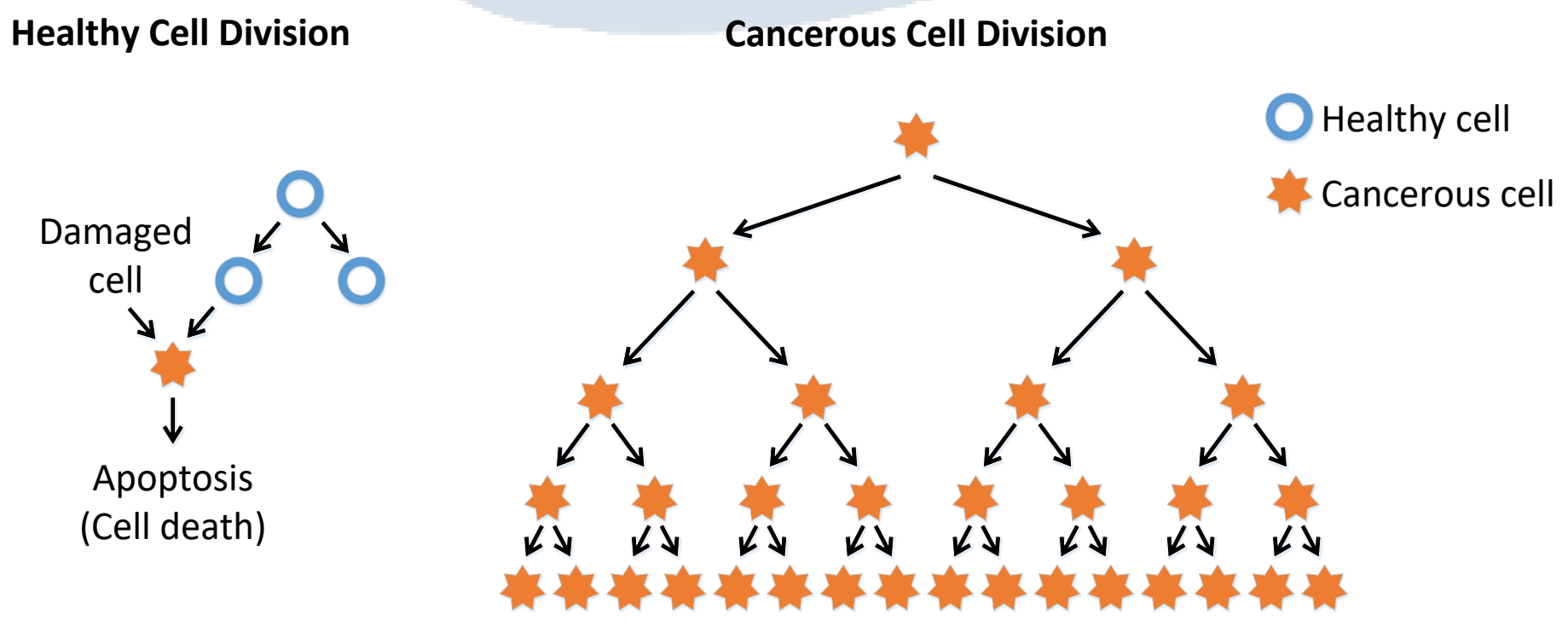

Figure 2: Normal cell division and cancer cell division process

These tumors are of two types as malignant and benign, while the malignant being the cancerous tumors, the benign being the non-cancerous tumors. The malignant tumors are life threatening and the benign tumors are usually not. But sometimes the benign tumors can also become dangerous if they grow a lot, or if they might become malignant after a certain amount of time. The malignant tumors can spread and destroy the surrounding tissues and cause also other tumors to develop. These cancer cells start replicating and don't behave like normal cells. They might break away and move through the blood vessels or lymphatic system and start growing somewhere else in the body. This process is termed as metastasis("Cancer," 2021) .

The cancers are classified by the type of cell that the tumor cells resemble ("The Hallmarks of Cancer," 2020), (*Prostate Cancer: Image Details - NCI Visuals Online*, 2001), ("Allee Effect," 2021), (Watson, 2015), (*Cancer Facts & Figures 2020 | American Cancer Society*, 2020).

1. Carcinoma: Cancers derived from epithelial cells. This group includes many of the most common cancers and include nearly all those in the breast, prostate, lung, pancreas and colon.

2. Sarcoma: Cancers arising from connective tissue (bone, cartilage, fat, nerve, etc.), each of which develops from cells originating in mesenchymal cells outside the bone marrow.

3. Lymphoma and leukemia: Cancers arose from hematopoietic (blood-forming) cells that leave the marrow and tend to mature in the lymph nodes and blood, respectively.

4. Germ cell tumor: Cancers derived from pluripotent cells, often presenting in the testicle or the ovary (seminoma and dysgerminoma, respectively).

5. Blastoma: Cancers derived from immature "precursor" cells or embryonic tissue.

Hanahan and Weinberg revealed five hallmarks which cancer cells had to bear:

1. Cancer cells stimulate their own growth (self-sufficiency in growth signals)

2. They resist inhibitory signals that might otherwise stop their growth (insensitivity to anti-growth signals)

3. They resist their programmed cell death (evading apoptosis)

4. They can multiply indefinitely (limitless replicative potential)

5. They stimulate the growth of blood vessels to supply nutrients to tumors (sustained angiogenesis)

6. They invade local tissue and spread to distant sites (tissue invasion and metastasis)

### *1.3.2. Colon cancer*

Colon cancer or the colorectal cancer is the type of cancer that begins in the large intestine or typically which is called as the colon. Colon is the final part of the digestive tract. Colon cancer mainly affects the older adults, but it also can happen at any age. It usually begins as small, noncancerous (benign) clumps of cells called polyps that form on the inside of the colon. With time, these polyps can become colon cancers. Polyps may be small and produce few, if any, symptoms. Because of this reason, doctors recommend regular screening tests to help prevent colon cancer by identifying and removing polyps before they turn into cancer. Oncologists have identified two main types of polyps as:

- Adenomatous polyps (adenomas)
- Hyperplastic polyps and inflammatory polyps (*Colon Cancer - Symptoms and Causes - Mayo Clinic*, 2020), (*Colon Cancer Treatment, Symptoms, Prevention & Survival Rate*, 2019)

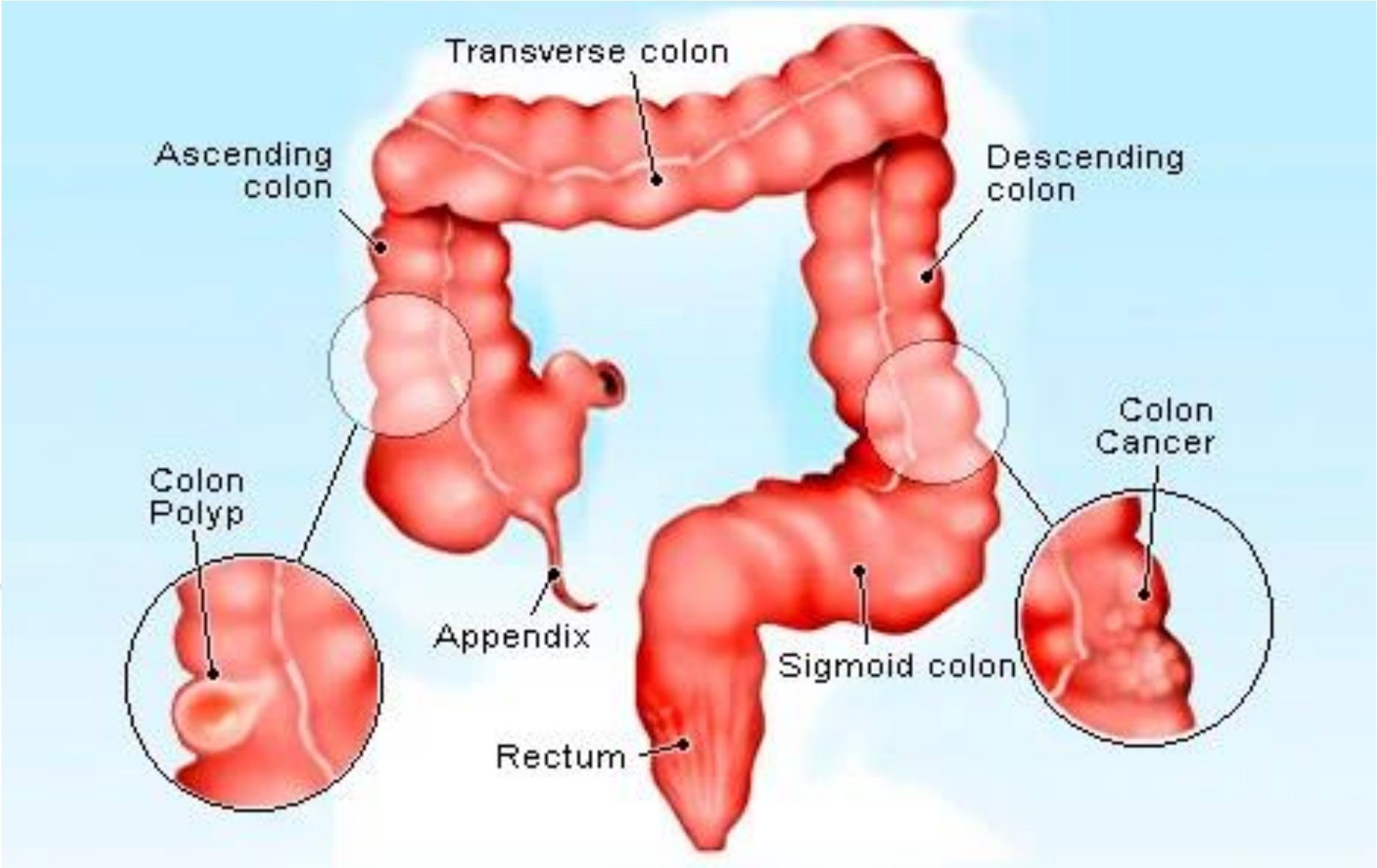

Figure 3: Colon polyps and colon cancerous state (*Colon Cancer Treatment, Symptoms, Prevention & Survival Rate*, 2019)

#### *1.3.2.1. Colon cancer statistics*

1. Colon cancer is the 2nd leading cause of cancer death among men and women combined in USA ("Facts about Colon Cancer & Rectal Cancer | Fight CRC," 2020), and 1st in Taiwan ("Facts about Colon Cancer & Rectal Cancer | Fight CRC," 2020).

2. According to statistics, 1 in 20 people will be diagnosed with colon cancer("Facts

about Colon Cancer & Rectal Cancer | Fight CRC,” 2020).

3. With the results of the cancer research analysis, there will be an estimated 147,950 new cases of colon cancer in 2020(“Facts about Colon Cancer & Rectal Cancer | Fight CRC,” 2020).

4. Research shows that 1 in every 3 people are not up to date with colon cancer screening (“Facts about Colon Cancer & Rectal Cancer | Fight CRC,” 2020).

5. 60% of the colon cancer deaths can be prevented with proper screening process from the beginning (“Facts about Colon Cancer & Rectal Cancer | Fight CRC,” 2020).

6. It is proved that about 1 in every 3 people with colon cancer have a family history of colon cancer (“Facts about Colon Cancer & Rectal Cancer | Fight CRC,” 2020).

### *1.3.3. Cancer Chemotherapy treatments*

Chemotherapy is the cancer treatment, which uses drugs to target rapidly dividing cells throughout the body. For colon cancer, chemotherapy treatments that are injected through veins is called as intravenous chemotherapy, and chemotherapy treatments that are given orally with tablets are called as oral chemotherapy, and chemotherapy treatments that are injected to skin surface is called as subcutaneous chemotherapy (*How Is Chemotherapy Used to Treat Cancer?*, 2021). These treatments kill cancer cells as well as they damage healthy cells. A person may have more than one chemotherapy drug, called a “combination”. Generally, combination works well than a single drug but cause more damage to the cells and more side effects. Common side effects include fatigue, hair loss, diarrhea, skin change, vomiting, neutropenia, anemia, Thrombocytopenia, Febrile Neutropenia etc. (*How Is*

*Chemotherapy Used to Treat Cancer?*, 2021), (*Chemotherapy - Mayo Clinic*, 2019). Use of drug combination would help lower the chance that the cancer may become resistant to any one chemotherapy drug and also will deliver successful positive results immediately.

A course of chemotherapy is made up of several cycles. In each of these cycles, there is a treatment period and then a rest period. Suppose if the oncologist prescribes the chemotherapy period is 3 days, then the following 25 days will be the resting period of the patient where he is not given any kind of drugs. During the resting period healthy cells repairs and the body recovers. The tumor cells cannot repair themselves, so they die. The number of cycles in the chemotherapy treatment and the length of each cycle depends on the treatment plan prescribed by the doctor. The plan prescribed by the doctor normally depend upon the some of the treatment deciding factors. These factors are experimented by the oncologist and the team after each chemotherapy cycle by conducting blood tests and several other medical tests (*How Is Chemotherapy Used to Treat Cancer?*, 2021), (*Chemotherapy - Mayo Clinic*, 2019), (*How Chemotherapy Works | Cancer in General | Cancer Research UK*, 2021). Based on the time period where the chemotherapy is given, they are classified as neoadjuvant therapy where the treatments are given before surgery or radiation therapy, and the adjuvant therapy where the treatments are given after the surgery or radiation therapy to help kill any remaining cancer cells in the body.

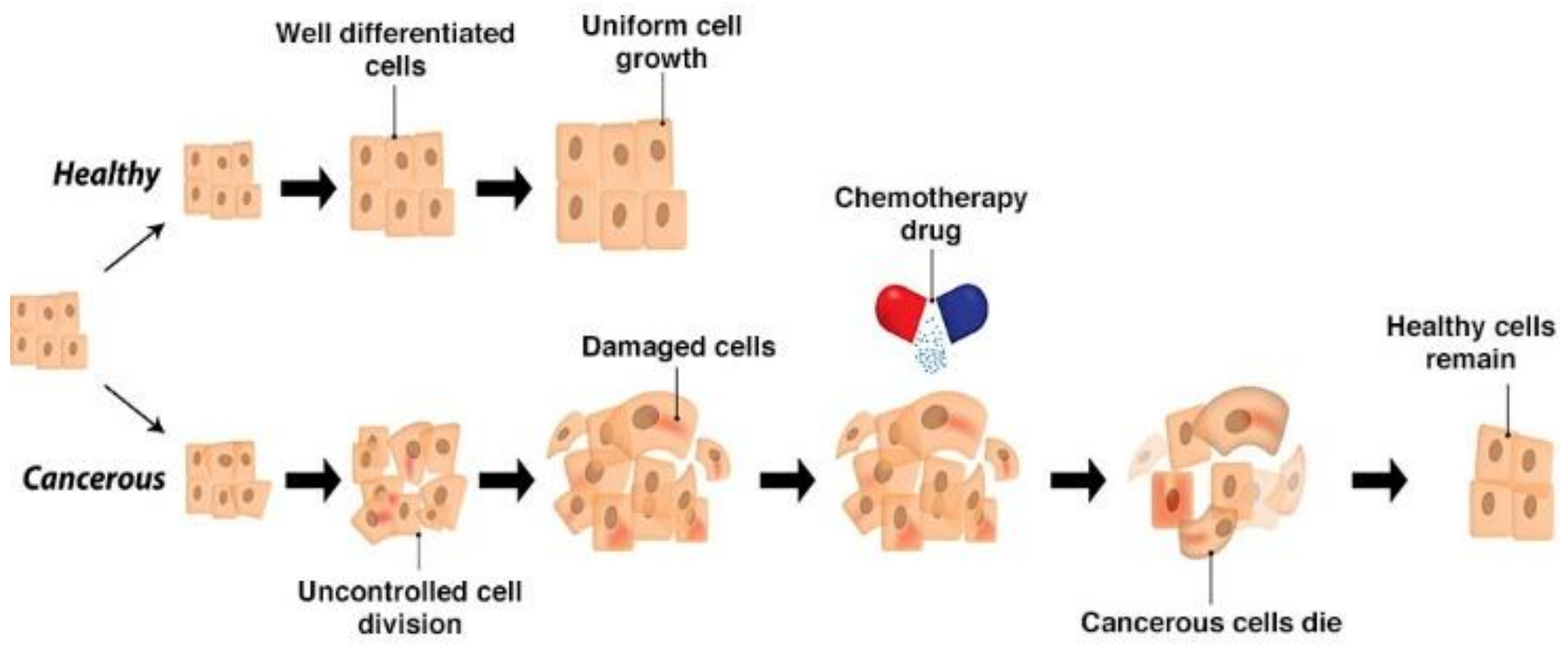

Figure 4: Behavior of cells during chemotherapy (Healthdirect, 2020)

### *1.3.3.1. Goals of chemotherapy*

When the oncologist prescribes to go with the chemotherapy treatments for a patient, there are 3 main goals he/she expects to fulfil as suggested by the American Cancer Society (*Chemotherapy - Mayo Clinic*, 2019). They are cure, control and palliation:

1. Cure: Chemotherapy is used to cure cancer, which means to destroy the cancer completely from the patient’ s body.

2. Control: In the situations where the complete curing is not feasible, Chemotherapy will be used to control the cancer. Here, oncologist will prescribe the treatments to shrink the tumors and/or stop the cancer from growing and spreading. Controlling the disease will help the patient feel better and live longer with increasing the life expectancy.

3. Palliation: Chemotherapy can also be used to ease symptoms caused by the cancer. This is called palliation, palliative chemotherapy, or treatment with palliative intent.

### *1.3.3.2. Planning the chemotherapy*

The oncologist will decide what drug or combination of drugs the patient will be prescribed, including what's kind of doses, how the drugs will be given, how often and the length of treatment. His decisions are majorly based on the type of cancer, where it is, how big it is, is it spread to other parts of the body, and how the treatment affects the normal body functions and overall health of the patient.

The literature in oncology gives multi-variant evidences for the studies done in order to reveal the prominent factors that affect the oncologists' decisions in successfully prediction of the dosing schedule (Fig. 4). References (Morimoto et al., 2010), (Baek & Yi, 2015), (Tabchi et al., 2017) gave an overview of the considerable factors that affected the choice of treatments. As a whole, these factors could be explained as follows:

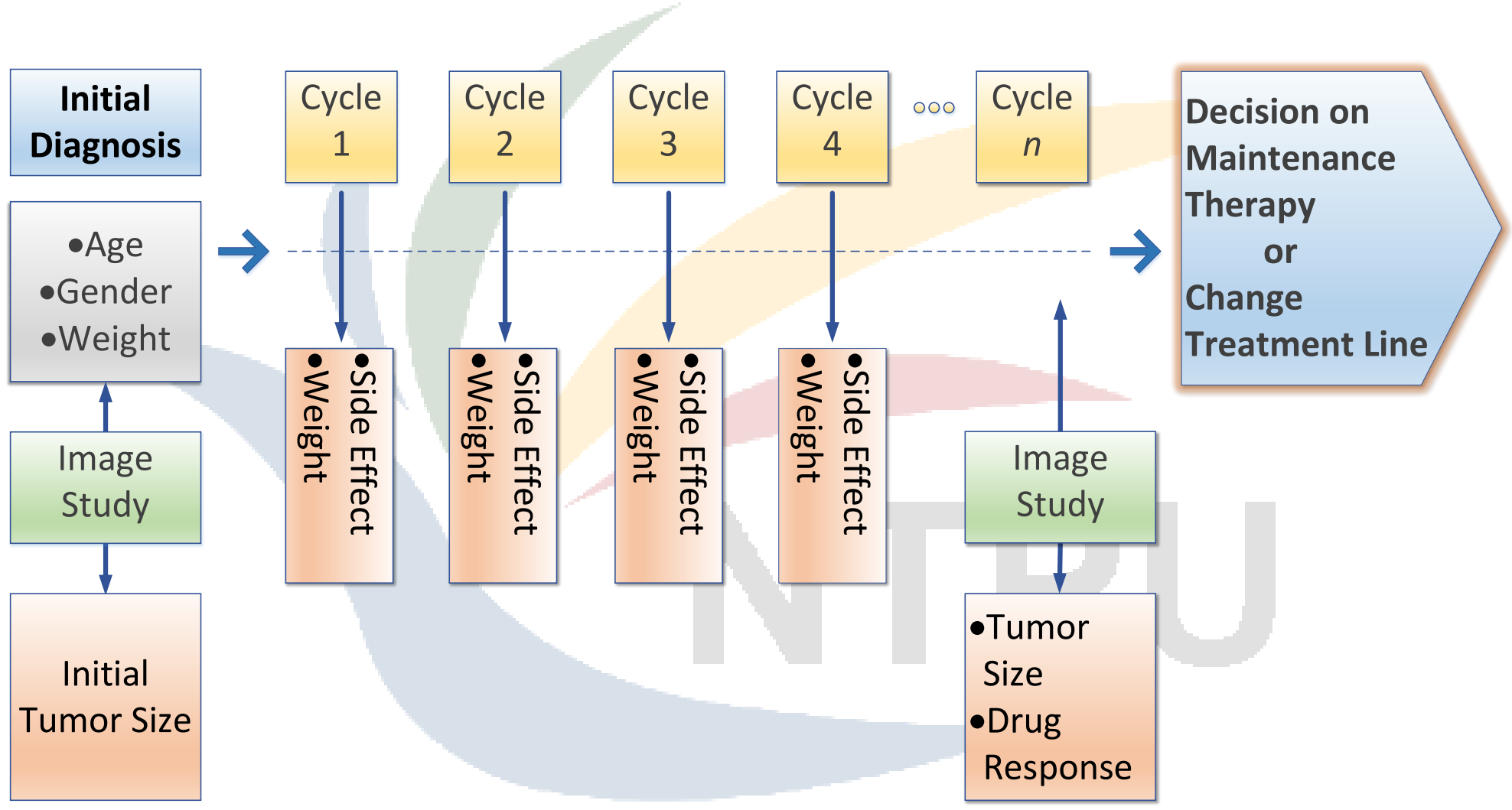

Figure 5: Decision flow on a typical chemotherapy

– Patient' s demographic characteristics: These are the variables related to the patient like his demographic factors (age, marital status, gender, *etc.*), disease status, ECOG PS measure, CCI, histology, and the psychological factors (patient wishes, acceptance, family influence, *etc.*)

– Physician factors: These are the variables related to the physician such as his clinical experience, responsibility, emotional care towards the patient, and the other particularly case-by-case considerations.

– Environmental factors: These are the external factors affecting for the oncologists' chemotherapy decisions such as the time availability, financial incentives, and drug toxic decisions.

#### *1.3.3.3. Line therapy*

In the clinical practice, chemotherapy treatments are given in separate lines of treatments and generally referred to as line therapy. First line therapy are the treatment regimen or regimens that are generally accepted by the medical establishment for initial treatment of a given type and stage of cancer. They are also termed as primary treatment or the induction therapy. The goal of the first line therapies is to cure the cancer if possible. If the first line therapy did not work well for the individuals, oncologists shift in to the second line therapy. The reasons to change line may be: the therapy have not worked, have had some limited efficacy, or have produced unacceptable side effects, damaged organs in the body, or jeopardized the patient's life. If the line therapy is succeeded, then oncologists will go with the maintenance therapy (Fig. 5).

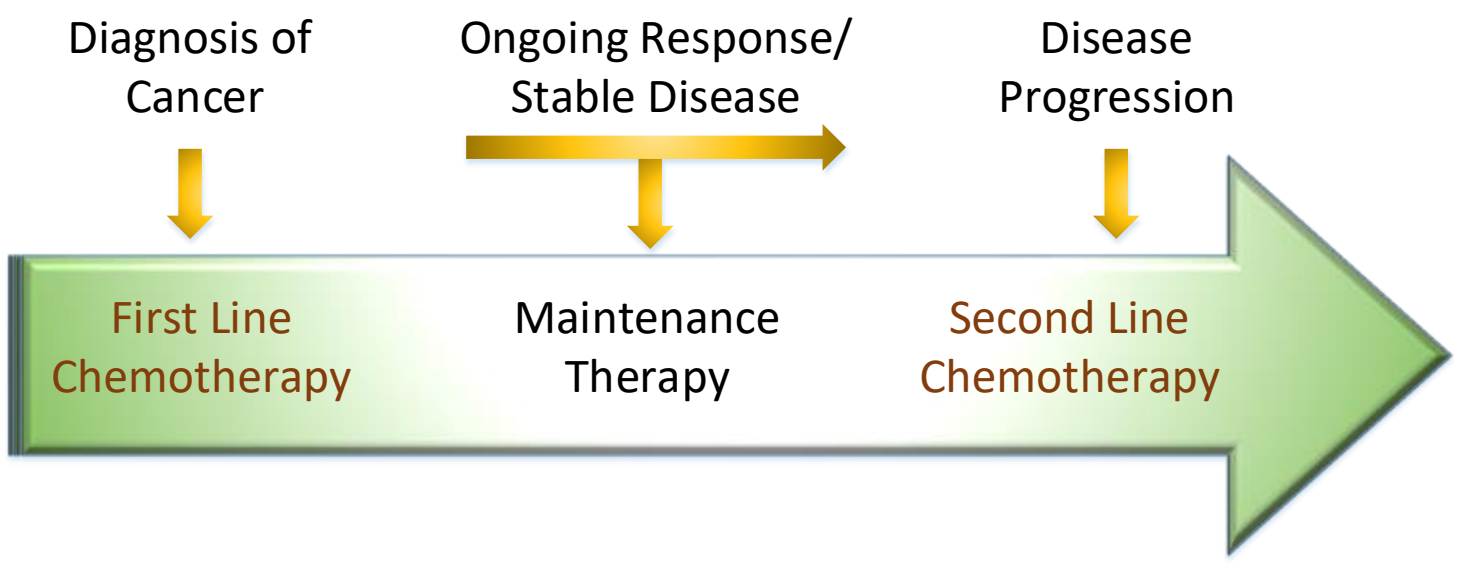

Figure 6: Line therapy in chemotherapy

Normally, the oncologists' decision to change the treatment or change the line of treatment depends on the drug response decided at the end of each chemotherapy cycle (Fig.4). The drug response is assessed based on the image study done normally after 3 months when the chemotherapy treatments are carrying on. The line therapy with the clinician's decisions are given illustratively in Fig. 6:

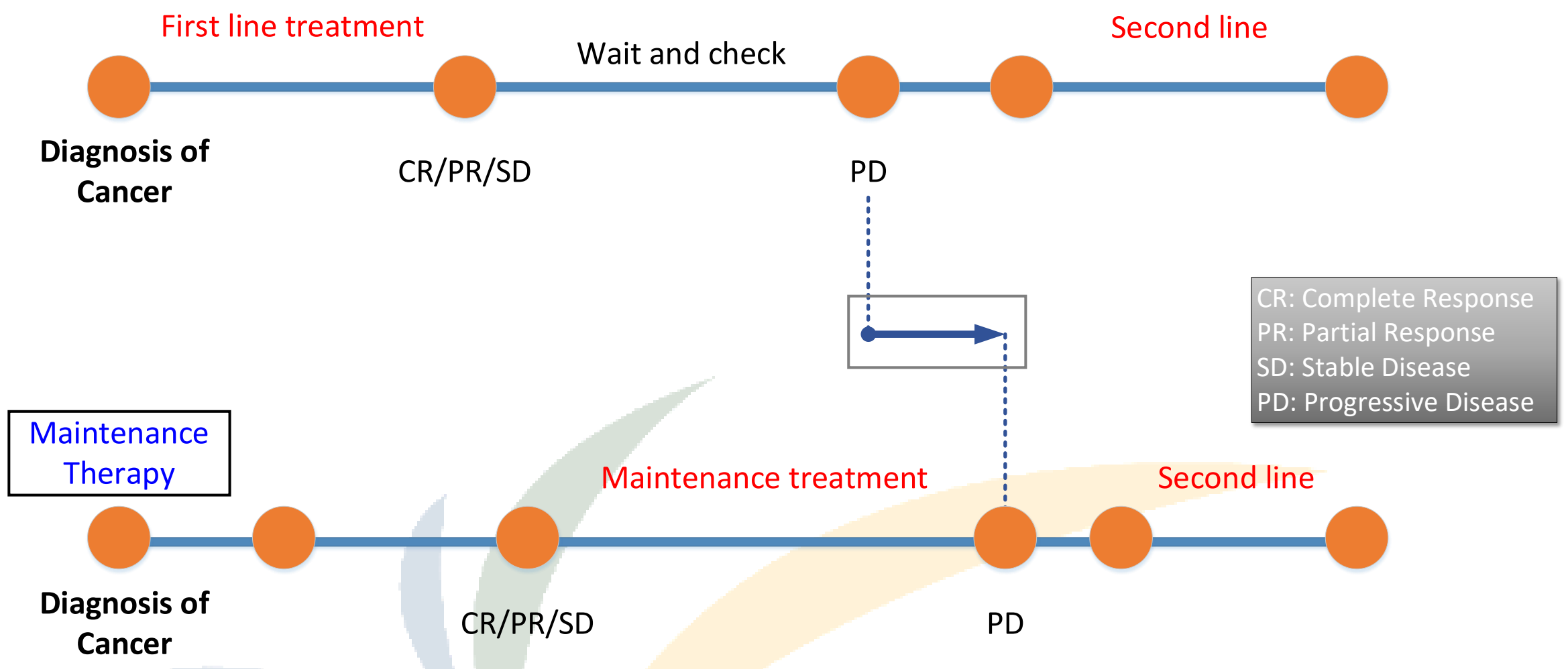

Figure 7: Classical procedure in line therapy treatments (Tabchi et al., 2017)

During the chemotherapy treatments, the anti-cancerous drugs will annihilate the cancerous cells and, at the same time, they will also destroy the healthy cells. This is the reason behind the potential side effects that a cancer patient experience (Fig. 4). In the chemotherapy, some cancerous cells will be resistant to the anti-cancer drug via a random mutation, and thus cannot be destroyed completely. Hence, a high level of drug dosage is often prescribed in the beginning of treatments. But, on the other hand, there will be unbearable side effects towards the patient. Indeed, the drug resistance and the toxicity has become two major problems during the traditional chemotherapy (Fig.4). To develop a successful treatment regimen, this tradeoff between the drug resistance, toxicity and cancerous cell reduction shall be addressed carefully (Bloem et al., 2016), (Brule, 2016).

## 1.4. Problem in brief

Based on the treatment schedules prescribed by oncologists in real clinical practice, this project intends to examine the optimal treatment schedule or the most suitable drug schedule to be given to colon cancer patients such that their overall survival rates will be enhanced, on the basis of Offline Reinforcement Learning concepts. The key idea is to model the real life oncology practice as a Markov Decision Process and provide insights on how to design new clinical trials for colon cancer, which can discover most cost effective, promising dynamic treatment regimens for sequential treatments.

### *1.4.1. Traditional Clinical Trials based Chemotherapy*

Clinical trials are the experiments or the observations done in clinical research. They are performed with a group of people aiming to evaluate a medical, surgical or behavioral intervention. It is the primary way that researchers find out if a new treatment, which can be a new drug or diet or a medical device, is safe and effective in people. Most of the times clinical trials are conducted to learn if a new treatment is more effective and or give less harmful side effects than the standard treatment. They generate data on safety and efficacy. Normally clinical trials are conducted after they have received the health authority or ethics committee approval in the country where the experiment is going to be held. In order to use a drug or a medication on human, it should get the approval of the U.S Food and Drug Administration (FDA). Usually before it is tested on human, scientists perform laboratory tests and studies on animals to test a potential therapy's safety and efficacy. If these studies show favorable results, the FDA gives approval for the intervention to be tested in humans.

Clinical trials consists of five phases. The FDA typically requires Phase I, II, and III trials to be conducted to determine if the drug can be approved for use.

- Preclinical trial investigates new drug applications. There they are testing on animals for toxicity. They further describes the manufacture’s plans for testing the drug in the clinical trials.

- A Phase I trial tests an experimental treatment on a small group of often-healthy people (20 to 80) to judge its safety and side effects and to find the correct drug dosage.

- A Phase II trial uses more people (100 to 300). While the emphasis in Phase I is on safety, the emphasis in Phase II is on effectiveness. This phase aims to obtain preliminary data on whether the drug works in people who have a certain disease or condition. These trials also continue to study safety, including short-term side effects. This phase can last several years.

- A Phase III trial gathers more information about safety and effectiveness, studying different populations and different dosages, using the drug in combination with other drugs. The number of subjects usually ranges from several hundred to about 3,000 people. If the FDA agrees that the trial results are positive, it will approve the experimental drug or device.

- A Phase IV trial for drugs or devices takes place after the FDA approves their use. A device or drug's effectiveness and safety are monitored in large, diverse populations. Sometimes, the side effects of a drug may not become clear until more people have taken it over a longer period.

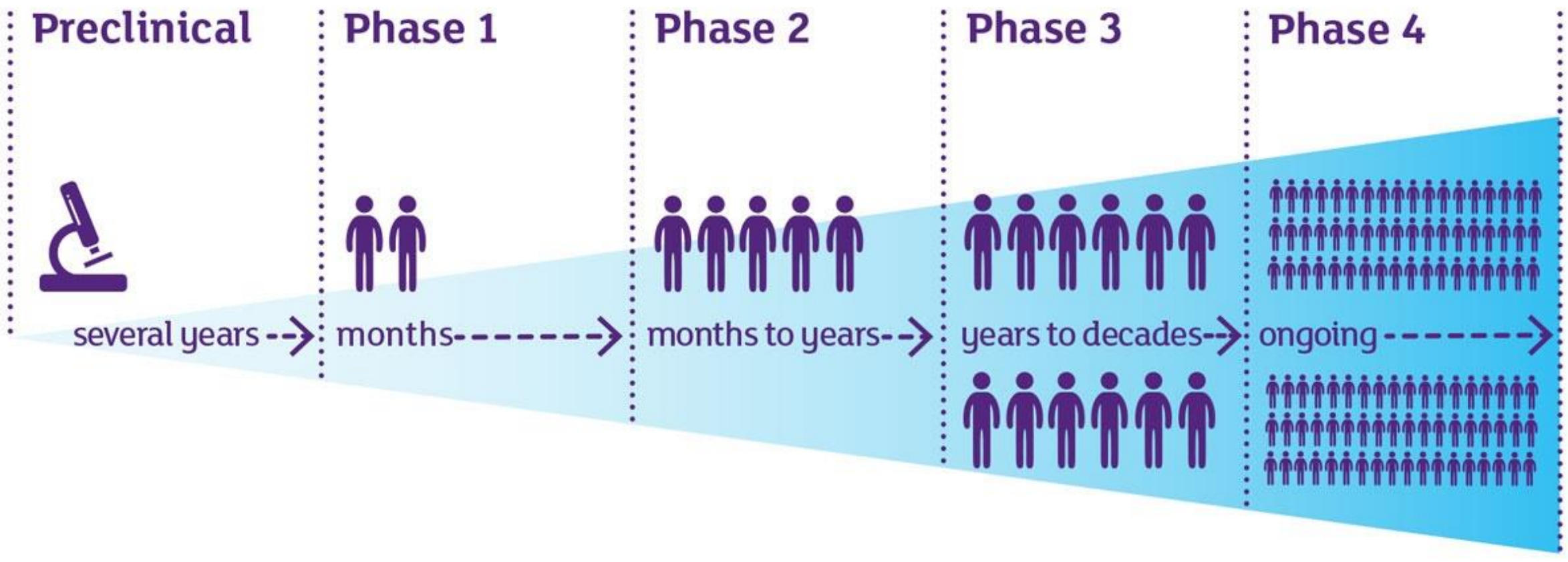

Figure 8: Stages in a traditional Clinical trial

### *1.4.2. Problem with Traditional Clinical Trial based Chemotherapy*

Clinical trials are more reliable to evaluate efficient chemotherapy treatment plans. However, they are limited by factors such as long trial times, high costs and difficulty in conducting such trials. Since cancer is a dreadful disease, it is needed to decide efficient treatments. So, deciding the optimal dosing schedule via a computerized approach is a timely and essential factor to ensure the safety and the overall survival of the cancer patients. With respect to the literature, enhancing the patient's survival probability, reduce the toxicities, reduce the chemotherapy costs associated with deciding the most suitable dosing schedule and the maintaining the dosing schedule, are the common goals most of the researchers have focused on as the goals of chemotherapy optimization problems. Thus our study aims at reducing these burdens by modeling the problem as a MDP and solving it via Data driven Reinforcement Learning means.

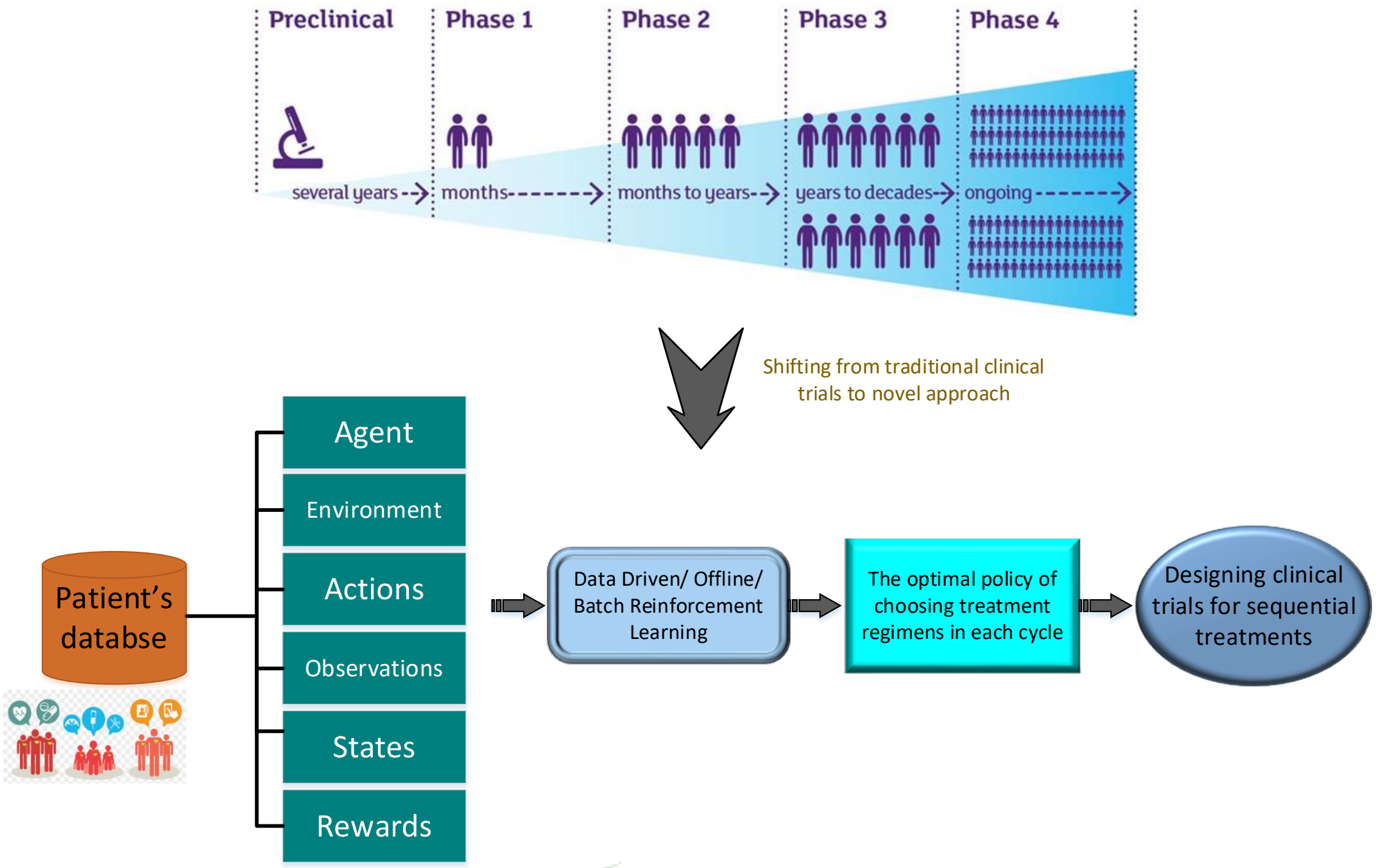

Figure 9: The computerized method as a solution instead of traditional clinical trials

## 1.5. Aim

The aim is to examine the optimal treatment schedule or the most suitable drug schedule to be given to colon cancer patients such that their overall survival rates will be enhanced.

## 1.6. Objectives

In order to reach the aim, following objectives are identified.

i. Comprehensively review the literature on developments in using reinforcement learning in medicine, dynamic treatment regimen planning, reinforcement learning in chemotherapy and the new directions of approaching offline reinforcement learning and its visible challenges.

ii. Design and develop the Online RL based approach for chemotherapy.

iii. Implement the Online RL based approach for chemotherapy using the proposed approach.

iv. Evaluation and validation of the developed Online RL based system.

v. Design and develop the Offline RL based approach for chemotherapy using the real clinical data, practices and knowledge.

vi. Implement the Offline RL based approach for chemotherapy using the proposed approach.

vii. Evaluation and validation of the developed offline RL based system.

### 1.7. Hypothesis

The hypothesis of the research is that Offline Reinforcement Learning can be adapted to solve the problem of cancer chemotherapy treatment planning. The inspiration behind this hypothesis is coming from how the online reinforcement learning have been adopted in finding optimal treatment schemes by using the mouse models which are problematic to approach in the real human treatment setting.

### 1.8. Solution

The proposed methodology is based on the offline reinforcement learning approach for the optimal treatment planning with the aid of real colon cancer patients' chemotherapy treatment data.

## 1.9. Structure of the thesis

The research thesis has been organized as follows.

**Chapter 1** has provided an overall introduction to the research project. This briefly explains the prolegomena, background and the motivation, research problems, aim, objectives, hypothesis, objectives and proposed solution.

**Chapter 2** includes literature review conducted on reinforcement learning in medicine, dynamic treatment regimen planning, reinforcement learning on chemotherapy and the new directions of approaching offline reinforcement learning and its visible challenges.

**Chapter 3** presents the technologies that were adopted to develop the solution in detailed.

In **Chapter 4** the approach has been taken to solve the identified problems. This includes the hypothesis, inputs, outputs, process related to the system, users, system requirements and the features of system.

**Chapter 5** is about the design of the proposed algorithm with the system architecture used.

**Chapter 6** is about how the researcher implemented the developed application.

**Chapter 7** refers to the evaluation done on the developed system.

**Chapter 8** will be continued with the conclusion and further work.

### 1.10. Summary

This chapter describes the full picture of the whole project showing research problem, aim, objectives, hypothesis and the novel solution. Next chapter will be on the comprehensive literature review of research on developments in using reinforcement learning in medicine, dynamic treatment regimen planning, reinforcement learning on chemotherapy and the new directions of approaching offline reinforcement learning and its visible challenges.

**CHAPTER 2**

# 2. LITERATURE SURVEY

## 2.1. Introduction

In chapter 01, we gave an introduction to the overall project. This chapter presents our critical review of research on developments in using reinforcement learning in medicine, dynamic treatment regimen planning, reinforcement learning on chemotherapy and the new directions of approaching offline reinforcement learning and its visible challenges. We have structure this chapter under several headings, namely, reinforcement learning on medicine, reinforcement learning on dynamic treatment regimen planning, reinforcement learning on chemotherapy, mathematical models of Cancer Chemotherapy treatments, Offline Reinforcement Learning, Offline Reinforcement Learning on Real life problems, Research findings and finally our designed problem definition.

## 2.2. Reinforcement Learning on medicine

The healthcare industry has always been an early adopter and a massive beneficiary of developments in technology. Better results have consistently been produced by the application of reinforcement learning to the healthcare system. As a subfield of machine learning, the sole objective of reinforcement learning is to endow the abilities of a person in behavioral decision-making by using the experience of contact with the environment around them and producing evaluative feedback. The reinforcement learning approach requires a progressive decision-making mechanism that is simultaneously sampled, evaluative and comes with delayed feedback, as opposed to the traditionally supervised learning approaches that focus on one-shot, detailed and supervised reward indicators. These specific features make the reinforcement learning methodology a suitable candidate in different healthcare spheres for the creation of prevailing solutions (Yu et al., 2020). In this way, a long and chronological process is also characterized by diagnostic decisions or treatment regimens.

The wide literature survey conducted prior to the implementation of this research yield greater details on the current approaches of researches in the medical domain. Reinforcement Learning has been actively approached in maintaining medical records, medical imaging and diagnostics, examining disease discovery and cell growing mechanisms, treating and predicting diseases and many more (Poolla, 2003), (Barros Netto et al., 2008), (Asoh et al., 2013), (Escandell-Montero et al., 2014), (Mahmud et al., 2018), (Liu et al., 2019), (Coronato et al., 2020). The studies approached by (Poolla, 2003), (Mahmud et al., 2018), (Jonsson, 2019), (Yu et al., 2020) collectively discuss about the application of Reinforcement Learning techniques in the healthcare issues where the data are so itinerant and dynamic. The authors provide a comprehensive survey on the application of Reinforcement Learning in health care domain ranging from dynamic treatment regimes in chronic disease and critical care, automated medical diagnosis from both structured and unstructured clinical data. (Jonsson, 2019) explains the applicability of the off-policy learning, since the trial and error nature of reinforcement learning is actually not practical and unsafe in the domain of medical setting.

The use of the concept of optimal control for modeling disease conditions has been initialized in the mid-1970s. (Schaefer et al., 2005). The first recorded chemotherapy treatment planning system was proposed as an optimal control problem by Swan *et al.* (George W. Swan, 1977). The research article from Schaefer *et al.* is an excellent summary where the MDPs has been introduced in making medical decision-making problems (Schaefer et al., 2005). The related works include: Chhatwal *et al.* (Chhatwal et al., 2010) developed the optimal biopsy decision model as a finite-horizon discrete-time MDP based on mammographic features and demographic factors, Shechter *et al* (Shechter et al., 2008) applied MDPs to develop the first HIV optimization models aimed at maximizing the expected lifetime or quality adjusted lifetime of a patient. Petousis *et al*. (Petousis et al., 2019) Represented the imaging-based screening process as a sequential decision-making and developed an MDP model in order to optimize the long-term cancer detection.(Padmanabhan et al., 2015) invented the concept of reinforcement learning to develop a closed-loop anesthesia controller using the bispectral index (BIS) as a control variable while concurrently accounting for mean arterial pressure (MAP). In (Jalalimanesh, Haghighi, et al., 2017) the researchers propose an agent based model of vascular tumor growth where they use multi-objective distributed Q-learning algorithm to find Pareto-optimal solutions for calculating the dynamic doses. Literature brings facts on treating anemia patients with reinforcement learning; (Martín-Guerrero et al., 2009), (Escandell-Montero et al., 2014) where (Escandell-Montero et al., 2014) proposes a methodology based on reinforcement learning (RL) to optimize ESA therapy with the fitted Q-iteration algorithm with data driven reinforcement learning. (Denton et al., 2009), (Asoh et al., 2013) describes treating diabetes patients with RL. Denton *et al*. (Denton et al., 2009) developed an MDP model to evaluate the optimal start time of statin treatment for the patients with diabetes and (Asoh et al., 2013) depicts that the process of medical treatment is a sequential interaction process between doctors and patients and apply the inverse reinforcement learning to medical records of diabetes treatment to explore the reward function that doctors have in their mind during their treatments.

Thus, the application of reinforcement learning in research related to healthcare setting have opened the door for a wide range of future research paths. With the aid of the agent based environment, it has simplifies timely disease detection, identifies therapy requirements, selects opportunities for clinical trials, highlights any gaps in the healthcare system and other important factors for several conditions. However, current techniques have resulted in many challenges to the application of reinforcement learning to the health care system. The major challenge is the impracticability of learning and evaluating purely observational data due to ethical and logistical reasons. Secondly, the partial observability seen in medicine where, the impossibility of observing everything taking place in an individual's body. Thirdly, the challenge of deciding the reward function that will balance temporary improvement with the overall lasting success. Fourthly, the issue with the non-stationary nature of health care data. With the time, treatment targets are likely to shift and develop in a complex manner that has not been seen previously in the training results. Fifthly, the uncertainty of the results produced by the reinforcement learning algorithms in the medical setting where there is often a risk that the generated results are not indicative of true or underlying causal processes (Poolla, 2003), (Dulac-Arnold et al., 2019), (Yu et al., 2020).

Therefore, in the field of healthcare, reinforcement learning is an exciting scope with its capacity to control ultimate behaviors within a particular context. Its introduction leads to a more thorough and precise treatment at reduced costs with satisfied patient outcomes. If well implemented, reinforcement learning is believed to deliver crucial results in the coming years and will have a significant effect on the treatment and control of prevalent chronic diseases in influencing patient-centered health knowledge with external factors, including weather and economic dynamics or exposure to pollution. Through using accessible genetic information to discover the best conceivable medical care methods, it can proficiently produce accurate medicine solutions tailored to human characteristics.

### 2.3. Reinforcement Learning on Dynamic Treatment Regimens (DTR)

A dynamic treatment regimen (DTR) consists of a set of decision rules that decide how to assess the treatment allocation to patients based on changing interventions and the background of covariates. The literature of Reinforcement Learning with Dynamic Treatment Regimens has been nourished with several interesting research carried out. The growing interest in generating personalized treatments with more evidence based or data driven approaches have facilitated the arena of treatment regimen to go beyond the limits of traditional treatment approaches. Studies done in this topic collectively propose analytical methods for clinical trial analysis and design that can identify drug formulations with substantial overall survival and toxicity improvements while reducing overall costs. Most of them have created a robust clinical trial database and use it to build new mathematical models that are able to predict the survival and new drug combinations from earlier studies. (Chakraborty & Moodie, 2013), (Kosorok & Moodie, 2015), (Bertsimas et al., 2016), (Krakow et al., 2017), (L. Wang et al., 2018), (Bertsimas, 2019), (N. Liu et al., 2019), (X. Liu et al., 2019). The major research objectives addressed under this topic are to compare two or more DTR in terms of their utility, to identify the optimal DTR or the optimal treatment sequence that result in the most favorable possible outcome and to successfully deal with the high dimensional state space and action space representation available with the medical data.

(L. Wang et al., 2018) is an interesting research of applying RL with DTR, which has

combined the benefits of supervised learning and reinforcement learning. They have applied an off-policy actor critic framework to handle the relations among medications. The actor network recommends time-varying medications according to the dynamic states of patients, where a doctor's decision supervisor provides the predictor signal to ensure safe conduct and leverages the expertise of physicians to speed up the learning process. The actor network recommends time-varying medications according to the dynamic states of patients. To promote or prevent the prescribed treatments, the critic network calculates the action value associated with the actor network. LSTM has been used to expand SRL-RNN to tackle POMDP by summarizing the entire historical observations to capture a more complete observation due to the absence of fully observed states in the real world. In order to develop the model for healthy complex care decision making, (X. Liu et al., 2019) propose a new privacy-preserving reinforcement learning system with experience replay. (S. Liu et al., 2020) has conducted a comprehensive survey which has concluded their findings as that RL has the potential to be a clinical decision support tool for the ICU patients. They highlights that the related research should have a meaningful reward design and the challenge with respect to the medical data acquisition, analysis, interpretation, and presentation of these data in a clinically relevant and usable format. A yet another engrossing research is (J. Wang et al., 2020) which propose a model based regularized actor critic framework by using the real world HER dataset, MIMIC-III. The authors present three data driven approaches for DTR, which are imitation learning, regularization to RL based on different criteria, and model based RL. (Kosorok & Moodie, 2015) in their compelling research incorporates a dynamic treatment regime developed using RL which involves the steps of data collecting by using sequential multiple assignment randomized trial (SMART) studies, formulating the reinforcement learning problem with state, action modeling, reward function defining, function approximating and finally validating. They have used the Fitted Q algorithm and highlights several important future research directions.

When reviewing the depth of the study, it was a major fact observed that the researchers have been adopted variants of traditional Q-Learning in their researches (Song et al., 2011), (Schulte et al., 2014). (Z. Zhang, 2019) used a variant of Q Learning with R programming language in DTR. (Song et al., 2011) approached a new reinforcement-learning framework called penalized Q-learning (PQ-learning). The proposed method is demonstrated with the data from a depression clinical trial study. And (Schulte et al., 2014) proposed Q and A learning algorithms for optimal dynamic treatment regimes. The recent research carried out by (J. Zhang & Bareinboim, 2019) with their extension of the research (J. Zhang & Bareinboim, 2020) presents the facts on using online RL methods in DTR with the newly invented algorithms as Reduce, OFU-DTR, PS-DTR. They demonstrate our algorithms on several dynamic treatment regimes, including randomly generated DTRs, and the survival model in the context of multi-stage cancer treatment.

### 2.4. Reinforcement Learning on Chemotherapy

In the innovative framework of medic/al diagnosis, researchers have continuously achieved immense results with the inspiration of artificial intelligence technologies. Over the past few decades, applications of RL has stepped towards a great position in the medical field as described in the previous sections. Marvelous applications arise also in the field of cancer and cancer treatments, for utilizing RL in deciding the optimal chemotherapy drug schedule. So far, many researchers have been marked several milestones in the field of chemotherapy treatment planning problem. Most of the studies focus on modeling the interactions among key factors such as overall survival, toxicity, cancerous cells, cost reduction, *etc*.

As an ahead foundation, three kinds of cell growth rates, not only for healthy cells but also for cancerous cells, have been proposed, *i.e.*, exponential growth, logistic growth and Gompertz growth rates. Based on the ahead foundation, a famous cell count ODE model, called CC model, was first introduced by de Pillis *et al*. to describe the effects of drugs on the progression in tumor (de Pillis et al., 2006). In (Hassani & Naghibi.S, 2010), an optimal drug schedule for patients in progressive cancer phase who take the drug through infusion pump was developed. The cancer state has been formulated with the de Pillis ODE equations and finally the authors have proposed a TD RL approach for solving the optimal control of chemotherapy drug dosage regimen. Actor critic approach in RL is used to decide the reward function in an infinite $(s, a)$ space. The critic estimates the value function while the actor updates the policy distribution in the direction suggested by the critic (Yauney & Shah, 2018) . Sbeity *et al*. (Sbeity, 2015) described a review of the optimization techniques used in cancer chemotherapy planning which also critically evaluated the chemotherapy treatments. In (Ahn & Park, 2011) , authors have studied the applicability of the natural actor critic (NAC) approach to the drug scheduling of cancer chemotherapy. In the NAC framework, the drug dose is regarded as the control input, and the reward signal is defined as a function of the control input and the cell populations of tumor cells, normal cells, and immune cells. In (Yazdjerdi et al., 2019) authors present an anti-angiogenic molecular targeted therapy which can be coupled with chemotherapy and radiotherapy. They have used Q-learning algorithm to develop the drug dosing closed-loop control.

In literature, most of the control methodologies presented for chemotherapy are based on the cancer growth models. The research done by Padmanabhan *et al*.(Padmanabhan et al., 2017a) illustrated a model-free method for the closed loop control on chemotherapy drug dosing. Together with a *Q* learning algorithm, authors have taken the de Pillis ODE equations in to reference to set the states of the model. A RL based work similar to (Padmanabhan et al., 2017a) which works on a tumor size model (namely TS model) simply based on the patient wellness and the tumor size was also developed by Zhao *et al.* (Zhao et al., 2009). The study estimates the states in the RL scheme relatively rough by only the tumor size and the overall patient health, and decomposes the reward function into three distinct parts, the survival status, the wellness effects, and the tumor size effect. Chareyron and Alamir (Chareyron & Alamir, 2009) presented one another study which implemented RL with action-derived rewards for chemotherapy and clinical trial dosing regimen selection. Padmanabhan *et al.* (Padmanabhan et al., 2017b) extended their research more experimentally on some special cases. They used a nonlinear four-state model to implement a RL based control agent for cancer chemotherapy. The attempt in (Beck et al., 2019) reviewed the current literature on reinforcement learning for strategic chemotherapy dosages, including a comparison between the TS model and CC model.

## 2.5. Mathematical models of Cancer Chemotherapy treatments

To achieve easily RL a success, a model as a solid reference of the environment is an important key. Several mathematical models for tumor chemotherapy together with their relating factors have been presented in literatures over the past few years (Martin et al., 1990), (De Pillis & Radunskaya, 2003), (Ergun, 2003), (de Pillis et al., 2006), (Chareyron & Alamir, 2009), (Isaeva & Osipov, 2009), (Engelhart et al., 2011), (Marios M. Hadjiandreou & Mitsis, 2014), (Jalalimanesh, Shahabi Haghighi, et al., 2017). No matter stochastic or deterministic, spatially resolved or not, or continuous or agent based, a basic mathematical model has to show the effects of its corresponding dynamic factors affecting the chemotherapy treatments. In the paper (De Pillis & Radunskaya, 2003) the authors first illustrate that the model's solutions with a time-varying drug term approach the system's solutions without the drug until treatment has ended. They present numerical experiments in which the device can be guided into a desirable basin of attraction by optimal control therapy, whereas conventional pulsed chemotherapy is not. Motivated by that study, in (de Pillis et al., 2006), they describe a primarily model of cancer growth with six different parameter set with a combination of chemotherapy and immunotherapy including six states, the tumor cell population, the total natural killer cells population, the total CD8+T cell population, the number of circulating lymphocytes, the chemotherapy drug concentration in the blood stream and the immunotherapy drug concentration in the blood stream. Six ODE equations have been proposed for these six states. In (de Pillis et al., 2009) the same authors update the model of (de Pillis et al., 2006), and then carefully identify appropriate values for the parameters of the new model according to recent empirical data.

The literature proves several instances where researchers have approached modeling compartment models including Pharmacokinetics; Pharmacodynamics; drug resistance; tumor volume and drug resistance (Iliadis & Barbolosi, 2000), (Marios M. Hadjiandreou & Mitsis, 2012), (Marios M. Hadjiandreou & Mitsis, 2014). (Marios M. Hadjiandreou & Mitsis, 2014) is a promising study performed, with a combination of mathematical modeling and optimal control techniques. They used Gompertz-type growth law and a pharmacokinetic-pharmacodynamics approach for modeling the effects of drugs on tumor progression in tumor bearing mice. The experiments have been done with colon cancer progression data in both untreated mice as well as mice treated with widely used anticancer agents. In (H. Murphy et al., 2016) authors examined all seven of the previously proposed ODE models in the presence and absence of chemotherapy. They derived equations for the maximum tumor size, doubling time, and the minimum amount of chemotherapy needed to suppress the tumor and used a sample data set to compare how these quantities differ based on choice of growth model. (Ghaffari & Karimi, 2008) is yet another interesting modeling research, which proposes an algorithm to design an optimal chemotherapy drug protocol with the target to optimize the number of rests, the application course and the drug dosage. They present new mathematical model employed in the form of ordinary differential equations governing cancer growth on a cell population level.

All of the studies we observed under the literature survey derived several important insights of the mathematical modeling of cancer growth and chemotherapy. Mainly, CC model and TS model, which were often used to describe the inner states of chemotherapy and derive further the alternatives, caught our attention for this study.

## 2.6. Offline Reinforcement Learning

Offline reinforcement learning (RL) relates solely to the problem of learning techniques from a large batch of previously collected data without investigation. This issue setting offers the promise of using such datasets to obtain policies without any costly or dangerous aggressive experimentation. In other words, Offline RL focusses on the problem of an agent to maximally exploit a static dataset. Moreover, since offline reinforcement learning disentangles exploration from exploitation, it can help provide uniform comparisons of reinforcement-learning algorithms' exploitation abilities.

In recent offline RL research and literature, it is evident that researchers have made considerable efforts to find correct algorithms in the task of finding optimal policy through the offline RL scheme. Several successful algorithm efforts have been made by researchers recently, in order to explore a good algorithm in finding the optimal policy under the Offline RL scheme. Some of the promising algorithms include; Fitted Q iteration (Ernst et al., 2005), Batch Constrained deep Q-Learning (BCQ) (Fujimoto et al., 2019), Bootstrapping Error Accumulation Reduction (BEAR) (Kumar et al., 2019), Random Ensemble Mixture (REM)(Agarwal et al., 2020), Implicit Reinforcement without Interaction at Scale (IRIS) (Mandlekar et al., 2020), Advantage Weighted Actor Critic (AWAC) (Nair et al., 2020), Behavior Regularized Actor Critic (BRAC) (Wu et al., 2019) Conservative Q-Learning (CQL) (Kumar et al., 2020).

The rise of Offline RL algorithms emerged with the introduction of Fitted Q-iteration by the researchers (Ernst et al., 2005) and (Lange et al., 2012). (Ernst et al., 2005) gives a description of the problem and, along with a taxonomy of many other algorithms, introduces Fitted Q Iteration as the "Q-Learning of Offline RL." Thereby, (Lange et al., 2012) mainly guide deep reinforcement learning that tackles neutral fitted Q-iteration and their proposed variant Deep fitted Q-iteration. One of the remarkable milestone in Offline RL findings is the Batch Constrained deep Q-learning (BCQ) (Fujimoto et al., 2019). The idea is to incorporate the normal Q-learning algorithm, but in the maximization step that is the $max_{a'}Q(s',a')$, rather than considering the maximum over all possible actions, the authors consider $a'$ such that $(s',a')$ that actually appear in the batch of data. That means authors suggest to eliminate the actions that are rarely appear in the dataset or unlikely to be selected by the behavior policy $\pi_{\beta}$. Further, (Fujimoto, Conti, et al., 2019) surveys some interesting emerging algorithms researchers have used for Batch RL namely; Quantile Regression DQN (QR-DQN), Random Ensemble Mixture (REM), Batch Constrained Deep Q-Learning (BCQ), Bootstrapping Error Accumulation Reduction Q-Learning (BEAR-QL), KL-Control, and Safe Policy Improvement with Baseline Bootstrapping DQN (SPIBB-DQN). Thereby, the authors suggest a new algorithm as the discrete version of the BCQ.

Recently, the Berkeley Artificial Intelligence Research (BAIR) has contributed to this research arena with a theoretical analysis that can help minimize the error propagation by carefully limiting the actions considering during Q-Learning. The Bootstrapping Error Accumulation Reduction (BEAR) algorithm is the practical algorithm they implemented. Under weaker assumptions, their theory demonstrates convergence properties, and they are able to relate their approach to sub optimality. The literature shows that BEAR is better when the data is collected by a weaker or a random policy, while BCQ is better when an expert collects the data.

In (Agarwal et al., 2020), authors propose the Random Ensemble Mixture(REM) algorithm, which uses an ensemble Q-network and forces Bellman consistency among the random convex combinations. This algorithm has online and offline versions. The IRIS or the Implicit Reinforcement without Interaction at Scale (Mandlekar et al., 2020) is yet another successful research carried out which is specialized for offline learning from large scale robotics datasets, where the demonstrations may be suboptimal. (Nair et al., 2020) in their paper describes the innovative algorithm named AWAC which is Advantage Weighted Actor Critic method which combines sample-efficient dynamic programming with optimum policy changes on probability, offering an easy and efficient platform that can exploit massive volumes of offline data and then easily fine-tune reinforcement learning policies online. It has approached the online fine-tuning from offline initialization.

The conservative Q-Learning (CQL) is the recent most successful effort levied by (Kumar et al., 2020) in the Offline RL research platform. It aims to address the limitations in Offline RL by learning a conservative Q-function such that the expected value of a policy under Q-function lower bounds its true value. With a simple Q-value regularizer that is straightforward to implement on top of existing deep Q-learning and actor-critic implementations, CQL augments the standard Bellman error objective. Their experimental results shows that the CQL algorithm substantially outperforms existing offline RL methods.

## 2.7. Challenges in approaching Online RL in real world problems

The mathematical models which have been used for the simulations in the literature are immature and do not adequately capture the complexity of cancer's progression in the human body. The ODEs in literature has been derived by using mice models with in vivo or in vitro experiments. The approached ODE bears the classical cell counts, tumor size and toxicity measures. However, there are multiple other factors, which oncologists consider when deciding the drug schedule for cancer patients such as the age, gender, ECOG PS measure, CCI score and other several practical clinical diagnosis measures. Thus the ODE models we present in this paper is still questionable to use with the real human life simulations.

Unlike with the traditional gaming applications of Reinforcement learning such as AlphaGo, Starcraft, or other board/video games, a vast range of simulations in which the agent makes interventions to learn the optimal policy is impracticable in the medical setting. Most notably, it would be inappropriate to use cancer patients for training purposes. Moreover, it would be expensive and would possibly take years to finish. Therefore, learning the optimal policy from observational historical data should be carried out which is termed as “Off-Policy evaluation” in literature.

Practically, Oncologists are almost never able to examine everything going on in the human body, unlike in many medical games. Medical experiments will review the basic measurements such as blood pressure, temperature, drug concentrations, but these are just cues and not the ground reality about the patient. Thus, the partial observability of medical data is a practical issue, which may concern the applicability of mathematical modeling.

In many real life applications of RL, deciding a proper reward function is problematic. In the field of oncology, it is necessary decide the reward function in a way that it will combine the short-term progress of the patient and the overall long-term progress. Due to the partial observability nature, the problem becomes even worse if the reward is to be decided based on the sequential variations with the observations.

The medical data is non-stationary and dynamic by its nature. The observations present in one patient is quite different from another patient with the same demographic factors. The symptoms will be recorded at odds with intervals, more than others or with several cyclic nature. Thus, the treatment objective of the oncologist will vary with respect to different individual patients. Practical clinical objectives will include, reducing the toxicity level, reducing the length of stay, increasing the performance status of the individual patients, which have not actively captured the attention of researchers who has approached RL in the chemotherapy decision problem.

## 2.8. Problem definition

Having considered the summary of the literature survey done on the above mentioned areas, we have noticed that little research has been conducted yet on the offline reinforcement learning arena. Also, considerably a less research has been done with model based RL approaches for the chemotherapy planning problem. We also notice many researchers have used only the traditional Q-Learning across the problem domains and they have many generalization and practical issues. Based on our critical review, we define our research problem as how to predict the optimal personalized chemotherapy treatment schedules which will also cope with the real clinical practices by using the real colon cancer data.

As the state of the art developments done on chemotherapy treatment planning using Reinforcement Learning, it was able to gather ample evidence on the applicability and limitations in the domain. In the same way, some inspiring clues that can be employed in to overcome the identified drawbacks of chemotherapy treatment planning using reinforcement learning could also be gathered.

### 2.9. Summary

In this chapter, we have given a critical review of research on developments in using reinforcement learning in medicine, dynamic treatment regimen planning, chemotherapy and the new directions of approaching offline reinforcement learning and its visible challenges

**CHAPTER 3**

# 3. TECHNOLOGY

### 3.1. Introduction

In Chapter 02, a critical review of literature review of reinforcement learning in dynamic treatment planning, chemotherapy, mathematical models have been presented. The literature review identified the research problem and potential technology for solving the research problem. As such, this chapter presents the technology adopted in our solution to planning of chemotherapy treatments in the online reinforcement learning setting based on Q-Learning and the offline reinforcement learning setting based on the Conservative Q-Learning (CQL) and Model Based Offline Policy Optimization (MOPO) algorithm.

## 3.2. Machine Learning algorithms used

### *3.2.1. Convolutional Neural Network (CNN) Classifier*

CNN architecture has been inspired by the basic structure and the functionality of the visual cortex of the human brain. A neuron in a layer will only be connected to a small region of the layer before it, instead of all the neurons in a fully connected layer. CNN are also known as convNet. CNN are not fully connected. The design of the conventional CNN includes the input layer, output layer and multiple hidden layers that can be the convolutional layer, Relu layer, pooling layer and finally the fully connected layer. The final convolution will often involves Backpropagation in order to more accurately weight the result. The objective of the convolution operation would be to extract the high-level features such as edges from the input image.

### *3.2.2. Random Forest (RF) Classifier*

The popular random forest classifier consists of tree-structured classifiers. It takes advantage of two powerful machine learning techniques, bagging the random feature selection. Each tree is trained on a bootstrap sample of the training data in bagging, and predictions are made by a majority vote of the trees. Random forest classifier is a forward step or a step away from bagging. When growing a tree, RF randomly chooses a subset of features to separate at each node rather than using all of them. RF uses out-of-bag (OOB) samples to conduct a type of cross-validation in parallel with the training process to test the random forest algorithm's prediction accuracy. RF uses out-of-bag (OOB) samples to perform a type of cross-validation in parallel with the training process to test the random forest algorithm's prediction accuracy. Specifically, each tree is grown using a special bootstrap sample during the training phase. Any sequences will be ‘left out’ of the sample, and others will be reused in the sample, since bootstrapping is sampling with substitution from the training results. The OOB sample is made up of the sequences that were left out.

## 3.3. Reinforcement Learning

### *3.3.1. Reinforcement learning preliminary*

Reinforcement learning (RL) (Sutton & Barto, 1998), (Szepesvári, 2010), (Powell, 2011), (den Hengst et al., 2020), (Dong et al., 2020) is the area of machine learning concerned with how the software agents take actions in the environment and learns via heuristic methods. It is different from the other computational approaches by its nature in learning through an agent while interacting with a corresponding environment, and in learning without the need of a special supervision or a complete model of the environment. In most of the cases, the action, which the agent has reacted to the environment, will result in an immediate reward as well as a delayed reward resulting from an ubiquitous searching in the state space by the agent. The agent performs the actions in the environment in a way that it would maximize the overall prospectively cumulative rewards. The agent is not aware of what action to perform but, instead, it must discover which action yield the most prospective total reward by a fashion of trial and error.

By contrast to the traditional supervised learning in which labeled input and output pairs have to be presented explicitly, and the traditional unsupervised learning in which deals about finding a hidden structure in a collection of unlabeled data, the RL paradigm shows multiple important features. One such feature, which is also a major challenge in RL, is the in-between exploration exploitation dilemma. The RL agents focus on finding a balance between exploration (trying different actions) and exploitation (improving the current action) without guidance labels. In order to obtain a considerable total reward, the RL agent must refer to the actions that it ever experienced in the past, and try to find a best way in producing the maximal total reward, *i.e.*, it must exploit what it has already experienced in obtaining the reward, and it must explore in all its effort to select the best action. This dilemma is more like a state-of-art problem that mathematicians are always keeping to provide a better solution with no explicit ending. The second feature of RL is the nature in considering the whole problem domain as a goal directed approach of the agent interacting with the environment. The goal centric approach of RL persists its goal pursuing even the goal changes varying time. With objective horizons proceeding ahead, RL can sustain the goal tracking dynamically instead of that from a traditional supervised learning, which keeps only steadily no change. The dynamics carry on one another RL important feature to adapt the evolving dynamic of a changing system. The adaptive nature conducts RL to be capable of a planning role in on-line decision-making or control. A RL agent with its explicit changing goals with respect to its environment form typically an interactive platform.

Theoretically, RL is invented to manipulate a dynamic system in making a set of subsequent decisions most beneficial to achieving its goal state, *i.e.*, the agent tries to learn and achieve the goal in an uncertain and potentially complex environment. Beyond the agent and the environment, there are four main sub-elements of a RL system: the policy, the reward signal, the value function, and the model of the environment (Sutton & Barto, 1998). Figure 1 describes the basic process routes of RL showing the relationship with the sub-elements:

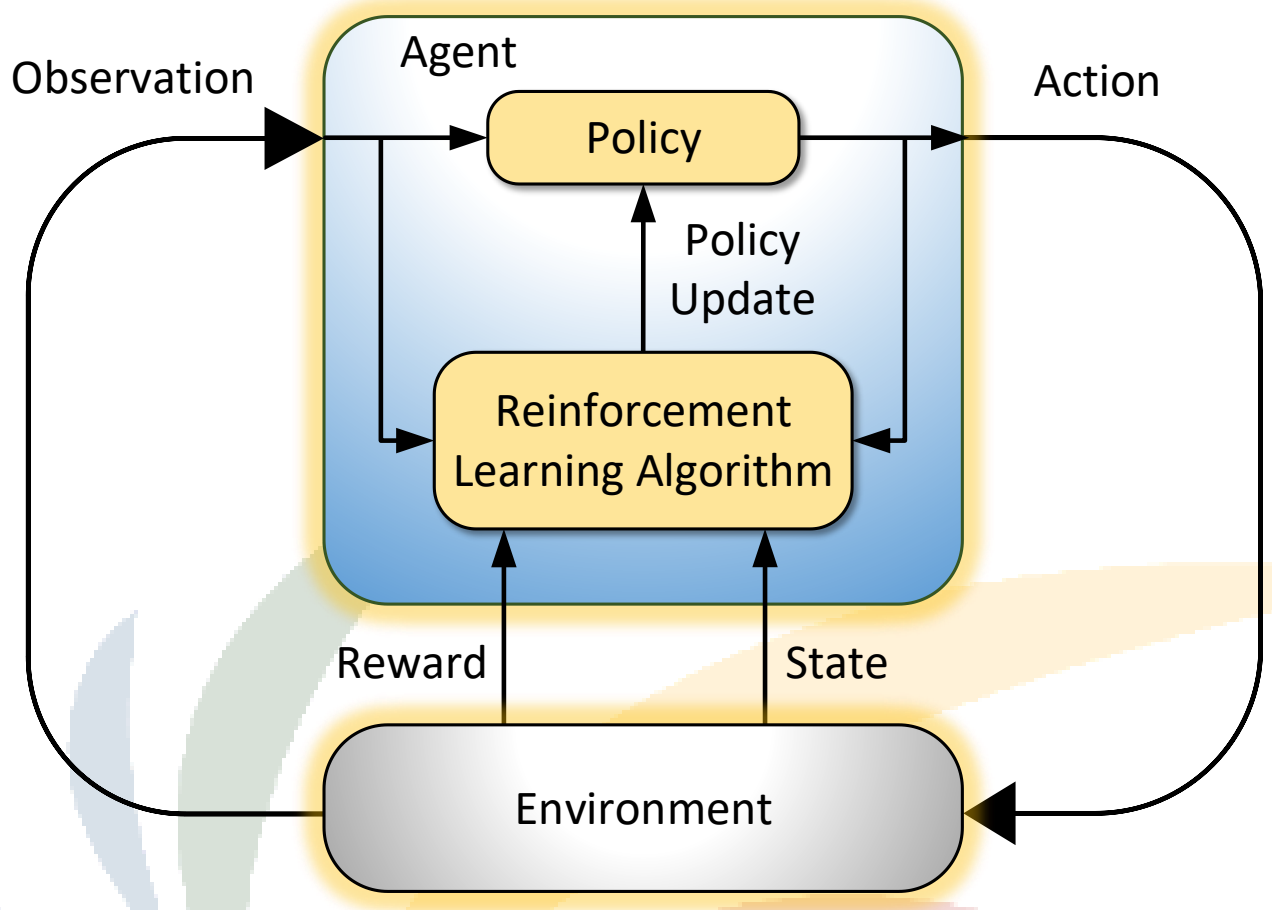

Figure 10: Basic reinforcement learning architecture

To describe the play of RL, let $S$ denote the set of environmental states and $A$ represent the possible actions. State gives a summary of history of the system status that used to determine the future evaluation. At each time horizon $t$, the agent observes the current state $s_t$ of the system and then performs an action $a_t$, which is generally selected from the predefined action space $A$. Thus, the state of the agent transforms to the new state $s_{t+1}$ from the current state $s_t$. For each action the agent is doing, a specific numerical reward $r_t$ would be assigned to the agent. The value of the reward $r_t$ determines whether the last action carried out is good or bad. Over the horizons of the process, the agent will manage to determine a proper action, state updates and the cumulative reward updates. The basic process of RL involves performing a sequence of actions, recording the relevant effects of those actions, estimating the relationship between the actions and the effects, and finally choosing the action that results in the most desirable effect. The agent is the player of the environment, which bears the primary goal of maximizing the cumulative rewards it receives in the lasting horizons $t = 0, 1, \dots, T$. The policy $\pi$ defines the agent's behavior at a given time horizon. It maps the states of the environment to the actions to be taken in the horizons. A return function $G$, summing up the cumulative rewards, determines the iterative headings of a RL agent. At each time horizon $t$ in $T$, the agent collects the reward by receiving a respective positive or negative score from the environment. The agent's sole objective is to maximize the return $G$ tentatively collected during the horizons $T$ by an estimating expectation. This return function constructs the primary basis for altering the policy $\pi$, where if an action $a_t$ selected by the policy $\pi$ resulted in a low return $G$, the policy $\pi$ might aggressively change to seek a more beneficial action $a_{t+1}$ to enrich the cumulative reward in the successive horizons $t + 1$ (Sutton & Barto, 1998).

The numerical reward $r_t$ the agent got could be expressed as a function $R$ which maps the scaler $r_t$ from the elements of current state $s_t$, current action $a_t$, and next state $s_{t+1}$ for $t = 0, 1, \dots, T$:

$$r_t = R(s_t \, , a_t, s_{t+1}) \tag{1}$$

Generally, the reward function $R(\cdot)$ could be a deterministic or stochastic function. RL is learning to map the processes from state space $S$ to action space $A$ depending aggressively on choosing to maximize the expected return with a fading discount rate $\gamma$, $0 \leq \gamma \leq 1$, the return, generally discounted return $G$, is expressed as:

$$G_T = r_t + \gamma r_{t+1} + \gamma^2 r_{t+2} + \gamma^3 r_{t+3} + \cdots + \gamma^T r_{t+T} = \sum_{k=0}^{T-t} \gamma^k r_{t+k} \qquad (2)$$

This discount rate $\gamma$ is a control to balance the agent's immediate reward and the future rewards by repeatedly weightings. While $\gamma$ is closer to one, all the rewards over the long term will be fully taken into account in the calculation of $G$. With $G$, the evolution of the environment mimics the behavior of the actual environment, and allows an inference to be made on how the environment will evolve. A value function $V$, defined especially from the expected return, which is cumulated by the agent in the span it concerned becomes a key for exploitation and exploration. In general, the value function $V$ indicates what deserves an expectation in a finite span.

A policy $\pi$ denotes namely the agent's behavior which maps from state $s$ to the action $a$. Due to the determination orientation, it comes therefore an exploration-exploitation dilemma. While the agent tries to maximize the same policy by doing a particular action in several trials and checking the overall reward during exploitation, it also looks for new possible policies and tries to get the best one to yield the maximal overall reward in exploration. It leads to a NP hard programming of the RL process, and makes it generally greedy. In the process, the agent must keep the learning balanced between the exploration and exploitation of the environment. Thus, the agent must know on when to explore or when to exploit. Referring to the general epsilon greedy action selection in RL terminology, the agent will take the greedy action with a probability $(1-\varepsilon)$ and the random action with a probability $\varepsilon$ in order to keep a balance with the exploration and exploitation (R.Sutton, 2017):

$a_t \leftarrow$

$$\begin{cases} arg_a \max Q_t(a), \text{selecting greedy action with probability } 1-\varepsilon \\ a \sim \text{Uniform}(\{a_1, \dots, a_k\}), \text{selecting random action with probability } \varepsilon \end{cases} \quad (3)$$

The value function allows agent to evaluate how "good" the police it adopted in a given state or a combination of a given state and a given action. It turns two kinds of definition, the state-value function $V_\pi$ and the action-value function $Q_\pi$, based on their corresponding different conditions shown below. The state-value function $V_\pi(s)$, which computes the expected return $G$ of being in the state $s$ following the policy $\pi$, is defined formally by:

$$V_\pi(s) = \mathbb{E}_\pi[G_t | s_t = s] = \mathbb{E}_\pi[\sum_{k=0}^{T-t} \gamma^k r_{t+k} \mid s_t = s], \text{for all } s \in S \quad (4)$$

where $\mathbb{E}_\pi$ takes the expected value of the conditional return following the policy $\pi$. The values of consecutive states satisfy a recursive relationship called Bellman equation:

$$V_\pi(s) = \sum_a \pi(a|s)( r(s,a) + \gamma \sum_{s'} P(s'|s,a) V_\pi(s')) \quad (5)$$

Similarly, the action-value function $Q_\pi(s,a)$, which calculates the expected return $G$ of being in the state $s$ and taking the action $a$, and thereafter following policy $\pi$, is also by:

$$Q_\pi(s,a) = \mathbb{E}_\pi[G_t | s_t = s, a_t = a] = \mathbb{E}_\pi[\sum_{k=0}^{T-t} \gamma^k r_{t+k} \mid s_t = s, a_t = a], \text{for all } s \in S \ \& \ a \in A \quad (6)$$

In short, an optimal policy would consist of the actions that lead to the greatest return over time. So, there is a straightforward relationship between the value function $V_\pi(s)$ and the action value function $Q_\pi(s,a)$ as follows:

$$V_\pi(s) = \sum_a \pi(a|s) Q_\pi(s,a) \quad (7)$$

$$Q_\pi(s,a) = r(s,a) + \gamma \sum_{s'} P(s'|s,a) V_\pi(s') \tag{8}$$

The optimal value function $V^*(s)$ is defined as the maximum amount of expected return that an agent can accumulate from a given state. The optimal value function in state $s$ is given by $V^*(s) = \max V_\pi(s)$, *i.e.*, the maximum value among the individual policies:

$$V^*(s) = \max V_\pi(s) = \max \mathbb{E}_\pi[\sum_{k=0}^{T-t} \gamma^k r_{t+k} \mid s_t = s] \tag{9}$$

To satisfy the optimal Bellman equation, the optimal values of consecutive states is as:

$$V^*(s) = \max(r(s,a) + \gamma \sum_{s'} P(s'|s,a) V_\pi^*(s')) \tag{10}$$

The optimal action value function $Q_\pi^*$ which is related to $V_\pi^*$ is thus followed and defined as:

$$V^*(s) = max_a\, Q^*(s,a) \tag{11}$$

$$Q^*(s,a) = r(s,a) + \gamma \sum_{s'} P(s'|s,a) V_\pi^*(s') \tag{12}$$

### *3.3.2. Model-free vs. Model-based*

Algorithms of RL are sought to help the agent in finding the optimal policy, which yields the greatest return in every episode. RL algorithms are of two types as model-based RL and model-free RL (Sutton & Barto, 1998), (Tordesillas & Arbelaiz, 2019). The difference between the model-based RL and model-free RL is with the transition probabilities and the reward function. In model-based RL, the transition probabilities and the reward function can be estimated empirically if we do not know the MDP of a particular problem. With the transition probabilities and the reward function, the MDP can be solved with the dynamic programming methods of policy iteration and value iteration. In Model-free RL, we ignore the model. The evolution depends only on sampling and simulation to estimate rewards even we do not know the inner operation of the system. In other words, the agent does not need to build a model to assess the environment in this fashion. The model-free methods are broadly divided in to two branches such as value-based and policy-based methods. The main difficulty of the model-free algorithms is that the unavailability of a ground truth representation of the environment. By contrast, the model based RL evaluates the environment states by a given model. The agent is then advantageous of the comfortable reference model to plan ahead the successive reactions.

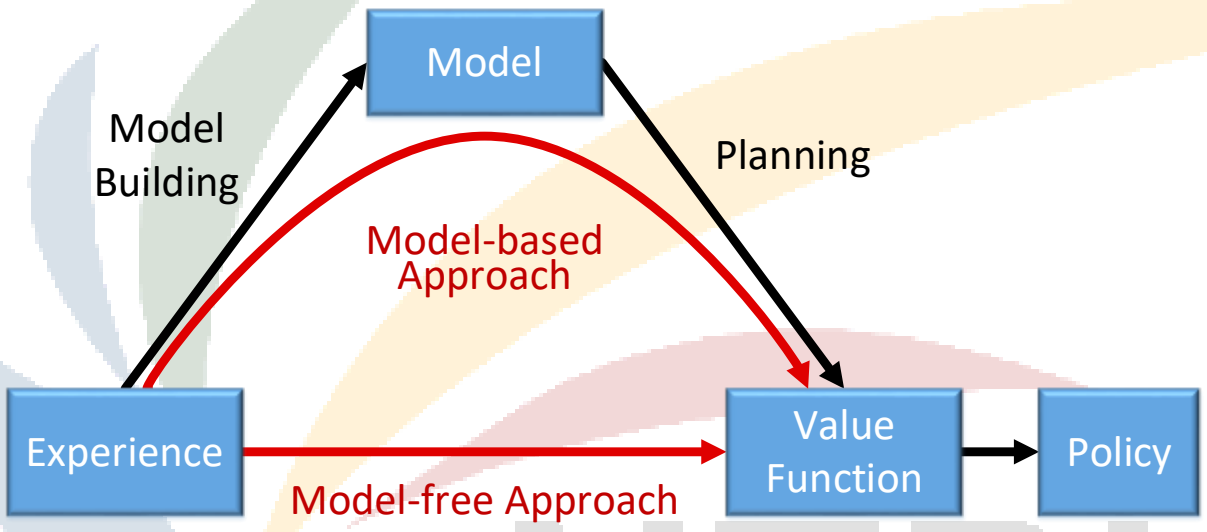

Figure 11: Difference between model-based RL and model-free RL (Sutton & Barto, 1998)

### *3.3.2.1. Model-free Q-Learning*

The $Q$ learning uses temporal differences (TD) to estimate the value of $Q^*(s_t, a_t)$. TD is an agent learning from an environment through episodes with no prior knowledge of the environment. It categorized the $Q$ learning a model-free, off-policy method for calculating an optimal policy for an RL agent. It uses a $Q$-function, $Q\colon S \times A \rightarrow \mathcal{R}$, to parameterize the expected discounted reward by corresponded state and action. The optimal $Q$-value for a state-action pair $(s_t, a_t)$ is given by a rearrangement of the Bellman equation where $s_t$ and $a_t$ represent the state and action at the current time horizon $t$:

$$Q_\pi(s_t, a_t) = \mathbb{E}[r_{t+1} + \gamma r_{t+2} + \gamma^2 r_{t+3} + \cdots \mid s_t, a_t] \tag{13}$$

The $Q$ values generated will be stored in a simple lookup table known as the $Q$-table, which will guide to select the best action at each state. The basic steps of $Q$ learning process are:

1. Initialize the $Q$-table
2. Observe the current state, $s_t$, and select an action $a_t$ by referencing the of $\varepsilon$-soft, $\varepsilon$-greedy or softmax action selection policies
3. Get the state $s_t$ resulted from the above action $a_t$, and measure the reward corresponded $r_t$
4. Update the $Q$-value of $s_t$, based on both the observed reward and the maximum reward possible from the next state, by using the learning rule with a learning rate $\alpha$ below:

$$Q_{new}(s_t, a_t) =$$

$$Q_{old}(s_t, a_t) + \alpha[r(s_t, a_t) + \gamma \max_a Q(s_{t+1}, a) - Q_{old}(s_t, a_t)] \tag{14}$$

5. Set the state to the new state, and repeat the procedures 2-5 until the terminal state arrived.

#### *3.3.2.2. Model-based RL*

From the aspect of MDP, RL is a looped solution which maximizes rewards for the actions repeatedly. An episode $\Sigma = \langle s_0, a_0, s_1, a_1, \ldots, s_T, a_T \rangle$ is produced under a probability $p_\theta$ generally written as:

$$p_\theta(\Sigma) = p(s_1) \prod_{t=0}^{T-1} \pi_\theta(a_t|s_t) \cdot p(s_{t+1}|s_t, a_t) \tag{15}$$

$$\max_{\theta} \mathbb{E}_{\Sigma \sim p_{\theta}(\Sigma)}[\sum_{t=0}^{T-1} r(s_t, a_t)] \tag{16}$$

where the policy $\pi_{\theta}(a_t|s_t)$ is a probability mapping from state $s_t$ to actions $a_t$, and $p(s_{t+1}|s_t, a_t)$ is the state transition probability. To elucidate alternatively the MDP from the aspect of control, where the episode $\Sigma$ is recognized as a sequence of state-control pairs $(\mathbf{x}_t, \mathbf{u}_\mathbf{t})$:

$$\langle s_0, a_0, s_1, \dots, s_T, a_T \rangle \rightarrow \langle \mathbf{x}_0, \mathbf{u}_0, \mathbf{x}_1, \mathbf{u}_1, \dots, \mathbf{x}_T, \mathbf{u}_T \rangle, \tag{17}$$

the decomposed terms $\pi_{\theta}(a_t|s_t)$ and $p(s_{t+1}|s_t, a_t)$ can thus be explained as "activator" and "mechanism of system", respectively. A RL corresponds to the fashion of model-free if the reinforcement focuses on the heuristics of the "activator" and makes the "mechanism of system" adapted passively. By contrast, a model-based RL corresponds to that focuses on evolving the "mechanism of system", and adapts the "activator" minorly. Here, we concentrated more on the elaboration of the term of $p(s_{t+1}|s_t, a_t)$. While $p(\cdot)$ represents the rate of a successful transition, the state transition mechanism $(s_{t+1}|s_t, a_t)$ can be recognized as a model of the evolving system, indeed the environment of the RL problem, behind. As termed as a model, the mathematics of model approximation is often abbreviated as:

$$s_{t+1} = f(s_t, a_t) \text{ or } \mathbf{x}_{t+1} = f(\mathbf{x}_t, \mathbf{u}_t), \tag{18}$$

Alternatively. With the model approximation, the RL problem can thus be turned to become,

$$\min_{u} \sum_{t=0}^{T-1} c(\mathbf{x}_t, \mathbf{u}_t, \mathbf{x}_{t+1}) \quad \text{s.t.} \quad \mathbf{x}_{t+1} = f(\mathbf{x}_t, \mathbf{u}_t) \tag{19}$$

where $c(\mathbf{x}_t, \mathbf{u}_t, \mathbf{x}_{t+1})$ denotes the cost to evolve the system from horizon $t$ to horizon $t+1$ which is analogously a reciprocal function of the reward function $R(\cdot)$ in RL, as we have already stated in (1).

The approximation can therefore be categorized as global approximation and local approximation. (J. Fu, 2020). In this topic, the two adopted models, the TS and CC models, will be generalized as global model to generate data to validate the RL approach for the cancer chemotherapy, and then be specifically reduced as a local model to learn cooperatively the RL scheme to suggest the optimal chemotherapy treatment.

#### *3.3.2.3. The paradigm of Dyna-Q*

Dyna Q is a powerful algorithm that combines Q-learning and Q-planning where planning is done online while interacting with the environment. Here, models are used to produce and improve a policy. Dyna architectures use machine learning algorithms to approximate the conventional optimal control technique known as Dynamic Programming (DP). Dyna architectures integrate trial-and-error reinforcement learning and execution-time planning into a single process operating alternately on the world and on a learned model of the world. Real or simulated experience improves both the model via model learning and value function/policy via direct RL. Model contains state, action, next state and reward tuples. Thus, the model can be both improved and queried to get to the next state in planning part. The possible relationships between experience, model, values, and policy are summarized in the following figure. The overall architecture of Dyna Q algorithm which is described in the reference book, "Reinforcement Learning: An Introduction" by Richard S. Sutton and Andrew G. Barto, is shown in Figure 3.

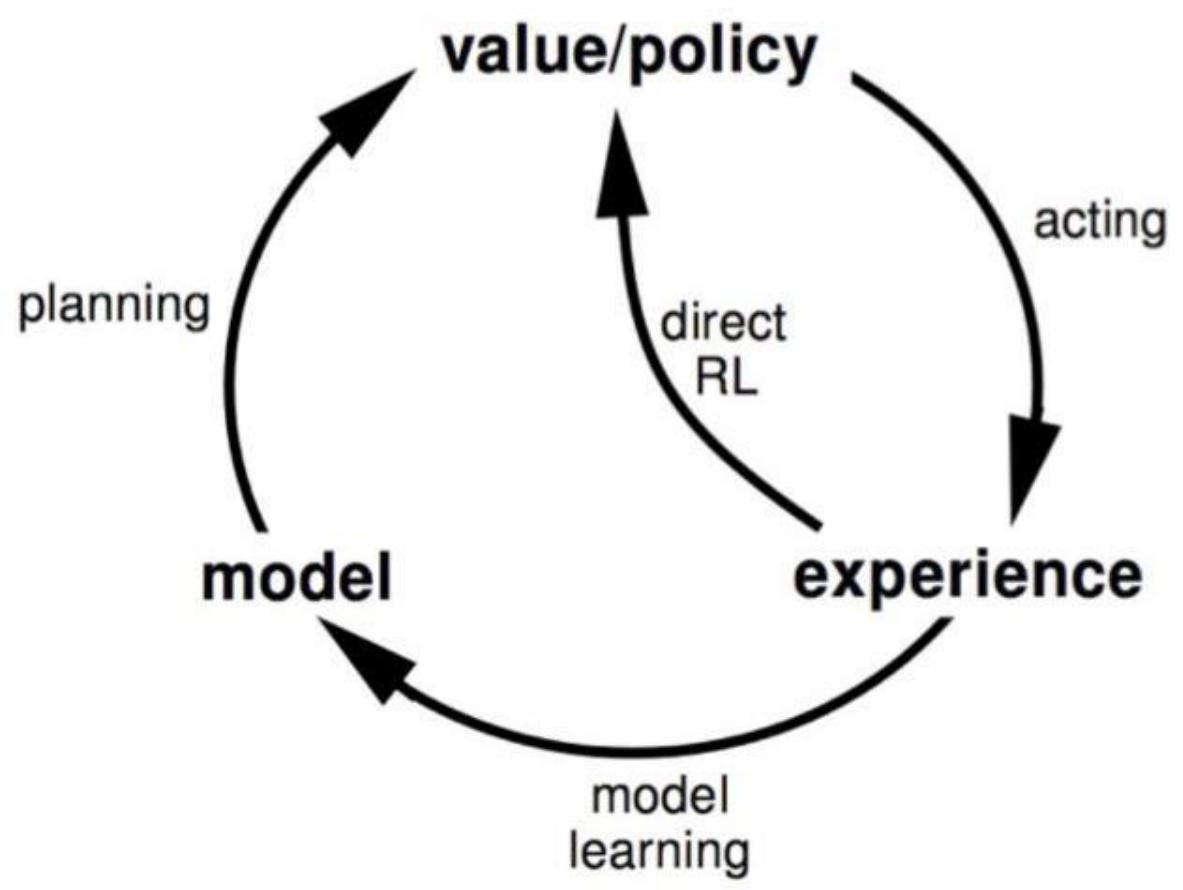

Figure 12: Relationship among Learning, Planning and Acting

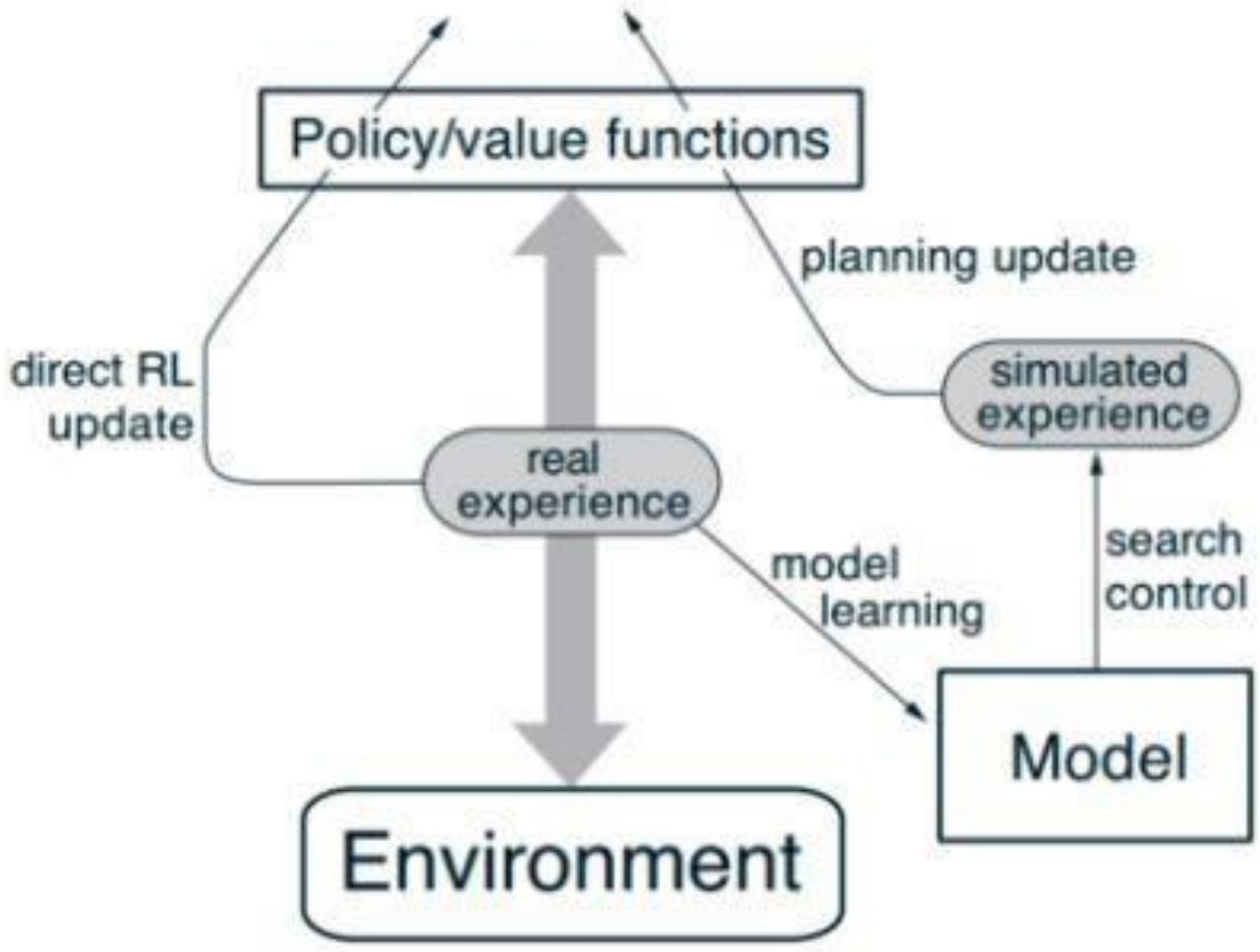

Figure 13: The General Dyna Architecture

The central column of the general Dyna architecture represents the basic interaction between the agent and environment, which gives rise to a trajectory of real experience. The arrow on the left side represents the direct RL update, which is operating on real clinical experience of the oncologist. The right side express the details of the model-based processes. The Chemotherapy RL model will learn from the real clinical experience and will give rise to simulated experiences with the intervention of the model. The term search control refers to the process of selecting the states and actions for the simulated experiences generated by the model.

### *3.3.3. Action selection mechanisms*

In reinforcement learning, action selection mechanisms have been utilized in order to cater with the exploration and exploitation dilemma. These action selection algorithms support agents for making the decision of action selection via taking advantage of what is already known about the environment or exploring further. Following are the action selection algorithms.

**Greedy** – In the greedy type of action selection, always the action with the highest estimated return will be picked as the next action. However, it is not considered as a smart strategy, since we would get stuck in a local minimum.

**e-epsilon decay** – At the beginning of reinforcement learning, the exploration rate will be defined as 'epsilon' that is set to 1. The reason is at the beginning, agent do not know much about the environment. Therefore, he will do much exploration. With the time, he will get to know more about the environment and will find the best way to do the action.. Thus, exploration will be reduced and he will go deeply on the best way more and more. Thus exploration rate would be minimized and will be 0 or near to 0 finally.

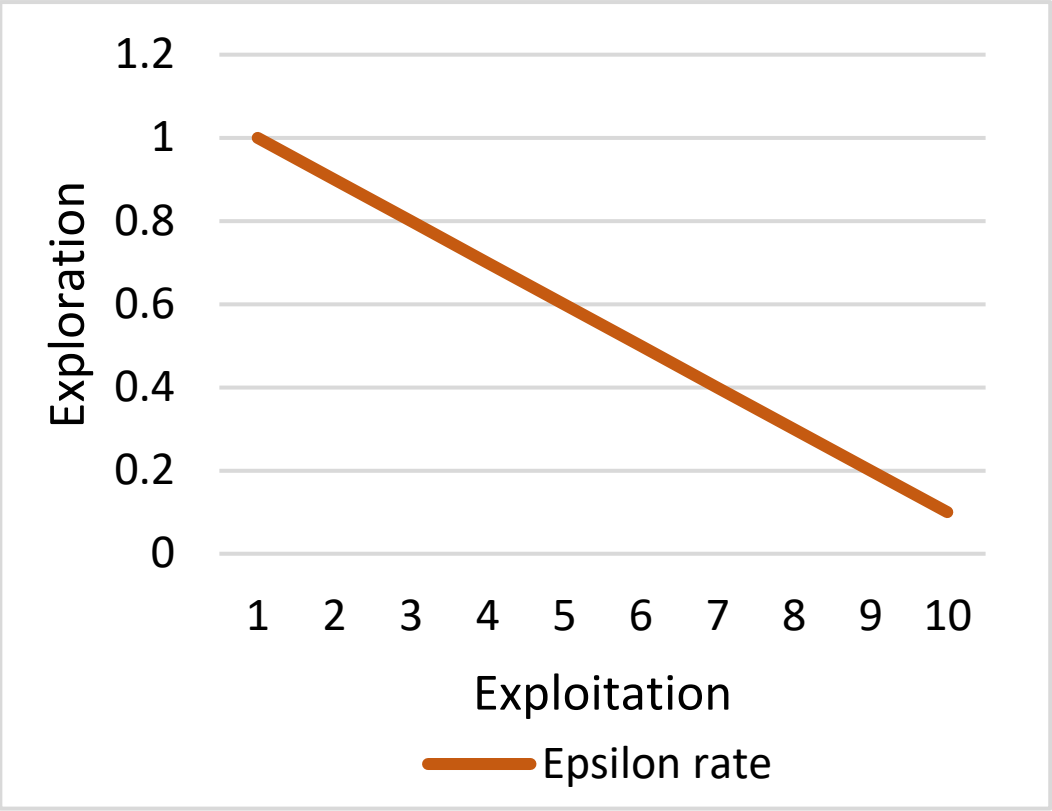

Figure 14: Exploration and exploitation behavior in e-epsilon decay

**Softmax** - In this type of action selection, action probabilities will be found as a graded function of estimated value. Common types of softmax methods are Gibbs or Boltzmann distribution. In this action selection method, exploration will automatically stop after some time.

### *3.3.4. Online, Off-policy and Offline RL*

In the classical RL setting, three basic variants are identified based on the nature and the behavior of the model building. They are online RL, off-policy RL and the offline RL. The online RL is the most common RL platform, which the data is collected each time and the policy is modified real time. It is the common paradigm in RL which assumes that an agent frequently interacts with the environment in an online manner, and learns using its own collected experiences. In the off-policy RL, the data is retained to update the policy, and, still, the new data will be collected periodically as the policy changes. In offline RL, data will be collected at one time as in the supervised learning approach and then will be used to choice the optimal policies without real-time updating the data collected. Therefore, the offline RL, also known as data driven RL or batch RL, bring forward the data driven approaches that carry potential to scale up learning tasks to several common real-world adaptive decision-tracking applications such as robotics, recommendation system, dialogue generation, autonomous driving, and safety-critical applications (Lange et al., 2012), (Gottesman et al., 2018), (Agarwal et al., 2020). In certain practical medical applications, the offline RL methods would promise a considerable advantage over the online RL and the off-policy RL. The offline RL approaches can be paired with small amounts of online fine tuning in realistic use, where the policy is deployed ahead with a well-arranged offline process and fine-tuned with additionally gathered data to develop online. Figure 3 shows the descriptive variations of the three variants of reinforcement learning.(Levine et al., 2020)

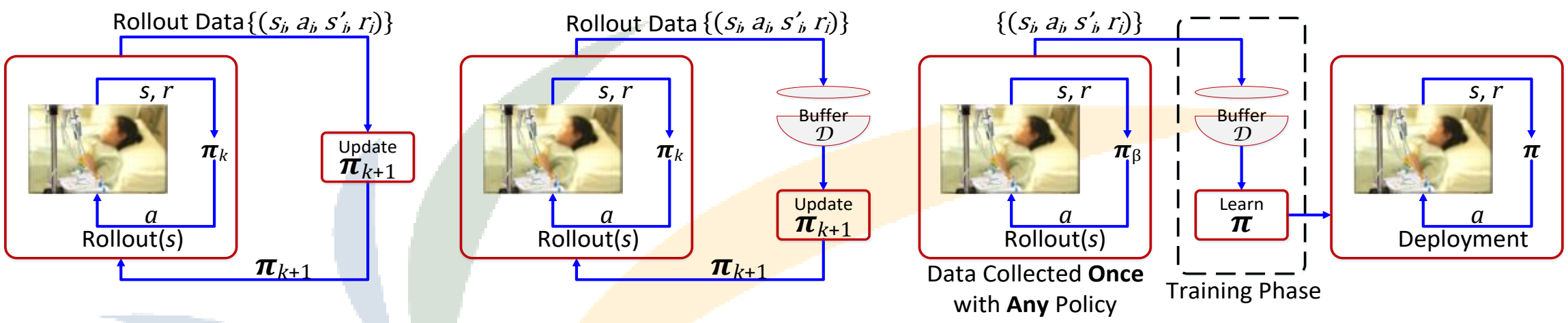

(a) Online RL (b) Off-policy RL (c) Offline RL

Figure 15: Three variants of RL

In this study, major focus will be laid towards the offline RL paradigm depending on the potential facts that the active exploration of individual patient's chemotherapy data is in practical prohibitive, costly and also dangerous. Thus, the agent will exploit the previously collected data on the patients who are receiving chemotherapy, and try to find the optimal schedule with the existing data. Recent literatures prove that the successful deep reinforcement learning algorithms are being adopted to the offline setting, and demonstrate promising potential in a number of domains; however, the offline fashion is still challengous due to the distributional shift between the offline training data and real states visited by the online policy. The most successful prior methods approached in the literatures are kinds of offline RL which constrains the policies to generalize the real unknown states from the available states (Fujimoto, Conti, et al., 2019), (Kumar et al., 2019), (Yu et al., 2019), (Fu et al., 2020). As confirmed by (Bazrafshan & Lotfi, 2016), (Fujimoto, Conti, et al., 2019), (Kumar et al., 2019), (Yu et al., 2019), (Fu et al., 2020) the facts that the offline RL would arise as a predominant methodology in solving problem associated with the medical dynamic treatments.

### *3.3.5. Offline Reinforcement Learning algorithms*

Offline reinforcement learning (RL) applies entirely to the issue of learning policies from a huge batch of data obtained previously without exploration. This problem setting provides the promise of using certain datasets without any expensive or risky aggressive experimentation to obtain policies. However, this becomes problematic and difficult, due to the distributional shift between the offline training data and those states visited by the studied policy.

In the recent research and literature on offline RL, it is visible that the researchers have made tremendous efforts in finding proper algorithms in the task of finding optimal policy via offline RL scheme. Among them Fitted Q iteration (Ernst et al., 2005), Batch Constrained deep Q-Learning (BCQ) (Fujimoto, Meger, et al., 2019a), Bootstrapping Error Accumulation Reduction (BEAR) (Kumar et al., 2019), Random Ensemble Mixture (REM)(Agarwal et al., 2020), Implicit Reinforcement without Interaction at Scale (IRIS) (Mandlekar et al., 2020), Behavior Regularized Actor Critic (BRAC) (Wu et al., 2019) Conservative Q-Learning (CQL) (Kumar et al., 2020) reserve promising domains occupying major research milestones in the literature. Among the wide area of the potential algorithms, this study focused on approaching the Conservative Q-Learning algorithm (CQL) due to features stated in the next section the algorithm encompassing.

#### *3.3.5.1. Conservative Q-Learning (CQL) algorithm*

The established Bellman update is as follows.

$$\delta(s,a) = ||\ Q(s,a) - (r(s,a)\ +\ max_{a'}Q(s',a'))\ ||^2 \quad (20)$$

This term mathematically express the Bellman error, which we are trying to minimize for the transitions in the form $(s, a, r, s')$. The process of Bootstrapping means we are approaching the best Q-value that our Q-function thinks it can get by taking the action a' and use that to the current estimate of the Q-value for the state s. Since our Q-function is not optimal (especially during early phases of training), this target value will surely be wrong sometimes, and will lead us to propagate errors to the estimate of the Q-value of s. This is known as the Bootstrapping error. In the online RL setting, errors in the Q-function estimation will cause the policy to take wrong actions. Thus, in order to mitigate the Bootstrapping error, we train the model for many episodes, experience the outcomes of taking those actions, and correct the mistakes eventually. However, in the offline RL setting, this is impossible due to two main reasons as follows.

- Because the policy does not interact with the environment and does not gather more data, therefore has no way to know it is propagating wrong values and correct them.
- Distribution shift problem: Off-policy evaluation can be difficult since the new policy's state-action visitation distribution (target policy) may differ significantly from that of the behavior policy.

Thus, the objective of the Q-function in offline RL is to minimize the empirical risk, the empirical error against the target policy, under the distribution of the states and actions induced by the behavior policy.

$$min_Q \quad E_{(s,a)\ \sim\ \pi_\beta(s,a)}[\ (Q(s,a) - y(s,a))^2\ ] \tag{21}$$

Where, $\pi_\beta(s,a)$ denotes the behavior policy and $y(s,a)$ denotes the target policy. The term $Q(s,a) - y(s,a)$ is the empirical error or the Bellman error we are attempting to minimize in the offline RL setting.

The simple meaning of the term "Conservative" refers to "Resistant to the deviation from established norms". That implies that this algorithm deviates from the accepted domain in solving a particular problem. CQL is a SAC-based data-driven deep reinforcement-learning algorithm, which achieves state-of-the-art performance in offline RL problems. CQL mitigates overestimation error by minimizing action-values under the current policy and maximizing values under data distribution for underestimation issue.

When performing naïve Q-learning on offline data, highly optimistic Q-values will occur. Because to minimize the bellman error, the Q values should be maximum. CQL addresses this issue with a simple addition to the objective function.

**In standard Q-learning the loss is:**

$$L = \ E_{s,a\sim D}[\delta\ (s,a)] = E_{s,a\sim D}\ [\ ||Q(s,a) - (r(s,a)\ + \gamma max_{a'} Q(s',a'))\ ||^2 \quad ] \tag{22}$$

$$where, \quad \delta\ (s,a)$$
$$= Q(s,a)$$
$$-\left(r(s,a) + \gamma max_{a'}Q(s',a')\right)\ is\ the\ Bellman\ Error$$

In CQL, following term will be added,

CQL, following term will be added,

$$L_{cql} = E_{s,a\sim D}[\delta\ (s,a)] + \alpha.E_{s\sim D,a\sim\pi}[Q\ (s,a)] \quad (23)$$

The term $\alpha.E_{s\sim D,a\sim\pi}[Q\ (s,a)]$ will push down the big Q-values.

By trying to minimize the Q-values of the state-actions our policy thinks are high, they gradually force down all those optimistic errors and force the Q-values to be no larger than what they actually should be. The authors prove that with a proper choice of alpha, the resulting Q-function can be bounded by the "true" Q-values (that are unknown), **and is thus a conservative estimate of those values.**

They empirically demonstrate that the predicted Q-values are lower than the ones gained by deploying the learned policy.

In fact, the resulting Q-values are a bit too conservative, and the authors propose another addition to the loss function:

$$L_{cql} = E_{s,a\sim D}[\delta\ (s,a)] + \alpha.E_{s\sim D,a\sim\pi}[Q\ (s,a)] - E_{s,a\sim D}[Q\ (s,a)] \quad (24)$$

$$E_{s,a\sim D}[\delta\ (s,a)] \rightarrow Standard\ Bellman\ Error$$

$$\alpha.E_{s\sim D,a\sim\pi}[Q\ (s,a)] \rightarrow will\ push\ down\ the\ big\ Q-values$$

$E_{s,a\sim D}[Q\ \ (s,a)]$

$$\rightarrow will\ push\ down\ the\ Q-values\ of\ state\ action\ pairs\ in\ data$$

This addition：

- Encourage the policy to stick to more familiar actions and making Q-values less conservative.
- Authors prove that the expected resulting Q-values are upper bounded by the “true” Q-function
- Demonstrate that this variant produces better results and more accurate Q-values.

### *3.3.5.2. Model Based Offline Policy Optimization (MOPO) algorithm*

The Model Based Offline Policy Optimization (MOPO) algorithm has been shown a great deal of simplicity and prior effective performances and effective results on the D4RL benchmarks. MOPO is a model-based RL approach for offline policy optimization. In order to produce new dynamics data with uncertainty penalties, MOPO leverages the probabilistic ensemble dynamics model.

In MOPO, the authors construct a conservative MDP by augmenting the reward function as follows.

$$\widetilde{r}(s,a) = \hat{r}(s,a) - \lambda u(s,a) \tag{25}$$

Where $\hat{r}$ is the learned estimate of the reward and $u$ is the estimated uncertainty for the model transition.

The authors specifically follow MOPO in using the maximum standard deviation across an ensemble of probabilistic dynamics models as our measure of uncertainty.

Therefore, they have decomposed the Q-function in this conservative MDP as:

$$Q^{\pi}(\mathrm{s,a}) = \hat{Q}_r^{\pi}(\mathrm{s,a}) - \lambda Q_u^{\pi}(\mathrm{s,a}) \tag{26}$$

Where $\hat{Q}_r^{\pi}$ represents our estimate of the expected discounted sum of rewards in the real MDP and $Q_u^{\pi}$ represents our expected discounted sum of uncertainty penalties.

Now at convergence, if our policy π deviates from the behavioral policy πβ that collected the data, then the authors expect for all states in the conservative MDP that:

$$\mathbb{E}[Q^{\pi}] \geq \mathbb{E}[Q^{\pi\beta}] \tag{27}$$

Thus, by plugging in our decomposition we get,

$$\mathbb{E}\left[\hat{Q}_r^{\pi}(\mathrm{s,a})\right] \geq \mathbb{E}[\hat{Q}_r^{\pi\beta}(\mathrm{s,a})] + \lambda\mathbb{E}[Q_u^{\pi}(\mathrm{s,a})] \tag{28}$$

While in theory, with well-calibrated uncertainty estimates and a proper tuning of λ, this should lead to only safe policy improvements over the behavioral policy, in practice it seems that MOPO is often unable to recover expert-level performance when it is provided in the offline dataset.

#### *3.3.5.3. Incorporating MOPO in the problem domain*

The ensemble dynamics model consists of $N$ probabilistic models $\{T_{\theta i}\}_{i=1}^{N}$. At each epoch, new transitions are generated via randomly picked dynamics model $T_{\theta}$.

$$s_{t+1}, r_{t+1} \sim T_{\theta}(s_t, a_t) \tag{29}$$

Where $s_t \sim D$ for the first step, otherwise $s_t$ is the previous generated observation, and $a_t \sim \pi\ (.\,|s_t)$. The generated $r_{t+1}$ would be far from the ground truth if the actions

sampled from the policy function is out-of-distribution. Thus, the uncertainty penalty regularizes this bias.

$$\tilde{r}_{t+1} = r_{t+1} - \lambda max_{i=1}^{N} \ |\Sigma_i(s_t, a_t)| \tag{30}$$

Where $\Sigma(s_t, a_t)$ is the estimated variance.

Finally, the generated transitions $(s_t, a_t, \tilde{r}_{t+1}, s_{t+1})$ are appended to dataset $D$.

This generation process starts with randomly sampled $n_transitions$ till $horizon$ steps.

### 3.4. Summary

This chapter provides an in-depth description of the technological background and theoretical aspect adopted in our solution to planning of chemotherapy treatments in the offline reinforcement learning setting based on the Conservative Q-Learning (CQL) and Model Based Offline Policy Optimization (MOPO) algorithm. Next chapter explains the Approach of the research with key components of the approach.

CHAPTER 4

# 4. APPROACH

## 4.1. Introduction

In chapter 03, the technology adopted in the solution to reinforcement learning based control of cancer chemotherapy treatments was discussed. This chapter presents the approach engaged by the researcher in driving towards a solution.

## 4.2. Hypothesis

The hypothesis of the research is that Offline Reinforcement Learning can be adapted to solve the problem of real clinical colon cancer chemotherapy treatment planning. The inspiration behind this hypothesis is coming from how the online reinforcement learning have been adopted in finding optimal treatment schemes by using the mouse models which are problematic to approach in the real human treatment setting.

## 4.3. Input

The input for the system would be the patient's clinical diagnosis measures evaluated at the end of each chemotherapy treatment course. They will include the tumor length, tumor width, their demographic factors such as age, gender, weight and height, patient's side effects, Drug response measure and the hospital standard patient scoring measures of the Charleson's Comorbidity index (CCI) and the Eastern Cooperative Oncology Group Performance Scale (ECOG PS) measure. With continuing the research broadly, the input data might be the data, perhaps taken from clinical trials.

## 4.4. Output

The output would be the personalized optimal treatment schedule, which contains the amount of drug amount and the corresponding relative dose indexes for the 6 types of Colon Cancer Chemotherapy treatment drugs Cetuximab, Bevacizumab, Irinotecan, 5FU, Xeloda and Oxaliplatin.

### 4.5. Process

The process followed or the technology adopted would be the offline reinforcement learning on the previous collected experiences.

First, the clinical diagnosis information of the colon patient or the observations would be collected and they will be inputted to the dataset. The information will be transformed in to transitions. The clinical observations would be the tumor size generated with the observed tumor length and tumor width, the demographic factors of age, gender, weight and height, patient's side effects, Drug response measure and the hospital standard patient scoring measures of the Charleson's Comorbidity index (CCI) and the Eastern Cooperative Oncology Group Performance Scale (ECOG PS) measure. For each of the transition, the respective real patient state would be labelled by using the expertise knowledge of the oncologist. The reward for each transition would be calculated by using the established reward function definition. A machine-learning model would be modelled for the determination of the real patient state. The model was fine-tuned with hyper parameter optimization and the final optimized best model was acquired.

Second, the replay buffer with the collected transitions of $\{(s_i, a_i, s_i', r_i')\}$ was finalized. The transitions was divided in to episodes based in the patient; meaning one episode would be equal to the sequential transitions associated with one single patient. Then the dataset will be inputted in to the colon cancer Reinforcement Learning module in order to predict the optimal treatment schedule. The Conservative Q-Learning (CQL) and the Model Based Offline Policy Optimization (MOPO) algorithms will be used in order to predict the optimal sequential policy associated. The Offline RL model was fine-tuned with hyper parameter optimization and the final optimized best model was acquired. The transitions would be generated considering the exploration and exploitation strategies.

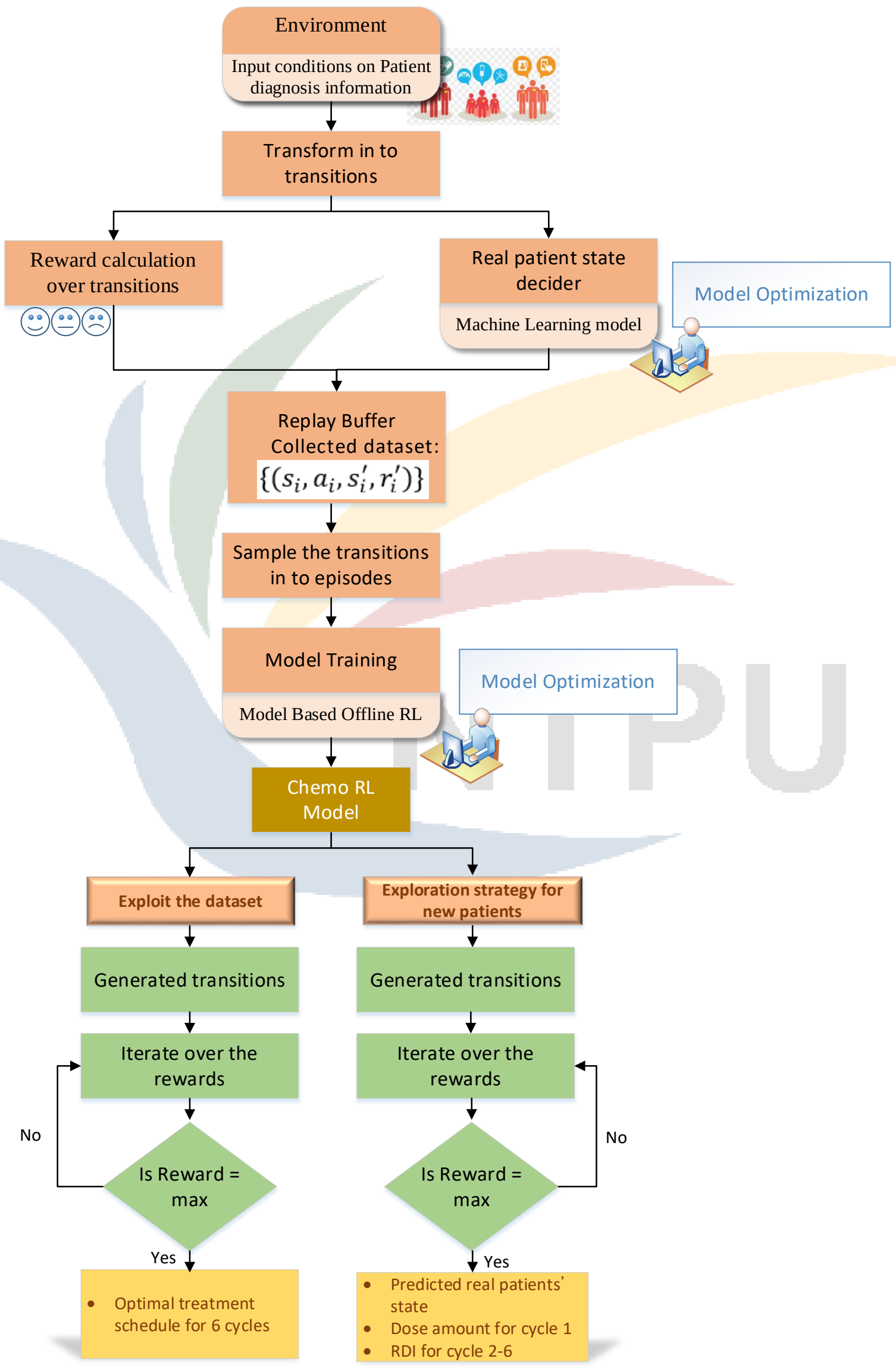

Figure 16: Proposed system flowchart

More descriptively, the aim we are intend to achieve in this research can be broadly described in the following simple flow chart.

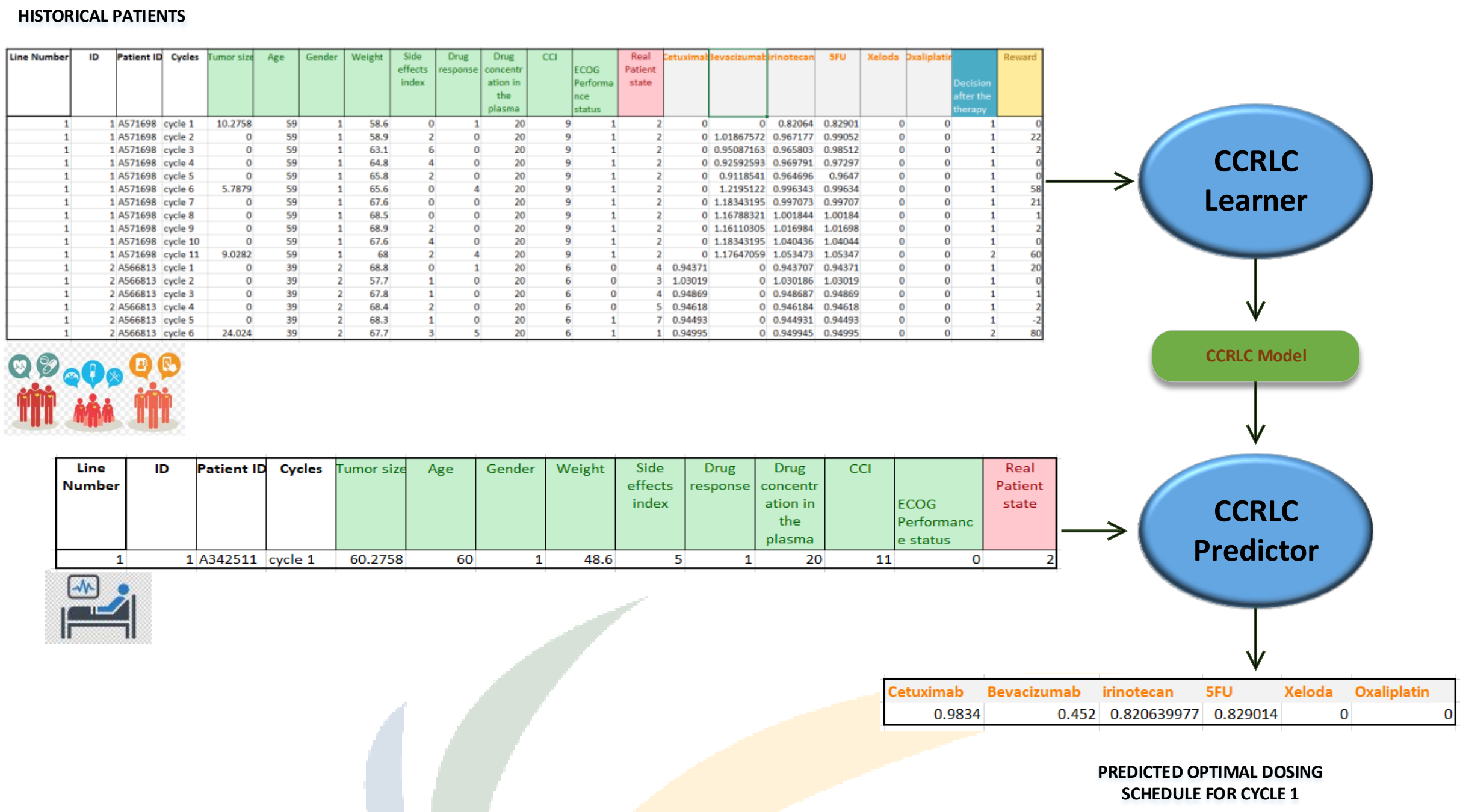

Figure 17: System basic aim

## 4.6. Features

The overall features of the system include the following.

- Integration of model-free and model-based Reinforcement Learning
- Determination of the real patient state via machine learning
- Optimized treatment scheduling for individual patients
- Development based on the Exploration and Exploitation strategy on the chemotherapy treatment planning domain
- Uncertain environment modelling (Can be modified/customized and adapt to the entirely new environments)
- Support for major agent behaviors

## 4.7. Users

The end users of this system would be researchers of reinforcement learning and the researchers of cancer related areas, in order to critically review the applicability of offline reinforcement learning in the Chemotherapy treatment-planning problem. The system outcomes would be greatly assisted on the decisions of the oncologists and the nurses in order to evaluate whether their decision on the Chemotherapy type of dose or the amount of dose are reliable or appropriate on the current condition of the individualized patients.

## 4.8. Summary

This chapter presented the hypothesis, input, output, features, users and process of the proposed solution. In particular, the researcher has explained the input processing concerning the with the proposed Reinforcement learning model. The proposed approach and the processes that has been used to convert the inputs to outputs have been described. The overall features of the new system also have been pointed out at the end of the approach section. The next chapter will present the design of the proposed solution.

CHAPTER 5

# 5. DESIGN

## 5.1. Introduction

The design section describes the design of the developed system by elaborating the overall system design, system model, system architecture, software architecture, modular architecture and simulation design. The modules of the system and their relationships have been elaborated using diagrams. In addition, what each module performs in the system is also described.

## 5.2. System Design

At the onset of the problem, Chemotherapy will be modelled as a Markov Decision Process (MDP).

## 5.3. System Model

The system model with the agent, observations, state definition, actions, rewards, time, environment are designed as follows on the basis of MDP. The Markov Decision Process (MDP) formally defines a reinforcement learning environment. Theoretically, there are two major types of MDP, based on the discrepancies between the agent state and the environmental state, such as the MDP and the POMDP (Partial Measurable Markov Decision Process).

**Agent state** – This is the internal representation of the agents or the knowledge used to determine the actions of the agent. He will also see the historical decisions. In this problem setting, the agent state will be the oncologist's expert knowledge.

**Environment state –** It is the private representation of the environment or the evidence used by the environment to see what happens next. In this problem setting, the environment state will be the dynamic nature of the patient which is characterized with the patient demographic factors such as the age, gender, weight and height, patient comorbidities, patient performance status etc. These factors will influence in the agent's sequential decision process.

The agent will observe the state of the environment directly in the full observability case, and here the state of the agent is the same as the state of the environment. Formally, this is a decision process for Markov. In the partial observability scenario, the environmental condition would be implicitly observed by the agent. In this problem setting, the oncologist will look at the type of patient, the status and determine the medication, the doses. The state of the agent is therefore not equal to the state of the environment. A partial observable Markov Decision Process is our case (POMDP).

The standard definition of the Markov Decision Process (MDP) is as follows.

- A Markov Decision Process is a tuple $< S, A, P, R, \gamma>$
- S is a finite set of states
- A is a finite set of actions
- P is a state transition probability matrix,

$$P^a_{ss'} = P\,[S_{t+1} = s' \mid St = s, A_t = a]$$

- R is a reward function, $R^a{}_s = E\,[R_{t+1} \mid S_t = s, A_t = a]$
- $\gamma$ is a discount factor $\gamma \in [0, 1]$.

### *5.3.1. System Model of the Online RL based approach*

The agent of the Online RL based approach is the doctor and the team. The observations will be the Normal cells, immune cells, tumor cells and the drug concentration available in the blood. Thus, the state space will be the cell concentration associated with these four observations of the normal cells, immune cells, tumor cells and the drug concentration in the blood. The action space will be constructed considering three types of patient types, the young patient, the pregnant woman with cancer and the elderly person with other types of critical diseases. The reward function will be designed as follows.

$T_t$ = Tumor size at time t

$V$ = Death of the patient

$$R(s,a) = \begin{cases} 15, & T_t \leq 0 \\ -50, & V = 1 \\ \frac{(T_t-(T_t+ T_{initial}))}{T_t}, & T_{t+1} = T_t \\ 20, & t = 0 \end{cases} \tag{31}$$

In the reward assignment, +5 points were assigned if the Total tumor cells were equal or lesser than 0. A reward of -50 points were assigned if the person is dead. The reward was calculated as a proportion of the decrease of tumor cells during the intermediate months of chemotherapy simulation. And, a reward of 20 points were assigned when the simulation process was finished after a period of 7 months which gives the meaning the patient has completed his chemotherapy treatments of the first line.

### *5.3.2. System Model of the Offline RL based approach*

The major consideration of this research is laid towards the offline RL based approach in order to implement a more realistic RL model with more real life applicative methodology. The basic RL models are described element wise as follows.

#### *5.3.2.1. Agent*

The agent is the oncologist and the team. Agent's goal is to enhance the patient's wellness, reduce the toxicity levels resulted by the sequential chemotherapy treatments, ensure speedy recovery of the patient and finally develop an optimal drug schedule.

#### *5.3.2.2. Observations*

The agent's observations are the clinical diagnosis factors measuring sequentially during the course of the Chemotherapy treatments. In the real clinical practice, Chemotherapy doses will be given with a time period of 2 weeks. Thus, the patient's clinical diagnosis factors are the observations in this problem setting. These observations include the tumor length, tumor width, Demographic factors of age, gender, weight and height, patient's side effects, Drug response measure and the hospital standard patient scoring measures of the Charleson's Comorbidity index (CCI) and the Eastern Cooperative Oncology Group Performance Scale (ECOG PS) measure which are recording sequentially.

#### *5.3.2.3. States*

The state will be the individual patient covariate which will describe the real state of the patient considering the current clinical observations the patient encompasses. The oncologist will determine the real patient state of the individual patient considering the patient's sequential clinical observations. The decided real patient state is taken as the State Space of the problem setting.

#### *5.3.2.4. Actions*

Actions are the methods of the Agent that help it to communicate and alter its environment and thus move between states. Each action taken by the Agent gives the environment a reward. The policy is responsible for determining which action to choose.

The actions are the chemotherapy treatments or the drugs as specified by the oncologist considering the recorded observations and the real patient status of the patient. The Relative Dose Intensity (RDI) for 6 type of frequently used drugs for Colon cancer (Cetuximab, Bevacizumab, Irinotecan, 5FU, Xeloda, Oxaliplatin) are taken as the Action Space of the problem setting.

#### *5.3.2.5. Reward Function*

Reward function will help in guiding the oncologist in assessing whether the action chosen at the Chemotherapy treatment course is desirable or not. The choice of the reward function should be delicate in the domains where multiple objectives should be balanced. It is useful to define the reward function in terms of a tradeoff between the costs associated with the treatments and the costs associated with the symptoms. Along with that reward function should be tackle the tradeoff between the toxicity and the tumor size. The precise choice of the reward function has a significant impact on learning rates in the RL problems and a poorly chosen reward function can lead to unexpected and undesirable results. Thus, the individual reward considering the sequential clinical measurements will be calculated at each cycle for every patient. The designed reward function is described as follows.

$R_T(s, a)$ = Individual reward for tumor size changes

$R_W(s, a)$ = Individual reward for weight changes

$R_E(s, a)$ = Individual reward for the overall side effects changes

$R_R$ (s, a) = Individual reward for drug response changes

$R_R$ (s, a) = Reward for dead

$T_t$ = Tumor size at time t

$W_t$ = Weight at time t

$E_t$ = Overall side effects index at time t

$R_t$ = Drug response at time t

$$R_T(s,a) = \begin{cases} -6, & T_{t+1} > T_t \\ 2, & T_{t+1} < T_t \\ 0, & T_{t+1} = T_t \\ 6, & T_{t+1} = 0 \end{cases} \tag{32}$$

$$R_W(s,a) = \begin{cases} 1, & W_{t+1} > W_t \\ -1, & W_{t+1} < W_t \\ 0, & W_{t+1} = W_t \end{cases} \tag{33}$$

$$R_E(s,a) = \begin{cases} -2, & E_{t+1} > E_t \\ 2, & E_{t+1} < E_t \\ 0, & E_{t+1} = E_t \end{cases} \tag{34}$$

$$R_R(s,a) = \begin{cases} 120, & R_{t+1} = \text{CR} \\ 60, & R_{t+1} = PR \\ 30, & R_{t+1} = SD \\ -60, & R_{t+1} = PD \end{cases} \tag{35}$$

$$R_{DEAD}(s,a) = -200 \tag{36}$$

Total reward functions will be the accumulation of the above individual reward functions.

$$R_{TOTAL}(s,a) = R_T(s,a) + R_W(s,a) + R_E(s,a) + R_R(s,a) + R_{DEAD}(s,a)$$

(37)

#### *5.3.2.6. Environment*

The environment is the series of states that the agent seeks to affect by its choice of actions. The environment will be modeled as a stochastic finite state machine with inputs (chemotherapy combination drug dosages decided by the oncologist) and outputs (clinical observations and the rewards). For the simulation process, the clinical database developed with real life patient's observations, actions, states and the computed rewards will be taken as the environment in the problem setting.

## 5.4. Clinical Database creation

In order to build the clinical trials database, stage 4 clinical trials of Colon cancer on the Oncology treatments database of the hospital were searched. The candidate clinical trials were also allowed to involve any other kinds of cancer treatment methods such as radiotherapy, surgery along with chemotherapy. Only the first line therapy treatments were taken in to consideration. In total, 40 patient's data who received Chemotherapy treatments from 2017 to 2020 were considered as the candidate subjects for the database creation.

Several variables including the descriptive information as reported by the oncologist were recorded individually for each of the patient. The data include patient demographic characteristics (age, gender, weight, height), Physical characteristics (side effects, CCI, ECOG PS measure), and Disease characteristics (Tumor primary site, Tumor length, tumor width, drug response) were collected after every chemotherapy treatment cycle they are undergoing. These data were already presented in the hospital database and the useful information cited by the oncologist were manually recorded in to the dataset. In addition, the final decision as prescribed by the oncologist for each patient after each cycle were also recorded which could be maintaining the therapy, going for second line treatments, going for surgery or radiation therapy. If the patient died in the course of the treatment, it was also recorded. Following descriptions will include the informative approach in revealing the observations and the decisive targets within the course of the chemotherapy cycles.

### *5.4.1. Deciding the Tumor size*

In the clinical practice, the tumor sizes are measuring based on image studies done with the Biopsy technique. Usually if colon cancer is found in a patient by any screening or diagnostic test, then it is biopsied during a colonoscopy. In a biopsy, the doctor removes a small piece of tissue with a special instrument passed through the scope. Here normally a part of the colon will need to be surgically removed to make the diagnosis [54]. These biopsy samples (from colonoscopy or surgery) are sent to the lab to look at them very closely. If cancer is found, other lab tests may also be done on the biopsy samples to help better classify the cancer and possibly find the specific treatment options.

In the clinical practice, image study using CT scans are performing for the purpose of getting a measure for the Tumor size. Usually this is done within 1-2 months of the chemotherapy cycles. The reason for not doing the image study too often is because frequent biopsy will be too painful to the patient and it will be too costly. Thus, clinically a measure for the tumor size can only be get after each 3-4 months during the course of the chemotherapy.

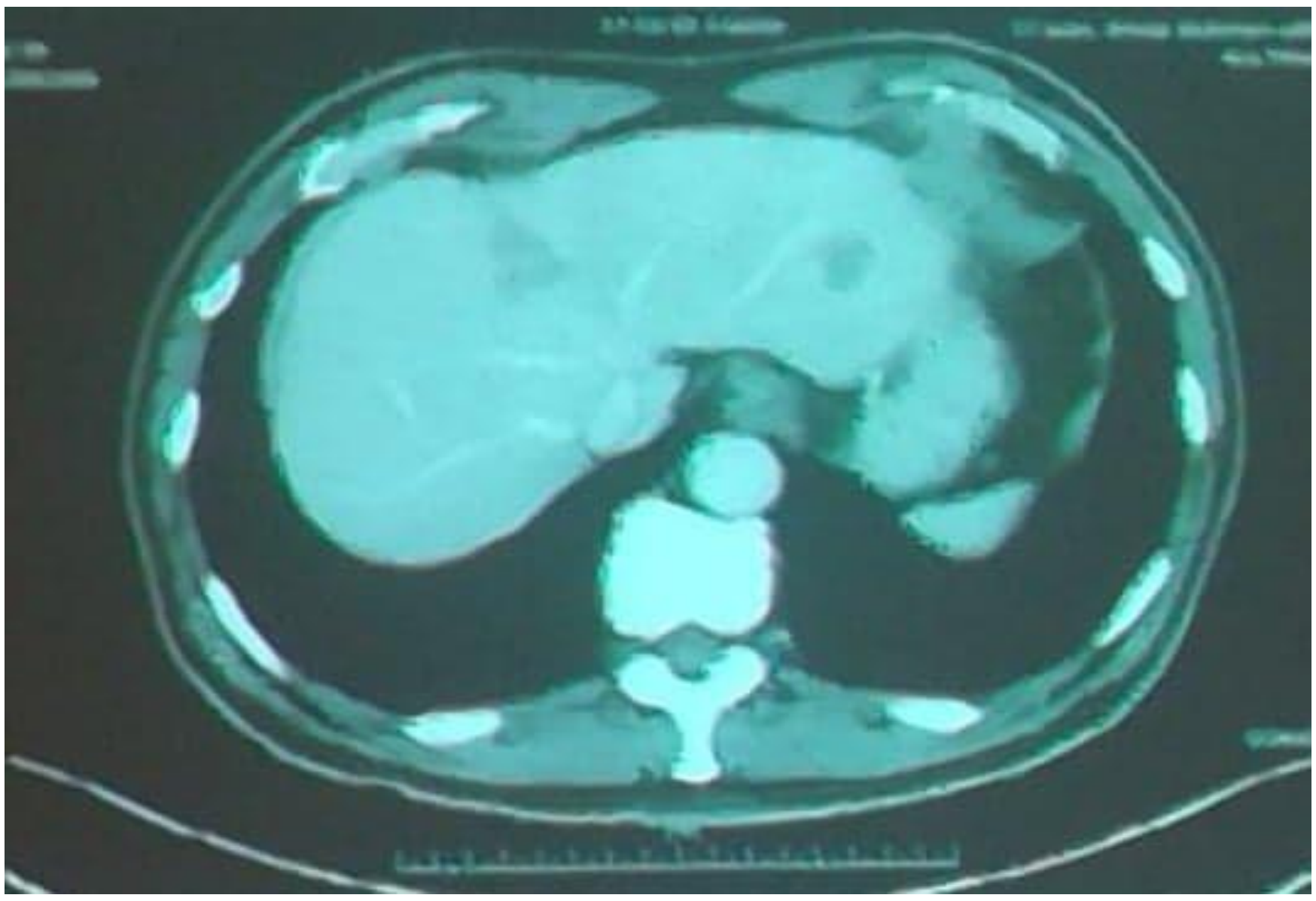

Figure 18: A colon cancer tumor CT scan image

In the clinical observations it is noticed that three kinds of tumor measurements are reveled via ultrasonography measurements as the maximum dimension, cranio-caudal dimension and the ventrodorsal dimension(depth) in which the cranio-caudal dimension and the ventrodorsal dimension are the most commonly approached distance metrics used.

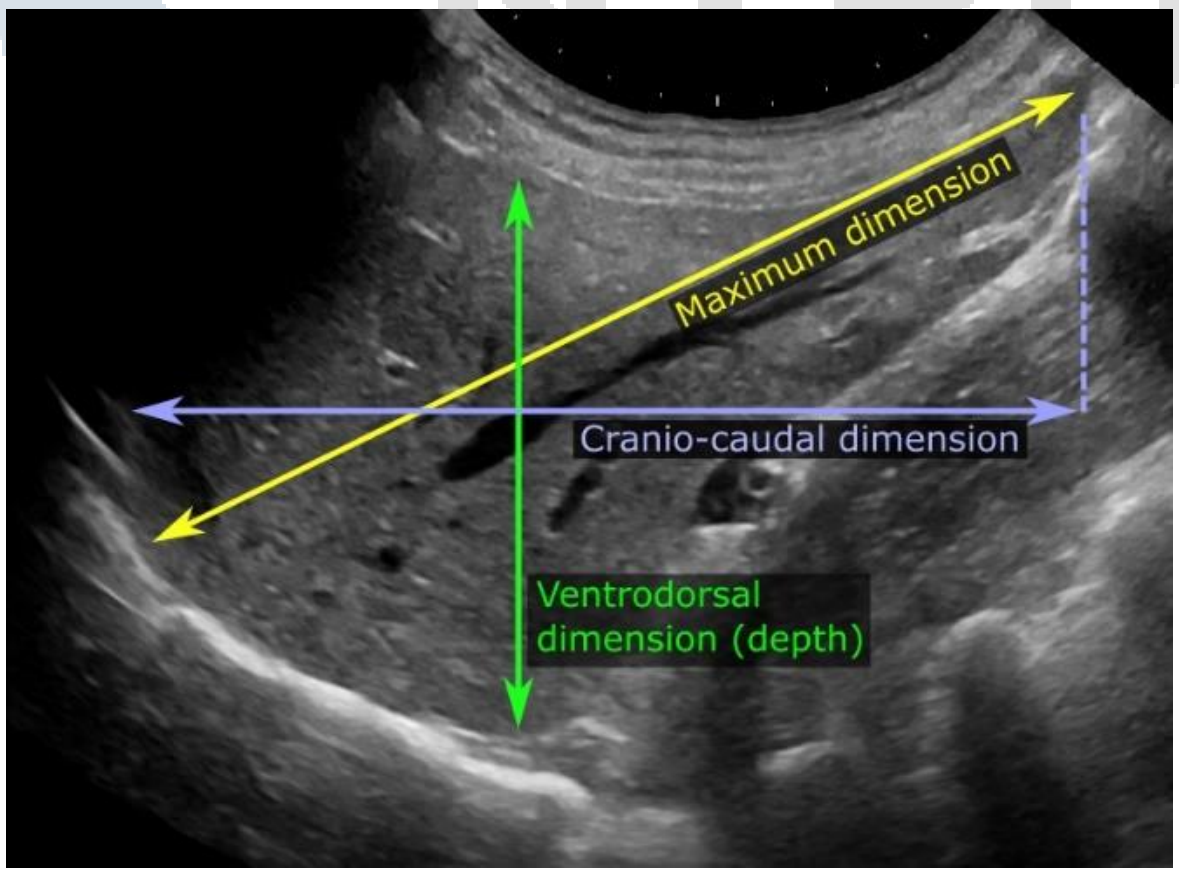

Figure 19: Tumor dimensions during colonoscopy

Thus, the tumor volume was obtained from the following standard equation.

L = Cranio-caudal dimension (Tumor Length)

D = ventrodorsal dimension (Tumor depth)

$$V = \frac{4}{3}\pi * \frac{L}{2} * \frac{L}{2} * \frac{D}{2} \quad (38)$$

$$V = \frac{4}{3}\pi L^2 D \quad (39)$$

A measure for the tumor volume can only be taken for the chemotherapy cycles around 6 cycles intervals. So, the problem of missing tumor volumes for the intermediate cycles was addressed through the linear interpolation technique assuming that the tumor size will be exponentially changed over time.

### *5.4.2. Deciding the Side effects index*

In literature, it is evident that the prevalence of a summary toxicity index leads to several important benefits in patient care. It is possible to promote the use of less toxic pharmaceutical agents. It is important to classify patients who are at high risk for toxicity in advance and take effective countermeasures. Changes over time in total toxicity may be detected. It is possible to model the results of various therapeutic interventions more accurately. It is possible to calculate the combined toxicity of a drug within the same patient. (Fries et al., 1990).

TABLE 1: SIDE EFFECTS VALUES AS DESIGNED BY THE HOSPITAL

| | Grade 0 | Grade 1 | Grade 2 | Grade 3 | Grade 4 |
|---|---|---|---|---|---|
| Fatigue | 0 | 1 | 1 | 2 | 2 |
| Skin changes | 0 | 1 | 1 | 2 | 2 |
| Hair loss | 0 | 1 | 1 | 1 | 1 |
| Vomiting | 0 | 2 | 4 | 6 | 8 |
| Diarrhea | 0 | 2 | 4 | 6 | 8 |
| Neutropenia | 0 | 2 | 4 | 8 | 16 |
| Anemia | 0 | 2 | 4 | 6 | 8 |
| Thrombocytopenia | 0 | 2 | 4 | 6 | 8 |
| Febrile Neutropenia | 0 | 3 | 6 | 12 | 24 |

As there are multiple possible side effects due to the treatment at each cycle, an overall side effects index will be computed for individual patient. That measure will represent how broad his overall side effects influence is and will tell facts on his survival rates at a nutshell.

An example of getting the overall side effects index can be shown as follows.

Suppose in a particular chemotherapy treatment cycle, the patients expressed the side effects as follows.

Fatigue – Grade 0

Skin changes – Grade 1

Hair loss – Grade 2

Vomiting - Grade 0

Diarrhea- Grade 3

Neutropenia – Grade 0

Anemia – Grade 0

Thrombocytopenia – Grade 0

Febrile Neutropenia - Grade 0

Then the overall side effects index will be calculated as:

Overall side effects index = (Score for Fatigue=grade 0) + (Score for Skin changes =grade 1) + (Score for Hair loss =grade 2) + (Score for Vomiting =grade 0) + (Score for Diarrhea =grade 3) + (Score for Neutropenia =grade 0) + (Score for Anemia =grade 0) + (Score for Thrombocytopenia =grade 0) + (Score for Febrile Neutropenia =grade 0)

### *5.4.3. Deciding the Drug Response*

The drug response is the result the patient shows after using a particular type of drug during the chemotherapy. In the clinical practice, this drug response is based on some conventional chemotherapy decisions hold by the oncologists. Accordingly, there are 4 type of drug response means.

**Complete response (CR)** - All of the cancer or tumor disappears, No evidence of disease. A tumor marker (if applicable) may fall within the normal range. At a situation with CR, oncologist will stop the chemotherapy treatments and will prescribe the patient to attend to the outpatient clinics to test his cancerous state.

**Partial response (PR)** - The cancer has shrunk by a percentage but disease remains. A tumor marker (if applicable) may have fallen but evidence of disease remains. At a situation with PR, oncologist will advise on continuing the chemotherapy treatments.

**Stable disease (SD)** - The cancer has neither grown nor shrunk, The amount of disease has not changed. A tumor marker (if applicable) has not changed significantly. At a situation with SD, oncologist will advise on continuing the chemotherapy treatments.

**Disease progression (PD)** - The cancer has grown, there is more disease now than before treatment. A tumor marker test (if applicable) shows that a tumor marker has risen. When the oncologist detect the PD, he will prescribe to change the line therapy or change the combination of drugs using by altering one or more drug types.

The acceptable method of deciding this response is based on the tumor size measures calculated after each 3-4 months during the chemotherapy course. Here following convention is used.

$$F(Response) = \begin{cases} 0.7 - 1, & if\ complete\ response\ (CR) \\ 0 - 0.7, & if\ partial\ reponse\ (PR) \\ 0.7 - 1.2, & if\ stable\ disease\ (SD) \\ > 1.2, & if\ disease\ progression\ (DP) \end{cases} \quad (40)$$

If the tumor reduction is 100%-70%, then it will be complete response. If the tumor size variation is 70%-120% then it will be stable disease. If it is between 0%-70%, then it will be termed as the partial response and if the tumor size is greater than 120%, then it is disease progression.

### *5.4.4. Deciding the CCI Index*

The Charlson Comorbidity Index is a method of categorizing comorbidities of patients based on the International Classification of Diseases (ICD). Each comorbidity category has an associated weight (from 1 to 6), based on the adjusted risk of mortality or resource use, and the sum of all the weights results in a single comorbidity score for a patient. A score of zero indicates that no comorbidities were found. The higher the score, the more likely the predicted outcome will result in mortality or higher resource use [55].

In our approach, the comorbidity index of each of the patient will be calculated separately by looking at the clinical diagnosis reports stored for each of the patient in the hospital clinical database. The score was calculated using online CCI calculator [56]. Following are the criteria presented in the website which we consider to calculate the CCI index for the individual patients.

- Age
- Myocardial infarction
- Cognitive Heart Failure
- Peripheral vascular disease
- Dementia
- Connective tissue disease
- Peptic ulcer disease
- Liver disease
- Diabetes mellitus

- Hemiplegia
- Chronic Kidney Disease
- Leukemia
- Lymphoma
- AIDS
- Is this a COVID-19 patient?

#### *5.4.5. Deciding the Action Space*

The set of actions should be specified such that it will reveal the potential interventions with the application of drug dosages. Reinforcement learning agents typically have either a discrete or a continuous action space (Sutton and Barto 1998). With a discrete action space, the agent decides which distinct action to perform from a finite action set. With a continuous action space, actions are expressed as a single real-valued vector. In our study, the action space is finalized to be in the discrete nature depending on the drug scheme prescribed by the oncologist in real clinical practice. Thus, the action space for the RL model for Colon cancer chemotherapy treatment is the 576 drug combinations from 40 patients receiving the 6 types of drugs as mentioning below.

The drugs used for the colon cancer patients in the intended cluster of patients are the Cetuzimab, Bevacizumab, Irinotecan, 5FU, Xeloda and Oxaliplatin. When deciding the drug amount the oncologists consider the Relative Dose Index (RDI), which is calculated as follows.

$$\text{Relative Dose Intensity (RDI)} = \frac{\textit{Relative Dose Intensity (RDI)}}{\text{Standard Dose Intensity (SDI)}} * 100 \qquad (41)$$

Where,

$$\text{Delivered Dose Intensity (DDI)} = \frac{\textit{Delivered total dose (mgm}-2)}{\textit{Actual time to complete chemotherapy with imputation for missed cycles (days)}} \quad (42)$$

$$\text{Standard Dose Intensity (SDI)} = \frac{\textit{Standard total dose (mgm}-2)}{\textit{Standard time to complete chemotherapy with imputation for missed cycles (days)}} \quad (43)$$

The standard amount of dosages for the 6 drugs are as follows.

$$\text{Cetuzimab dose} = \text{RDI} * 500\ mgm^{-2} * \text{Body Surface Area} \quad (45)$$

$$\text{Bevacizumab dose} = \text{RDI} * 5\ mgkg^{-1} * \text{Body Weight} \quad (46)$$

$$\text{Irinotecan dose} = \text{RDI} * 180\ mgm^{-2} * \text{Body Surface Area} \quad (47)$$

$$\text{5FU dose} = \text{RDI} * 400\ mgm^{-2} * \text{Body Surface Area} \quad (48)$$

$$\text{Xeloda dose} = \text{RDI} * 1250\ mgm^{-2} * \text{Body Surface Area} \quad (49)$$

$$\text{Oxaliplatin dose} = \text{RDI} * 85\ mgm^{-2} * \text{Body Surface Area} \quad (50)$$

### *5.4.6. Deciding the State Space*

It is a common feature in medical outcome measures that the measures are naturally expressed mostly with a mix of discrete categorical and continuous variables, which may vary over a wide or even unknown range. State spaces in medical problems are typically composed with high dimensional, heterogeneous dimensions, which gives rise to practical problems in the state space representation. Some of these practical problems can be summarized as:

1. The high dimensional nature will emerge the need for a single state that sufficiently

represents the statistics for the problem domain.

2. The heterogeneous nature of the variables will cause the domain to consist of some irrelevant state variables.

3. The "curse of dimensionality" problem, which occurs because of the high dimensional state space.

4. The need to approximate the continuous state variables.

In a statistical context, a state representation is adequate if it provides enough details to fully define the necessary parameters for the related distributions. In the domain of RL, a state should be appropriate to determine the allocation of potential incentives and state transitions. In our problem, we express the RL policy as a mapping from the state to the action. Therefore, if the state space definition lacks the sufficient information, the predicting policy may lack the crucial information that the overall study intends to achieve. Thus, a careful judgement over the determination of the state space should be taken prior to the RL algorithm initialization.

We rarely know the state variables are appropriate for our MDP in medical areas, and so we need to rely on a certain degree of intuition to direct the choice of state variables. We would also always allow the conservative decision to add variables of the state that are in fact meaningless. In an RL problem, these unrelated state variables can become a source of error or decreased data efficiency. Some techniques are considered to be relatively stable in the presence of uninformative state variables, so these techniques are of considerable use when in question.

In our problem domain, it was decided to map the patient's clinical observation set over a particular cycle in to one single real patient state. The process was done under the supervision and careful judgement of the oncologist with his domain knowledge acquired through the years of clinical practice over the years of treatment. This process required the insight into specific limitations and the behavior of the relevant observations.

Thus, a patient real state at a particular time over the chemotherapy cycles will be,

$$S(t) = T_t + \ Age_t + Gender + \ W_t + \ E_t + \ R_t + CCI + P_t \tag{51}$$

$T_t$ = Tumor size at time t

$Age_t$ = Patient's age at time t

$Gender$ = Patient's gender

$W_t$ = Patient's weight at time t

$E_t$ = Side effects index at time t

$R_t$ = Drug response at time t

$CCI$ = CCI index of the patient

$P_t$ = ECOG PS measure at time t

Thus, the state space for the RL model for Colon cancer chemotherapy treatment is the 576 real patient status decided for the 40 patients decided with clinical guidelines and the domain expertise knowledge of the oncologist.

The procedure in this function approximation methodology will be explained theoretically under the Model Based State Estimation (MBSE) module.

## 5.5. System Architecture

### *5.5.1. System Architecture of the Online RL approach*

The proposed primary approach is based on a range of available ordinary differential equation (ODE) based tumor behavior models, majorly the cell count model (CC model) and the tumor size model (TS model). These ODE models are capable to generate data real time with inserted initial state conditions. Our approach will be to determine the optimal chemotherapy treatment schedule with respect to the patient's conditions by modeling first the problem as a finite MDP model to follow up the RL techniques.

1. Modeling chemotherapy as MDP
2. Confirming cancer growth models and their ODE equations
3. Validating offline the methodological applicability of RL by $Q$ learning
4. Building an online model-based RL for chemotherapy treatment
5. Determining the optimal drug regimen

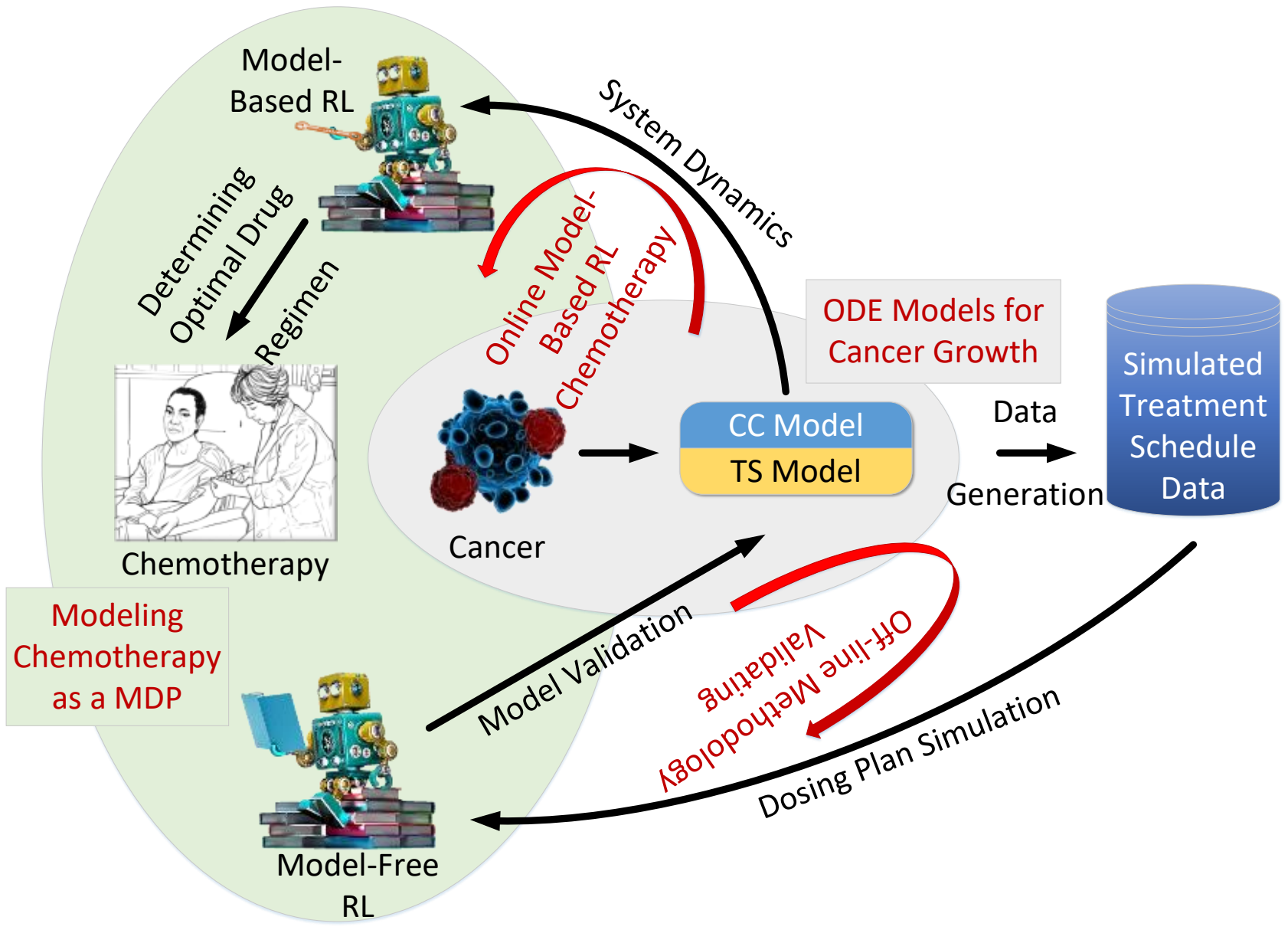

Figure 20: System Architecture of Online RL approach

### 5.5.2. *System Architecture of the Offline RL approach*

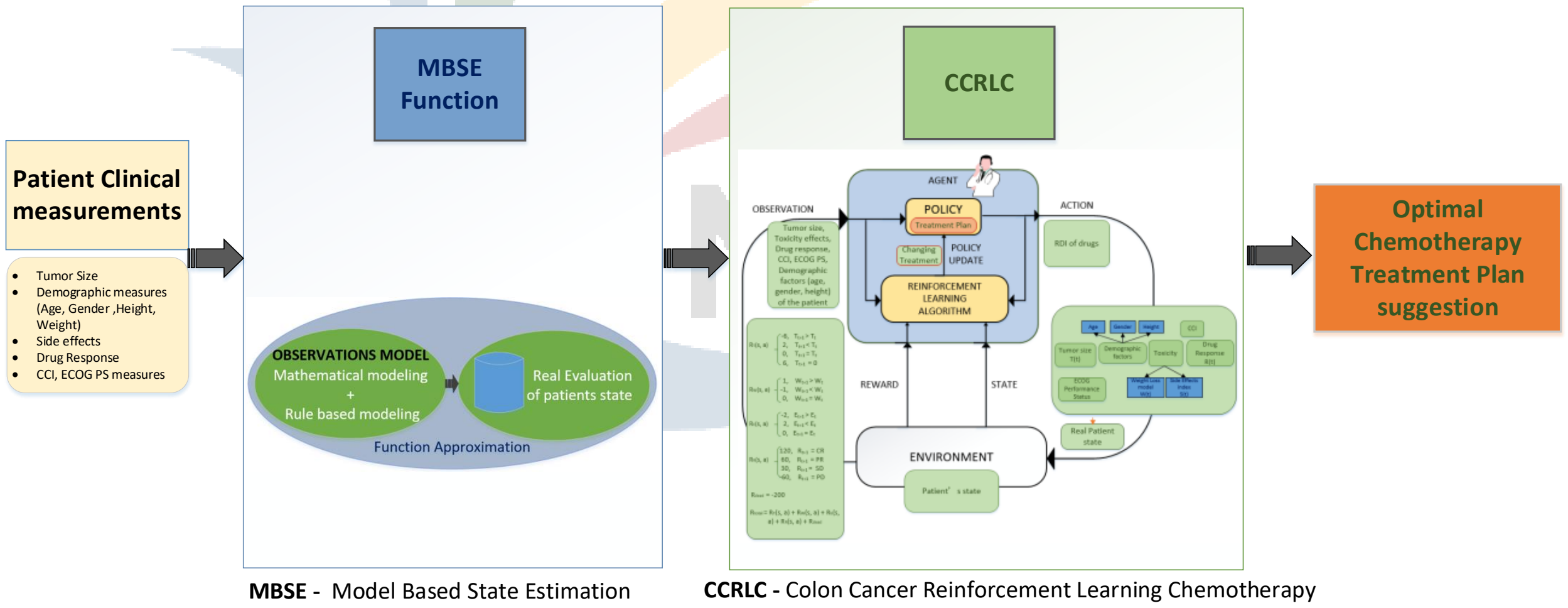

Figure 21: System Architecture of Offline RL approach

The proposed main solution includes 2 major components interconnected as shown in the Figure 21. The major components consist MBSE module and the CCRLC module.

## 5.6. Modular Architecture

The proposed solution is viewed as a modular design as proposed in below figure. It consists of major three tiers as Presentation Layer, Application Layer and Database Layer. The researcher identified major five modules that operate within these layers as, Data acquisition module, Data preprocessing module, Model Based State Estimation (MBSE) module, Colon Cancer Reinforcement Learning Chemotherapy (CCRLC) module and Result logging module. The five major modules include five sub modules as categorized in the Figure 5.2. The design of the modules is being discussed in the next sub sections.

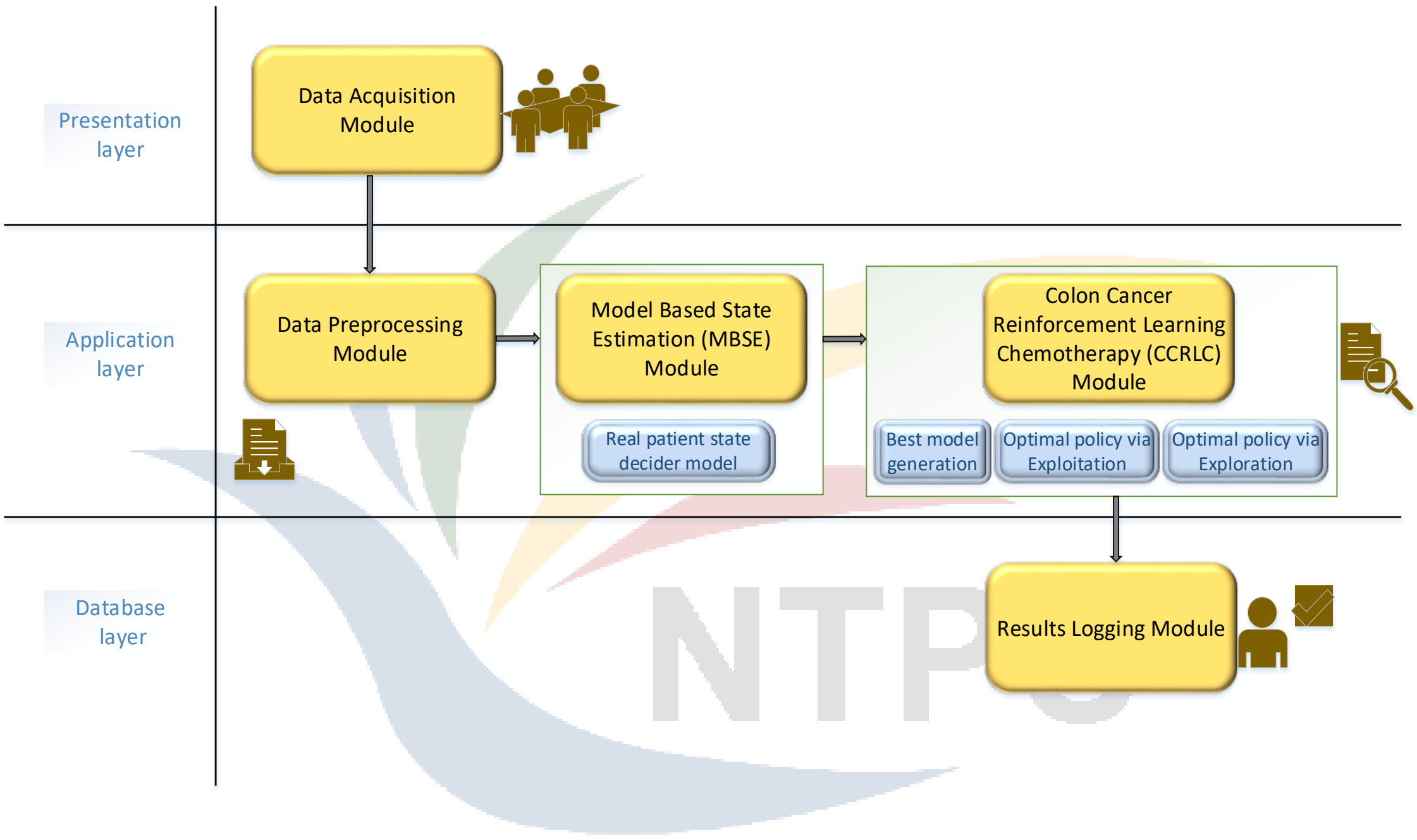

Figure 22: Modular architecture

### 5.6.1. *Data acquisition module*

In the Data acquisition module, the patient clinical measurements of tumor size, patient's demographic measures of age, gender, height, weight, overall side effects index, drug response measure and the CCI and ECOG PS measure would be determined. The clinical diagnosis observations would be inputted to the Model Based State Estimation (MBSE) module in order to predict the real patient state, which will be varying between 0-10.

### 5.6.2. *Data preprocessing module*

The first step in any machine learning related project is to preprocess the data such that it will be suitable for the data processing step with machine learning algorithms. In this module, missing values associated with the columns were identified and managed in an appropriate manner. We used the data preprocessing techniques of feature creation, feature subset selection, instance selection, handling missing values and attribute transformation. and obtained the final clean dataset which can be approached in the model based state estimation module.

### 5.6.3. *Model Based State Estimation (MBSE) module*

The first step in implementing a problem associated with medical domain, with the RL framework is to decide how to summarize the history of a patient $H$ into some more manageable covariate that we describe as the state representation $s$. This challenge is understood in machine learning to select a representation because it requires determining how to represent the history of a patient in a manner that preserves adequate knowledge for the task. It is important that this representation provide for any factors that may confuse the policy's estimation of results (Robins, 2000). Thus, we must incorporate any variables that have a causal effect on both the care decisions and the outcome observed (Gottesman et al., 2018).

In our problem in deciding the optimal chemotherapy treatment scheme, the confounding variables will represent the information that will influence a decision and will have an effect on the outcome of that decision. These variables include the patient's historical medical records, tumor length, tumor width, their demographic factors such as age, gender, weight and height, patient's side effects, drug response measure and the hospital standard patient scoring measures of the CCI and the ECOG PS measure. Thus, a careful mapping of all the state transitions in to a single value which determines the real patient state was initiated with the support of the domain knowledge the oncologist has via his clinical treatment experiences over years.

Here two models was constructed on the basis of the CNN classifier and the RF classifier.

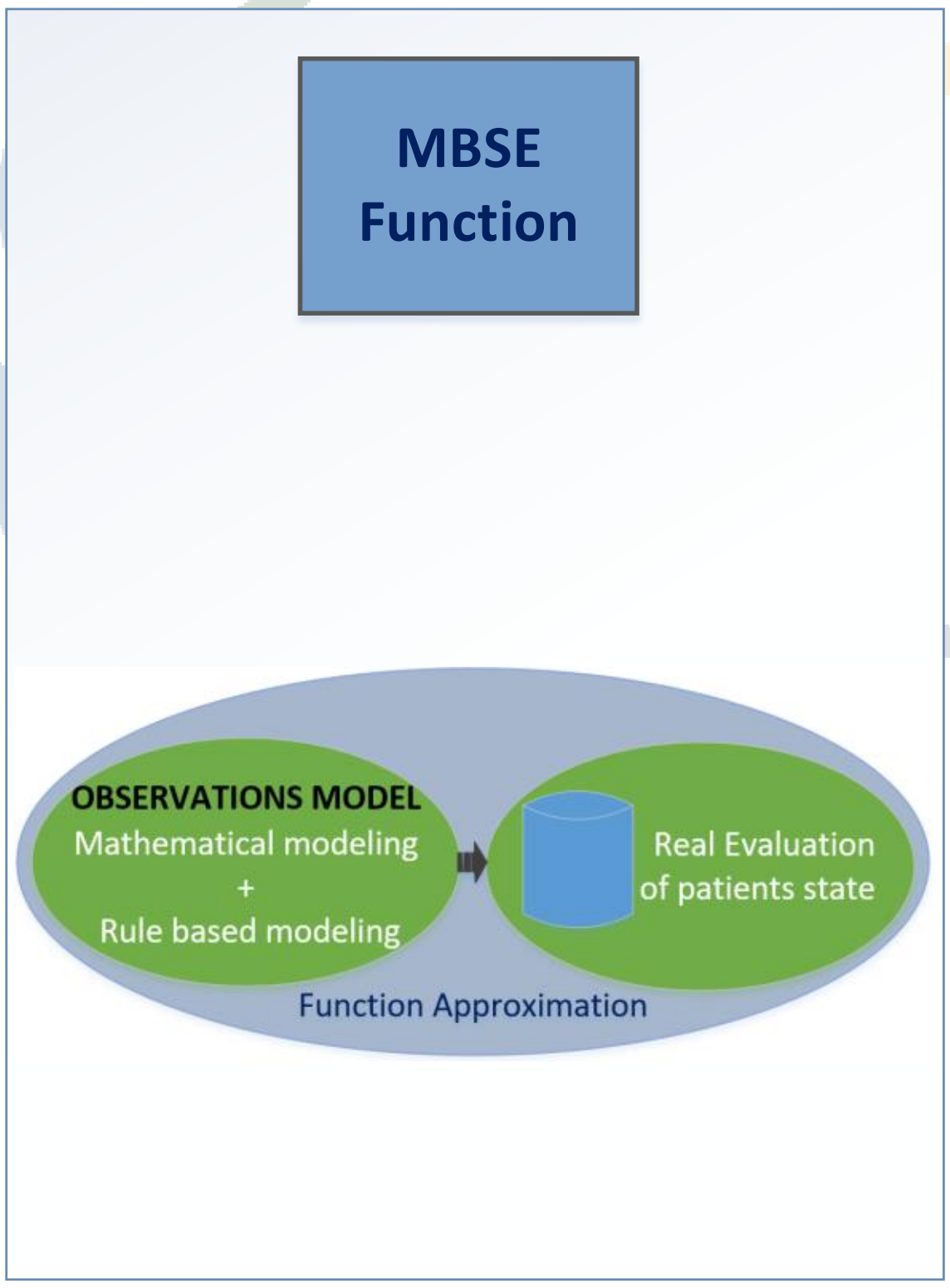

Figure 23: Model Based State Estimation (MBSE) module

### 5.6.4. *Colon Cancer Reinforcement Learning Chemotherapy (CCRLC) module*

This section is the main second sub module of the overall project. This encompass with the procedure of deciding the optimal treatment schedule or the amount of drugs doses to be given to the new patient by considering the observations set identified at the initial coarse of the diagnosis. The agent will decide the combination of the drugs to be prescribe to the patient in cycle 1 based on the initial diagnosis. Then the reward he get will be computed based on the suggested reward function. That reward will be the value of the action or the value of the combination therapy given according to the identified observation set. The combination therapy to be prescribed in the cycle 2 will be decided based on the toxicity effects, ECOG PS measure, Tumor size variations resulted after the cycle 1 treatment results. Again, the reward of taking that combination therapy would be calculated. Accordingly, the process will proceed by agent again deciding the combination therapy based on the observation set resulted from the prior cycle. Thus this is a sequential decision making process where the agent is trying to maximize the overall reward at the end of the chemotherapy cycles of the patient by carefully selecting the combination therapy drugs and the suitable drug dosages.

Here we approached the offline RL algorithm of Conservative Q-Learning (CQL) algorithm and the Model Based Offline Policy Optimization (MOPO) algorithm in order to reveal the optimal treatment schedule.

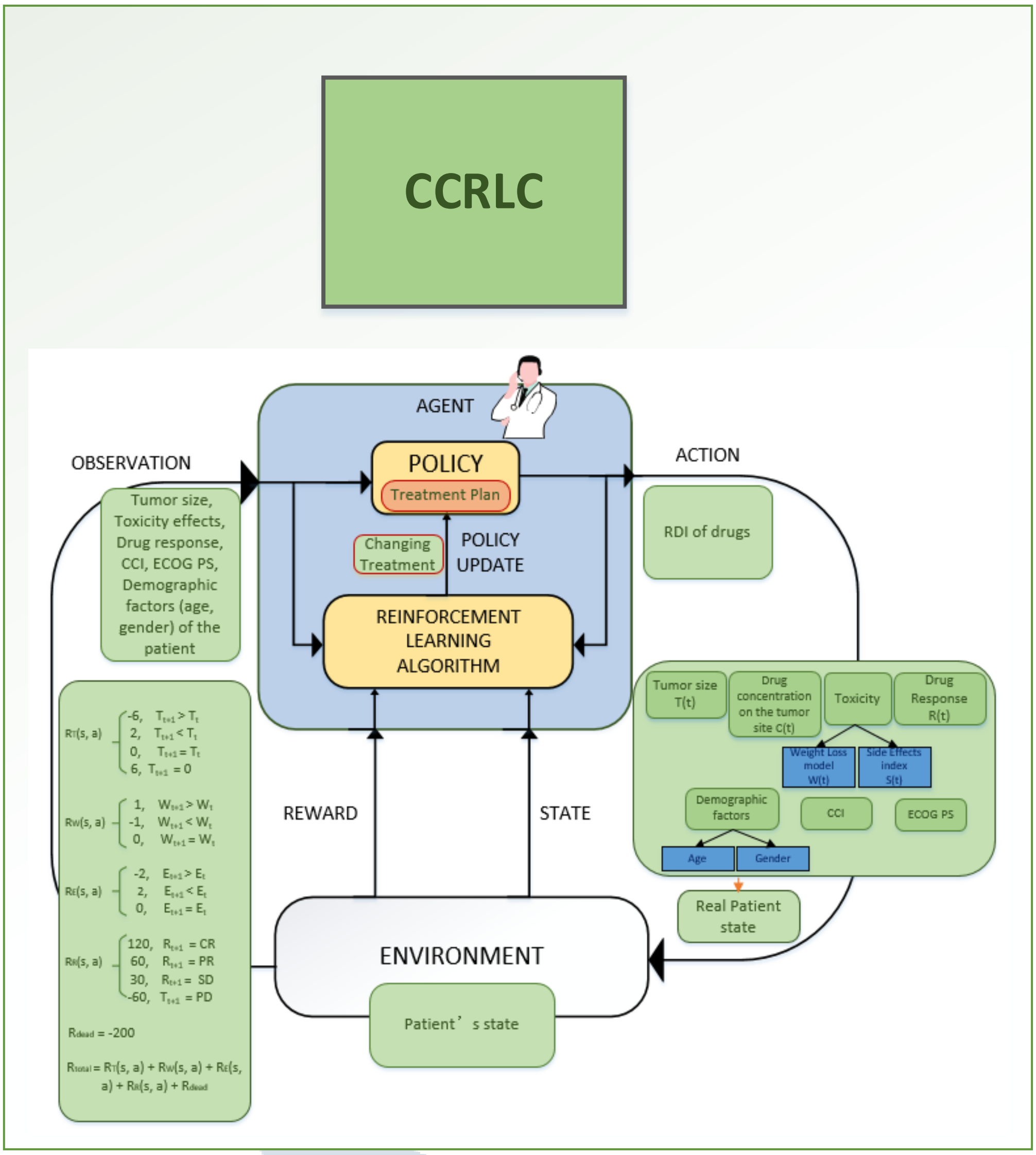

Figure 24: Colon Cancer Reinforcement Learning Chemotherapy (CCRLC) module

### 5.6.5. *Results logging module*

In this module, the optimized optimal treatment schedule would be retrieved and saved in a suitable manner that enables reloading when necessary.

## 5.7. Proposed Dyna-Q architecture with CQL and MOPO algorithms

Model based reinforcement learning has an agent who is trying to understand the world and create a model to represent it. It refers to learning the optimal behavior indirectly by learning a model of the environment by taking actions and observing the outcomes that will include the next state and the immediate reward. The main advantage of the model based architecture is to plan ahead by thinking ahead.

The basic steps of the model based planning method in reinforcement learning can be depicted as follows.

1. Run base policy $\pi_0(a_t|s_t)$ (e.g., random policy) to collect $D = \{(s, a, s')_i\}$

2. Learn dynamics model $f(s, a)$ to minimize $\Sigma_i \ \ || \ f(s_i, \ a_i) - s_i' \ ||^2$

3. Plan through $f(s, a)$ to choose actions

4. Execute those actions and add the resulting data $\{(s, a, s')_j\}$ to $D$

Repeat the step 2 and the process goes on like that.

Thus when a new clinical observation enters or a new patient with unseen patient diagnosis data enters, the model should be able to decide the real patient state, the next transitions, the next predicting rewards and thereby should decide the optimal policy from the transitions and the rewards.

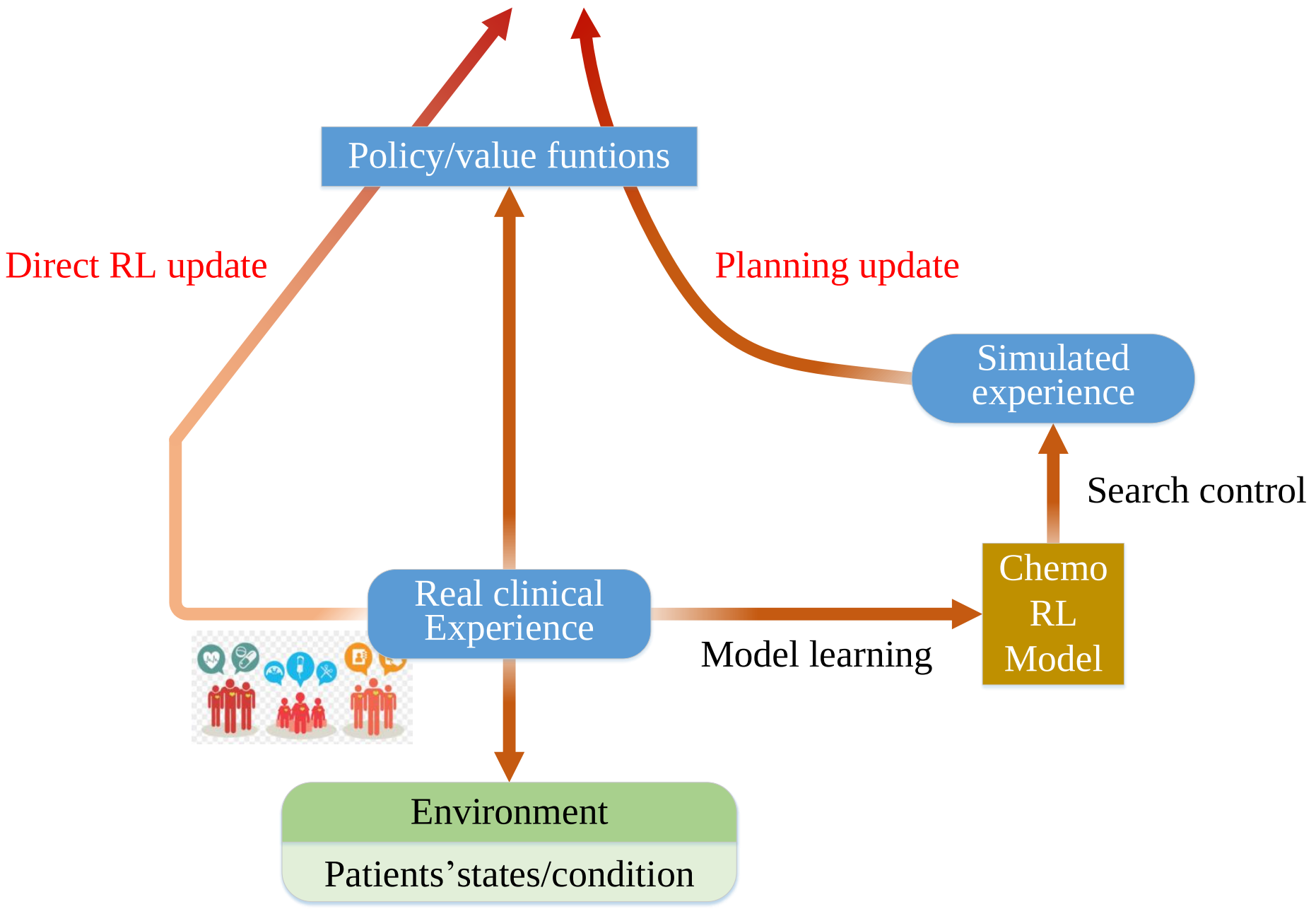

Figure 25: The Chemo RL Dyna –Q architecture

As depicted in the above figure, the Direct RL update will be achieved with using the Conservative Q-Learning algorithm, while the Planning update will be achieved with using the Model Based Offline Policy Optimization (MOPO) algorithm.

## 5.8. Summary

This chapter presented the design process of the proposed system. Here, firstly the overall system design, system model, system architecture, software architecture, modular architecture and simulation design were presented. Then all the identified modules with their interconnections and functions have been elaborated along with diagrams. In the next chapter, we discuss about the System Implementation.

**CHAPTER 6**

# 6. IMPLEMENTATION

## 6.1. Introduction

Chapter 05 presented the Design of the proposed system. Chapter 06 presents the Implementation of the system. Firstly, this chapter presents details on how development of the simulation was done and how each of the module were developed. Then, it is followed by the tools and technologies adopted in the implementation.

## 6.2. Implementation of Online RL based approach

The implementations related to the online RL approach is directly inspired from the studies done by (Ernst et al., 2006) and the simulations done by (Beck et al., 2019) and (Padmanabhan et al., 2017b). I used python programming language and the Pycharm Professional IDE for our simulation process. The traditional Q-Learning process were adopted in the simulation. The basic model was simulated with the initial state values as Natural cells concentration equaling to 0.6, Tumor cells equaling to 0.9 and Immune cells as 0.4. The process was ran for 11,000 episodes where the exploration episodes are 10,000 and exploitation episodes are 1,000.

### *6.2.1. Deciding the action space based on the patient types*

Under the section of the simulation results numerical examples to validate our proposed online RL approach for the chemotherapy drug dosing is presented. In the real clinical scenario, an oncologist will consider numerous aspects when deciding the treatment schedule for the cancer patients. These factors include the age, gender, patients' other critical illnesses and special cases like whether the patient is pregnant. Some of the special cases where an oncologist considers in balancing the effects from chemotherapy to the normal cells, tumor cells, immune cells are described in the study done by Padmanabhan et al. (Padmanabhan et al., 2017b). Normally, natural cells and immune cells are age dependent. Young patient's natural cells and the immune cells growth rates are higher than that of the elderly people. Thus, when deciding the treatment strategy for the young patients, oncologist tend to reduce the number of cancerous cells with less attention to the normal cells and immune cells. The reason is young patients body will regenerate the normal cells and immune cells more quickly than that of the elderly patients. In addition, another case is if the cancer is in a vital part of the body such as the brain, then oncologist will pay more attention on the normal cells and take measures to prevent the destroying of the normal cells and reducing the cancerous cells. In addition, in the case of pregnancy, the oncologist will pay more attention to give lower doses before the delivery of the baby to ensure the safety of the fetus and vary the dose amount later appropriately.

Paying attention to the above clinical decisions, we categorized our action space for three types of patients as young patients, pregnant women with cancer and elderly people with other critical illnesses. The simulative model was ran for several times considering the decided action spaces definition. Finally, the validation process was decided to continue under 3 main methods as model validity for 3 types of patient environments, model validity through parameter variation of the ODE equations and model validity by Artificial generation of clinical data.

### *6.2.2. Model validity for 3 types of patient environment*

The three types of cases we considered in our simulation were young patients with cancer, pregnant woman with cancer and elderly person with other critical illnesses. The simulation results related to these 3 kinds of patients are shown in the evaluation section.

### *6.2.3. Model validity through parameter variation of the ODE equations*

For the purpose of ensuring the robustness of the proposed RL controller, we used the trained optimal RL model for the simulated patients by varying the parameters in the ODE equations. In the first case, we consider the original parameter values suggested in the primary study done by De Pills. (De Pillis & Radunskaya, 2003). In the second and third cases, we use simulated patients with -12% and +15% parameter variations with respect to the nominal model parameter values. The new values and the nominal model values are shown in the following table.

TABLE 2. CC MODEL ODE PARAMETER VARIATION

| **Parameter** | **Nominal model** | **-12% parameter variation** | **18% parameter variation** |
|---|---|---|---|
| a1 | 0.32 | 0.2816 | 0.3776 |
| a2 | 0.38 | 0.3344 | 0.4484 |
| a3 | 0.1 | 0.088 | 0.118 |
| b1 | 0.2 | 0.176 | 0.236 |
| b2 | 0.2 | 0.176 | 0.236 |
| c1 | 0.5 | 0.44 | 0.59 |
| c2 | 0.4 | 0.352 | 0.472 |
| c3 | 0.585 | 0.5148 | 0.6903 |
| c4 | 0.6 | 0.528 | 0.708 |
| d1 | 0.25 | 0.22 | 0.295 |
| d2 | 1 | 0.88 | 1.18 |
| r1 | 1.4 | 1.232 | 1.652 |
| r2 | 1 | 0.88 | 1.18 |
| s | 0.37 | 0.3256 | 0.4366 |
| alpha | 0.35 | 0.308 | 0.413 |
| rho | 0.01 | 0.0088 | 0.0118 |

### 6.2.4. *Artificial generation of clinical data*

In order to evaluate the ability of the implemented Chemotherapy RL model to compute good chemotherapy schedules, we applied the Q-learning algorithm on artificially generated clinical data. For this process, we approached 15 simulated patients for each type of patients. Thus, totally we generated 45 simulated patients such that there will be 15 patients for each of the three different types of patients. The patients were generated by altering the initial state variables of natural cells, tumor cells and immune cells. In all the instances, the drug concentration would be zero when the treatment is starting.

The generation procedure of the clinical data is an iterative process. At the first iteration, we considered 5 young patients and ran the model once for each patient and found the optimal treatment schedule for each patient. The action space would be the original action space we decided for the young patients. The mean optimal policy will be computed for the first 5 set of patients. Next, at the second step of the iterative process, we considered a new set of 5 young patients. The action space will be constructed bearing the effect of both the original action space and the optimal policy generated via the first 5 set of young patients. Here we considered a weighted scale for the two type of actions as 80% weightage for the optimal policy found at the 1st iteration and 20% weightage for the original action space. Then, the optimal policy will be computed for the second 5 set of patients. Finally, for the third step of the iterative process, we considered a new set of 5 patients. For the action space, we considered the weighted scale similar to the second step. That means new action space would be the 80% effect from the optimal policy resulted from the second set of 5 patients and 20% effect from the original action space. Finally, combined trajectories would be generated considering the effect of all the 15 patients.

The main reason behind this iterative process of clinical data generation is to add the real expertise knowledge application of the oncologist in to the simulated model. The drug dosages in the clinical practice is not always administered randomly, but it would at least partially bears the expertise knowledge from the clinicians which may already confirmed as good chemotherapy strategies. In addition, by using the optimal policy results that have already acquired, better decisions can be made. As result, with a small number of clinical trials, we can more quickly converge up to the most effective optimal chemotherapy schedule by approaching this iterative process.

### 6.3. Implementation of Offline RL based approach modules

The main hypothesis of the research is that Offline Reinforcement Learning can be adapted to solve the problem in planning chemotherapy treatment schedules for colon cancer patients. Implementations would be discussed under the five modules as the data acquisition module, data preprocessing module, Model Based State Estimation module (MBSE), Colon Cancer Reinforcement Learning Chemotherapy module (CCRLC) and the results logging module.

### *6.3.1. Data acquisition module*

In the data acquisition module, the metadata with respect to below parameters are obtained from the user.

- Tumor size
- Age
- Gender
- Height
- Weight
- Overall side effects index
- Drug response measure
- CCI score
- ECOG PS measure

These data should be related to the clinical diagnosis of stage 4 metastasis, colon cancer patients who are receiving chemotherapy treatments under the line 1 treatment policy.

### *6.3.2. Data preprocessing module*

The dataset used in the offline RL approach was constructed from the raw data collected from the Cheng Hsing General Hospital, Shipai, Taiwan. All together 40 colon cancer patient's data who are in stage 4 metastasis stage and receiving treatments under the line 1 treatment policy are accumulated together in the dataset. The patients are ranging from the age of 39-84 years. There are 17 male patients and 23 female patients. Dataset contains 23 columns and 575 rows containing the transitions data related to the 40 patients. The columns include the ID, Patient ID, Line number, 'Cycles', 'Tumor length',,'Tumor

Width', 'Tumor size', 'Age', 'Gender', 'Weight', 'Height','Side effects index', 'Drug response','Drug concentration in the plasma', 'CCI','ECOG PS measure','Real Patient state', 'Cetuximab dosage', 'Bevacizumab dosage', 'Irinotecan dosage', '5FU dosage','Xeloda dosage', 'Oxaliplatin dosage', 'Decision after the therapy'.

The first step after loading the data in to the implementation module was to clean and preprocess the data such that they are suitable for the targeted data processing steps. We used the data preprocessing techniques of feature creation, feature subset selection, instance selection, handling missing values and attribute transformation. Feature creation is the approach of creating new attributes that can capture the important information in a data set much more efficiently than the original attributes. Here two new columns were appended to the data file as the 'Reward' and the 'Cumulative reward'. These values were computed separately for every patient along their transitional behavior shown in the dataset. Feature subset selection is the selection of the most appropriate features, which will be useful in the machine learning process. This was achieved along with two methods as removing redundant features and irrelevant features and removing one of two features that have a correlation higher than 0.9 with a correlation matrix. The first method was done by removing the other columns keeping only the columns of 'Tumor size', 'Gender', 'Weight', 'Height', 'Side effects index', 'CCI Index', 'ECOG PS measure and 'Real Patient State' in the Model Based State Estimation module. The second method was done by drawing the correlation matrix. There we removed the age column, which has a greater correlation with the CCI index column. When calculating the CCI index, age also becomes one factor. So, keeping only the CCI index column is more relevant.

TABLE 3. CORRELATION BETWEEN THE PATIENT'S CLINICAL OBSERVATIONS AND THE REAL PATIENT STATE

| | Real patient state | Tumor size | Age | Gender | Weight | Height | Side effects index | CCI Index | ECOG PS |
|---|---|---|---|---|---|---|---|---|---|
| Real patient state | 1 | | | | | | | | |
| Tumor size | -0.006 | 1 | | | | | | | |
| Age | -0.176 | 0.149 | 1 | | | | | | |
| Gender | -0.069 | -0.166 | -0.38 | 1 | | | | | |
| Weight | -0.216 | -0.108 | 0.035 | -0.349 | 1 | | | | |
| Height | 0.009 | 0.091 | 0.052 | -0.578 | 0.462 | 1 | | | |
| Side effect index | -0.581 | -0.050 | 0.060 | -0.074 | 0.103 | -0.008 | 1 | | |
| CCI Index | -0.394 | 0.101 | 0.795 | -0.272 | 0.251 | 0.201 | 0.160 | 1 | |
| ECOG PS | -0.453 | -0.125 | 0.165 | 0.058 | 0.145 | 0.137 | 0.223 | 0.462 | 1 |

The data preprocessing method Instance selection is the selection of the most appropriate rows or the instances, which will be useful in the machine learning process. Here we used the z-score to remove the outliers that can be presented in the dataset. Then we handled the missing values in the tumor size column by approaching the linear interpolation technique with the polynomial method of order 2. Then, we used the data preprocessing method of Attribute transformation which involves the mapping the entire set of values of given attribute to a new set of replacement values such that each old value will be identified with one of the new values. That was done by using the standardization (Feature scaling) and normalization was used in the Model Based State Estimation module prior to the classification process.

After the primary data-preprocessing step, the clean data suitable for the application of the machine learning algorithms were saved in the form of .excel file.

### *6.3.3. Model Based State Estimation (MBSE) module*

In this module, a supervised learning network to predict the variation of the oncologist's decisions on the chemotherapy treatment was implemented. The input features will be the patient's clinical diagnosis measures which will include the tumor size, age, gender, weight, height, side effects index, CCI score and ECOG PS measure and the target feature will be the real patient's state.

TABLE 4: IMBALANCED DATA STRUCTURE IN THE DATASET

| ***Real patient state*** | **instances** |
|---|---|
| 0 | 0 |
| 1 | 0 |
| 2 | 0 |
| 3 | 2 |
| 4 | 20 |
| 5 | 42 |
| 6 | 99 |
| 7 | 105 |
| 8 | 139 |
| 9 | 121 |
| 10 | 29 |

Figure 26: Pie chart for the Distribution of Real Patient State Instances

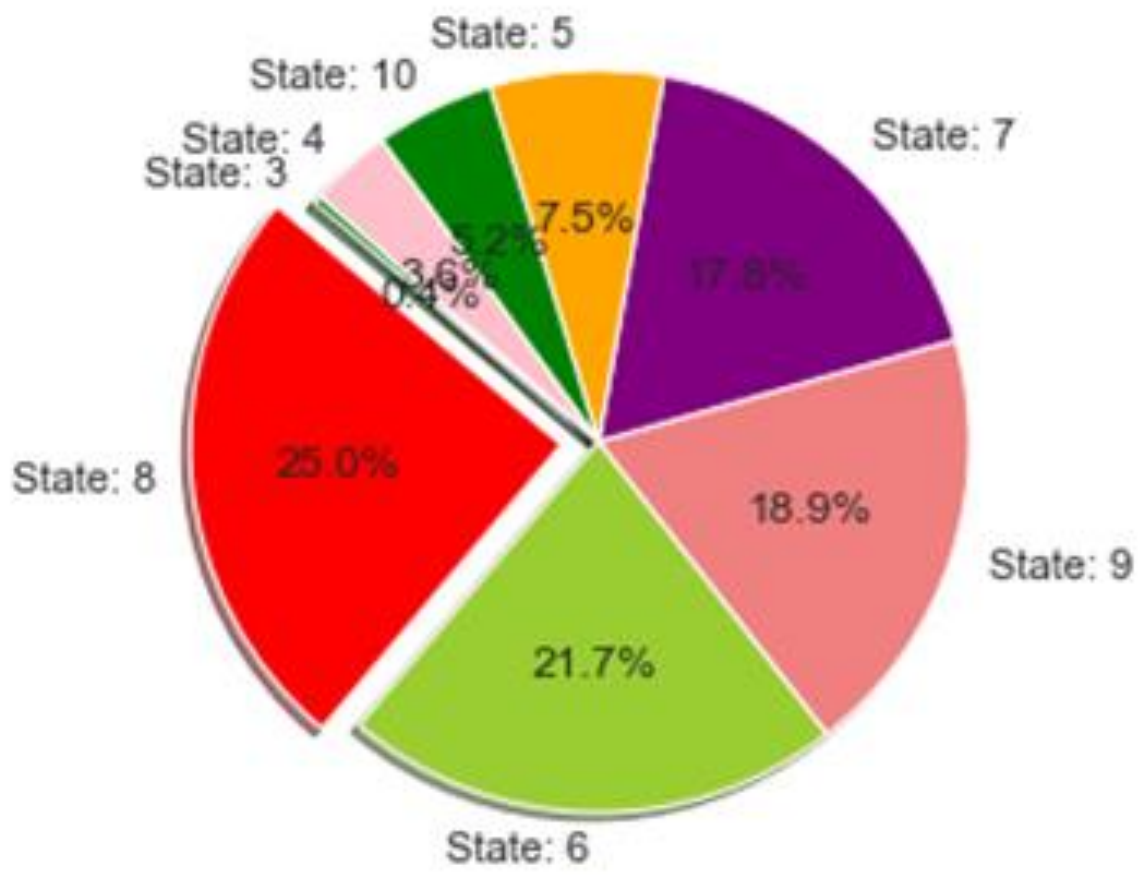

Originally, the data distribution of our dataset over the classes is imbalanced where the instances for separate real patient status are dispersed like above. Two basic classification models by using the Artificial Neural Network and the Random Forest Classifier. Both of these modeled were implemented in Jupyter notebook on anaconda navigator platform with python version 3.6.

#### *6.3.3.1. Real patient state decider using CNN classifier*

There is a considerable difference of the instances presenting in each of the class and even there is not at least 1 instance for the classes 0-2. Using unbalanced dataset with major deviations in the instances will make bias the network for the majority classes. Thus, prior to feeding in to the CNN classifier we upsampled the data such that each class will have the instances similar to the majority class.

Here we adopted a simple CNN architecture as shown in the figure 2. The network contains three-layer convolution process, one max pooling layer and two fully connected layers. The network is fully described below in the algorithmic format.

Algorithm 1. CNN architecture pseudocode

**Algorithm 1**. CNN architecture pseudocode

**CNN** *(trainX, trainY, testX, testY, learningrate = 0.001,epoch = 100, batchsize = 1)* :

*inputs = shape(datapoints, channels)*

*model* ← *Conv1D(filters = 32, kernels = 2, activation = ReLU)(input)*

*model* ← *Conv1D(filters = 64, kernels = 2, activation = ReLU)(model)*

*model* ← *Dropout(0.10)(model)*

*model* ← *MaxPooling(model,pool_size=1)*

*model* ← *Conv1D(filters = 64, kernels = 2, activation = ReLU)(model)*

*model* ← *Dropout(0.10)(model)*

*model* ← *Flatten()(model)*

*model* ← *Dense(neurons = 256, activation = ReLU)(model)*

*model* ← *Dense(neurons = 11, activation = softmax)(model)*

*model.compile(loss = sparse_categorical_crossentropy, optimizer = adam, learningrate)*

*model.fit(trainX, trainY,epoch, batchsize)*

*Accuracy = model.evaluate(testX, testY)*

**return** *Accuracy*

First the dataset containing the 7 features of 'Tumor size', 'Gender', 'Weight', 'Height', 'Side effects index', 'CCI Index', 'ECOG PS' are inputted to the network.

The CNN model constructed by stacking three convolution layers followed by two fully connected layers, i.e., two dense layers one with the Relu activation function and the other with the softmax activation function. The three convolutional layers are constructed with one single convolutional layer, followed by one max pooling layers. A window size of 2 and stride size of 1 were applied for all the convolutional and max pooling layers. The pool size of the max pooling layer is 1. A dropout of 10% was applied for convolutional layers except for the first convolutional layer. Two dense layers was positioned at the end of the network. One dense layer is followed by a relu activation layer and the final dense layer is followed by a softmax layer

The epochs and training batch size of the three models were set identically as 100 and 1 respectively. For all three models, Rectified Linear Unit (ReLU) was chosen as the activation function for all convolutional layers. The sparse categorical cross entropy was chosen as the loss function and the Adaptive Moment Estimation (ADAM) optimizer was used in optimization for all models. The metrics used to evaluate the training process is the accuracy.

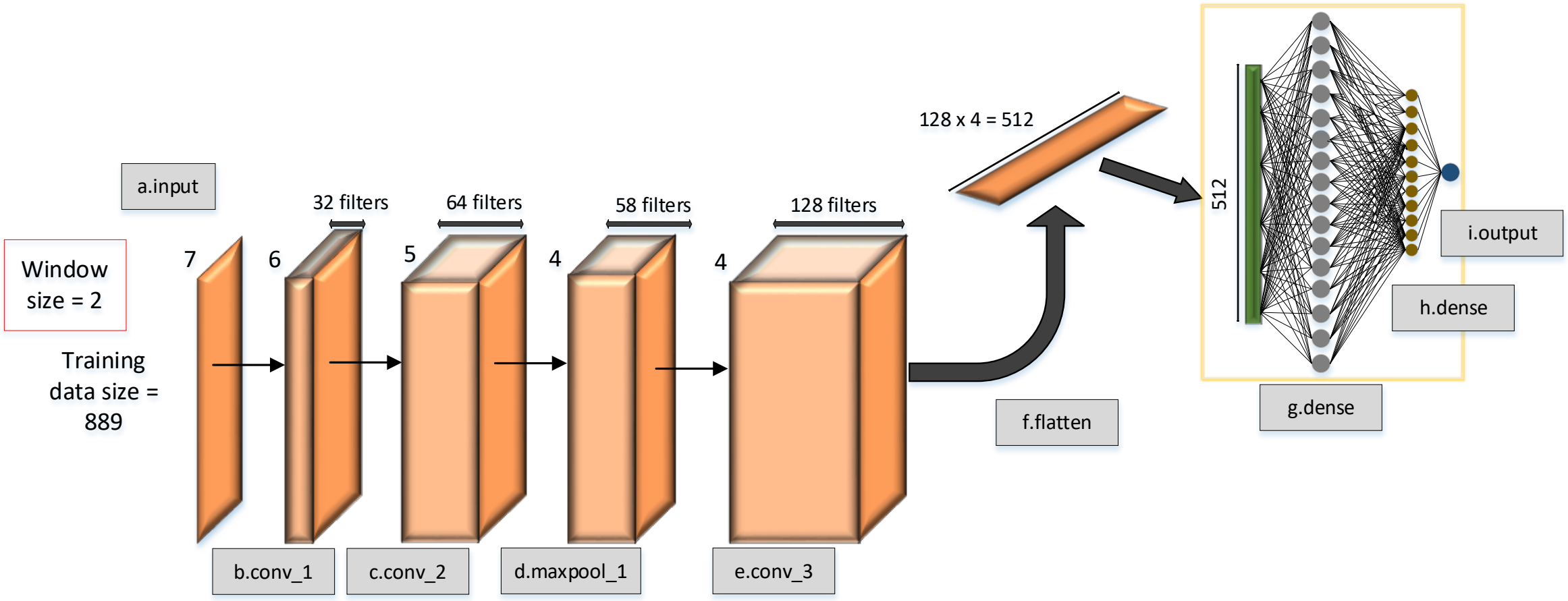

Figure 27: ID CNN architecture for the real patient state classification

Table 5 described the summary of the designed CNN model.

TABLE 5: SUMMARY OF THE 1D CNN MODEL

| **Layer No.** | **Layer type** | **Filters size** | **Kernel size** | **Region size** | **Strides** | **Padding** | **Output size** |
|---|---|---|---|---|---|---|---|
| 1 | 1D – Conv | 32 | 1 x 2 | - | 1 | - | 32 x 1 x 6 |
| 2 | 1D – Conv | 64 | 1 x 2 | - | 1 | - | 64 x 1 x 5 |
| 3 | 1D - MaxPool | 58 | - | - | 1 | - | 58 x 1 x 5 |
| 4 | 1D – Conv | 128 | 1 x 2 | - | 1 | - | 128 x 1 x 4 |
| 5 | Flatten | - | - | - | - | - | 1 x 512 |
| 6 | Dense | - | - | 256 | - | - | 1 x 256 |
| 7 | Dense | - | - | 11 | - | - | 1 x 11 |

### *6.3.3.2. Real patient state decider using Random Forest Classifier*

First the dataset containing the 7 features of 'Tumor size', 'Gender', 'Weight', 'Height', 'Side effects index', 'CCI Index', 'ECOG PS' are inputted to the random forest classifier.

As discussed in the technology section, random forest is an extension upon bagging and like bagging, random forest focus on selecting bootstrap samples from the training dataset and then fits a decision tree on each. However, when the dataset is imbalance or when there is a significant probability that a bootstrap sample contains a very few or none of the minority class, it will result in a poor performance for predicting the minority class. Thus, we approached to use the Weighted Random Forest Classifier by adding the class_weight= 'balanced' argument when defining the RF model which takes a dictionary with a mapping of each class value to the weighting. The application of this argument is used to use the inverse weighting from the training dataset, giving focus to the minority class. By following this approach, algorithms can be modified in a way such that learning will perform to bias towards the classes that have fewer examples in the dataset. It is termed as the cost sensitive learning.

Before the training process, we focused on using the hyper parameter tuning which relies more on the experimental results than the theory and this selects the best method to determine the optimal settings to try many different combinations to evaluate the performance of each model. This process took a considerable computational process. In order to perform the tuning, we created a parameter grid to sample from during the training as follows.

TABLE 6: PARAM GRID USED FOR GRID SEARCH WITH THE RF CLASSIFIER

| ***Parameter*** | **Parameter meaning** | **Values used in the grid** |
|---|---|---|
| n_estimators | The number of trees in the forest | 200, 500, 1000, 1250, 1500 |
| max_features | The number of features to consider when looking for the best split | 'auto', 'sqrt', 'log2' |
| max_depth | The maximum depth of the tree | 7, 8, 9, 10, 11, 12 |
| criterion | The function to measure the quality of a split | 'gini', 'entropy' |

On each iteration, the algorithm will choose different combination of the features. All together, there will be 5*3*6*2 = 180 settings and the model will run for all the possible combinations in order to yield the parameter settings with the best model performance or the best hyper parameters.

#### *6.3.3.3. Performance evaluation of Real patient state decider*

For the CNN model and the RF model, following performance evaluation metrics were calculated separately.

$$Accuracy = \frac{TP+TN}{TP+TN+FP+FN} \quad (52)$$

$$Precision = \frac{TP}{TP+FP} \quad (53)$$

$$Recall = \frac{TP}{TP+FN} \quad (54)$$

$$F1\ score = \frac{2\ \times Precision\ \times Recall}{Precision+Recall} \quad (55)$$

$$k = \frac{po-pe}{1-pe} \quad (56)$$

$$Jaccard\ score = \frac{Number\ of\ observations\ in\ both\ sets}{Number\ in\ either\ set} \quad (57)$$

Where *TP*, *FP*, *TN*, *FN* represent the true positive, false positive, true negative and false negative, respectively. *Po* and *pe* represent the actual agreement and the chance agreement, respectively.

### *6.3.4. Colon Cancer Reinforcement Learning Chemotherapy (CCRLC) module*

As described in the Design section, the clinical data will be collected for each consecutive transition of the patient's chemotherapy treatment period. After determination of the real patient state via the MBSE module, CCRLC module will focus on predicting the optimal sequence of drugs by following the offline reinforcement learning approach.

Figure 3 describes about the basic implementation steps followed. First data collected will be stored as the Replay buffer. These already collected data of the chemotherapy treatment cycles will follow the off policy framework during the training process. In the meantime, new data relating to the chemotherapy treatment cycles for the new patients will also be appended to the Replay buffer. That particular data will follow the on policy framework during the training process. During training, it will compute the State action values for each state action pairs. These state action pairs will be the labelled data that will be appended to the training buffer. During this process, the model will always try to minimize the bellman error associated with the training threads.

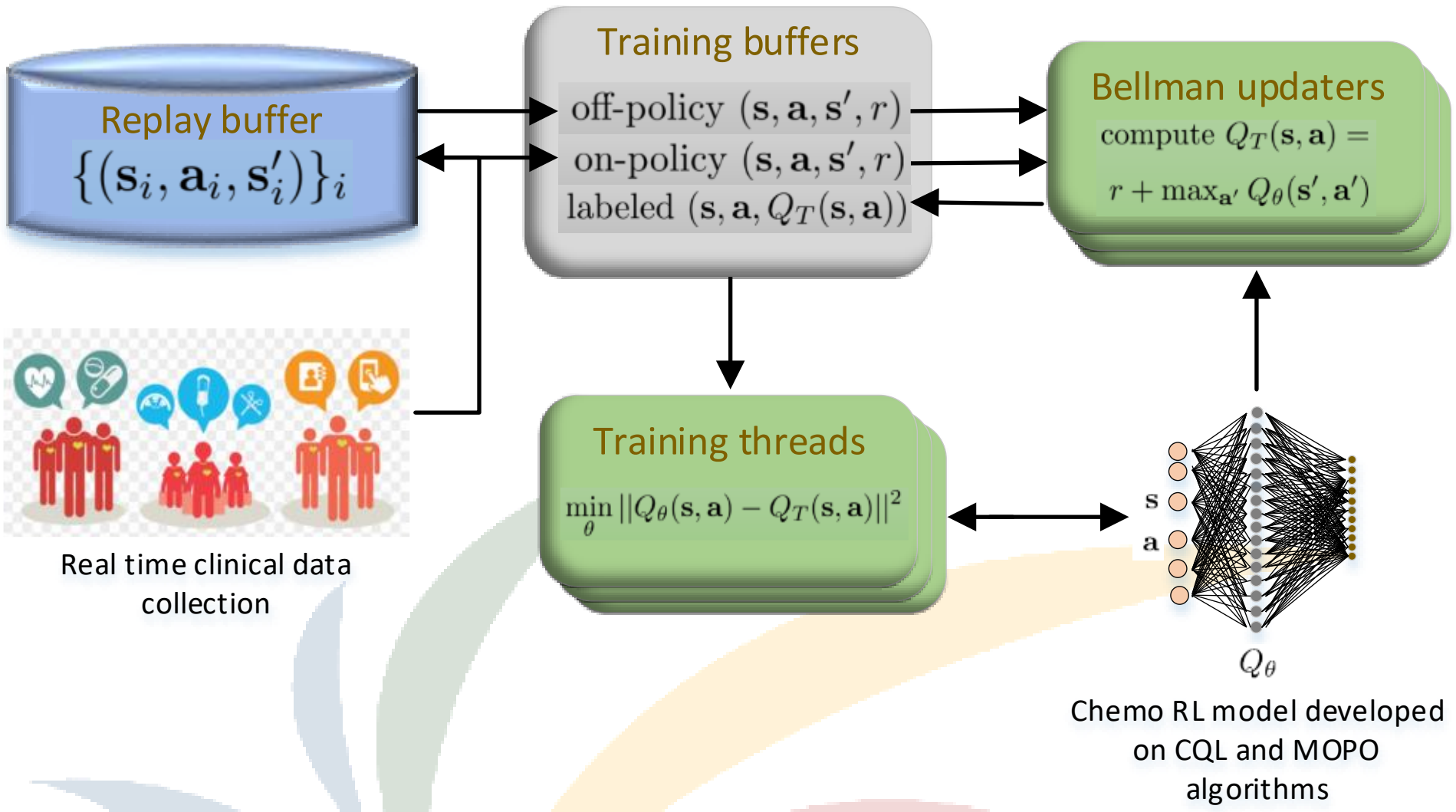

Figure 28: Implementation steps of Offline RL approach for Chemotherapy (Levine, 2019)

The implementations were carried out in Jupyter notebook on Anaconda navigator platform. The D3rlpy; which is a data driven deep reinforcement-learning library was used here. D3rlpy provides the state of the art offline reinforcement learning algorithms with Scikit learn compatible APIs. We used the offline reinforcement learning algorithms of Conservative Q-Learning (CQL) and the Model Based Offline Policy Optimization (MOPO). In the data-collecting module, we collected 40 patient’s data. Among them in order to have only the patients with similar treatment schedules, we removed the patients with less than 6 cycles of chemotherapy treatment period and more than 15 cycles of chemotherapy treatment period. Thus, 34 patients were remained and among them, we partitioned 7 patients for the testing purpose and 27 patients for the training purpose.

First, the dataset containing the selected 27 patients' clinical observations, decided real patient states, rewards, and actions or the dose schedules were loaded in to the code base. Then, these data will be converted in to the form of a MDP dataset and later in to transition objects. The data will be scaled by using the Standard Scaler. Then the model based model by using the CQL and the MOPO algorithms will be developed.

#### *6.3.4.1. Best model generation*

Our goal of approaching towards the offline reinforcement learning is to enable the application of RL methods in real world scenarios where logged data can be used. Here we focus on selecting the hyper parameters such that it would select the best out of a collection of specified policies that are tuned with different hyper parameters. We learn a set of $N$ N policies using the CQL and MOPO integrated model (Chemo RL model) with different hyper parameters and attempted to pick the best policy $\pi^*$.

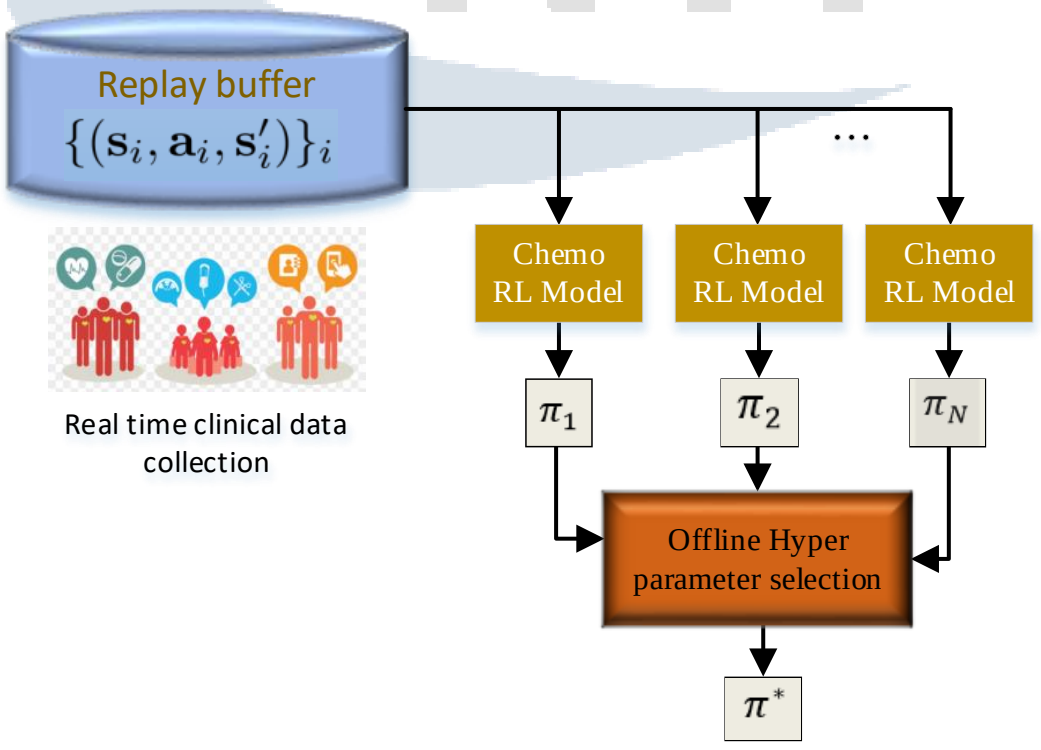

Figure 29: Offline hyper parameter selection (Paine et al., 2020)

In order to perform the tuning, we created a parameter grid to sample from during the training as follows.

TABLE 7: PARAM GRID USED FOR GRID SEARCH WITH THE CQL AND MOPO ALGORITHMS

| ***Parameter*** | **Parameter meaning** | **Values used in the grid** |
|---|---|---|
| actor_learning_rate | Learning rate for policy function | 1e – 05, 2e – 05, 3e – 05, 1e – 04, 2e -04, 3e – 04 |
| critic_learning_rate | Learning rate for Q functions | 1e – 05, 2e – 05, 3e – 05, 1e – 04, 2e -04, 3e – 04 |
| temp_learning_rate | Learning rate for temperature parameter of Soft Actor Critic (SAC) | 1e – 05, 2e – 05, 3e – 05, 1e – 04, 2e -04, 3e – 04 |
| alpha_learning_rate | Learning rate for α | 1e – 05, 2e – 05, 3e – 05, 1e – 04, 2e -04, 3e – 04 |
| gamma | Discount factor | 0.97, 0.98, 0.99 |
| Initial_alpha | Initial α value | 10.0, 5.0, 4.0 |
| tau | Target network synchronization coefficiency | 0.03, 0.05, 0.08 |

This process took a considerable computational process. On each iteration, the algorithm will choose different combination of the features. All together, there will be 6*6*6*6*3*3*3= 34,992 settings and the model will run for all the possible combinations in order to yield the parameter settings with the best model performance or the best hyper parameters. The running process took 1.5 hours – 2 hours in the Intel(R) Core(TM) i7-7700 CPU machine with the Nvidia geforce rtx 2070 GPU.

During the training process, we used the 'td_error' as the scorer or the strategy to evaluate the performance of the cross-validated model on the test set. The reason behind that is the model will run in such a way that it will focusing on minimizing the 'td_error' with the epochs, which is the ultimate goal to be achieved during offline Q Learning process.

The trained model was then used to predict the optimal drug schedule for the 7 testing patient set which was separated prior to the training process and for new patients with unseen clinical observations. For each patient instance, the drug policy for 6 cycles were generated under the decision of the oncologist. The action space is the RDI for the 6 types of drugs Cetuximab, Bevacizumab, Irinotecan, 5FU, Xeloda and Oxaliplatin. Here we made a series of combination of the six types of drugs adhering to some principles that oncologists at Taiwan follow. These drug rules have been imposed by the National Health Insurance (NHI). Following are the three main drug rules as imposed by NHI regulations.

1. Bevacizumab and Cetuximab can be used as substitutes. In the first cycle of the chemotherapy treatment period, many patients are given Bevacizumab, which needs to be approved by the NHI which takes some time. For the patients who did not be able to get the approval, Cetuximab is given in the first course until he gets the official approval.

2. 5FU and Xeloda can be used as substitutes.

3. Irinotecan and Oxaliplatin can be used as substitutes.

Here we applied a little adjustment to the suggesting optimal policy such that the policy will get the effect from both the behavior policy (The policy expressing by the dataset) and the target policy (The policy we are finding for). Thus, the optimal policy would be the combination of a 40% of the effect from the behavior policy and a 60% of the effect from the target policy. This would be resulted in making an adjustment from offline RL on the basement drug amount. This adjustment is meaningful in real life clinical sense because the dose schedule suggesting from the model has also considered the oncologist's previous supervised clinical decision.

#### *6.3.4.2. Optimal policy via Exploitation*

The term exploitation refers to RL algorithm trying to find the optimal policy within the dataset without further adding the new data or observations. The agent will focus on obtaining higher rewards in the short term with the information he already have. Thus, oncologist will focus primarily to find for a best policy hidden among the available clinical records.

So, the model was used to predict the optimal drug schedule for the 7 testing patient set. We also computed the predicted reward for each cycle he will be getting by following the optimal policy suggested. The Evaluation chapter will contain the results achieved.

#### *6.3.4.3. Optimal policy via Exploration*

The term exploration refers to RL algorithm trying to find the optimal policy when new data or observations enters. The agent will focus on obtaining higher rewards in the long term by further exploring the environment. Thus, oncologist will focus secondarily to find the best policy when new patients with previously unobserved clinical observations are entered to the clinical records. This process will contains two stages.

- Stage 1: Chemo RL model will predict the real patient status and the dose amount for cycle 1. For this purpose, the algorithm will consider all the 40 patients data irrespective of the cycle length. (During the exploitation process, we only considered 34 patients, removing the patients who are having too shorter and too longer chemotherapy cycles).
- Stage 2: Chemo RL model will predict the Relative Dose Index (RDI) for cycles 2-6. For this purpose, the algorithm will consider all the 40 patients data and the predicted cycle 1 dose amount from the stage 1.

So, the model was used to predict the optimal drug schedule when new clinical observations entered to the medical records by creating virtual patient instances. We also computed the predicted reward he will be getting by following the optimal policy suggested. The Evaluation chapter will contain the results achieved.

#### 6.3.5. *Results logging module*

Here the optimal treatment schedules for the two strategies exploitation of the dataset and the exploration for new observations for a treatment time span of 6 cycles will be saved as .jpg images locally.

## 6.4. Tools and Technologies

### 6.4.1. *Integrated Development Environments (IDEs) Used*

#### 6.4.1.1. *JetBrains-PyCharm (IDE)*

PyCharm is an Integrated Development Environment (IDE) used in computer programming, specifically for the Python language. PyCharm is cross-platform, which works with Windows, macOS and Linux versions. Some of the major features PyCharm that supported in this development are; Coding Assistance and analysis, Project and code navigation, Integrated python debugger, Python refactoring and Version control integration etc.

#### *6.4.1.2. Anaconda navigator*

Anaconda Navigator is a desktop graphical user interface (GUI) that comes with the Anaconda® distribution which helps you to run programs and control conda packages, environments, and channels without having to use command-line commands. Packages can be found on Anaconda.org or in a local Anaconda Repository using Navigator. It's compatible for Windows, Mac OS X, and Linux.

### *6.4.2. Languages used*

Python 3.6 is used as the main programming language for the application.

### *6.4.3. Libraries used*

#### *6.4.3.1. D3rlpy*

D3rlpy is a data-driven deep reinforcement-learning library that is simple to use. It includes out-of-the-box scikit-learn-style APIs that include state-of-the-art data-driven deep reinforcement learning algorithms. Unlike other RL libraries, the given algorithms are said to be tweaked to achieve incredibly powerful efficiency. The D3rlpy package offers both continuous control algorithms and Discrete control algorithms. Continuous control algorithms include the Behavior Cloning (BC), Deep Deterministic Policy Gradients algorithm (DDPG), Twin Delayed Deep Deterministic Policy Gradients algorithm (TD3),

Soft Actor Critic Algorithm (SAC), Batch-Constrained Q-learning algorithm (BCQ), Bootstrapping Error Accumulation Reduction algorithm, Conservative Q-Learning algorithm (CQL), Advantage-Weighted Regression algorithm (AWAC), Policy in Latent Action Space algorithm (PLAS) and the Policy in Latent Action Space algorithm with perturbation layer (PLASwithPerturbation). Discrete control algorithms include Behavior Cloning algorithm for discrete control (DiscreteBC), Deep Q-Network algorithm (DQN), Double Deep Q-Network algorithm (DoubleDQN), Soft Actor-Critic algorithm for discrete action-space (DiscreteSAC), Discrete version of Batch-Constrained Q-learning algorithm (DiscreteBCQ), Discrete version of Conservative Q-Learning algorithm (DiscreteCQL) and the Discrete version of Advantage-Weighted Regression algorithm (DiscreteAWR). The package also provides Scikit learn capability, data augmentation methods, data preprocessing techniques, network architectures, data logging facility, and online & offline model training capability.

#### *6.4.3.2. Tensorflow*

TensorFlow is an end-to-end open source platform for machine learning initially developed by Google. TensorFlow is popular in its applications and usage in develop and train machine learning models. In our application, Tensorflow 2.0.0 is used in developing and training the Deep Q Network.

#### *6.4.3.3. Matplotlib*

Matplotlib is the general plotting library for the Python programming language and it is used in visualizing purposes in the proposed solution.

### *6.4.4. Hardware used*

#### *6.4.4.1. Graphics Processing Unit (GPU)*

Intel® UHD Graphics 620 GPU is used as resource requirement.

### 6.5. Summary

This chapter presented the implementation process of the developed system. Here, the details on implementation of the modules and the tools and technologies adopted have been presented in detailed. The modular designs identified in the Design chapter were implemented by means of introducing software architectures and this chapter presented the implementation details in detailed. In the next chapter, discuss about the Evaluation of the developed system.

**CHAPTER 7**

# 7. RESULTS AND EVALUATION

## 7.1. Introduction

Chapter 05 presented the Design of the proposed system. Chapter 06 presented the Implementation of the system. This chapter presents details on how development of the simulation was done and how each of the module were developed. Then, it is followed by the tools and technologies adopted in the implementation.

## 7.2. Experimental design

The hypothesis of the research is that a simulated model by using Online Reinforcement Learning and a more real life applicative model by using Offline Reinforcement Learning approaches can be adapted to solve the problem.

## 7.3. Online RL based approach results

### *7.3.1. Model validity for 3 types of patient environment*

As mentioned in the implementation section, we approached three types of patients for the model validity process as the young patient, pregnant woman with cancer and the elderly person with other types of diseases.

#### *7.3.1.1. Simulations for the young patient*

Young patients' body is more resistant for the chemotherapy drugs and also the side effects of chemotherapy. His/her body will regenerate the normal cells, immune cells more easily and effectively than the other types of patients. Thus, oncologist will consider to eradicate the tumor cells by giving a higher range of drug amount. Therefore for the simulative model of the young patients, we used the action space as 0.7, 0.73, 0.74, 0.75, 0.77, 0.78, 0.80, 0.82, 0.85, 0.87, 0.9, 0.95, 1 which is ranging in between 0.7 – 1. Figure 30 shows the state responses of a young patients when the chemotherapy drug is administered using the implemented RL based controller, which will includes the Normal cells, Tumor cells, Immune cells and Drug concentration on blood variations for a random policy. The random policy associated with this result is [1, 0.8, 0.87, 1, 1, 1, 1] for the 7 months we are expecting as the chemotherapy treatment period.

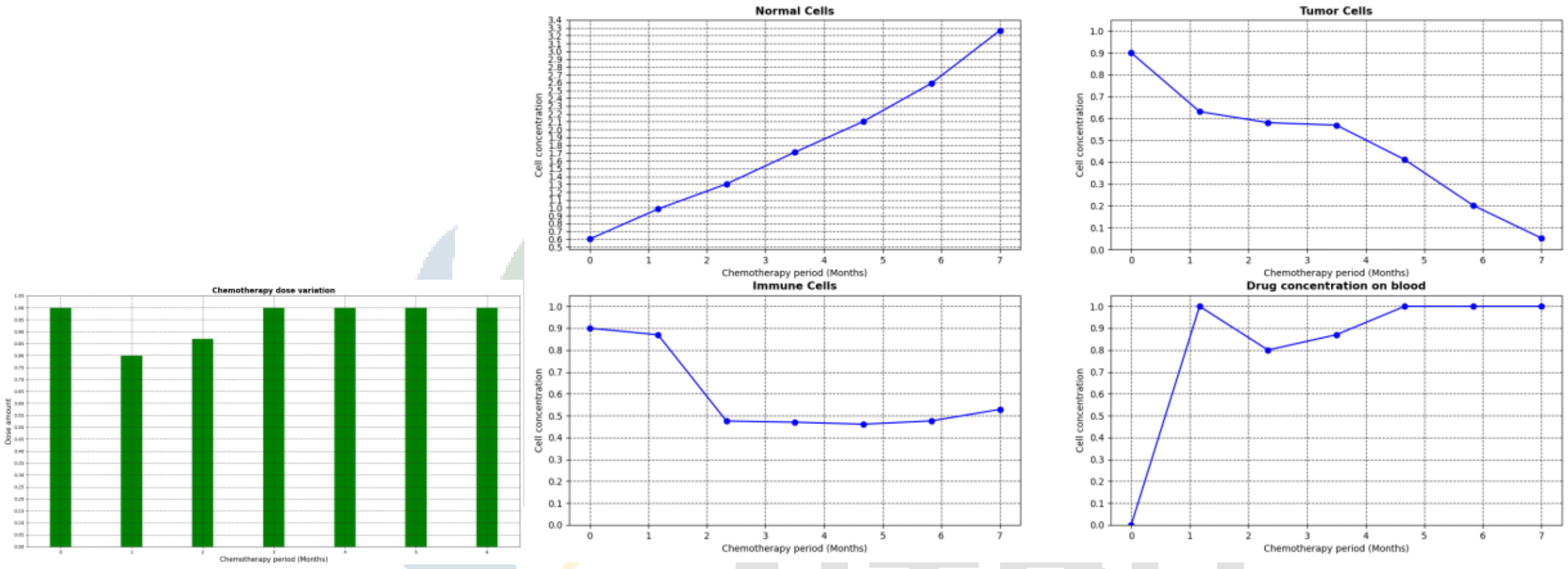

Figure 30: Random policy for a young patient

Then we re ran the model for several times and compared the state behaviors with several randomized policies. Finally, the optimal policy which yield the highest return over the chemotherapy period was acquired at episode number 8033. The optimal dosing schedule for young patients is [0.95, 0.77, 0.9, 0.95, 0.95, 0.95, 1] and the states behavior for the young patients is as follows in the figure 31, for the optimal policy. It can be seen that with the chemotherapy, the tumor cells count is decreasing, normal cells are increasing. It is seen that with the start of chemotherapy treatment, the immune cells have been decreased as a result of the drug effect and later the number of immune cells in being increasing with time. The drug concentration on blood is too increasing and maintain a constant pattern with a little drop after the 2nd chemotherapy cycle.

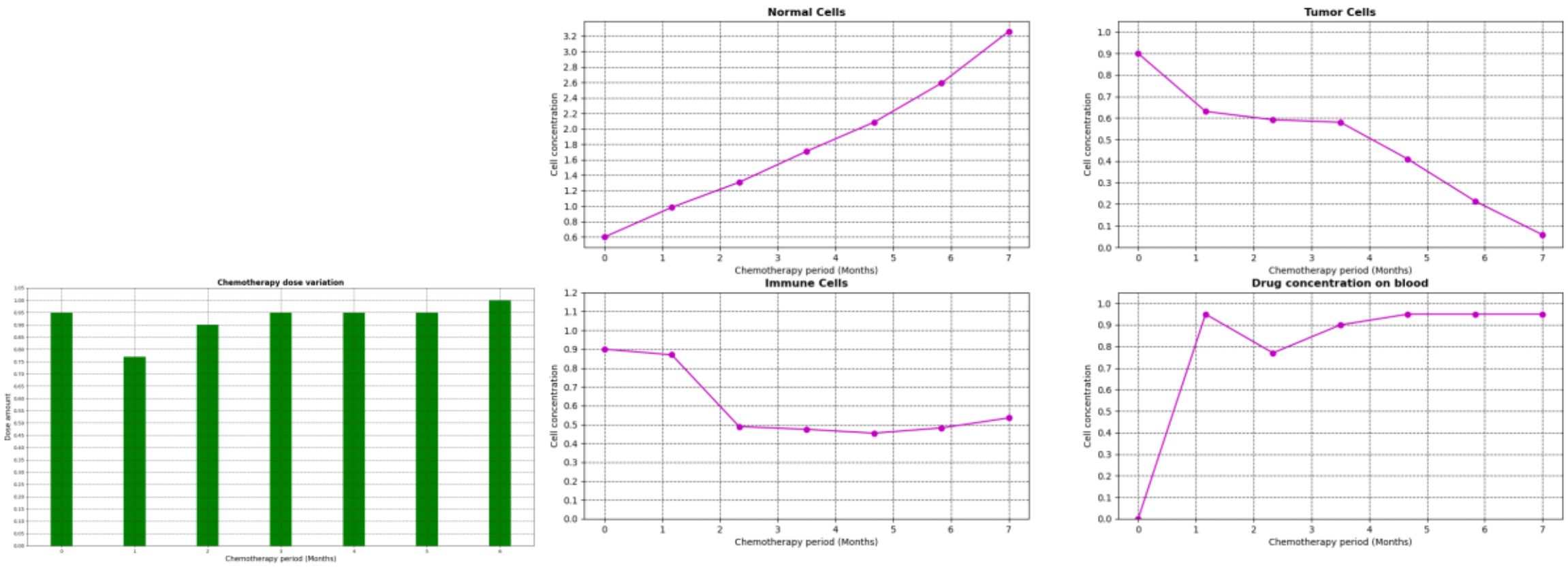

Figure 31: Optimal policy for a young patient

#### *7.3.1.2. Simulations for the pregnant woman with cancer*

For the second type of patient, we consider a pregnant woman with cancer. As mentioned in the section 6.1, the oncologist will try to keep the drug dosage in a minimum level in order to maintain the safety of the child. Later after the child birth, he will increase and change the dosage appropriately considering the mother's condition and the side effects she might encounter. Thus we approached some what varying dose level as 0.1-0.4 when she is pregnant and 0.8-0.9 of dose amount after she delivers the baby. The action space was finalized as 0,0.01,0.02,0.03,0.04,0.06,0.08,0.1,0.15,0.2,0.25,0.3,0.35,0.4, 0.8,0.9 for the pregnant woman with cancer. Figure 32 shows the state responses of a pregnant woman when the chemotherapy drug is administered using the implemented RL based controller, which will includes the Normal cells, Tumor cells, Immune cells and Drug concentration on blood variations for a random policy. The random policy associated with this result is [0.9, 0.02, 0.2, 0.1, 0.2, 0.9, 0.8] for the chemotherapy treatment period.

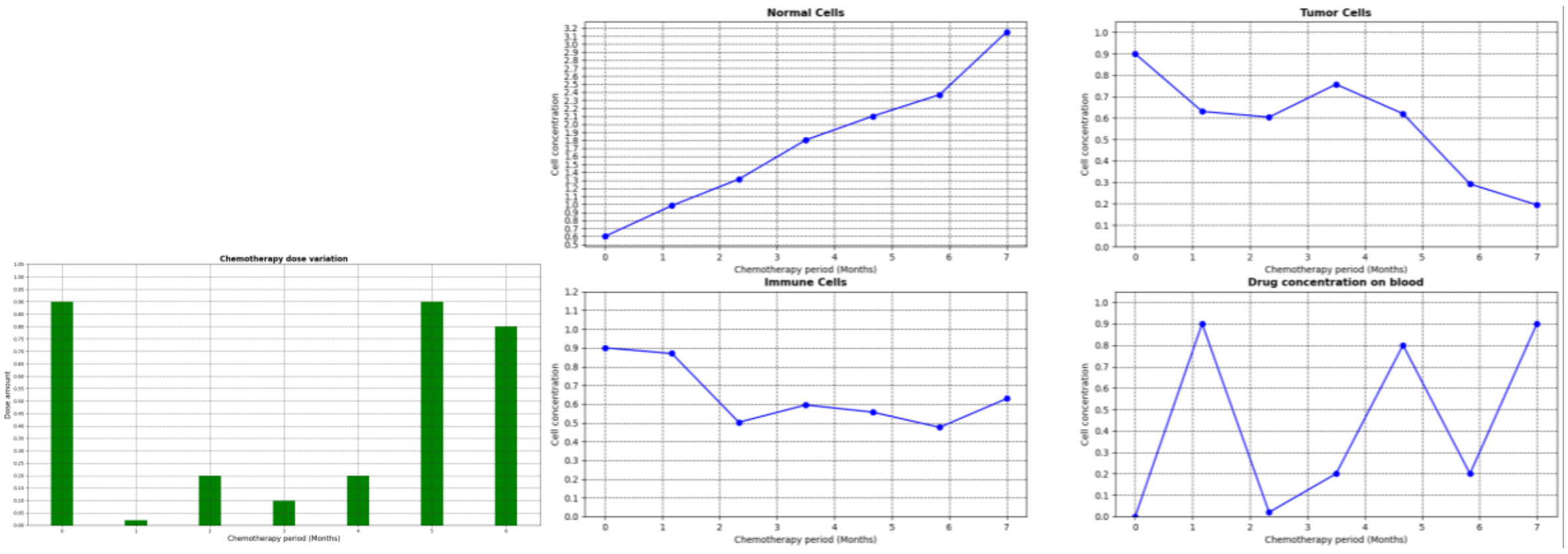

Figure 32: Random policy for a pregnant woman with cancer

Then we re ran the model for several times and compared the state behaviors with several randomized policies. Finally, the optimal policy which yield the highest return over the chemotherapy period was acquired at episode number 9517. The optimal dosing schedule for pregnant woman with cancer is [0.8, 0.02, 0.25, 0.35, 0.45, 0.9, 0.9 ] and the states behavior for the pregnant woman is as follows in the figure 33, for the optimal policy. It can be seen that with the chemotherapy, normal cells are increasing. However, around the period of the child delivery, the tumor cells count has been risen a bit because of using the lesser amount of chemotherapy drug amount. It is visible that the immune cells have been decreased because of the drug effect and later the number of immune cells has been increased. The drug concentration on blood is fluctuating very much since the input drug dosage to the body is too deviating because of the child delivery process.

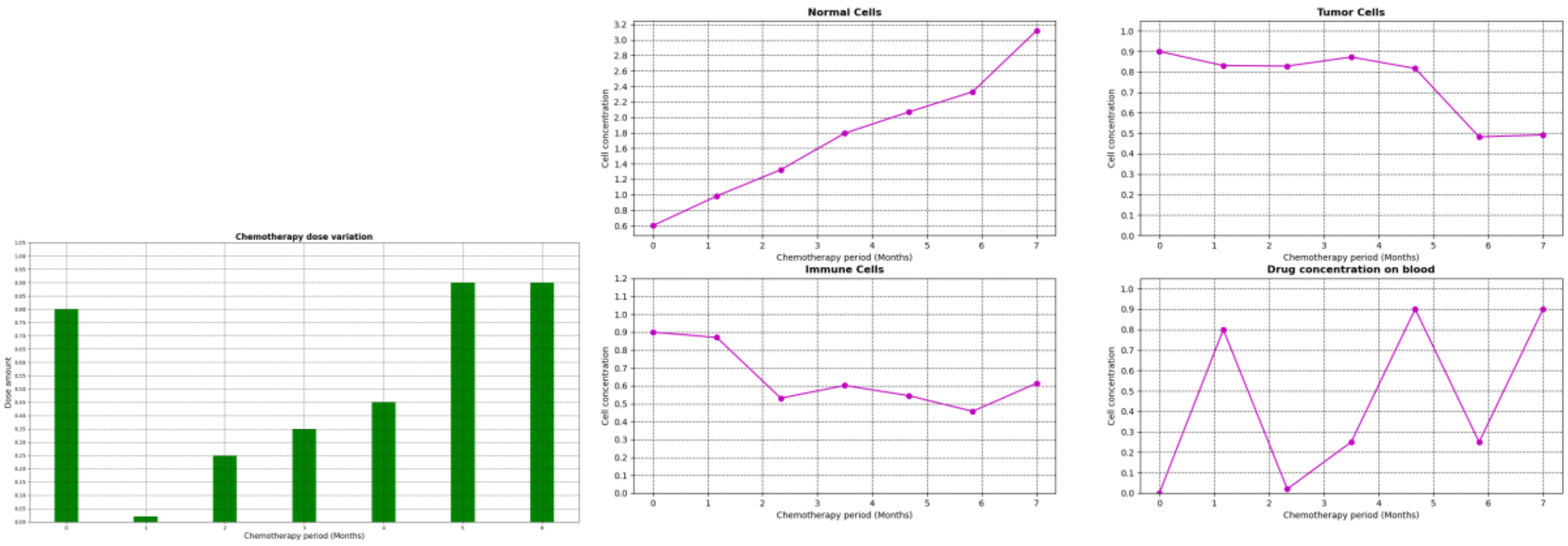

Figure 33: Optimal policy for a pregnant woman with cancer

#### *7.3.1.3. Simulations for the elderly person with other types of diseases*

For the third type of patient, we consider an elderly patient with other types of illnesses. As mentioned in the section 6.1, the oncologist will try to keep the drug dosage in a moderate level due to the fact that the elderly person has a low normal cells growth rate and a considerable high rate of side effects will also occur since he might be taking medicine parallel to cure his other illnesses too. Thus, we approached a medium dose level as 0.5-0.75. The action space was finalized as 0.5,0.52,0.53,0.54,0.57,0.6,0.65,0.68,0.7,0.72,0.75 for the elderly patient with other types of illnesses. Figure 34 shows the state responses of an elderly patient with other types of illnesses when the chemotherapy drug is administered using the implemented RL based controller, which will includes the Normal cells, Tumor cells, Immune cells and Drug concentration on blood variations for a random policy. The random policy associated with this result is [ 0.75, 0.57, 0.54, 0.6, 0.75, 0.75, 0.75 ] for the chemotherapy treatment period.

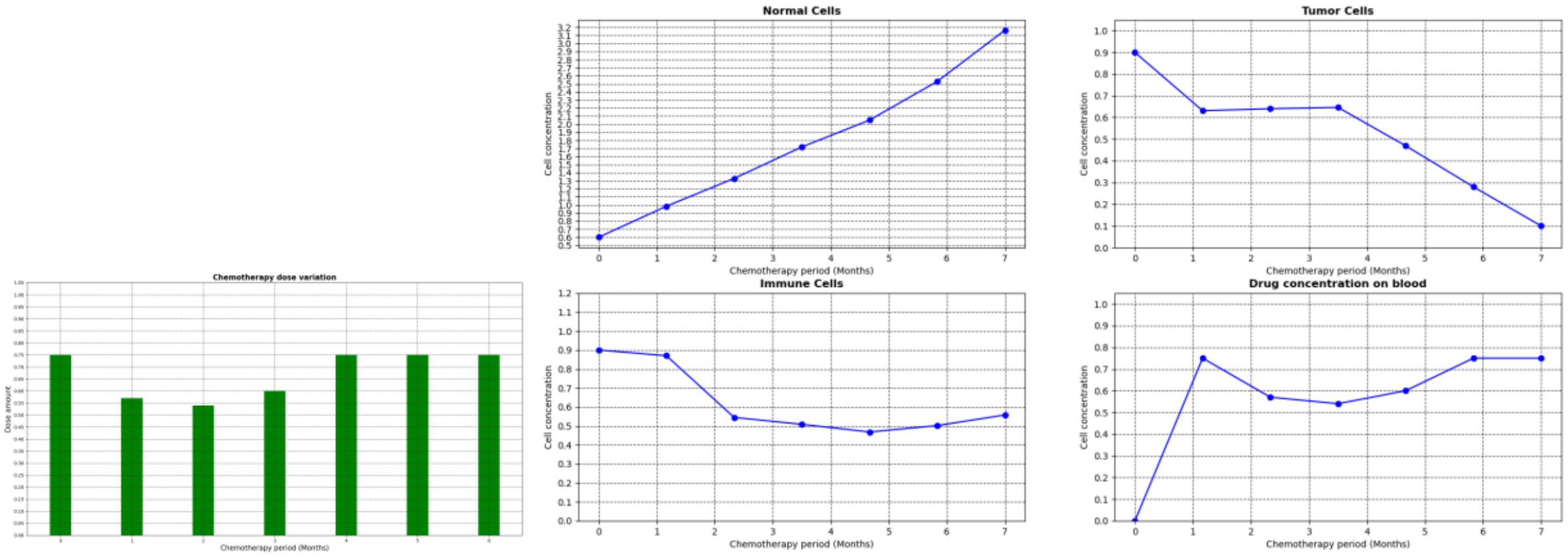

Figure 34: Random policy for an elderly person with other types of illnesses

Then we re ran the model for several times and compared the state behaviors with several randomized policies. Finally, the optimal policy which yield the highest return over the chemotherapy period was acquired at episode number 9022. The optimal dosing schedule for the elderly patient with other types of illnesses is [ 0.68, 0.5, 0.52, 0.68, 0.68, 0.68, 0.68 ] and the states behavior for the elderly patient with other types of illnesses is as follows in the figure 35, for the optimal policy. It can be seen that with the chemotherapy, normal cells are increasing. Tumor cells are decreasing with a little deviation around the $2^{nd}$ cycle. Immune cells are decreasing as a result of the chemotherapy dosage and later it is increasing with time. Drug concentration on blood is increasing first and maintains a constant trend after a little drop around the $2^{nd}$ chemotherapy treatment cycle.

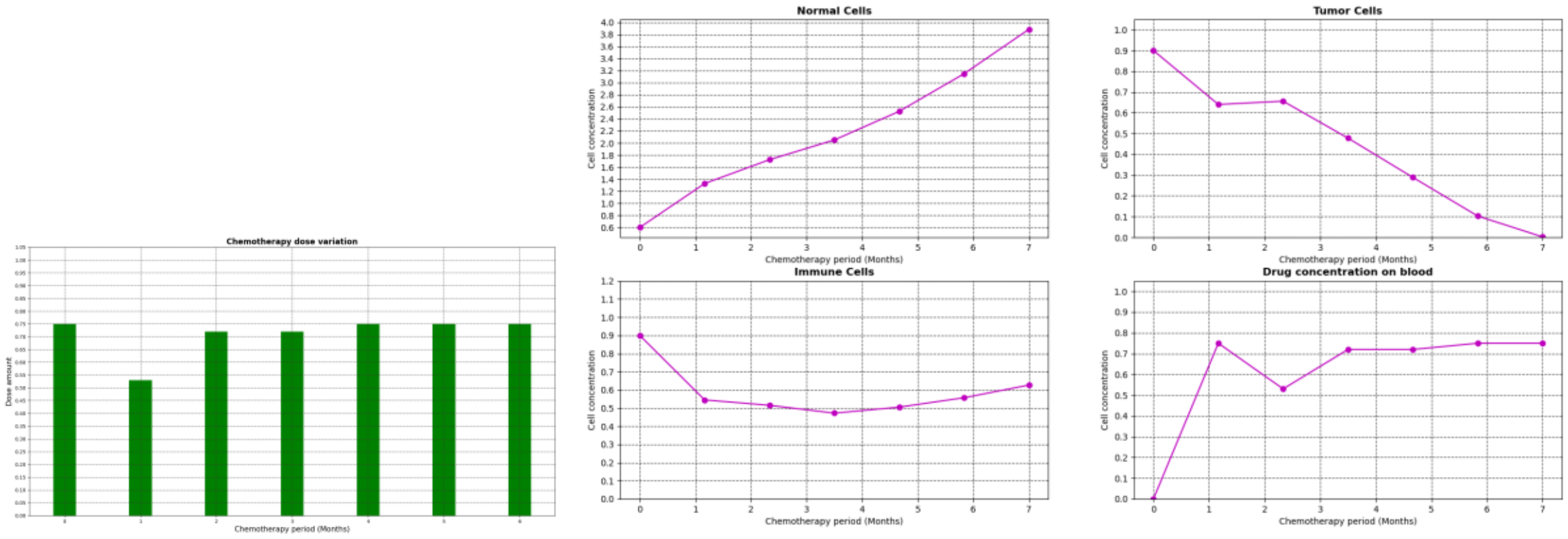

Figure 35: Optimal policy for an elderly person with other types of illnesses

### *7.3.2. Model validity through parameter variation of the ODE equations*

As mentioned in the implementation section, we generated artificial simulated data by changing the parameters of the ODE equations by -12% and +18%. We then used this modified ODE equations to generate the states changes for the three pre decided types of patients. Following figures illustrates their behavior.

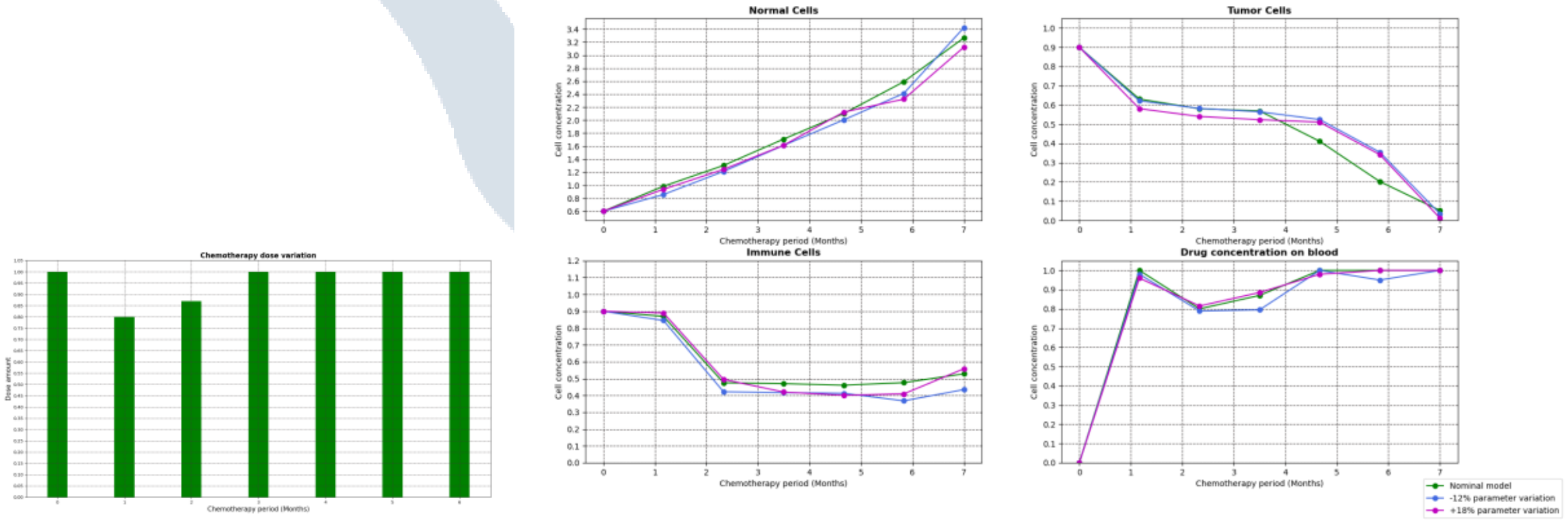

Figure 36: Response for the three different ODE models, for the random policy [ 1, 0.8, 0.87, 1, 1, 1, 1 ] of a young patient.

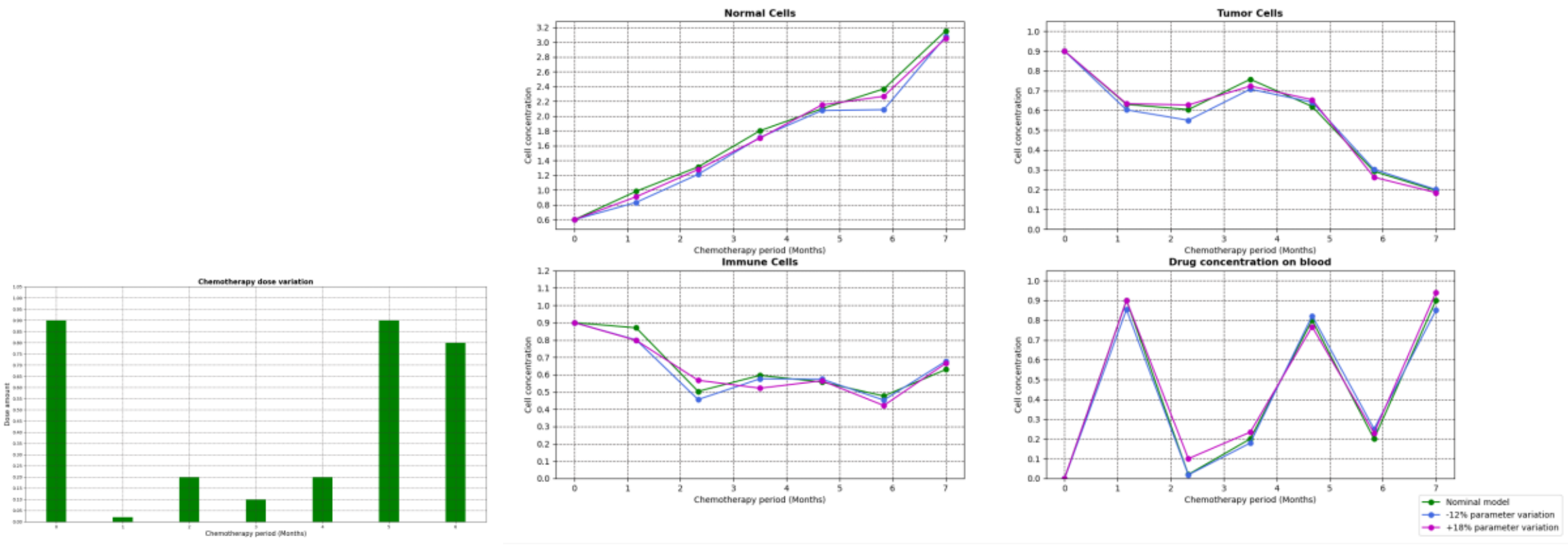

Figure 37: Response for the three different ODE models, for the random policy [ 0.9, 0.02, 0.2, 0.1, 0.2, 0.9, 0.8] of a pregnant woman with cancer.

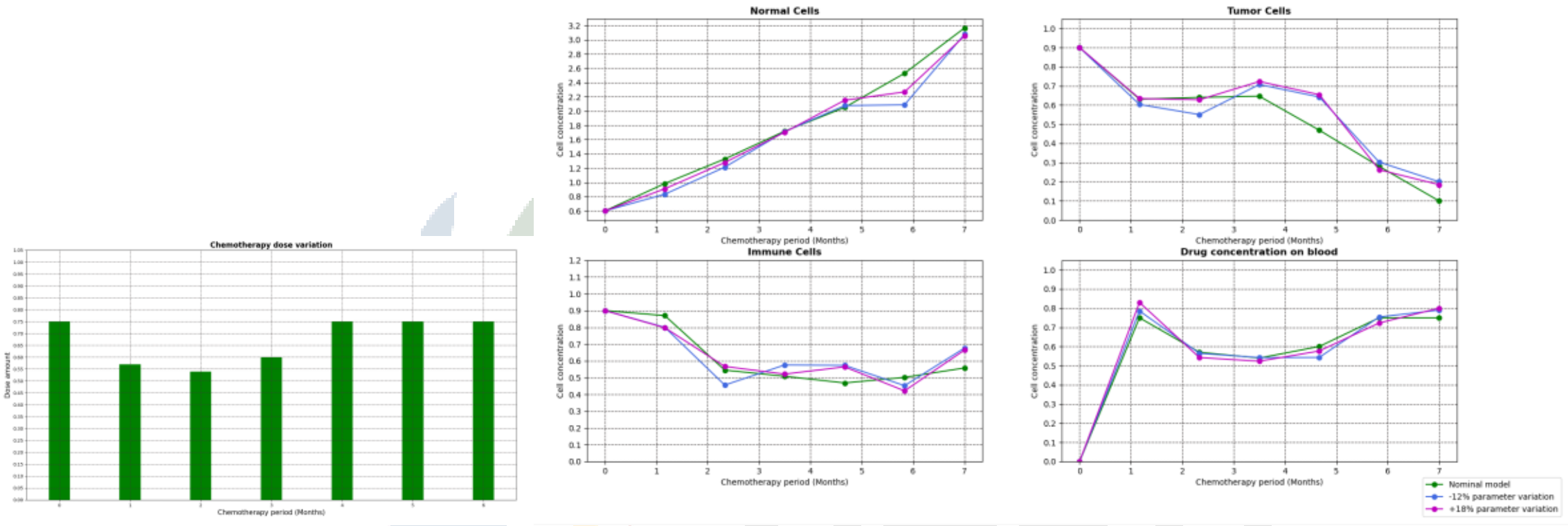

Figure 38: Response for the three different ODE models, for the random policy [ 0.75, 0.57, 0.54, 0.6, 0.75, 0.75, 0.75 ] of an elderly person with other types of illnesses.

### *7.3.3. Artificial generation of clinical data*

As explained in the implementation section, we generated 45 simulated patients belong to the three patient types as young patients, pregnant woman with cancer and the elderly person with other types of illnesses.

Following figure shows the optimal dose schedule variation for the 15 simulated young patients behave for the 7 cycles.

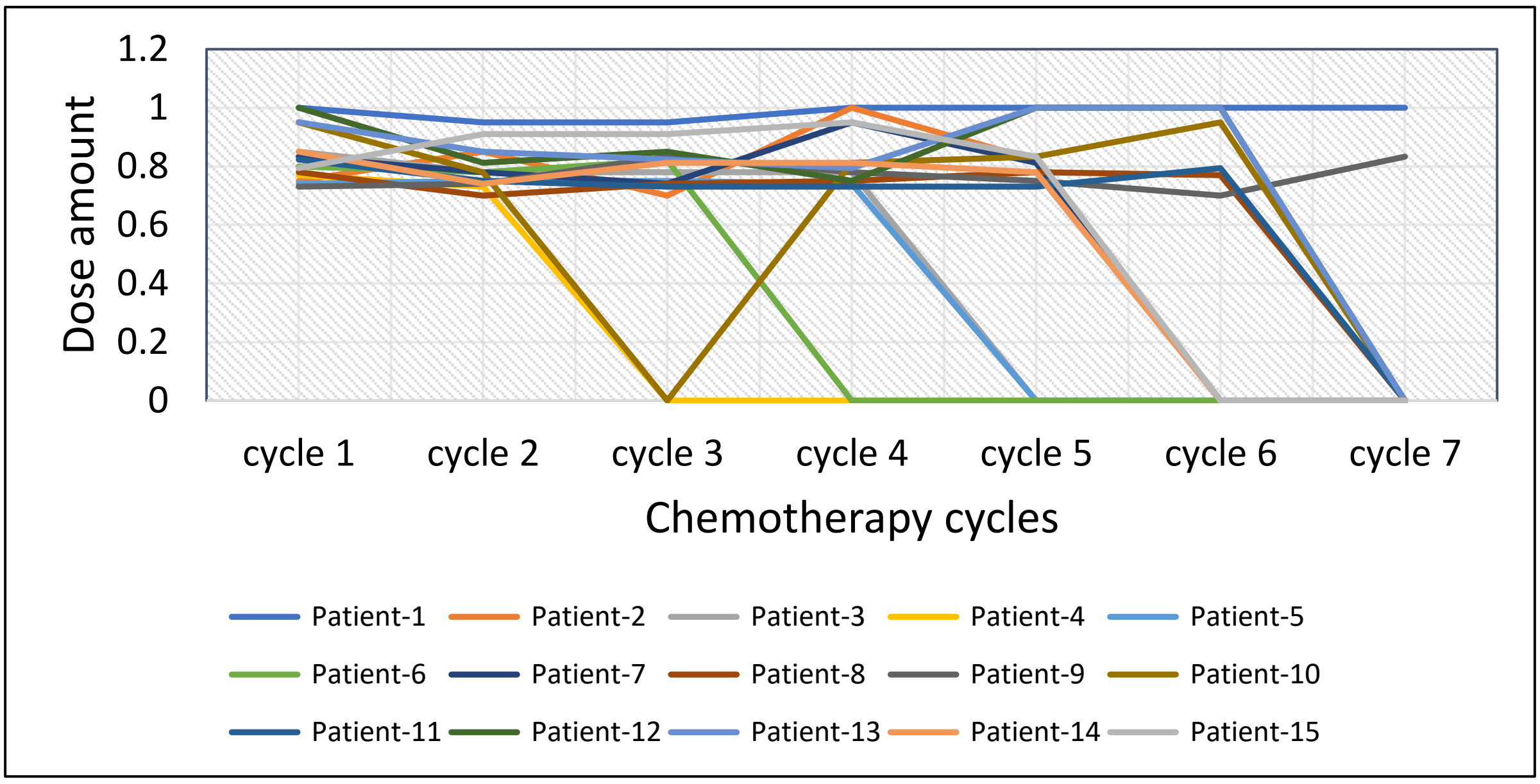

Figure 39: Optimal dose schedule for the 15 simulated young patients

Then we calculated how the mean optimal dose schedule will behave for the 15 young patients. The predicted optimal policy was [ 0.8413,0.7934,0.8029,0.8344,0.8489,0.8876,0.9165 ]. This optimal dosing schedule and the states behaviors are shown in the figure 40 graphically.

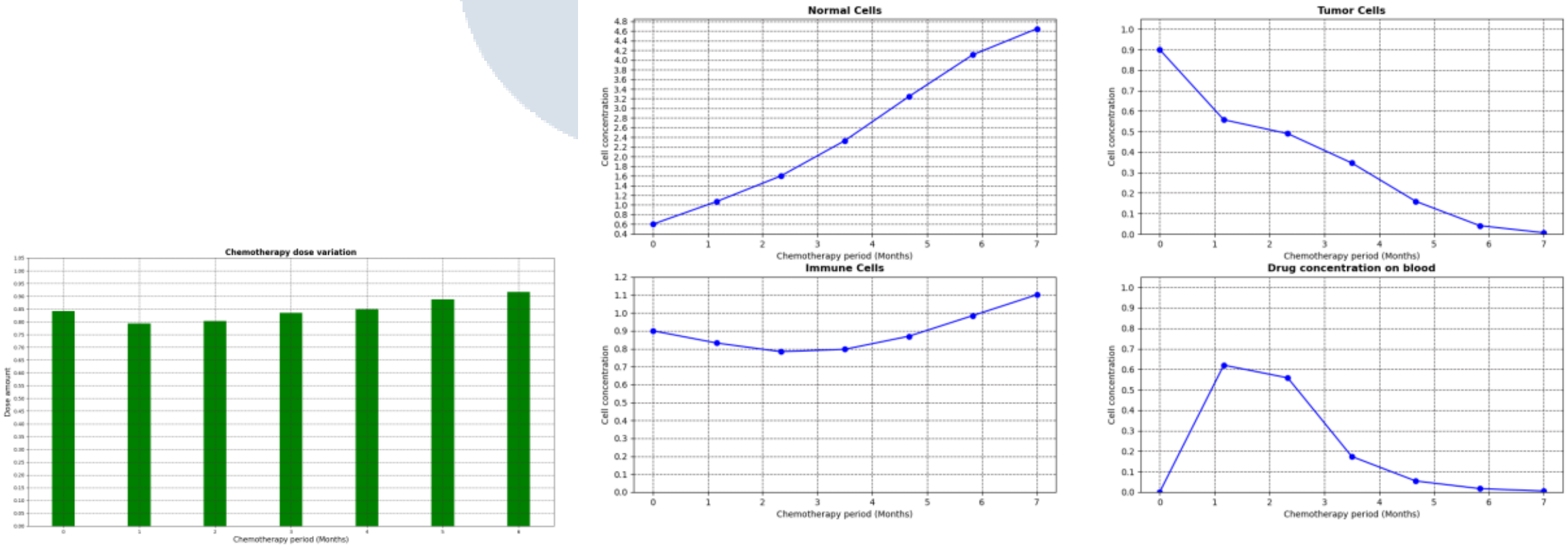

Figure 40: Mean Optimal dose schedule and the states behavior with the 15 simulated young patients

It is highlighted that the trend with the normal cells is increasing at an increasing trend.

This behavior is because young patient's normal cells are reproducing t a much faster rate than that of other patients. The tumor cells are decreasing along with time due to the effect from the chemotherapy drug dosage. The immune cells are decreasing first due to the effect of the chemotherapy and increasing later with the time since young person's body is more tolerant to the drugs. The drug concentration on the blood varies similar to a pattern of the input drug dosage.

Similarly, we simulated the iterative process for the pregnant woman with cancer and the elderly person with the other type of illnesses.

Following figure shows the optimal dose schedule variation for the 15 simulated pregnant woman with cancer behave for the 7 cycles.

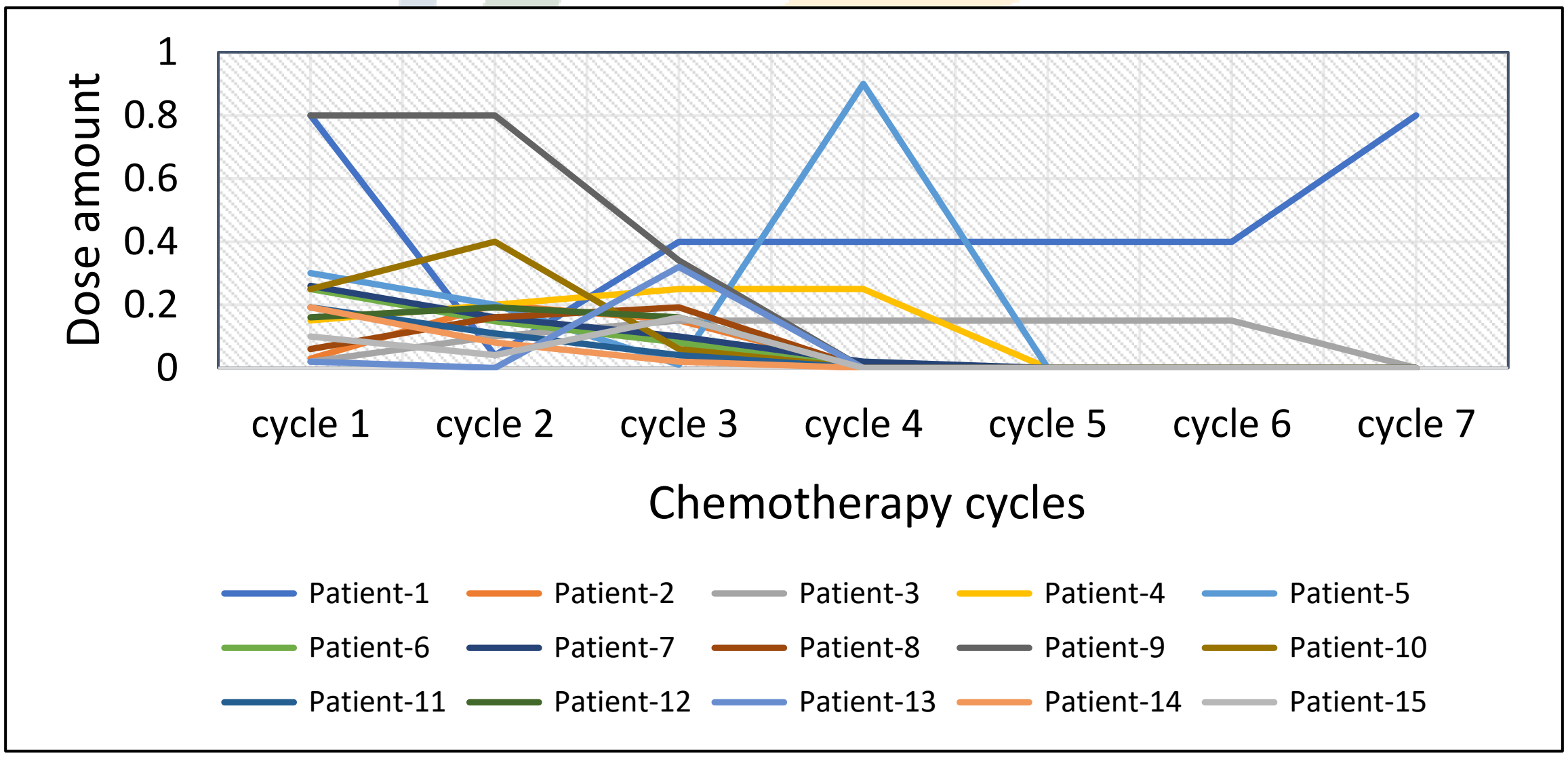

Figure 41: Optimal dose schedule for the 15 simulated pregnant woman with cancer patients

Then we calculated how the mean optimal dose schedule will behave for the 15 pregnant woman with cancer. The predicted optimal policy was [0.238933333, 0.188666667, 0.162133333, 0.344, 0.275, 0.275, 0.8 ]. This optimal dosing schedule and the states behaviors are shown in the figure 21 graphically.

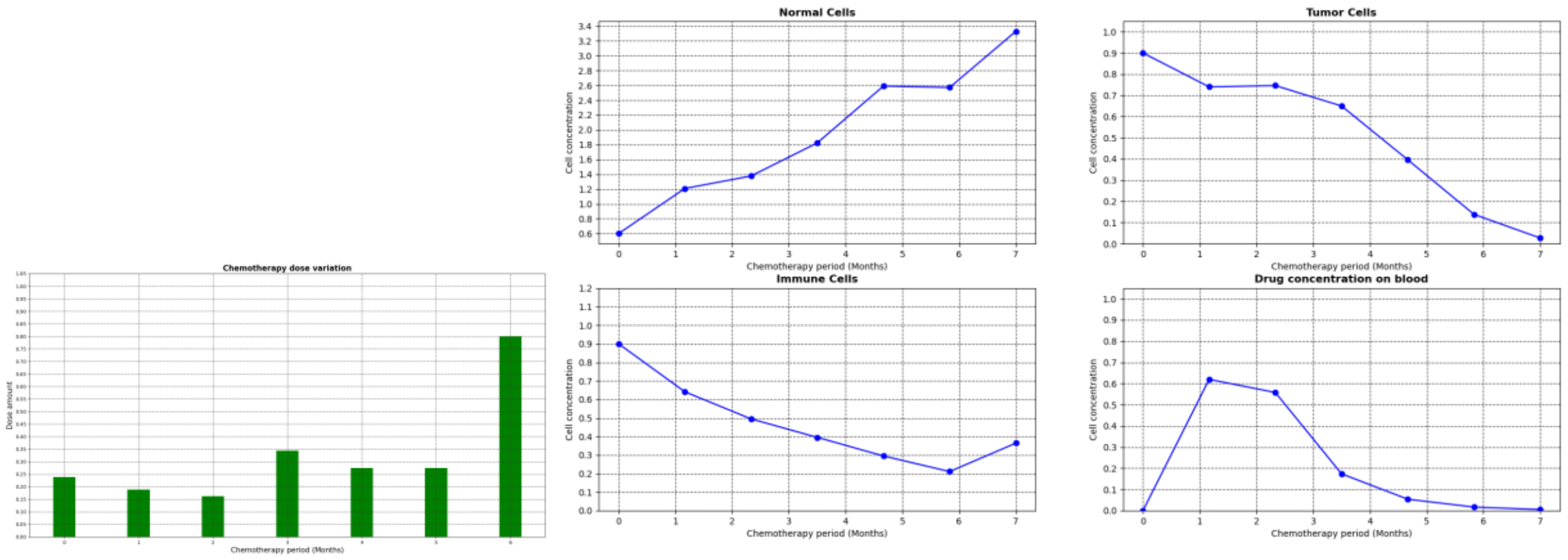

Figure 42: Mean Optimal dose schedule and the states behavior with the 15 simulated pregnant patients

It is seen that the trend with the normal cells is increasing, but in an fluctuating manner with several little changes. This can be because the drug dosage has many ups and downs in its trend. The tumor cells are decreasing along with time, but the trend has a decreasing pattern. The immune cells are decreasing first due to the effect of the chemotherapy and increasing only during the 6th month. Immune cells show a decreasing pattern during her pregnancy period and show an increasing pattern after the delivery of the baby. The drug concentration on the blood increases first but later decreases with the treatment period.

Following figure shows the optimal dose schedule variation for the 15 simulated elderly patients with other type of illnesses behave for the 7 cycles.

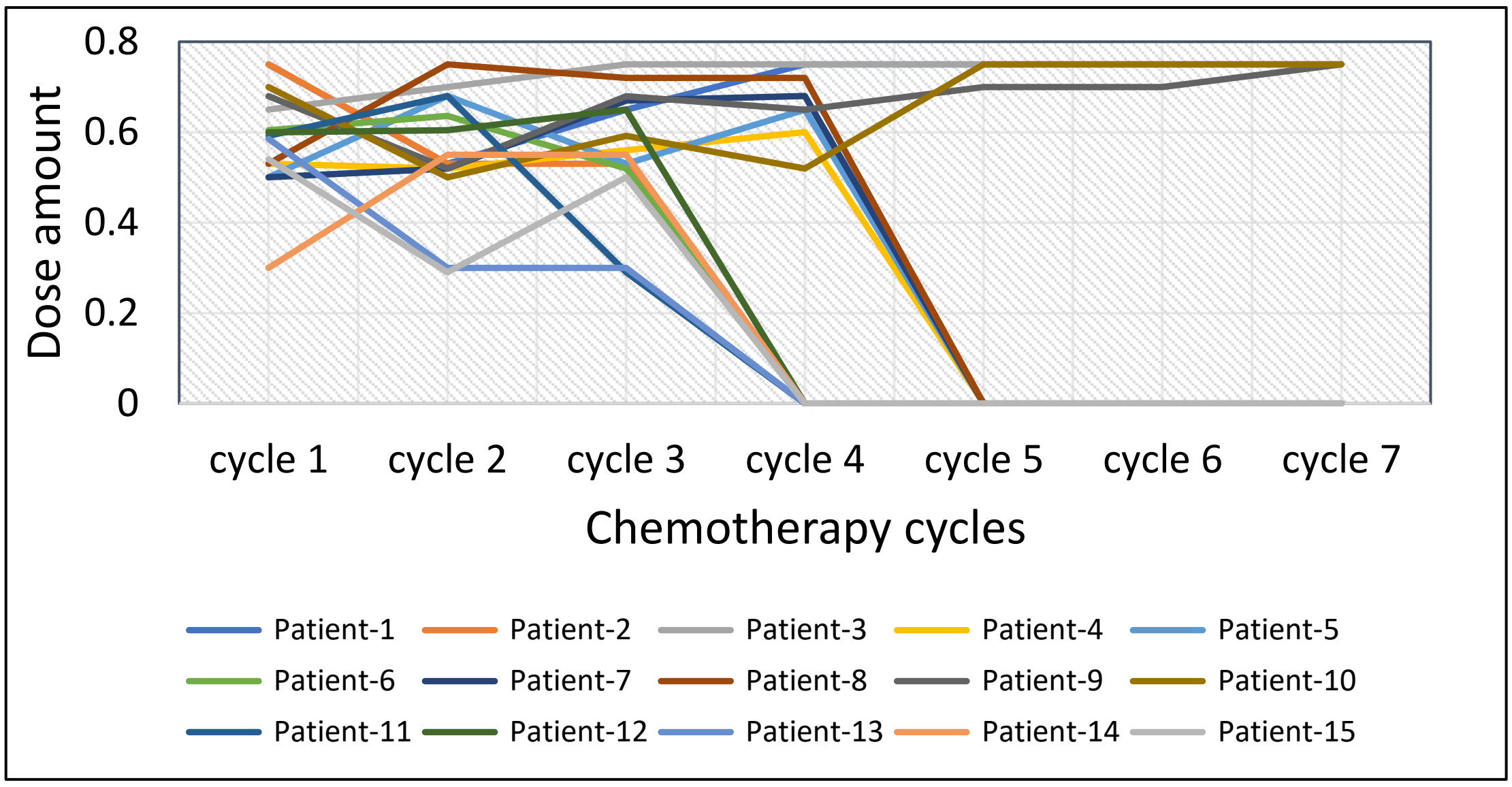

Figure 43: Optimal dose schedule for the 15 simulated elderly patients with other type of illnesses

Then we calculated how the mean optimal dose schedule will behave for the 15 elderly patients with other type of illnesses. The predicted optimal policy was [ 0.587413333, 0.554, 0.56613333, 0.657777778, 0.7375, 0.7375, 0.75 ]. This optimal dosing schedule and the states behaviors are shown in the figure 21 graphically.

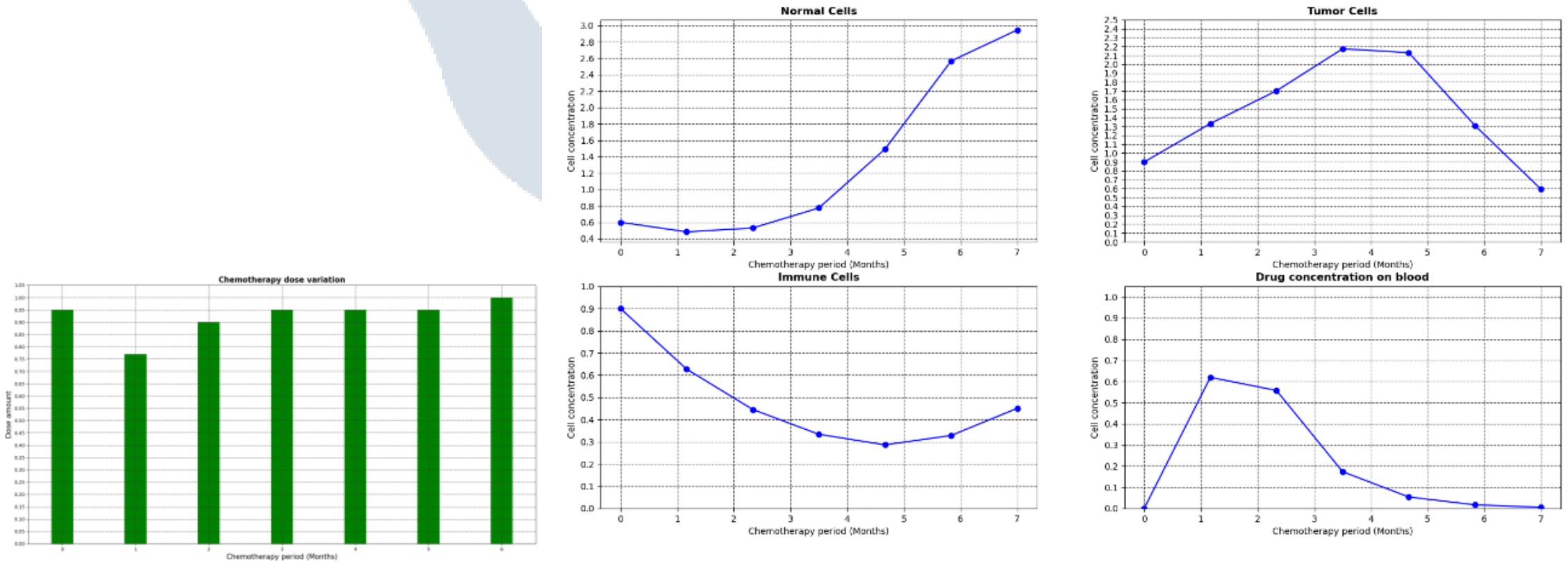

Figure 44: Mean Optimal dose schedule and the states behavior with the 15 simulated elderly patients

It is seen that the trend with the normal cells is increasing. The tumor cells are increasing along with time, and decreasing later. This can be due to the reason that the oncologist considers less consideration towards the tumor cells paying more attention on the growth of the normal cells in the case of the elderly patients. The immune cells are decreasing first due to the effect of the chemotherapy and increasing around during the 5$^{th}$ month and the graphs shows exponential behavior. The drug concentration on the blood increases first but later decreases with the treatment period.

## 7.4. Offline RL based approach results

### *7.4.1. Model Based State Estimation (MBSE) module simulation results*

This section includes the training and testing results achieved by using the CNN classifier and the RF classifier comparatively.

#### *7.4.1.1. Results of Real patient state decider using CNN classifier*

Following the CNN architecture design we have presented in the implementation section, the CNN classifier was trained and tested with the testing dataset. Prior to the finalizing of the final CNN architecture design we ran the model for several times by altering the activation functions and the optimizers in order to choose the model parameters that yields the best performance by carefully checking the training accuracy and the testing accuracy associated with the each test run. Following tables shows how the training and testing accuracies varies with the changing model parameters.

TABLE 8. TRAINING ACCURACY RESULTS FOR CHANGING OPTIMIZER AND ACTIVATION FUNCTION IN THE DESIGNED CNN ARCHITECTURE

| | | Optimizer | | |
|---|---|---|---|---|
| | | RMSProp | Admax | Adam |
| **Activation function** | Sigmoid | 59.634 % | 74.96 % | 83.97 % |
| | Relu | 75.95 % | 88.33 % | 93.67 % |

TABLE 9. VALIDATION ACCURACY RESULTS FOR CHANGING OPTIMIZER AND ACTIVATION FUNCTION IN THE DESIGNED CNN ARCHITECTURE

| | | Optimizer | | |
|---|---|---|---|---|
| | | RMSProp | Admax | Adam |
| **Activation function** | Sigmoid | 56.74 % | 64.61 % | 76.40 % |
| | Relu | 74.72 % | 80.34 % | 82.59 % |

From the results shown under the table 1 and table 2, it is seen that approaching Relu activation function with the Adam optimizer yields the best performance in terms of the training accuracy and the validation accuracy resulting during the training process. Therefore, the CNN architecture was decided to construct using these selected optimizer and the activation function. With several experimentally checking, a final training accuracy 93.67 % together with a final validation accuracy 82.59 % was achieved. We also stored the training accuracies, validation accuracies, training losses and the validation losses along with the with the number of iterations.

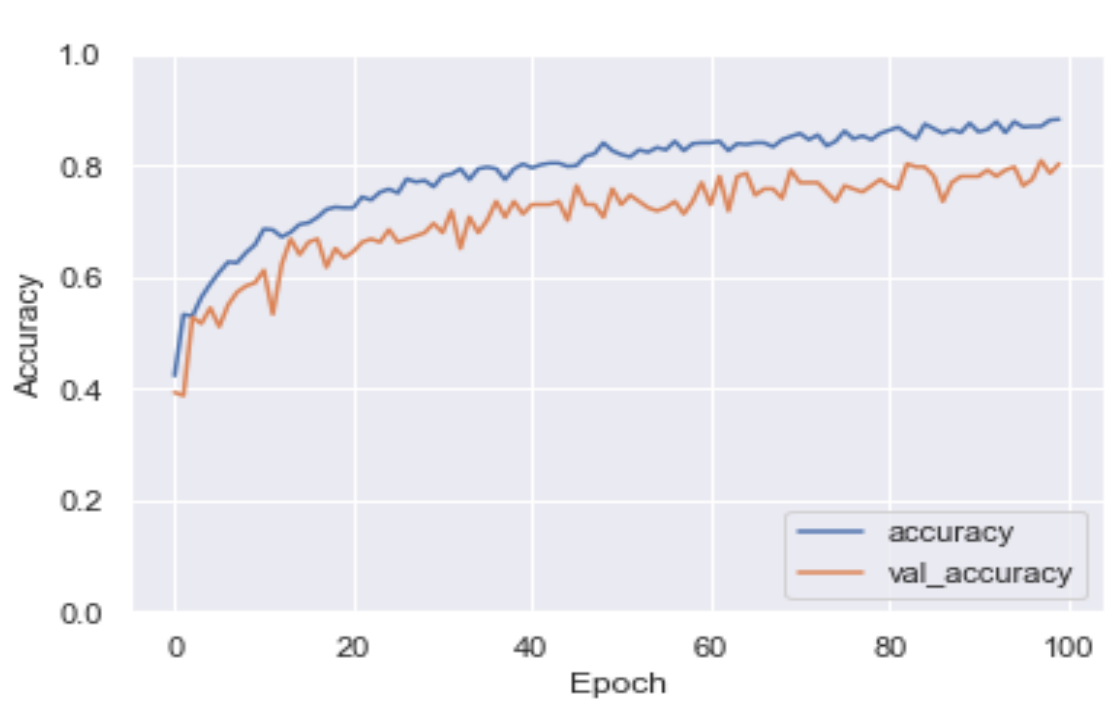

Figure 46: Training and validation accuracies of the proposed model

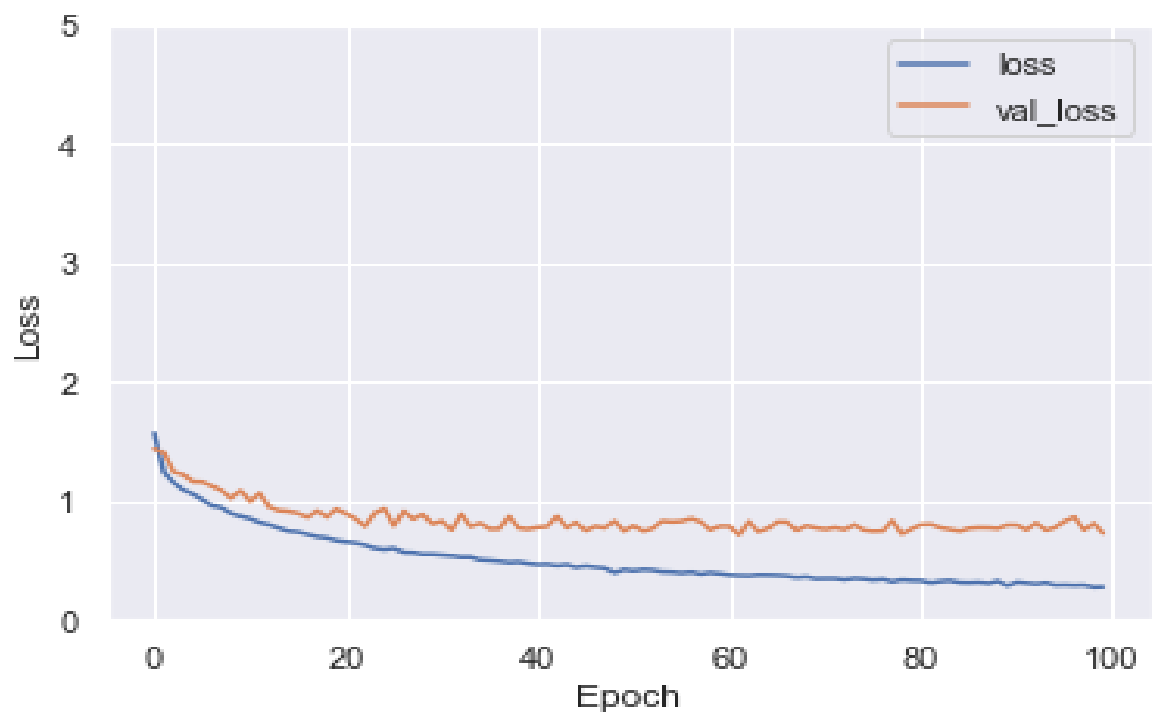

Figure 45: Training and validation accuracies of the proposed model

As shown in the figure 45, it is seen that both the training and validation accuracies are being increasing and then converging with the number of iterations. Similarly, as shown in the figure 46, is seen that both the training and validation losses are gradually decreasing with the iterations to converge to the approximation within respective precise ranges. The relative lower validation loss guarantees that no overfitting happened to the converged model.

In addition to the accuracies, a confusion matrix, which helps to graphically check the true label and the predicted label more comparatively, has also been constructed. The confusion matrix contains information about the actual and predicted classifications done by a classification system and the performance of such systems is commonly evaluated using the data in the matrix.

TABLE 10. CONFUSION MATRIX OF THE CNN MODEL

| | | Predicted Class | | | | | | | | |
|---|---|---|---|---|---|---|---|---|---|---|
| | | 3 | 4 | 5 | 6 | 7 | 8 | 9 | 10 | Recall (%) |
| Actual Class | **3** | **29** | 0 | 0 | 0 | 0 | 0 | 0 | 0 | 100.00 |
| | **4** | 0 | **24** | 0 | 0 | 0 | 0 | 0 | 0 | 100.00 |
| | **5** | 0 | 3 | **24** | 4 | 0 | 1 | 0 | 0 | 72.70 |
| | **6** | 0 | 0 | 2 | **18** | 6 | 0 | 1 | 0 | 63.00 |
| | **7** | 0 | 1 | 0 | 0 | **21** | 5 | 0 | 0 | 76.90 |
| | **8** | 0 | 0 | 2 | 2 | 3 | **12** | 4 | 0 | 52.20 |
| | **9** | 0 | 0 | 0 | 0 | 0 | 0 | **30** | 4 | 88.20 |
| | **10** | 0 | 0 | 0 | 0 | 0 | 0 | 0 | **27** | 100.00 |
| **Precision (%)** | | 100.00 | 85.70 | 85.70 | 70.80 | 66.70 | 66.70 | 85.70 | 87.10 | **82.10** |

In the above confusion matrix the bold numbers in the diagonal indicate correctly classified instances and the bottom right bold number indicates the overall accuracy which is 82.10%. We also calculated the performance measures of average recall score, precision score, f1_score, cohen kappa score, jaccard score in order to formulate a better comparison of the accuracy metrics associated with the testing process.

TABLE 11. PERFORMANCE METRICS OF THE CNN MODEL

| Metric | Average value |
|---|---|
| Average recall score | 82.10 |
| Average Precision score | 81.75 |
| Average f1_score | 81.55 |
| Cohen kappa score | 79.46 |
| Jaccard score | 70.90 |

#### *7.4.1.2. Results of Real patient state decider using RF classifier*

This section describes the results we acquired by using the Random Forest classifier following the implementation steps described in the implementation section. First data was loaded to the classifier, defined the model by using the param grid to choose the best performing hyper parameters. The chosen hyper parameters after the tuning process are as bellows.

TABLE 12. HYPER PARAMETERS RESULTED FOR THE RF CLASSIFIER

| Parameter | Parameter value of the best model |
|---|---|
| n_estimators | 1000 |
| max_features | 'auto' |
| max_depth | 12 |

| criterion | 'gini' |
|---|---|

The RF model was then defined by using the above parameters and trained. The training accuracy was 88.1% and the testing accuracy was 92.37%.

In addition, a confusion matrix, has also been constructed for the results achieved via RF classifier.

TABLE 13. CONFUSION MATRIX OF THE RF MODEL

| | | Predicted Class | | | | | | | | |
|---|---|---|---|---|---|---|---|---|---|---|
| | | 3 | 4 | 5 | 6 | 7 | 8 | 9 | 10 | **Recall (%)** |
| **Actual Class** | **3** | **37** | 0 | 0 | 0 | 0 | 0 | 0 | 0 | 100.00 |
| | **4** | 0 | **30** | 0 | 0 | 0 | 0 | 0 | 0 | 100.00 |
| | **5** | 0 | 0 | **29** | 1 | 0 | 0 | 0 | 0 | 96.70 |
| | **6** | 0 | 0 | 0 | **26** | 4 | 1 | 0 | 0 | 83.90 |
| | **7** | 0 | 1 | 0 | 1 | **19** | 3 | 0 | 0 | 79.20 |
| | **8** | 0 | 0 | 0 | 2 | 4 | **12** | 5 | 0 | 52.20 |
| | **9** | 0 | 0 | 0 | 0 | 0 | 1 | **23** | 0 | 95.80 |
| | **10** | 0 | 0 | 0 | 0 | 0 | 1 | 0 | **27** | 95.80 |
| **Precision (%)** | | 100.00 | 96.80 | 100.00 | 86.70 | 70.40 | 66.70 | 82.10 | 100.00 | **89.20** |

In the above confusion matrix the bold numbers in the diagonal indicate correctly classified instances and the bottom right bold number indicates the overall accuracy which is 89.20%. Similarly to the CNN classifier, We also calculated the performance measures of average recall score, precision score, f1_score, cohen kappa score, jaccard score and ROC_AUC score in order to formulate a better comparison of the accuracy metrics associated with the testing process.

TABLE 14. PERFORMANCE METRICS OF THE RF MODEL

| Metric | Average value |
|---|---|
| Average recall score | 89.24 |
| Average Precision score | 89.16 |
| Average f1_score | 89.01 |
| Cohen kappa score | 87.65 |
| Jaccard score | 82.44 |
| ROC_AUC Curve | 99.78 |

### *7.4.2. Colon Cancer Reinforcement Learning Chemotherapy (CCRLC) module simulation results*

This section includes the results achieved in the CCRLC module.

First, as mentioned in the MBSE module, we selected the optimal hyper parameters for the CQL and MOPO integrated model. The chosen hyper parameters after the tuning process are as bellows.

TABLE 15. HYPER PARAMETERS RESULTED FOR THE CQL & MOPO INTEGRATED MODEL

| **Parameter** | **Parameter value of the best model** |
|---|---|
| actor_learning_rate | 2e-05 |
| critic_learning_rate | 2e-05 |
| temp_learning_rate | 3e-05 |
| alpha_learning_rate | 2e-05 |
| gamma | 0.99 |
| Initial_alpha | 5.0 |
| tau | 0.08 |

The CQL and MOPO integrated model was then defined by using the above parameters and trained.

### *7.4.3. Visualization of the optimal drug policy*

This section includes the resulted drug optimal drug policies in a more detailed, visualized manner.

#### *7.4.3.1. Optimal policy via Exploitation results*

As described in the implementation section, we first used the trained model to generate the optimal policies for individual patients via exploitation. Thus, we used the model for the 7 test patient set which includes the patient No. 1, patient No. 7, patient No. 15, patient No. 25, patient No. 26, patient No. 37 and patient No. 40. For each patient, 4 drug combinations for a period of 6 cycles were generated. In order to check the correctness of the predicted drug amounts, a reasonable comparison between the given dose and the predicted optimal dose was carried out. The combination of drug was decided based on the type of drugs the oncologist has given to individual patients.

Following is the drug schedule decided for the patient No. 37 comparing with the given dose presented in the dataset.

TABLE 16. GIVEN DOSE COMPARISON WITH THE PREDICTED OPTIMAL DOSE FOR THE CYCLE 1: PATIENT NO. 37

<table>
<tr><th>Given doses for the cycle 1</th><th>Predicted optimal dose for the cycle 1</th></tr>
<tr><td>
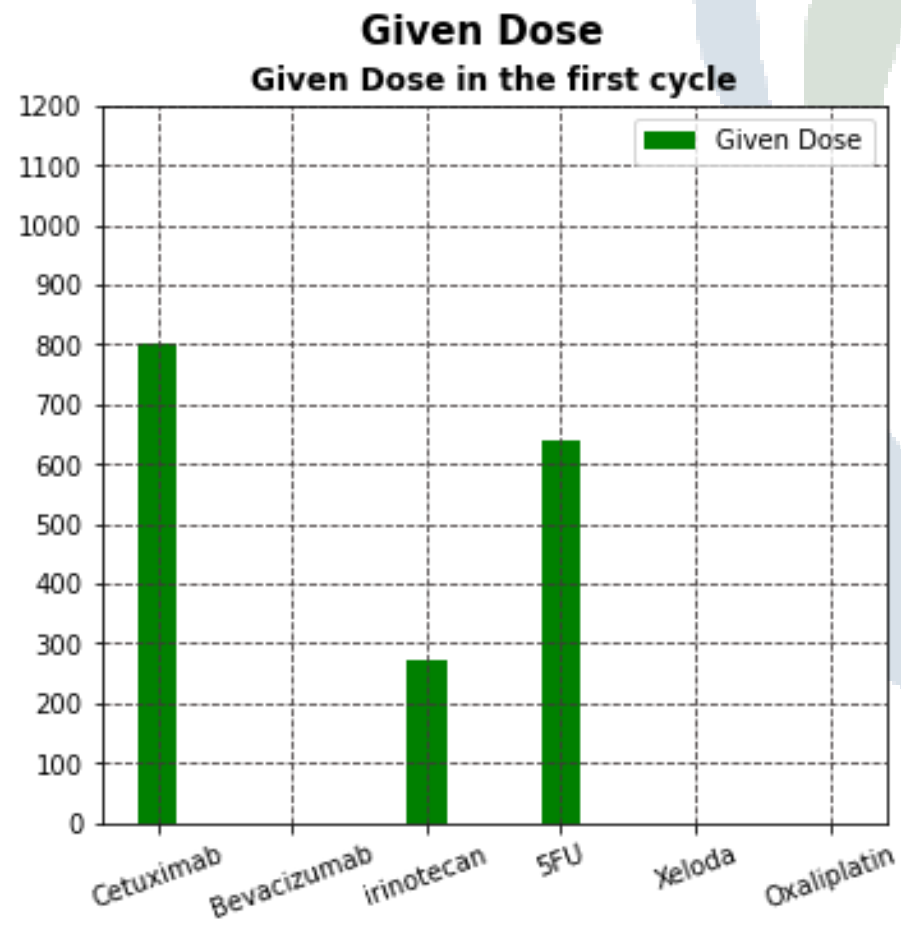

<br>Given Cetuximab dose: 800 mg<br>Given Irinotecan dose: 270 mg<br>Given 5FU dose: 640 mg
</td><td>
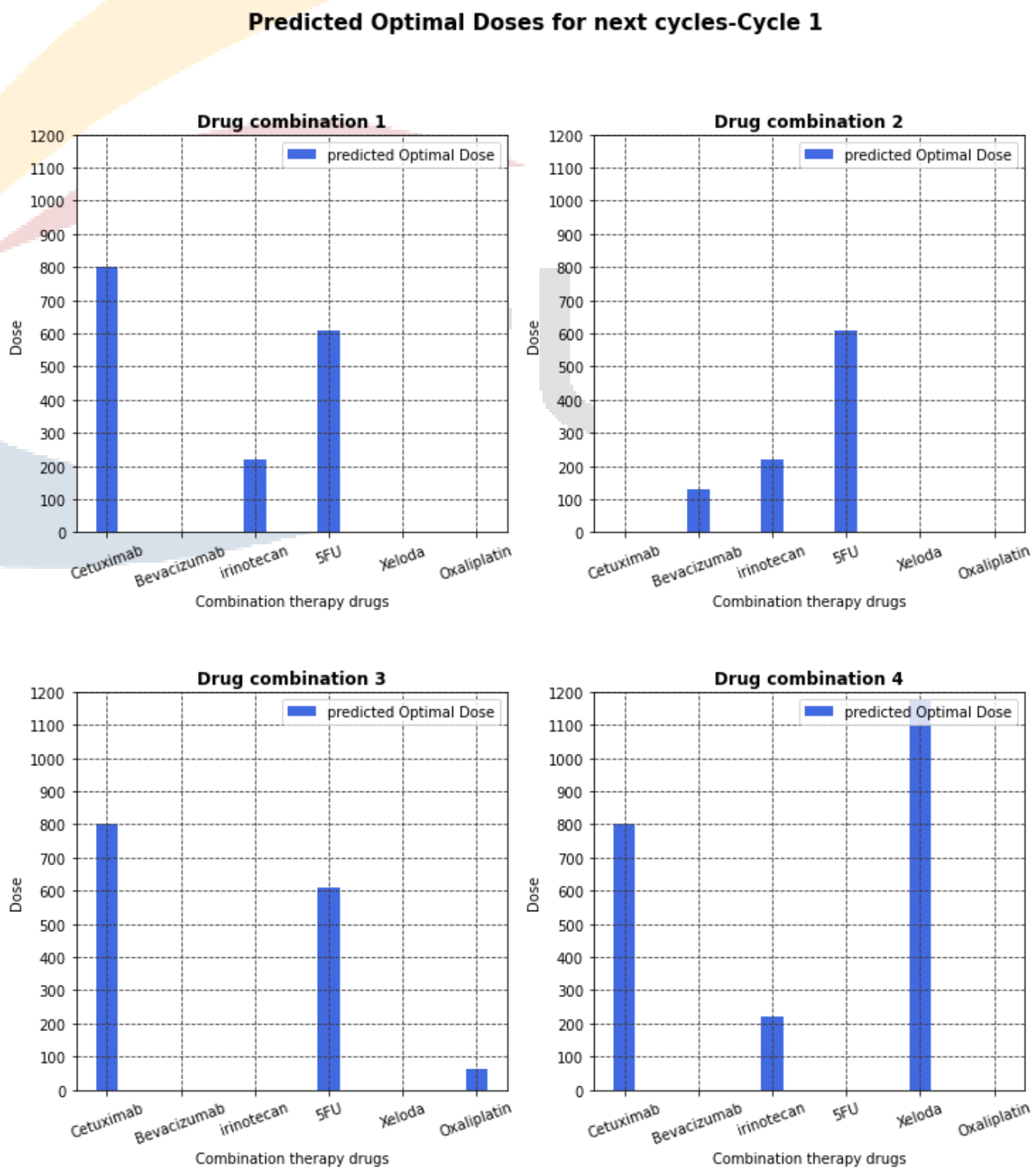

</td></tr>
</table>

| | | |
|---|---|---|
| | **Drug combination 1:**<br>Predicted Cetuximab dose: 800 mg<br>Predicted Irinotecan dose: 210 mg<br>Predicted 5FU dose: 603 mg | **Drug combination 2:**<br>Predicted Bevacizumab dose: 130 mg<br>Predicted Irinotecan dose: 210 mg<br>Predicted 5FU dose: 603 mg |
| | **Drug combination 3:**<br>Predicted Cetuximab dose: 800 mg<br>Predicted 5FU dose: 603 mg<br>Predicted Oxaliplatin dose: 65 mg | **Drug combination 4:**<br>Predicted Cetuximab dose: 800 mg<br>Predicted Irinotecan dose: 210 mg<br>Predicted Xeloda dose: 1175 mg |

Description: The patient No. 37 has been given the combination chemotherapy drug schedule as Cetuximab: 800 mg, Irinotecan: 270 mg, 5FU: 640 mg. The predicted optimal dose schedule contains 4 combinations as shown above. The drug combination 1 is the combination chemotherapy drug schedule similar to the given drug schedule. When comparing the Drug combination 1 with the given dose it is visible that it is almost similar. The proposed Chemo RL model have suggested similar Cetuximab dose, -22.2% Irinotecan dose (210-270/270) and -5.7% 5FU dose (603-640/640) which can be acceptable in the clinical sense.

TABLE 17. GIVEN DOSE COMPARISON WITH THE PREDICTED OPTIMAL DOSE FOR THE CYCLE 2: PATIENT NO. 37

| Given doses for the cycle 2 | Predicted optimal dose for the cycle 2 |
|---|---|

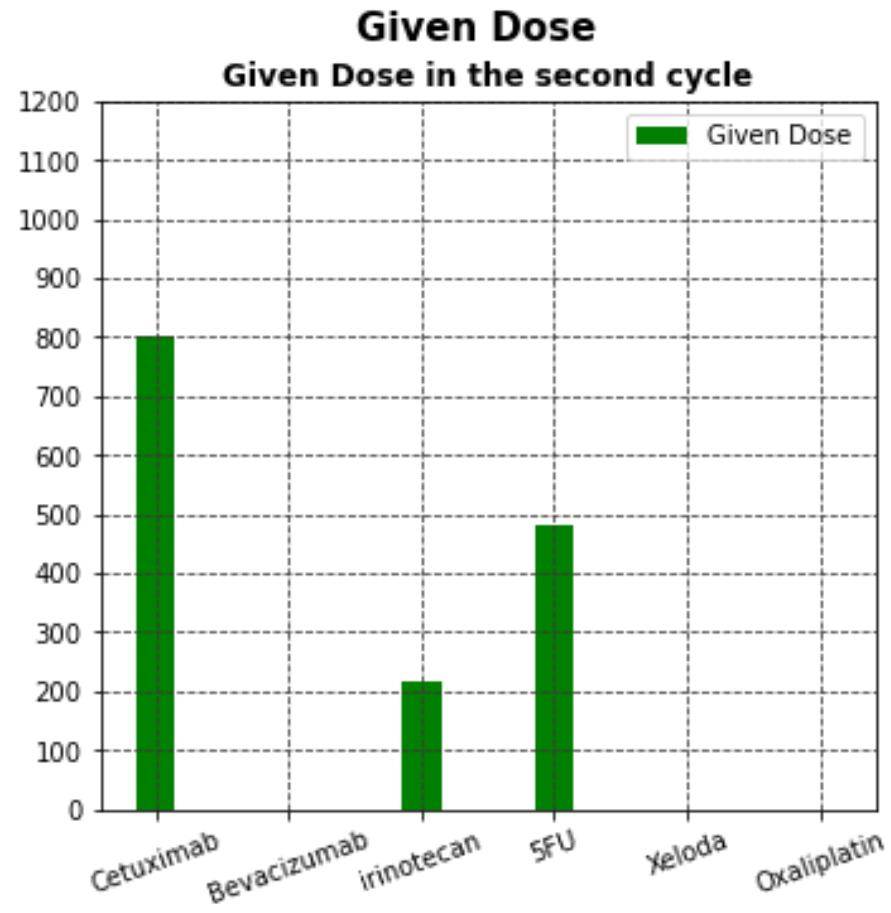

Given Cetuximab dose: 800 mg

Given Irinotecan dose: 216 mg

Given 5FU dose: 480 mg

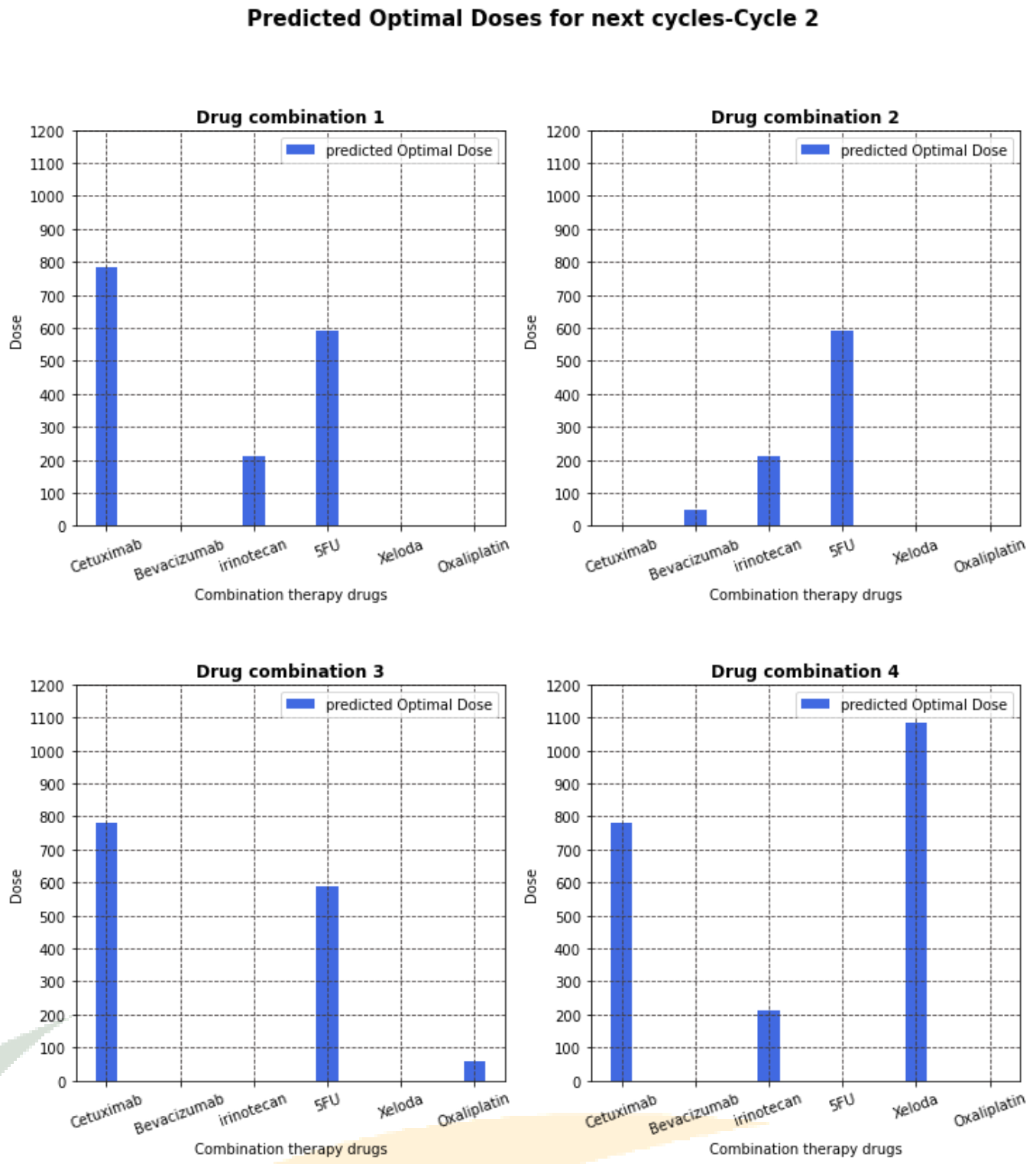

| | |
|---|---|
| **Drug combination 1:**<br>Predicted Cetuximab dose: 785 mg<br>Predicted Irinotecan dose: 210 mg<br>Predicted 5FU dose: 596 mg | **Drug combination 2:**<br>Predicted Bevacizumab dose: 65 mg<br>Predicted Irinotecan dose: 210 mg<br>Predicted 5FU dose: 596 mg |
| **Drug combination 3:**<br>Predicted Cetuximab dose: 785 mg<br>Predicted 5FU dose: 596 mg<br>Predicted Oxaliplatin dose: 65 mg | **Drug combination 4:**<br>Predicted Cetuximab dose: 785 mg<br>Predicted Irinotecan dose: 210 mg<br>Predicted Xeloda dose: 1085 mg |

Description: During the second cycle of the patient No. 37, the oncologist has prescribed the combination chemotherapy drug schedule as Cetuximab: 800 mg, Irinotecan: 216 mg, 5FU: 480 mg, which has only a little difference to the given dose at cycle 1. The predicted optimal dose schedule contains 4 combinations as shown above. The drug combination 1 is the combination chemotherapy drug schedule similar to the given drug schedule. When comparing the Drug combination 1 with the given dose it is visible that it is almost similar. The proposed Chemo RL model have suggested -1.875% Cetuximab dose (785-800/800), -2.78% Irinotecan dose (210-216/216) and +24.16% 5FU dose (596-480/480) which can be acceptable in the clinical sense.

TABLE 18. GIVEN DOSE COMPARISON WITH THE PREDICTED OPTIMAL DOSE FOR THE CYCLE 3: PATIENT NO. 37

<table>
<tr><th>Given doses for the cycle 3</th><th>Predicted optimal dose for the cycle 3</th></tr>
<tr><td>
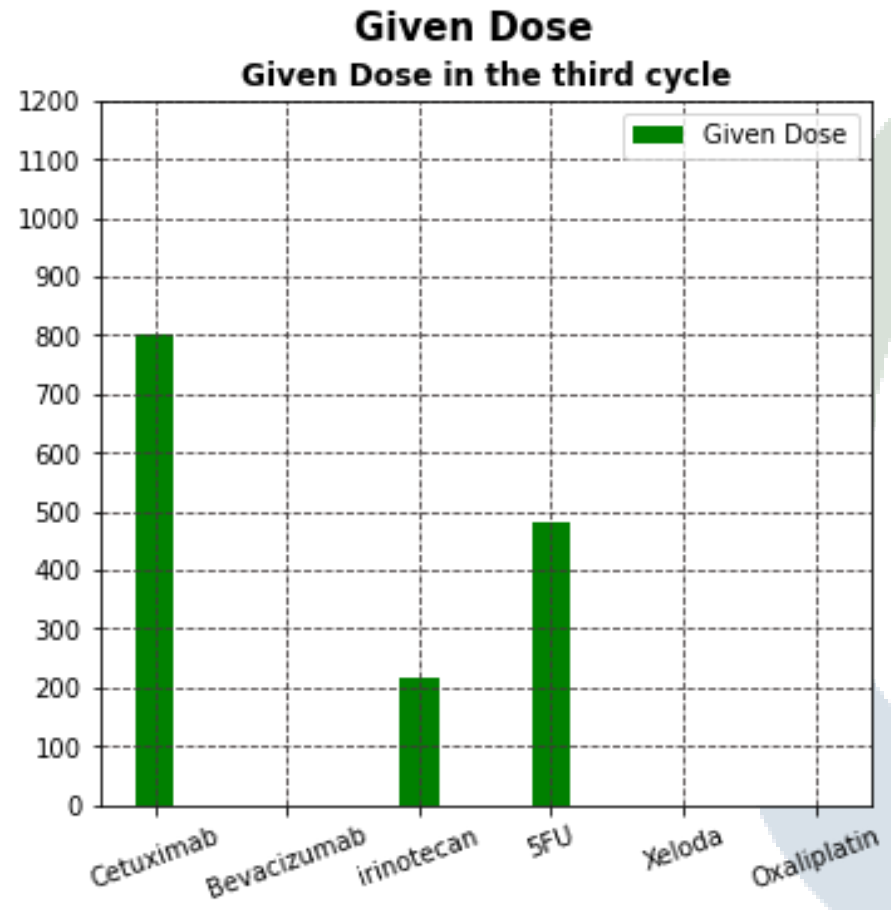

<br>Given Cetuximab dose: 800 mg<br>Given Irinotecan dose: 216 mg<br>Given 5FU dose: 480 mg
</td><td>
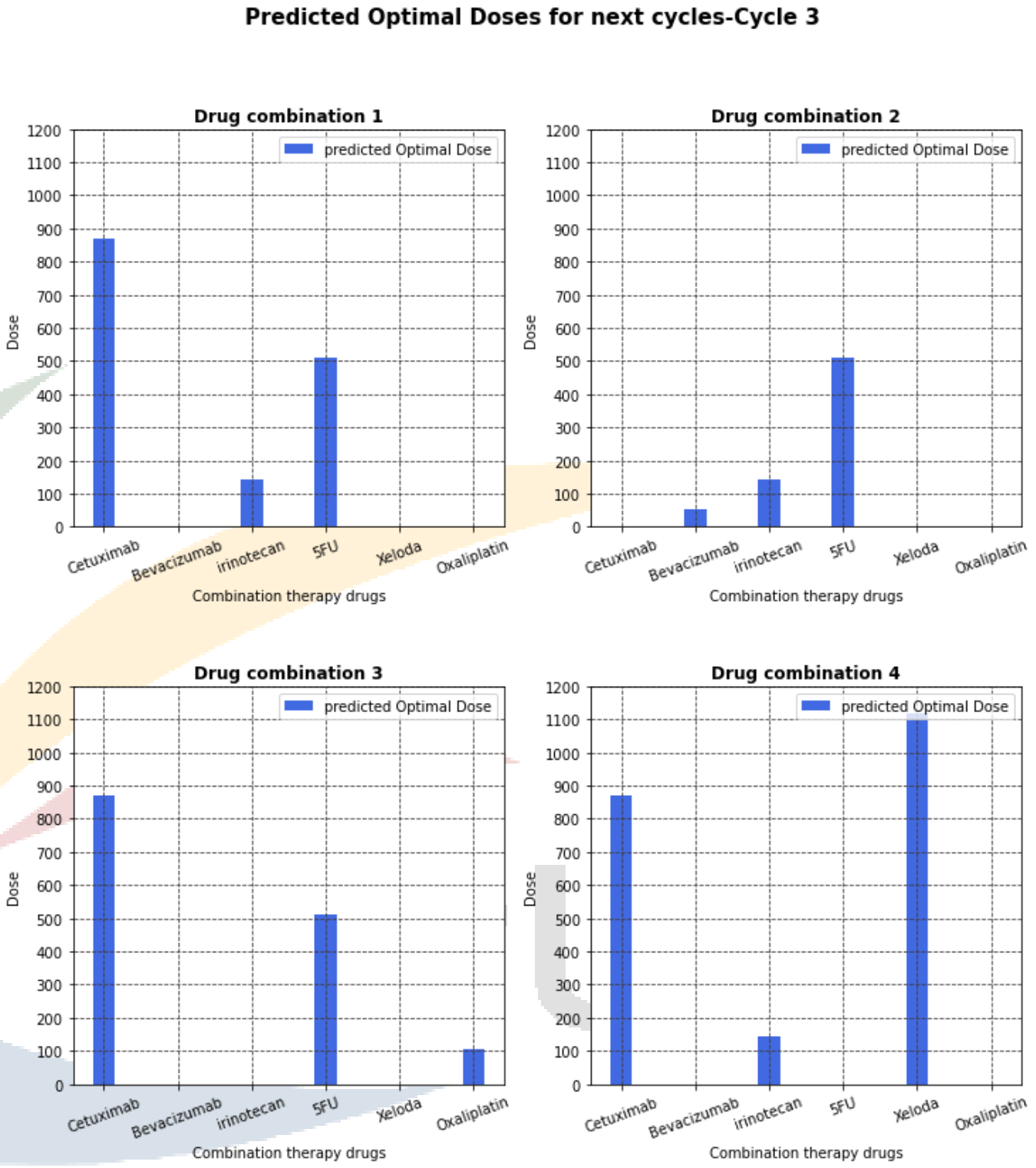

<br>Drug combination 1:<br>Predicted Cetuximab dose: 875 mg<br>Predicted Irinotecan dose: 145 mg<br>Predicted 5FU dose: 505 mg
<br>Drug combination 2:<br>Predicted Bevacizumab dose: 60 mg<br>Predicted Irinotecan dose: 145 mg<br>Predicted 5FU dose: 505 mg
<br>Drug combination 3:<br>Predicted Cetuximab dose: 875 mg<br>Predicted 5FU dose: 505 mg<br>Predicted Oxaliplatin dose: 102 mg
<br>Drug combination 4:<br>Predicted Cetuximab dose: 780 mg<br>Predicted Irinotecan dose: 145 mg<br>Predicted Xeloda dose: 1110 mg
</td></tr>
</table>

Description: During the third cycle of the patient No. 37, the oncologist has prescribed the combination chemotherapy drug schedule as Cetuximab: 800 mg, Irinotecan: 216 mg, 5FU: 480 mg, which is almost similar to the given dose at cycle 2. The predicted optimal dose schedule contains 4 combinations as shown above. The drug combination 1 is the combination chemotherapy drug schedule similar to the given drug schedule. When comparing the Drug combination 1 with the given dose it is visible that it is almost similar. The proposed Chemo RL model have suggested +9.375% Cetuximab dose (875-800/800), -32.8% Irinotecan dose (145-216/216) and +5.2% 5FU dose (505-480/480) which can be acceptable in the clinical sense.

TABLE 19. GIVEN DOSE COMPARISON WITH THE PREDICTED OPTIMAL DOSE FOR THE CYCLE 4: PATIENT NO. 37

| **Given doses for the cycle 4** | **Predicted optimal dose for the cycle 4** |
| --- | --- |
| 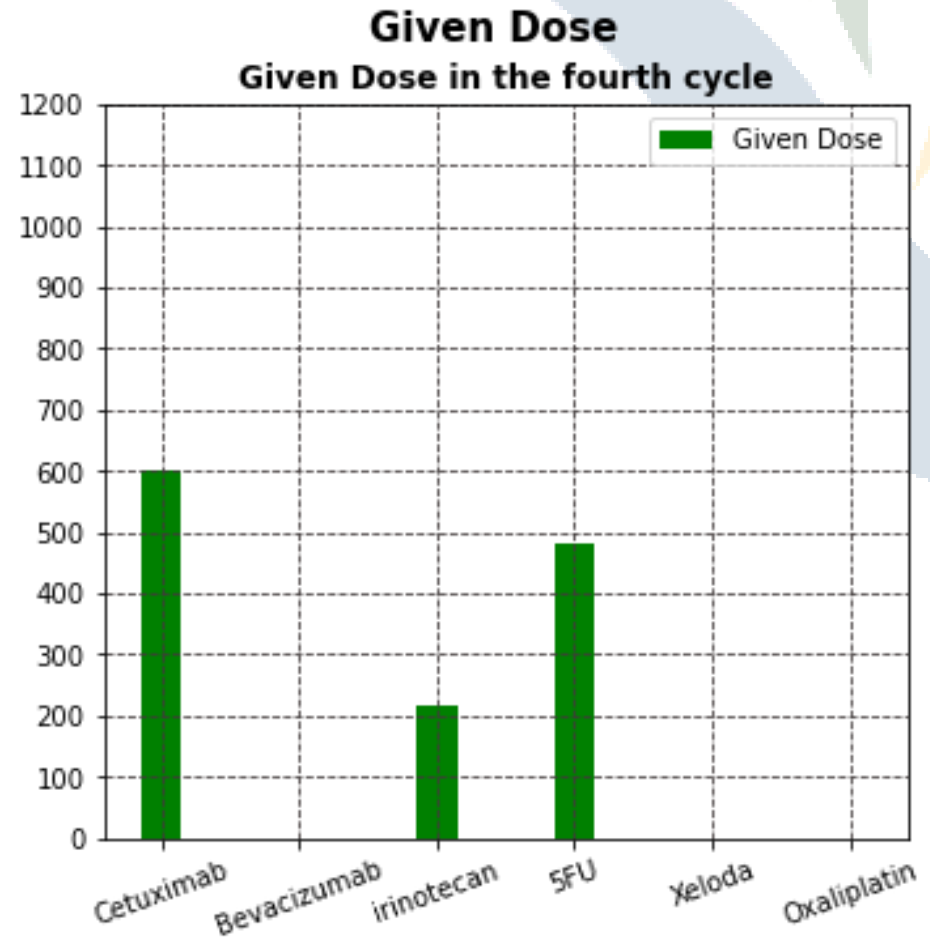<br><br>Given Cetuximab dose: 800 mg<br>Given Irinotecan dose: 216 mg<br>Given 5FU dose: 480 mg | 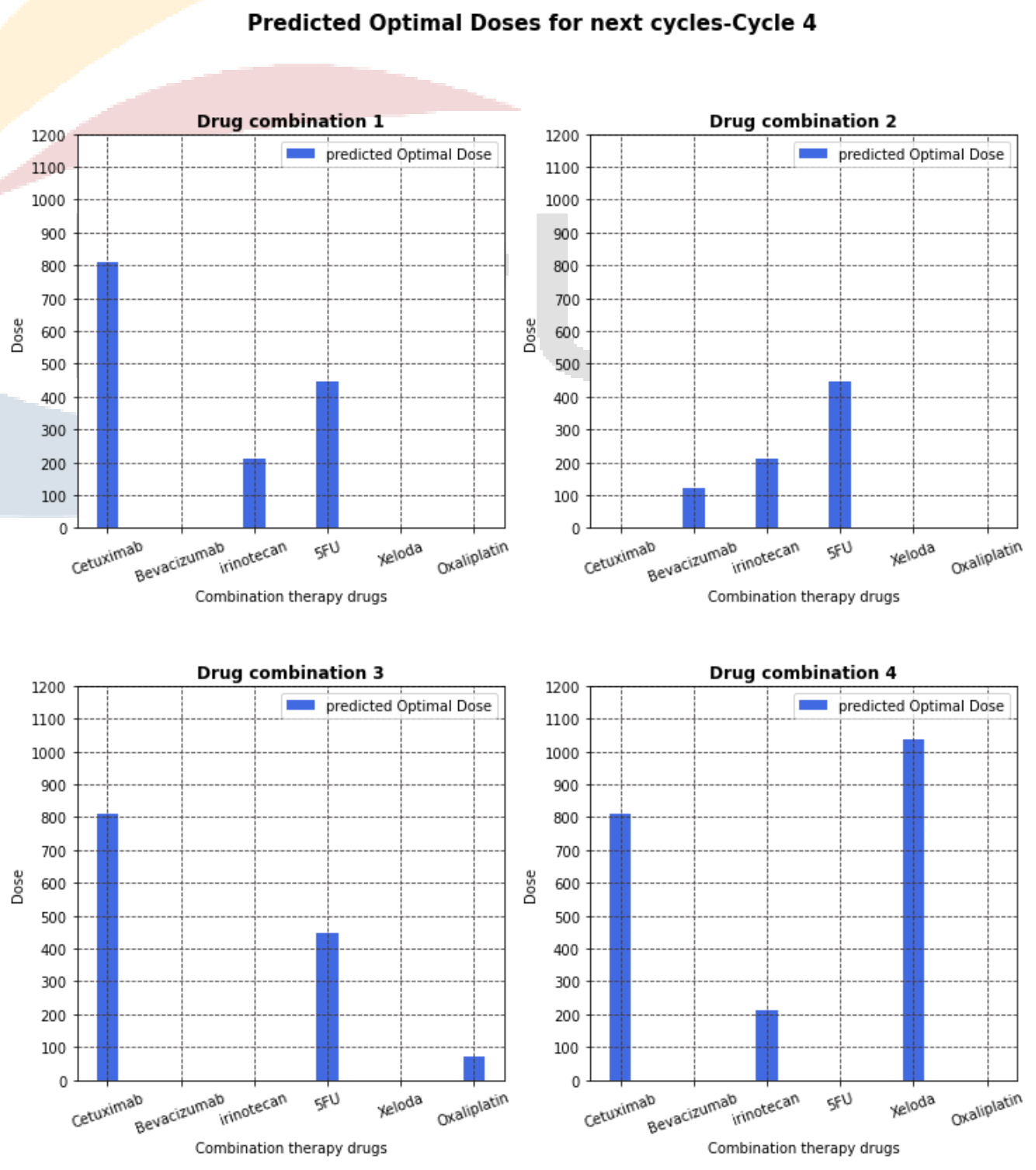<br> |
| | **Drug combination 1:**<br>Predicted Cetuximab dose: 805 mg<br>Predicted Irinotecan dose: 210 mg<br>Predicted 5FU dose: 450 mg |
| | **Drug combination 2:**<br>Predicted Bevacizumab dose: 110 mg<br>Predicted Irinotecan dose: 210 mg<br>Predicted 5FU dose: 450 mg |
| | **Drug combination 3:**<br>Predicted Cetuximab dose: 805 mg<br>Predicted 5FU dose: 450 mg<br>Predicted Oxaliplatin dose: 84 mg |
| | **Drug combination 4:**<br>Predicted Cetuximab dose: 805 mg<br>Predicted Irinotecan dose: 210 mg<br>Predicted Xeloda dose: 1030 mg |

Description: During the fourth cycle of the patient No. 37, the oncologist has prescribed the combination chemotherapy drug schedule as Cetuximab: 800 mg, Irinotecan: 216 mg, 5FU: 480 mg, which is almost similar to the given dose at cycle 3. The predicted optimal dose schedule contains 4 combinations as shown above. The drug combination 1 is the combination chemotherapy drug schedule similar to the given drug schedule. When comparing the Drug combination 1 with the given dose it is visible that it is almost similar. The proposed Chemo RL model have suggested +0.0625% Cetuximab dose (805-800/800), -2.78% Irinotecan dose (210-216/216) and -6.25% 5FU dose (450-480/480) which can be acceptable in the clinical sense.

TABLE 20. GIVEN DOSE COMPARISON WITH THE PREDICTED OPTIMAL DOSE FOR THE CYCLE 5: PATIENT NO. 37

| **Given doses for the cycle 5** | **Predicted optimal dose for the cycle 5** |
|---|---|
| | |

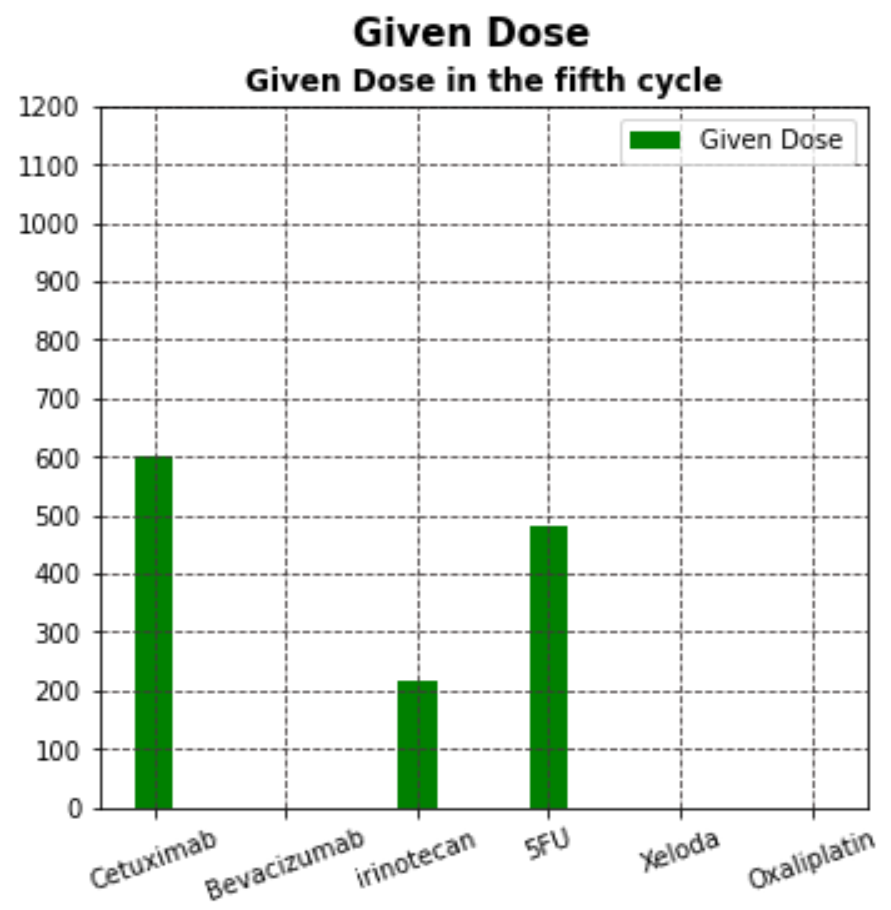

Given Cetuximab dose: 800 mg

Given Irinotecan dose: 216 mg

Given 5FU dose: 480 mg

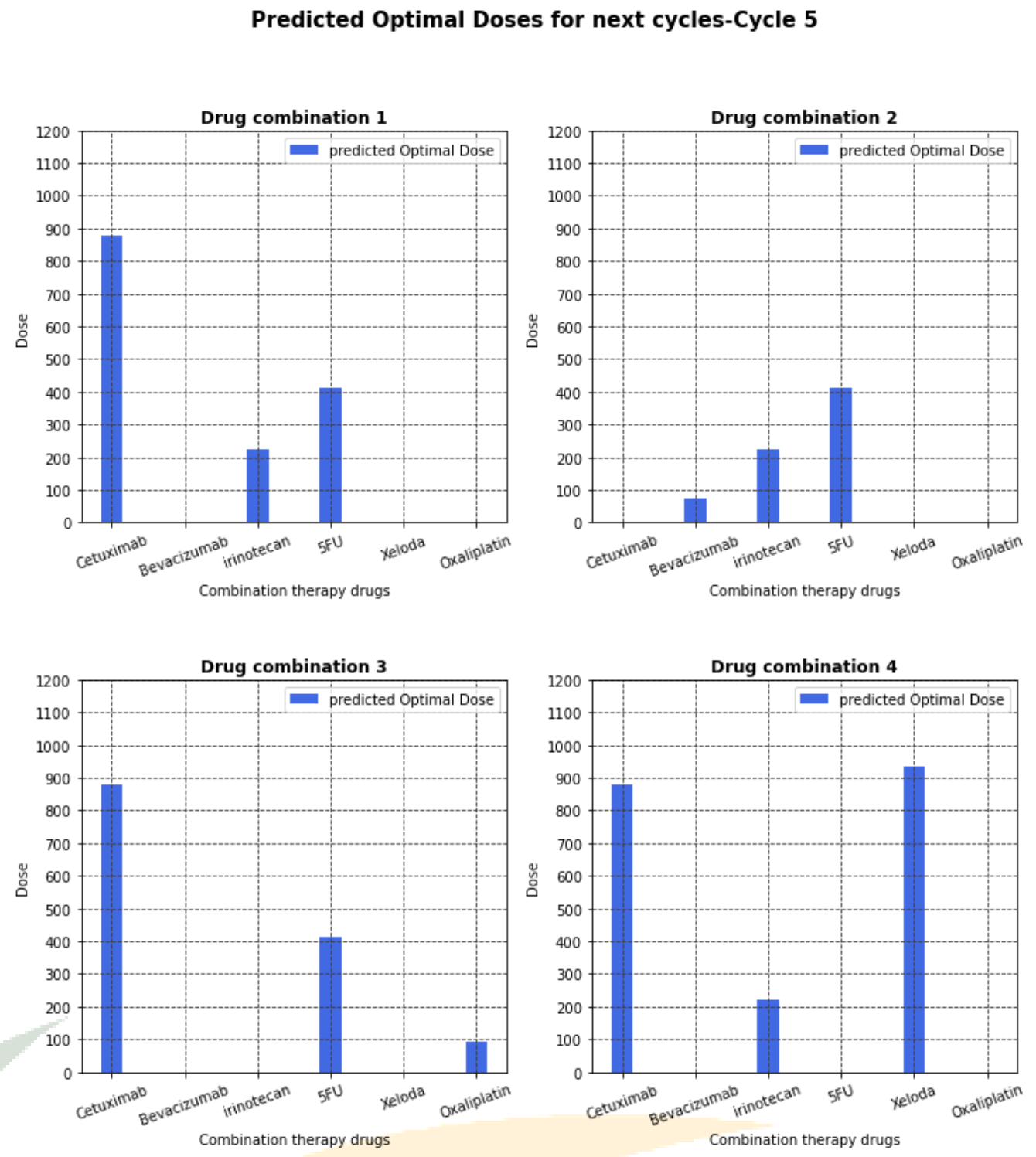

| | |
|---|---|
| **Drug combination 1:**<br>Predicted Cetuximab dose: 875 mg<br>Predicted Irinotecan dose: 210 mg<br>Predicted 5FU dose: 410 mg | **Drug combination 2:**<br>Predicted Bevacizumab dose: 80 mg<br>Predicted Irinotecan dose: 210 mg<br>Predicted 5FU dose: 410 mg |
| **Drug combination 3:**<br>Predicted Cetuximab dose: 885 mg<br>Predicted 5FU dose: 410 mg<br>Predicted Oxaliplatin dose: 90 mg | **Drug combination 4:**<br>Predicted Cetuximab dose: 885 mg<br>Predicted Irinotecan dose: 210 mg<br>Predicted Xeloda dose: 940 mg |

Description: During the fifth cycle of the patient No. 37, the oncologist has prescribed the combination chemotherapy drug schedule as Cetuximab: 800 mg, Irinotecan: 216 mg, 5FU: 480 mg, which is almost similar to the given dose at cycle 3. The predicted optimal dose schedule contains 4 combinations as shown above. The drug combination 1 is the combination chemotherapy drug schedule similar to the given drug schedule. When comparing the Drug combination 1 with the given dose it is visible that it is almost similar.

The proposed Chemo RL model have suggested +0.011% Cetuximab dose (875-800/800), -2.78% Irinotecan dose (210-216/216) and -14.5% 5FU dose (410-480/480) which can be acceptable in the clinical sense.

TABLE 21. GIVEN DOSE COMPARISON WITH THE PREDICTED OPTIMAL DOSE FOR THE CYCLE 6: PATIENT NO. 37

| Given doses for the cycle 6 | Predicted optimal dose for the cycle 6 |
|---|---|
| 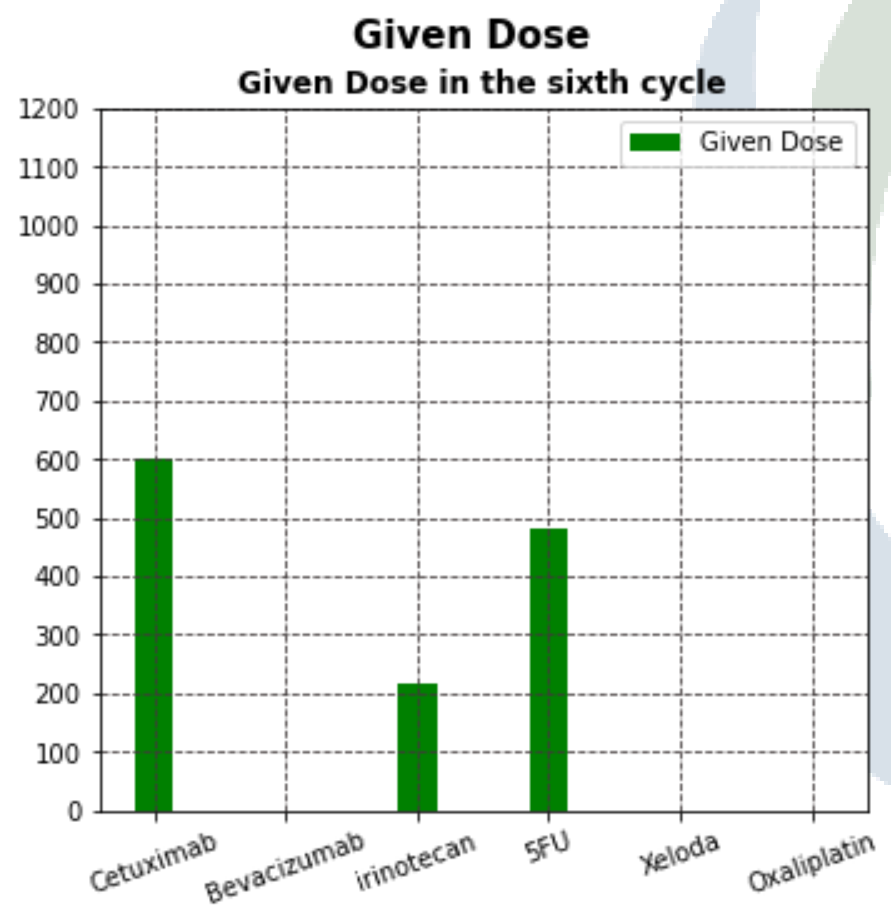<br>Given Cetuximab dose: 800 mg<br>Given Irinotecan dose: 216 mg<br>Given 5FU dose: 480 mg | 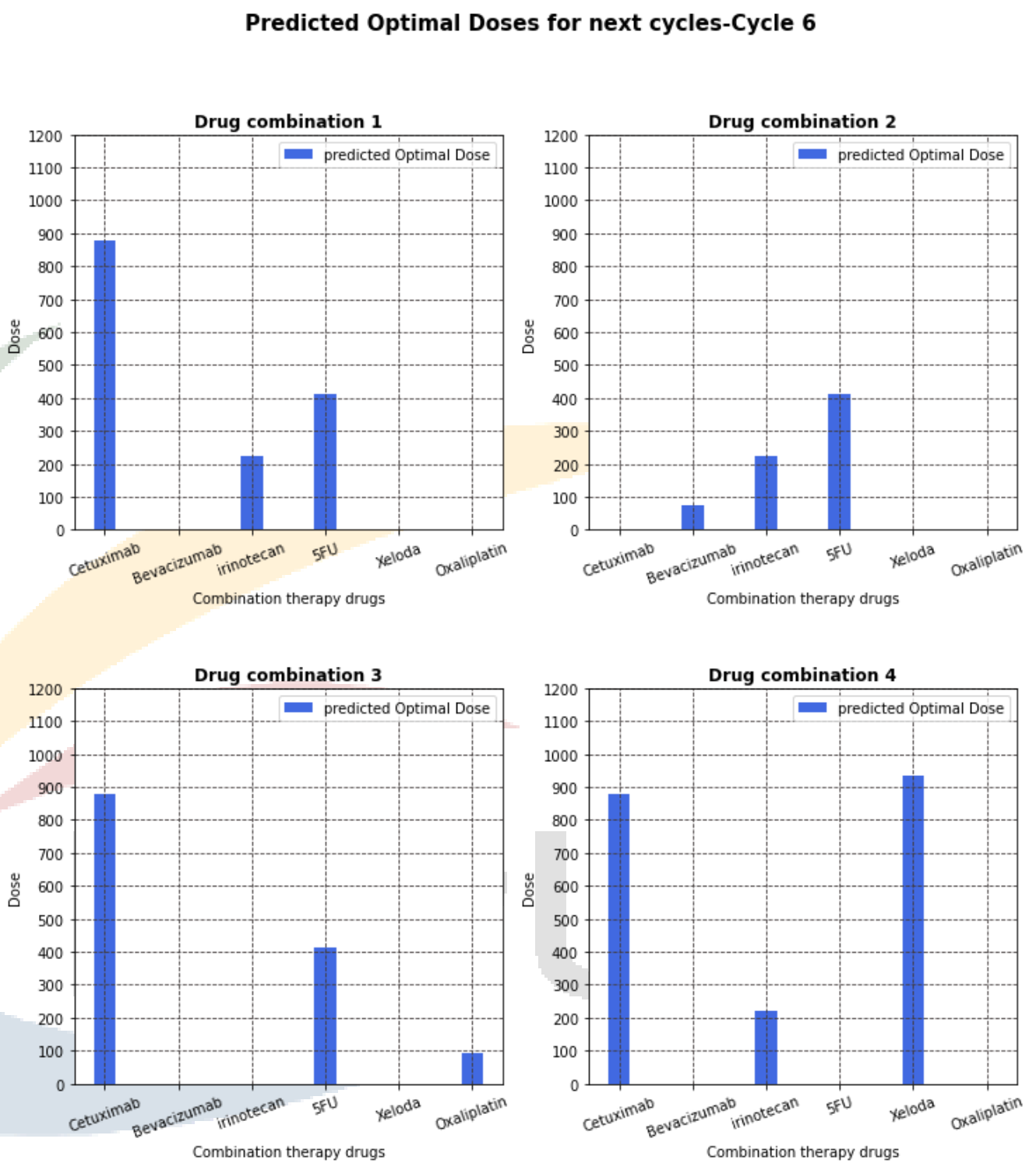<br>**Drug combination 1:**<br>Predicted Cetuximab dose: 875 mg<br>Predicted Irinotecan dose: 210 mg<br>Predicted 5FU dose: 410 mg<br>**Drug combination 2:**<br>Predicted Bevacizumab dose: 80 mg<br>Predicted Irinotecan dose: 210 mg<br>Predicted 5FU dose: 410 mg<br>**Drug combination 3:**<br>Predicted Cetuximab dose: 888 mg<br>Predicted 5FU dose: 410 mg<br>Predicted Oxaliplatin dose: 90 mg<br>**Drug combination 4:**<br>Predicted Cetuximab dose: 888 mg<br>Predicted Irinotecan dose: 210 mg<br>Predicted Xeloda dose: 940 mg |

Description: During the sixth cycle of the patient No. 37, the oncologist has prescribed the combination chemotherapy drug schedule as Cetuximab: 800 mg, Irinotecan: 216 mg, 5FU: 480 mg, which is almost similar to the given dose at cycle 3. The predicted optimal dose schedule contains 4 combinations as shown above. The drug combination 1 is the combination chemotherapy drug schedule similar to the given drug schedule. When comparing the Drug combination 1 with the given dose it is visible that it is almost similar. The proposed Chemo RL model have suggested +0.011% Cetuximab dose (875-800/800), -2.78% Irinotecan dose (210-216/216) and -14.5% 5FU dose (410-480/480) which can be acceptable in the clinical sense.

Apart from the above explanations, it is seen that the drug dose values of consecutive cycles of all the 4 drug combinations behaves in the same trend or sometimes with little negligible variations. In the real clinical setting, oncologist will not make major deviations in the prescribing drug doses for consecutive cycles. Thus, the drug dosing pattern suggested by our Chemo RL model is more compatible with the real clinical practices.

Following the above mentioned same process as we tested for the Patient No. 37, we also predicted the optimal drug policy for the other 6 patients in the 7 test patients set. And, it was concluded that the consecutive cycles predict the same or even little deviating values for the drug doses for 6 cycles of all the individual patients. More test cases done with changing the training size when running the model are mentioned in the 1.4 section in this chapter.

#### *7.4.3.2. Optimal policy via Exploration results*

As described in the implementation section, secondly we used the trained model to generate the optimal policies by searching for new patient's unseen clinical observations. At this initial stage, we created two virtual patient instances as patient No.41 and patient No.42. Following table shows the patient No.41 virtual clinical observations.

TABLE 22. PATIENT NO. 41 VIRTUAL CLINICAL OBSERVATIONS

| **Virtual clinical observations** | **value** |
|---|---|
| Tumor size | 1.08183361 $cm^3$ |
| Age | 60 years |
| Gender | Male |
| Weight | 51.2 kg |
| Height | 160 |
| Side effects index | 3 |
| Drug Response | Baseline |
| CCI index | 8 |
| ECOG PS measure | 0 |

We approached the two stages as mentioned in the implementation section. Following are the results achieved for the two stages.

**Stage 1: Predicting the real patient status and dose amount for cycle 1.**

| **Predicted real patient state (MBSE)** | **10** |
|---|---|

TABLE 23. PREDICTED OPTIMAL DOSE FOR CYCLE 1 FOR VIRTUAL CLINICAL OBSERVATIONS: PATIENT NO. 41

<table>
<tr><th>New patient’s virtual clinical observations</th><th>Predicted optimal dose for the cycle 1</th></tr>
<tr><td>
Patient's observations<br><br>
Tumor size: 1.08183361 cm3<br>
Age: 60 years<br>
Gender: Male<br>
Weight: 51.2 kg<br>
Height: 160 cm<br>
Side effects index: 3<br>
Drug response: Baseline<br>
CCI Index: 8<br>
ECOG performance status: 0<br><br>
Predicted real patient status: 10
</td><td>
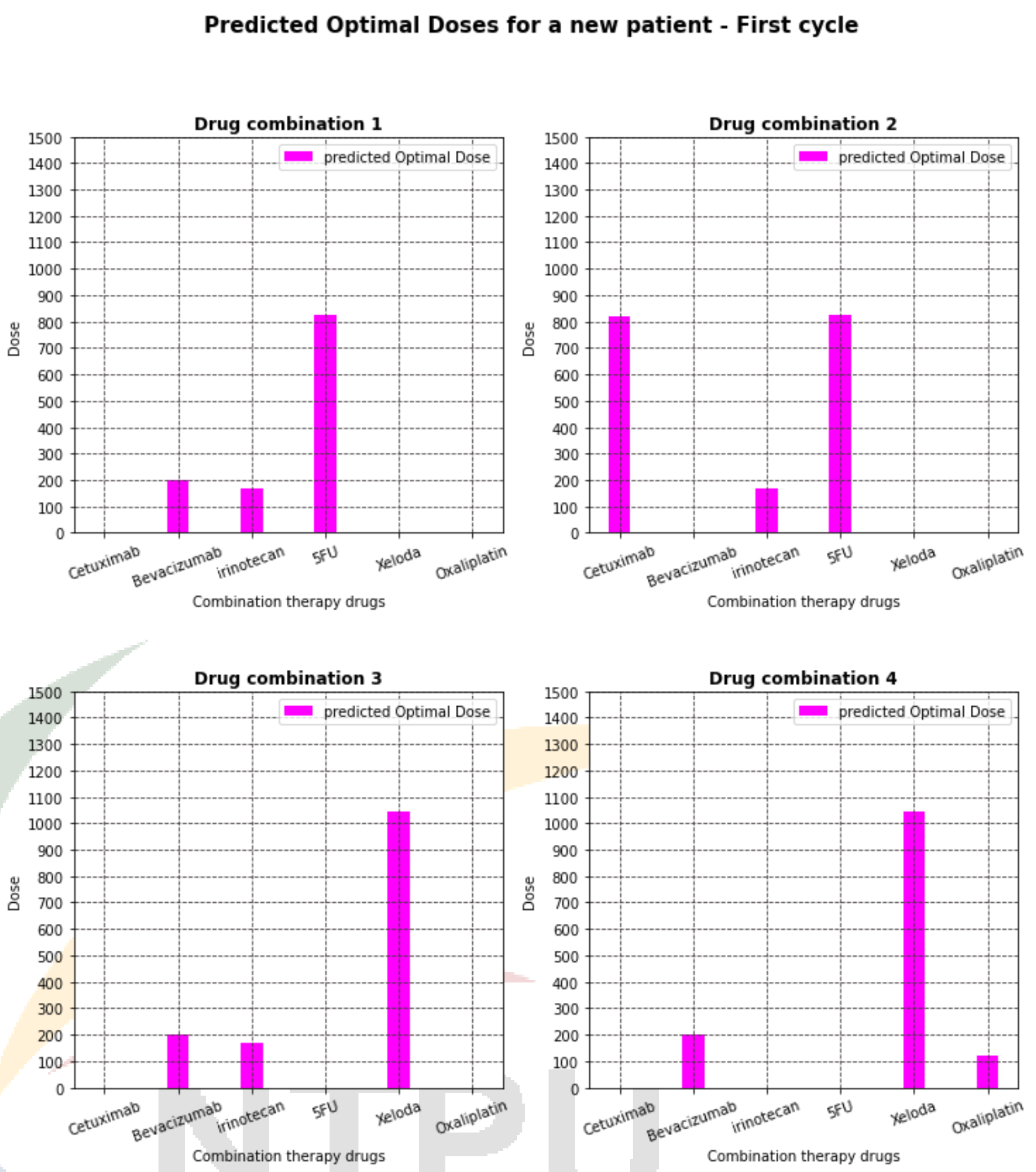

<br>
Drug combination 1:<br>
Predicted Bevacizumab dose: 200 mg<br>
Predicted Irinotecan dose: 180 mg<br>
Predicted 5FU dose: 810 mg<br><br>
Drug combination 2:<br>
Predicted Cetuximab dose: 808 mg<br>
Predicted Irinotecan dose: 180 mg<br>
Predicted 5FU dose: 810 mg<br><br>
Drug combination 3:<br>
Predicted Bevacizumab dose:888mg<br>
Predicted 5FU dose: 180 mg<br>
Predicted Xeloda dose: 1050 mg<br><br>
Drug combination 4:<br>
Predicted Cetuximab dose: 200 mg<br>
Predicted Xeloda dose: 1050 mg<br>
Predicted Oxaliplatin dose: 120 mg
</td></tr>
</table>

Description: When a new patient with the above clinical diagnosis measures enters, the Chemo RL model predicts his real patient status as 10 and the above mentioning drug combinations for his cycle 1 drug plan.

**Stage 2: Predicting the RDI for cycle 2-6.**

TABLE 24. PREDICTED RDI FOR CYCLE 2: PATIENT NO. 41

**Predicted optimal RDI for the cycle 2**

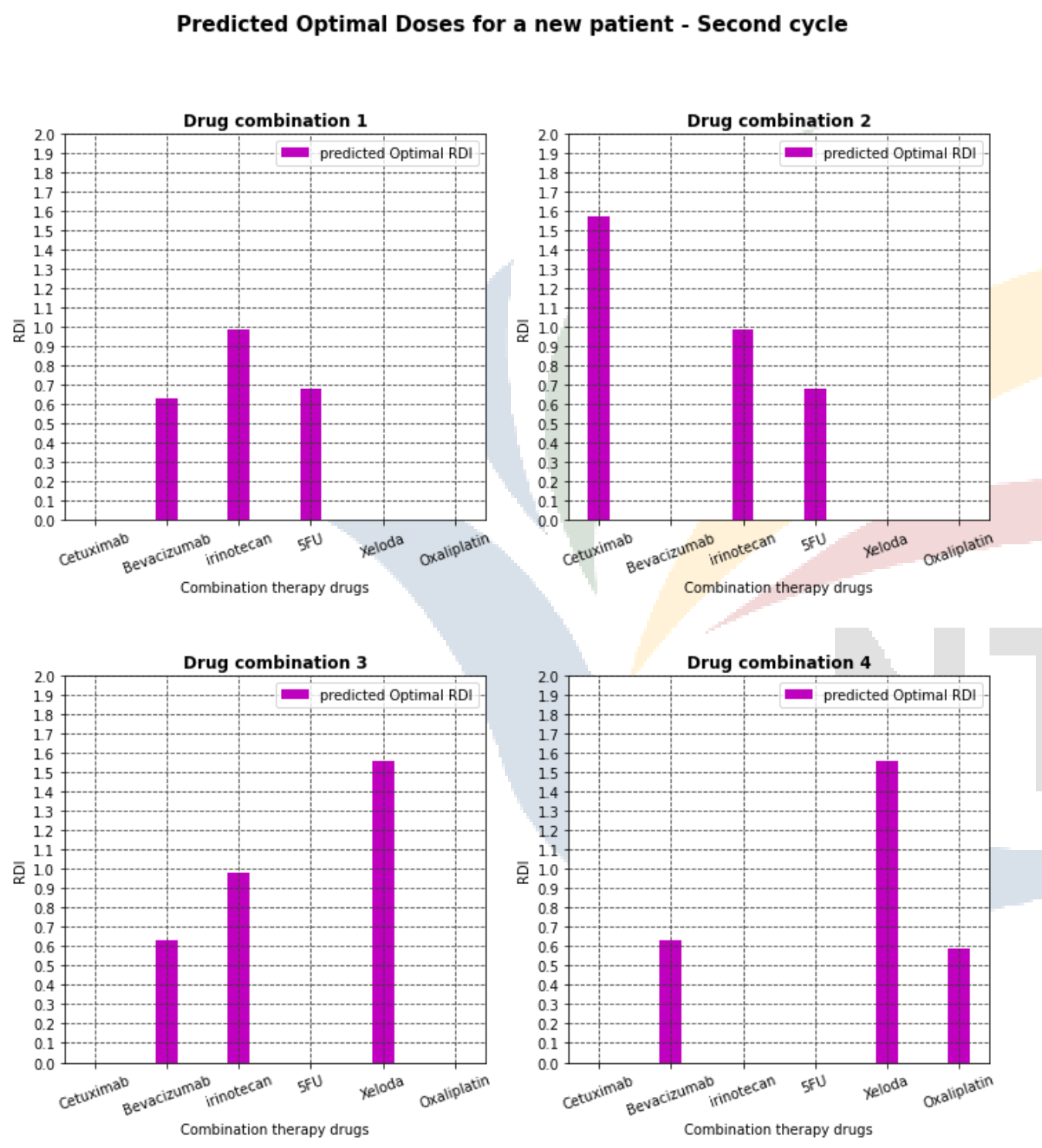

**Drug combination 1:**
Predicted RDI for Bevacizumab: 0.625
Predicted RDI for Irinotecan: 0.980
Predicted RDI for 5FU: 0.680

**Drug combination 2:**
Predicted Cetuximab dose: 1.587
Predicted Irinotecan dose: 0.990
Predicted 5FU dose: 0.680

**Drug combination 3:**
Predicted Bevacizumab dose: 0.630
Predicted 5FU dose: 0.988
Predicted Xeloda dose: 1.556

**Drug combination 4:**
Predicted Cetuximab dose: 0.634
Predicted Xeloda dose: 1.556
Predicted Oxaliplatin dose: 0.590

Description: The Chemo RL model predict the above four combinations of RDI for the cycle 2.

TABLE 25. PREDICTED RDI FOR CYCLE 3: PATIENT NO. 41

**Predicted optimal RDI for the cycle 3**

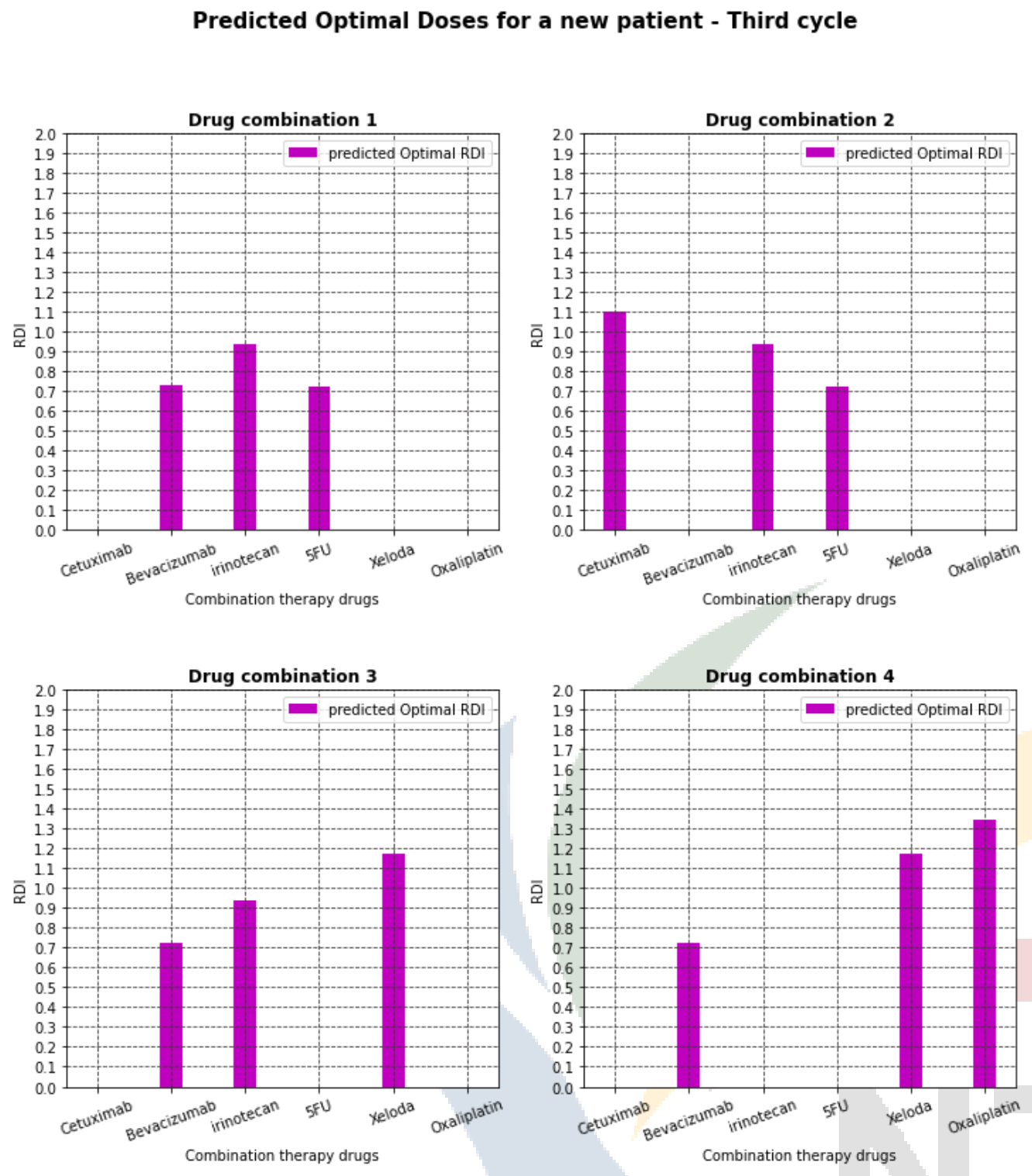

**Drug combination 1:**
Predicted RDI for Bevacizumab: 0.730
Predicted RDI for Irinotecan: 0.940
Predicted RDI for 5FU: 0.735

**Drug combination 2:**
Predicted Cetuximab dose: 1.100
Predicted Irinotecan dose: 0.946
Predicted 5FU dose: 0.715

**Drug combination 3:**
Predicted Bevacizumab dose: 0.718
Predicted 5FU dose: 0.948
Predicted Xeloda dose: 1.180

**Drug combination 4:**
Predicted Cetuximab dose: 0.714
Predicted Xeloda dose: 1.187
Predicted Oxaliplatin dose: 1.375

Description: The Chemo RL model predict the above four combinations of RDI for the cycle 3.

TABLE 26. PREDICTED RDI FOR CYCLE 4: PATIENT NO. 41

| **Predicted optimal RDI for the cycle 4** | |
|---|---|
| 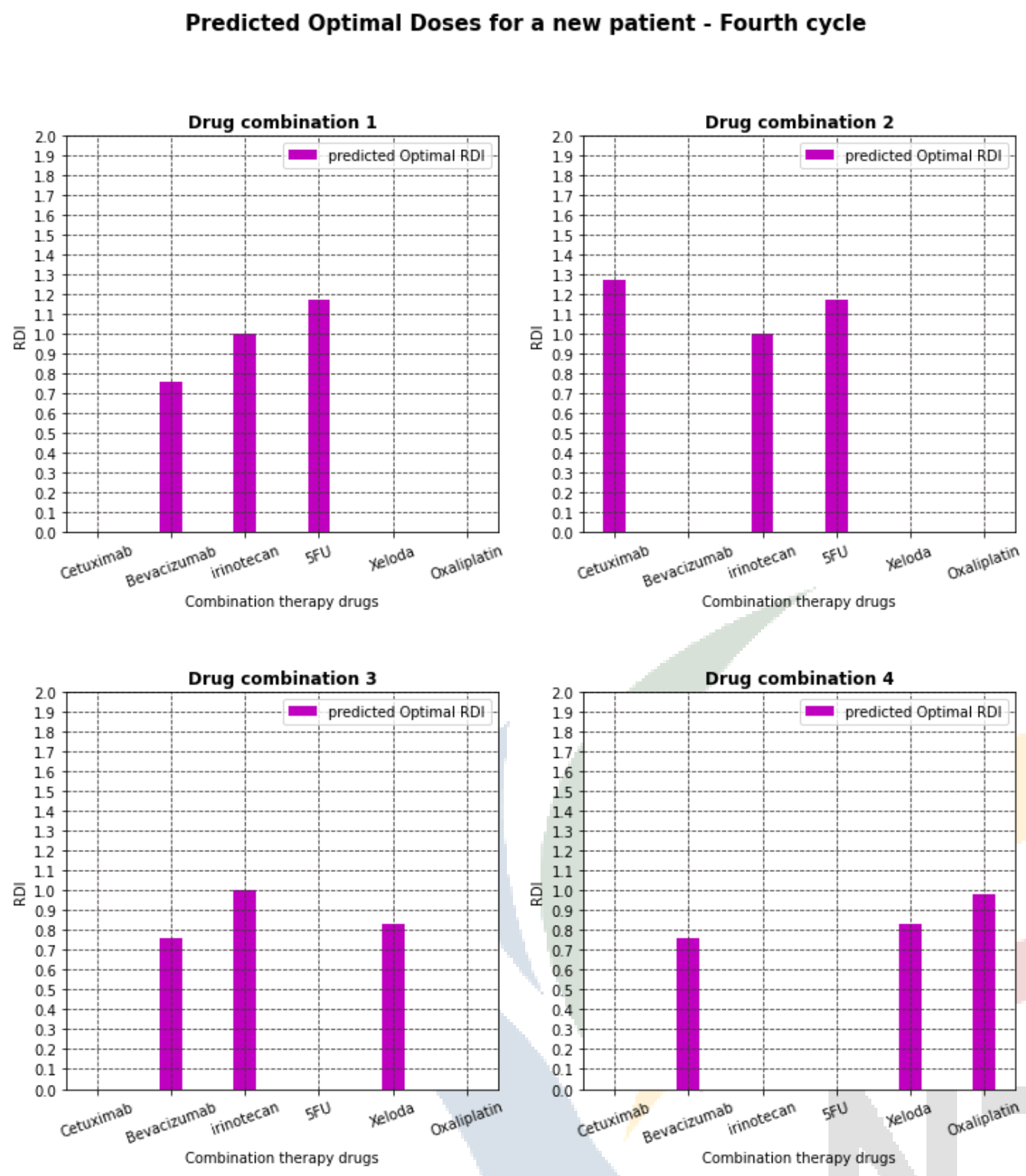 | **Drug combination 1:**<br>Predicted RDI for Bevacizumab: 0.761<br>Predicted RDI for Irinotecan: 1.0<br>Predicted RDI for 5FU: 1.189<br><br>**Drug combination 2:**<br>Predicted Cetuximab dose: 1.278<br>Predicted Irinotecan dose: 1.0<br>Predicted 5FU dose: 1.180<br><br>**Drug combination 3:**<br>Predicted Bevacizumab dose: 0.753<br>Predicted 5FU dose: 1.0<br>Predicted Xeloda dose: 0.834<br><br>**Drug combination 4:**<br>Predicted Cetuximab dose: 0.750<br>Predicted Xeloda dose: 0.827<br>Predicted Oxaliplatin dose: 0.984 |

Description: The Chemo RL model predict the above four combinations of RDI for the cycle 4.

TABLE 27. PREDICTED RDI FOR CYCLE 5: PATIENT NO. 41

**Predicted optimal RDI for the cycle 5**

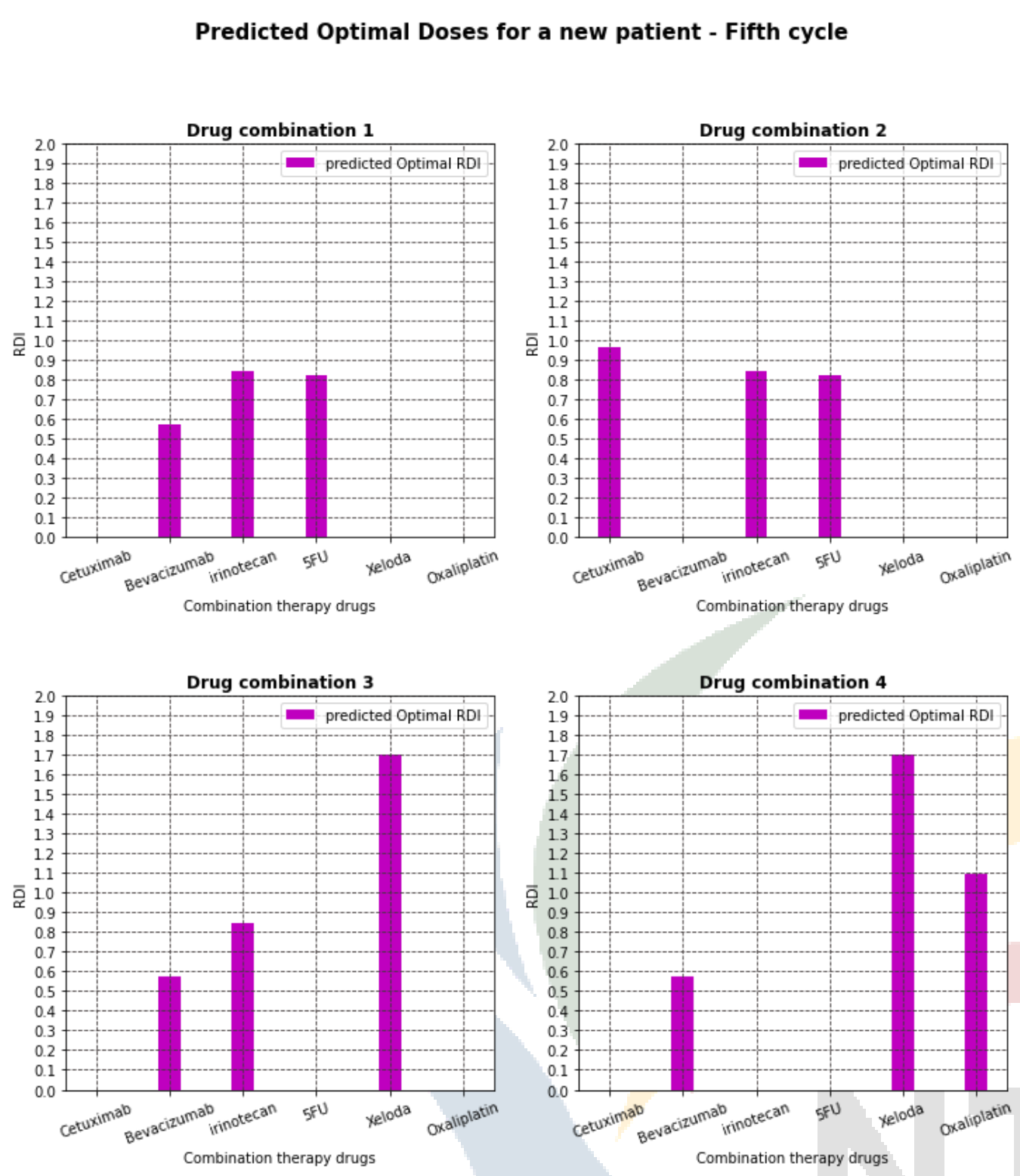

**Drug combination 1:**
Predicted RDI for Bevacizumab: 0.579
Predicted RDI for Irinotecan: 0.835
Predicted RDI for 5FU: 0.818

**Drug combination 2:**
Predicted Cetuximab dose: 0.967
Predicted Irinotecan dose: 0.854
Predicted 5FU dose: 0.838

**Drug combination 3:**
Predicted Bevacizumab dose: 0.576
Predicted 5FU dose: 0.845
Predicted Xeloda dose: 1.700

**Drug combination 4:**
Predicted Cetuximab dose: 0.576
Predicted Xeloda dose: 1.7
Predicted Oxaliplatin dose: 1.1

Description: The Chemo RL model predict the above four combinations of RDI for the cycle 5.

TABLE 28. PREDICTED RDI FOR CYCLE 6: PATIENT NO. 41

| Predicted optimal RDI for the cycle 6 | |
|---|---|
| 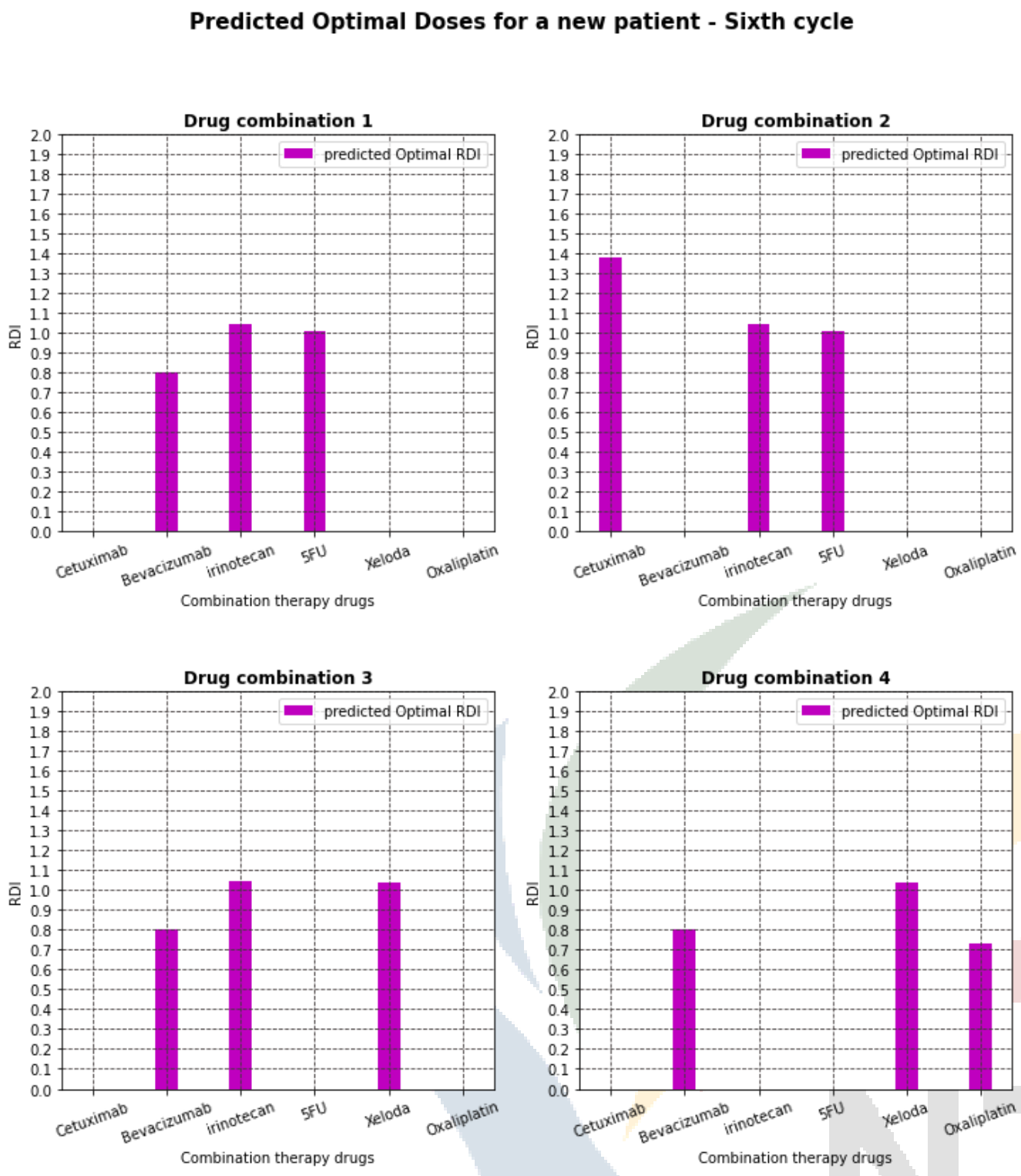  | **Drug combination 1:**<br>Predicted RDI for Bevacizumab: 0.8<br>Predicted RDI for Irinotecan: 1.051<br>Predicted RDI for 5FU: 1.0<br><br>**Drug combination 2:**<br>Predicted Cetuximab dose: 1.380<br>Predicted Irinotecan dose: 1.054<br>Predicted 5FU dose: 1.0<br><br>**Drug combination 3:**<br>Predicted Bevacizumab dose: 0.8<br>Predicted 5FU dose: 1.06<br>Predicted Xeloda dose: 1.04<br><br>**Drug combination 4:**<br>Predicted Cetuximab dose: 0.8<br>Predicted Xeloda dose: 1.045<br>Predicted Oxaliplatin dose: 0.735 |

Description: The Chemo RL model predict the above four combinations of RDI for the cycle 6.

From the above visual explanation, it is clear that the drug dose values of consecutive cycles of all the 4 drug combinations behaves in the same trend or sometimes with little variations. The dosing amount in mg can be obtained with the equations numbered 14-19, stated in the Design chapter under 5.4.5 section. Thus, the drug-dosing pattern suggested by our Chemo RL model is compatible with the real clinical practices when considering the new patients also.

We then conducted a validation approach to check whether the predicting results under the exploration strategy is correct. Thus, we evaluated the stage 1 mentioned above with some of the existing patients in the dataset, assuming they are new patients. We predicted the optimal dosing schedule for the cycle 1 and compared the already available given drug dose for the patient. Following table 29 shows the evaluation done with the patient No.13 assuming the patient is a new patient.

TABLE 29. GIVEN DOSE COMPARISON WITH THE PREDICTED OPTIMAL DOSE FOR THE CYCLE 1: PATIENT NO. 13

<table>
<tr><th>Given doses for the cycle 1</th><th>Predicted optimal dose for the cycle 1</th></tr>
<tr><td>
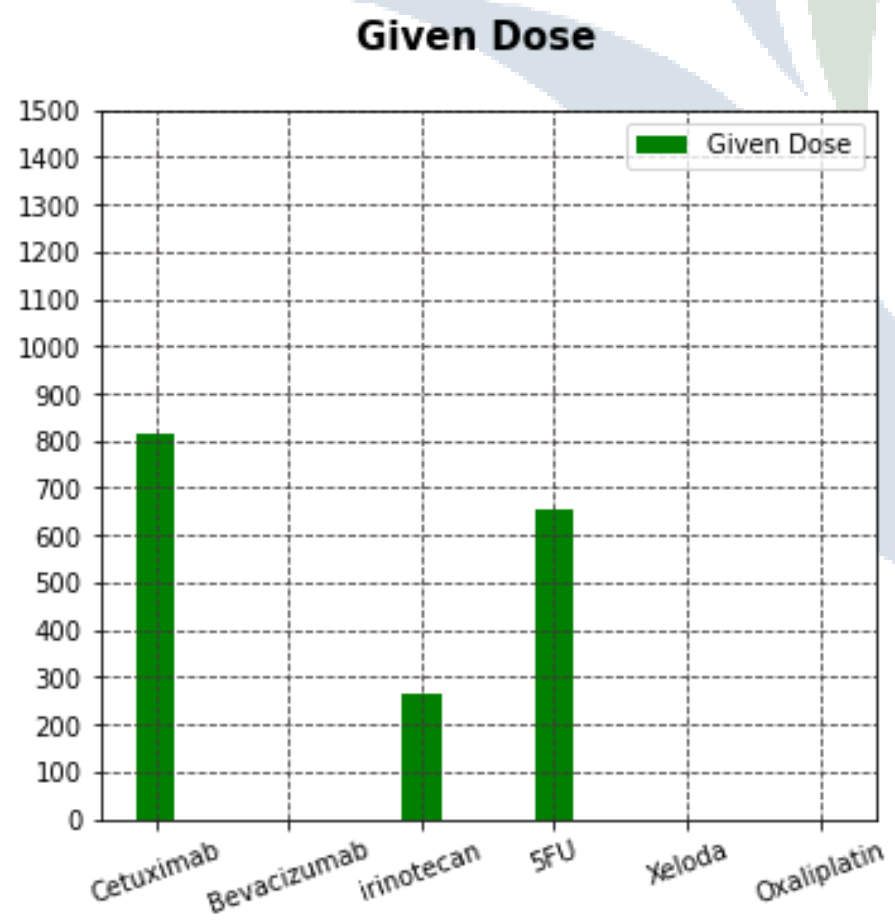

<br>Given Cetuximab dose: 810 mg<br>Given Irinotecan dose: 216 mg<br>Given 5FU dose: 640 mg
</td><td>
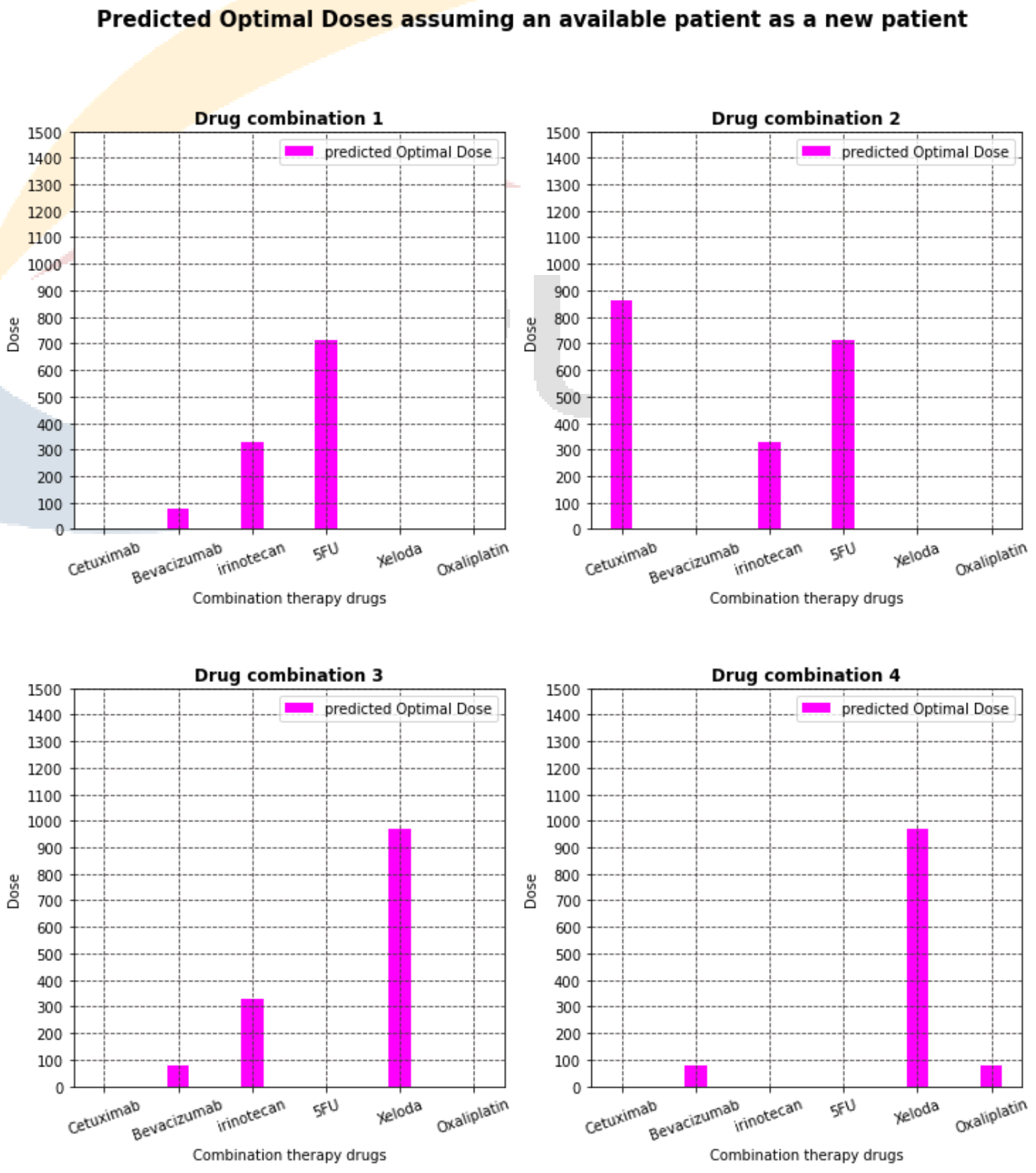

</td></tr>
</table>

| | |
|---|---|
| | **Drug combination 1:**<br>Predicted Bevacizumab dose:90 mg<br>Predicted Irinotecan dose: 320 mg<br>Predicted 5FU dose: 710 mg<br><br>**Drug combination 2:**<br>Predicted Cetuximab dose: 875 mg<br>Predicted Irinotecan dose: 320 mg<br>Predicted 5FU dose: 710 mg<br><br>**Drug combination 3:**<br>Predicted Bevacizumab dose:90 mg<br>Predicted Irinotecan dose: 320 mg<br>Predicted Xeloda dose: 980 mg<br><br>**Drug combination 4:**<br>Predicted Bevacizumab dose:90 mg<br>Predicted Xeloda dose: 980 mg<br>Predicted Oxaliplatin dose: 90 mg |

Description: During the first cycle of the patient No. 13, the oncologist has prescribed the combination chemotherapy drug schedule as Cetuximab: 810 mg, Irinotecan: 216 mg, 5FU: 640 mg. The predicted optimal dose schedule contains 4 combinations as shown above. The drug combination 1 is the combination chemotherapy drug schedule similar to the given drug schedule. When comparing the Drug combination 2 with the given dose it is visible that it is almost similar. The proposed Chemo RL model have suggested +0.011% Cetuximab dose (875-800/800), +48.15% Irinotecan dose (320-216/216) and +10.9% 5FU dose (710-640/640) which can be acceptable in the clinical sense.

### *7.4.4. Visualization of the reward function*

#### *7.4.4.1. Reward via Exploitation results*

Then, we plotted how the reward of the predicted optimal drug dose and the reward of the given drug dose behaves for the patients available in the dataset. In this following figure 47 and 48, the red line shows the rewards for separate cycles which the RL agent is predicting for the optimal policy. And the green line shows the actual reward for separate cycles based in the reward function we generated. So, as shown in the figure, the RL algorithm shows a higher reward in each cycle and higher cumulative reward, for the optimal policy compared to the actual dose schedule given to the patient.

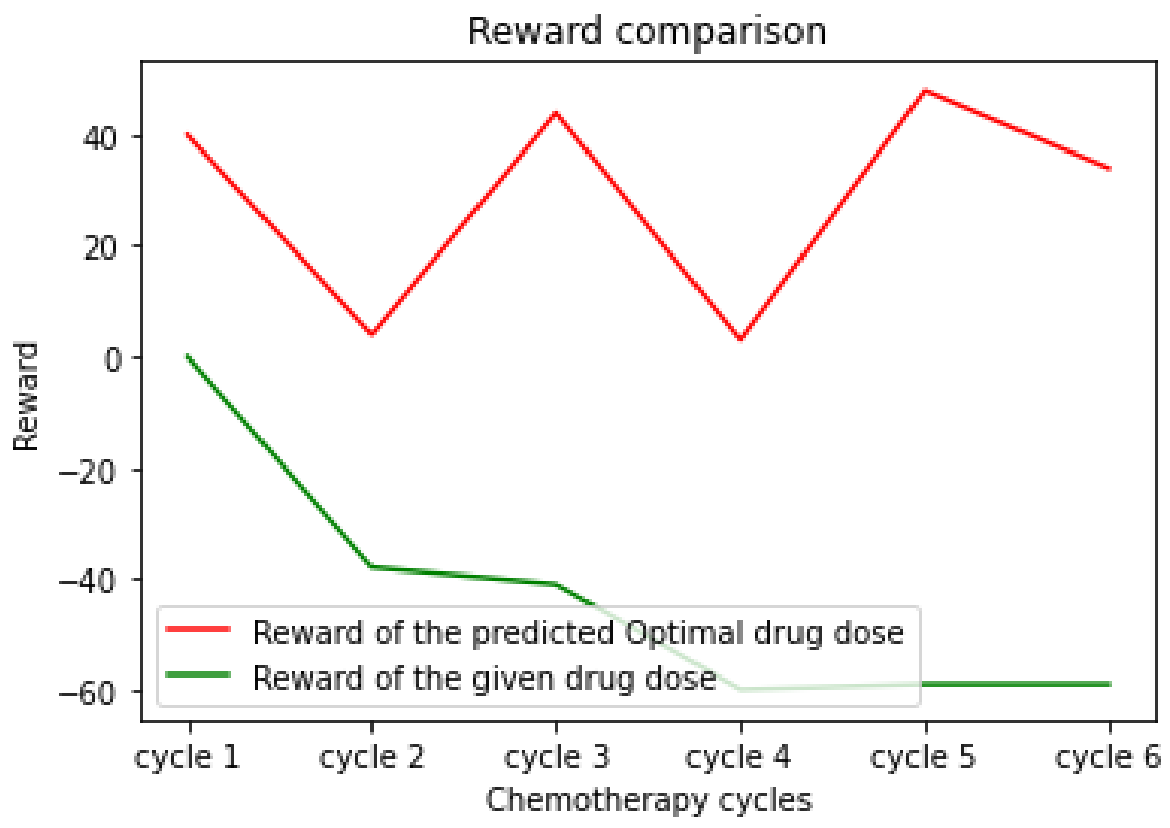

Figure 47: Reward comparison for the 6 cycles: patient No. 37

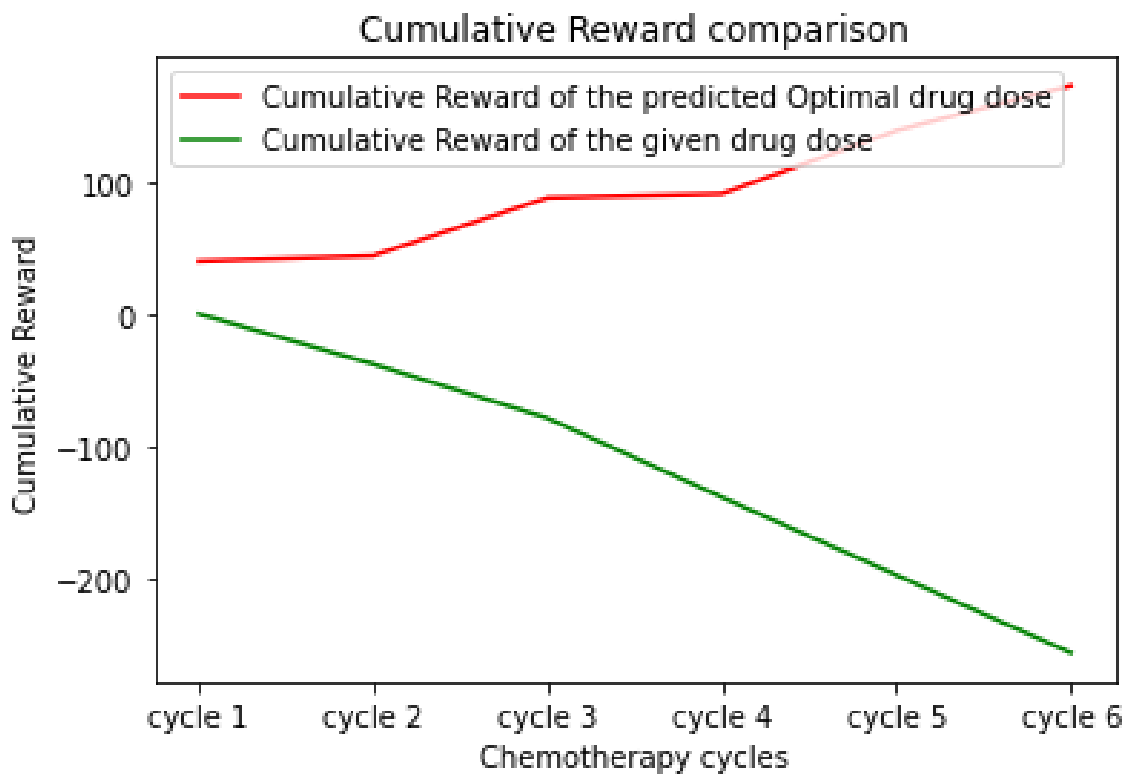

Figure 47: Cumulative Reward comparison for the 6 cycles: patient No. 37

#### *7.4.4.2. Reward via Exploration results*

Also, we plotted how the reward of the predicted optimal drug dose and the reward of the given drug dose behaves when a new patient enters. We checked this behavior by assuming the new patient is already available in the dataset. Thereby we compared the reward for the predicted optimal drug and the reward actually given. In this following figure 49, the red line shows the rewards for separate cycles which the RL agent is predicting for the optimal policy and the green line shows the actual reward for separate cycles based on the reward function we generated. So, the RL algorithm shows a higher reward for one cycle ahead, for the optimal policy compared to the actual dose given to the patient in the first cycle.

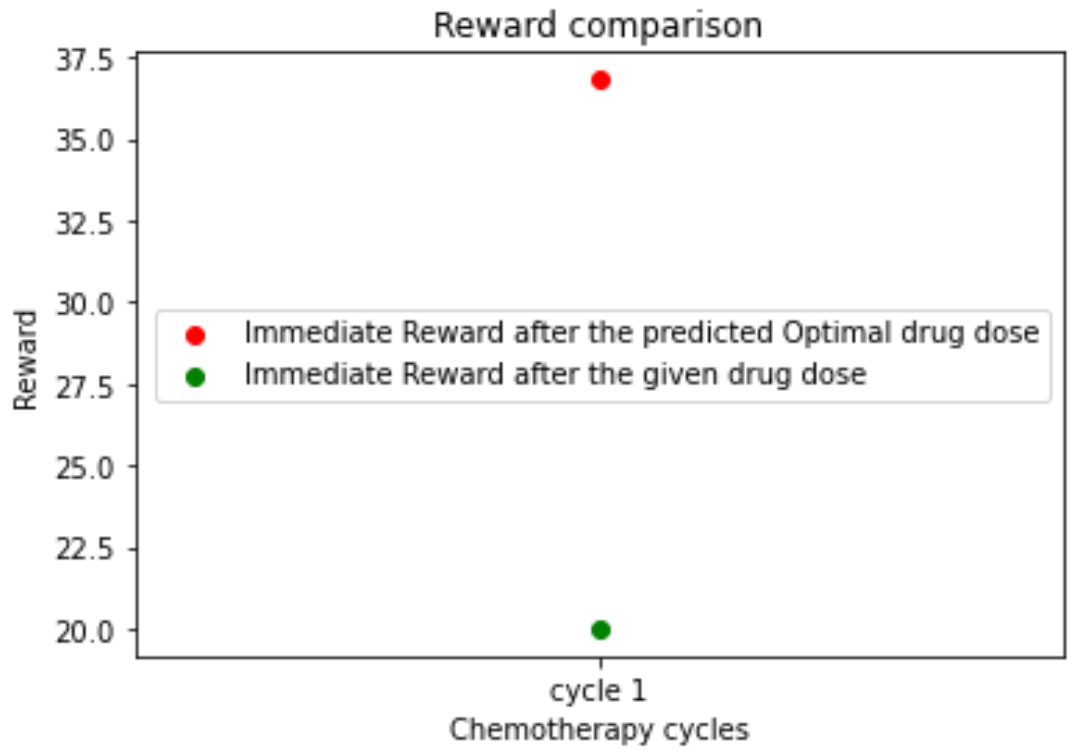

Figure 48: Reward comparison for one cycle ahead: patient No. 37

### 7.5. Behavior of the prediction error with the training size

When training the Chemo RL model, we followed a special approach by training separate models by increasing the number of training samples, which is the training patient set. Five separate models were trained by changing the number of training patients for these separate models as 5 patients, 10 patients, 15 patients, 20 patients and finally 27 patients. For each of the separate models, the testing patient set, which was mentioned in the 7.4.3.1 section, was used to calculate the dose change between the given dose by the oncologist and the optimal predicted dose through the Chemo RL model.

The results acquired with the trained model with 27 patients are as in the below table.

TABLE 30. TESTING RESULTS FOR THE SIX DRUGS WITH THE 27 PATIENTS MODEL

| **Patient No.** | **Mean Cetuximab dose change per patient** | **Mean Bevacizumab dose change per patient** | **Mean Irinotecan dose change per patient** | **Mean 5FU dose change per patient** | **Mean Xeloda dose change per patient** | **Mean Oxaliplatin dose change per patient** | **Overall Mean change for the drugs** |
|---|---|---|---|---|---|---|---|
| 1 | 0 | 65.3333333 | 77.1666667 | 114.666667 | 0 | 0 | |
| 7 | 167.5 | 0 | 47.5 | 156.833333 | 0 | 0 | |
| 15 | 0 | 80.3333333 | 53.1666667 | 90.8333333 | 0 | 0 | |
| 25 | 0 | 38.1666667 | 61.1666667 | 127.333333 | 0 | 0 | |
| 29 | 104.83333 | 0 | 75 | 167.166667 | 0 | 0 | |
| 37 | 49.333333 | 0 | 50.5 | 115.833333 | 0 | 0 | |
| 40 | 0 | 32.6666667 | 37.3333333 | 0 | 124 | 0 | |
| **Mean error for the dataset (27 patients) per each drug** | **107.22222** | **54.125** | **57.4047619** | **128.7777778** | **124** | **0** | **86.882** |

Similarly, we recorded the errors and the mean errors by changing the number of training patients as 20, 15, 10 and 5. In addition, the following table 21, shows the summarized error checking results for all the altered patient models.

TABLE 31. TESTING RESULTS FOR THE SIX DRUGS WITH ALL THE FIVE PATIENT MODELS

| Patient No. | Mean Cetuximab dose change per patient | Mean Bevacizumab dose change per patient | Mean Irinotecan dose change per patient | Mean 5FU dose change per patient | Mean Xeloda dose change per patient | Mean Oxaliplatin dose change per patient | Overall Mean change for the five drugs |
|---|---|---|---|---|---|---|---|
| **5 patients** | 247 | 83.7083333 | 82.83333 | 191.27778 | 345.5 | 0 | 190.063889 |
| **10 patients** | 242.38889 | 69.2916667 | 66.619047 | 183.527778 | 331.5 | 0 | 178.6654762 |
| **15 patients** | 173.94444 | 68.4583333 | 73.238095 | 160.722222 | 342.5 | 0 | 163.772619 |
| **20 patients** | 125.72222 | 66.125 | 66.071428 | 141.305556 | 251 | 0 | 130.044813 |
| **27 patients** | 107.22222 | 54.125 | 57.404761 | 128.777778 | 124 | 0 | 86.8824404 |

The reason for the Oxaliplatin dose being zero is because, only one patient has given oxaliplatin in the collected dataset and that patient's data have been removed while removing the data related to too low and too long chemotherapy cycles from the raw dataset.

We then plotted the error behavior with the training size for each of the drugs and considering the overall error.

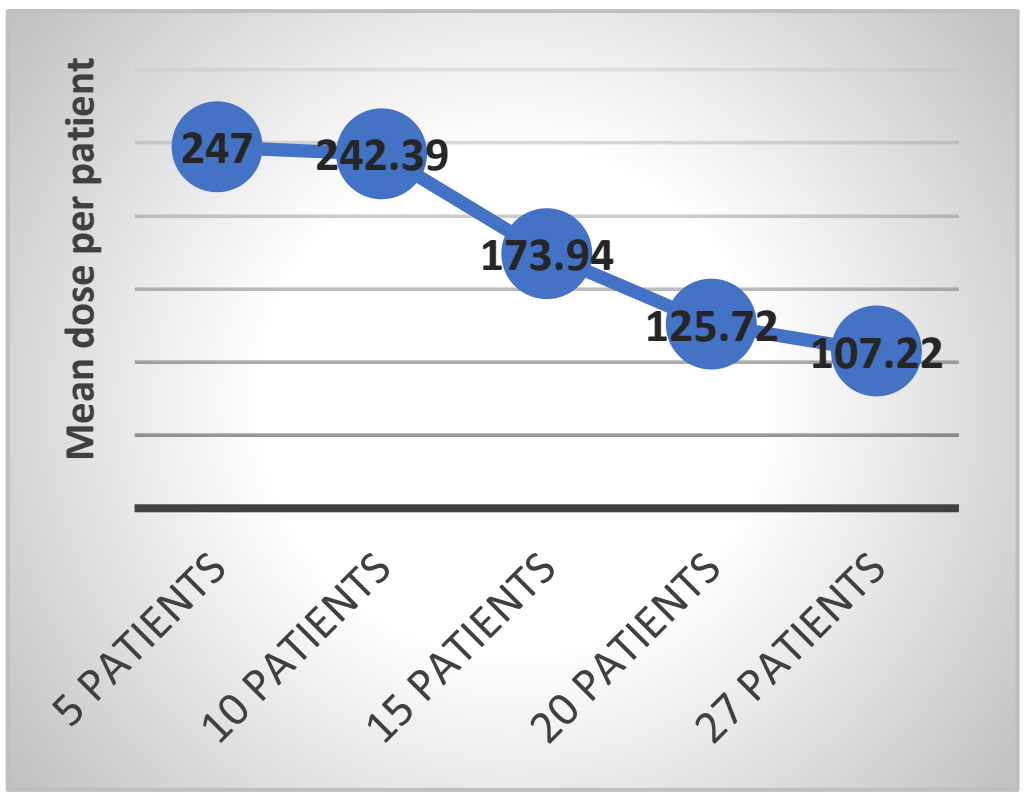

Figure 49: Mean Cetuximab dose change per patient

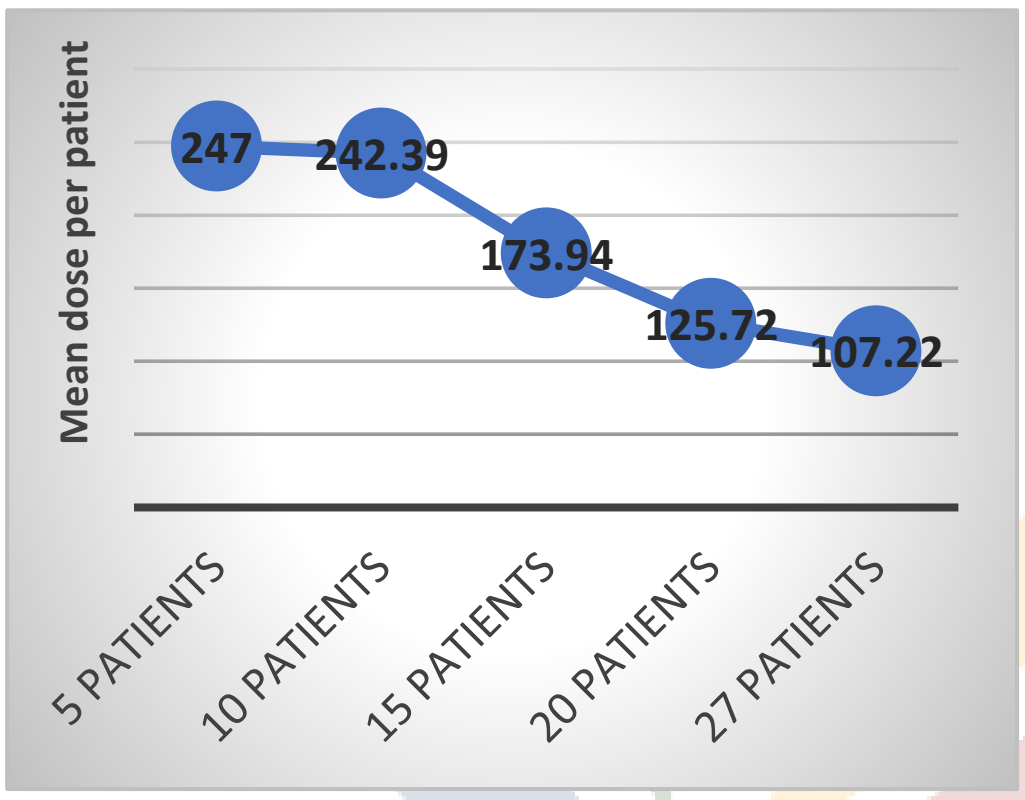

Figure 50: Mean Irinotecan dose change per patient

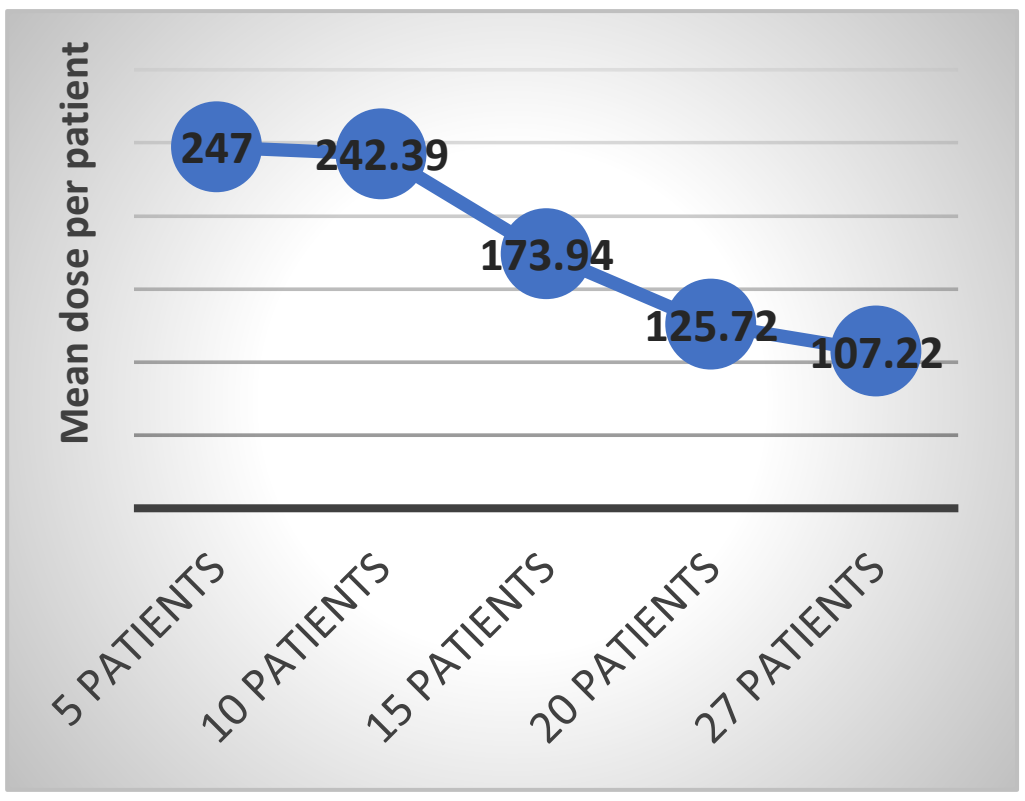

Figure 51: Mean Bevacizumab dose change per patient

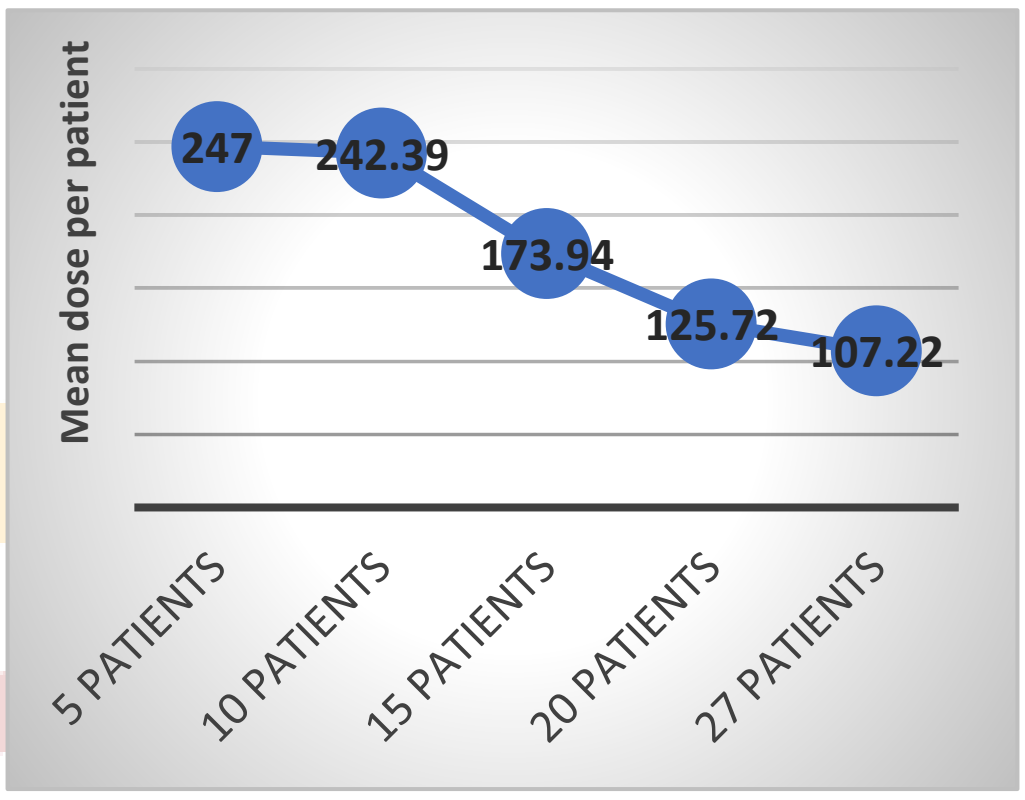

Figure 52: Mean 5FU dose change per patient

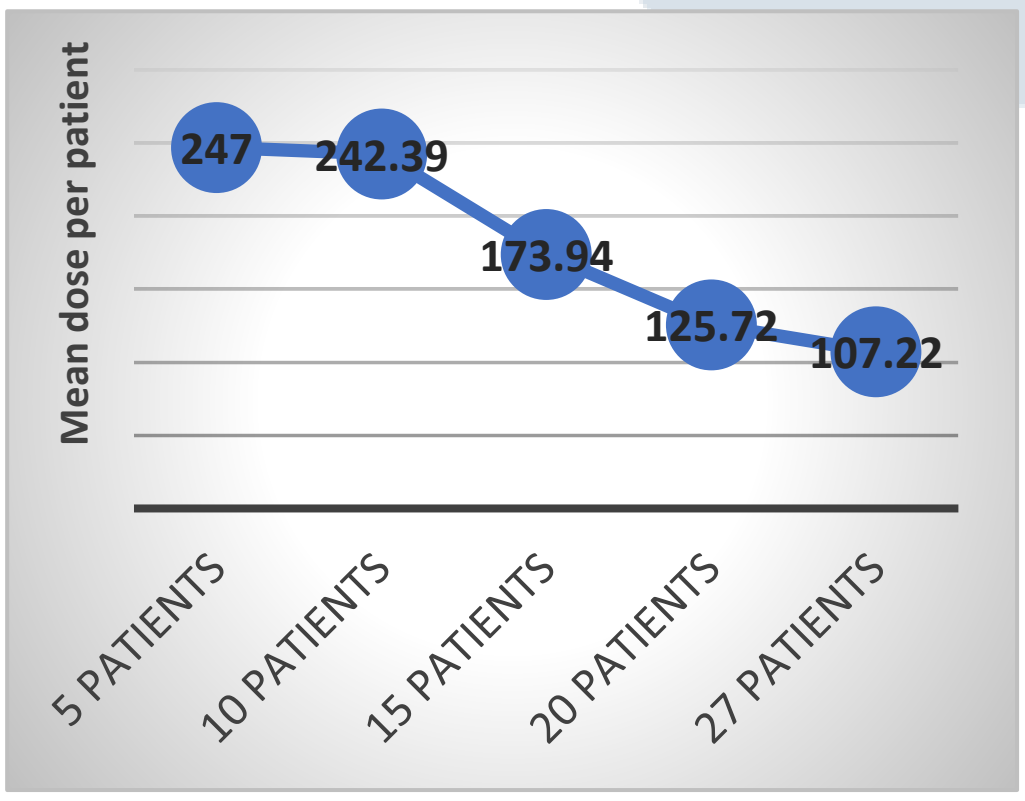

Figure 53: Mean Xeloda dose change per patient

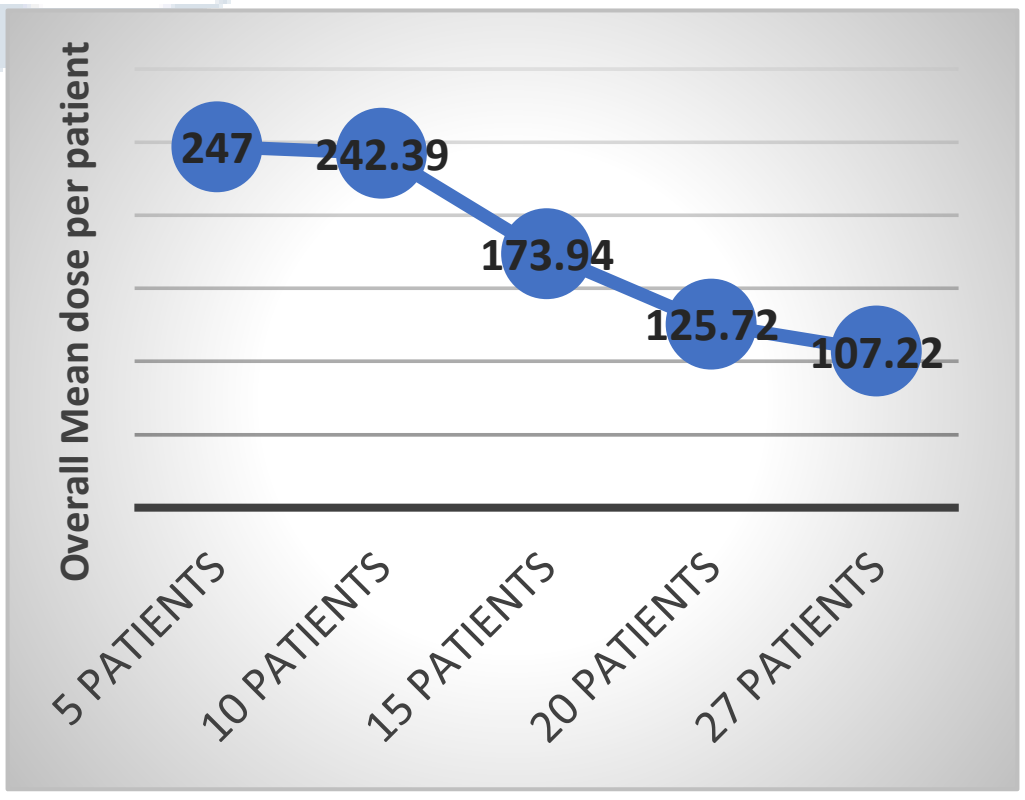

Figure 54: Overall mean dose change for the five drugs

As it is visible from figure 50-55, it can be concluded that the mean dose for the individual drugs and the overall mean dose for the five drugs are decreasing when increasing the training patient size. Thus, we can assure that when the training samples are increasing with the time, more accurate measurements for the dosing plan can be acquired. If this proposed Chemo RL model is decided to be used as a supportive measurement for the oncologists to decide the chemotherapy dosing schedule, they can assure more accuracy when more patient's data are entering to the model.

## 7.6. Behavior of Scores related to CQL and MOPO

The d3rlpy platform that we used for the development process provides scoring functions, which can evaluate with the test episodes during training. So in order to further tune our Chemo RL model and to evaluate the overall training process, we approached the following scorers related to the CQL and MOPO algorithms.

TABLE 32. SCORERS FOR THE CQL AND MOPO ALGORITHMS

| **Scorer** | **Simple meaning** |
|---|---|
| ***Scorers for the CQL algorithm*** | |
| TD_error | Average TD error (in negative scale) |
| Value_scale | Average value estimation (in negative scale) |
| Discounted_advantage | Average of discounted sum of advantage (in negative scale) |
| Initial_state_value_estimation_scorer | Mean estimated action-values at the initial states. |

| *Scorers for the MOPO algorithm* | |
|---|---|
| MOPO_loss | The MOPO loss observed during the testing process |
| Observational error | MSE of observation prediction (in negative scale) |
| Reward error | MSE of reward prediction (in negative scale) |
| Variance | Prediction variance of ensemble dynamics (in negative scale) |

### *7.6.1. TD_error*

This metric returns the average TD error. It suggests how the Q functions are overfitting to the training sets. Q functions are overfitting if the TD error is large.

$$\mathbb{E}_{s_t,a_t,r_{t+1},s_{t+1}\sim\ D}\ [Q_\theta(s_t,a_t) - \ (r_t\ +\ \gamma max_a Q_\theta(s_{t+1},a))^2] \qquad (58)$$

Following figure 56 and figure 57 explains how the TD error behaves during the training process.

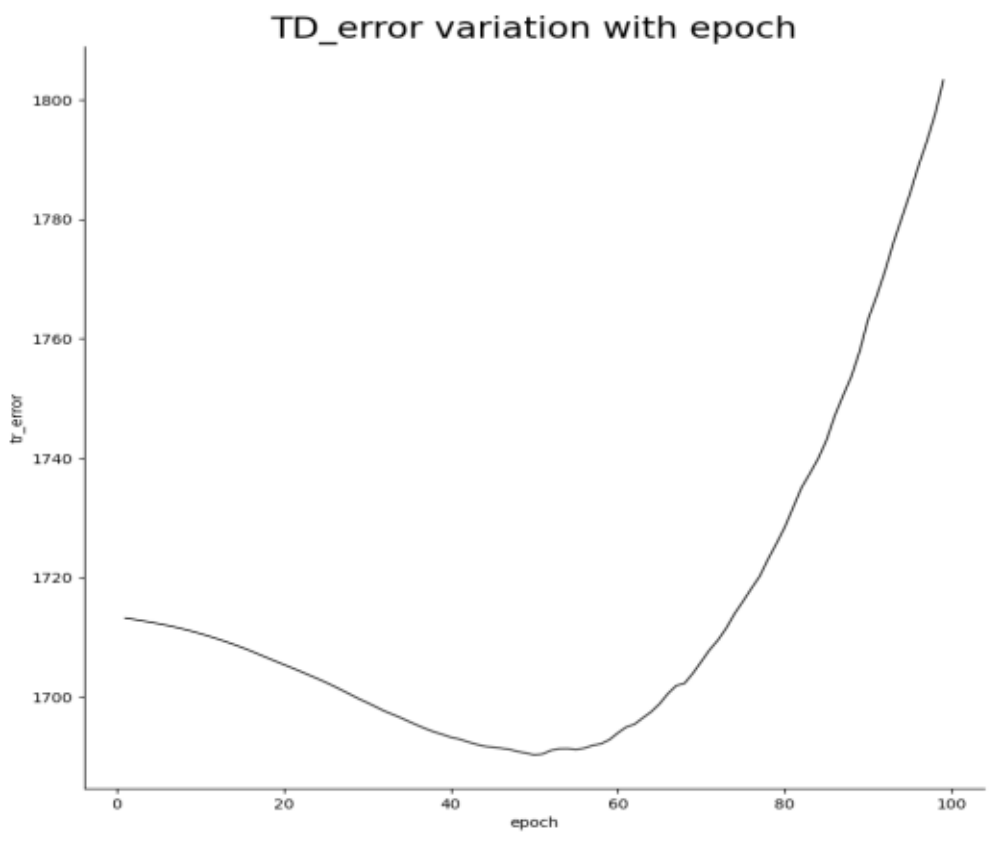

Figure 56: TD_error variation with the epochs

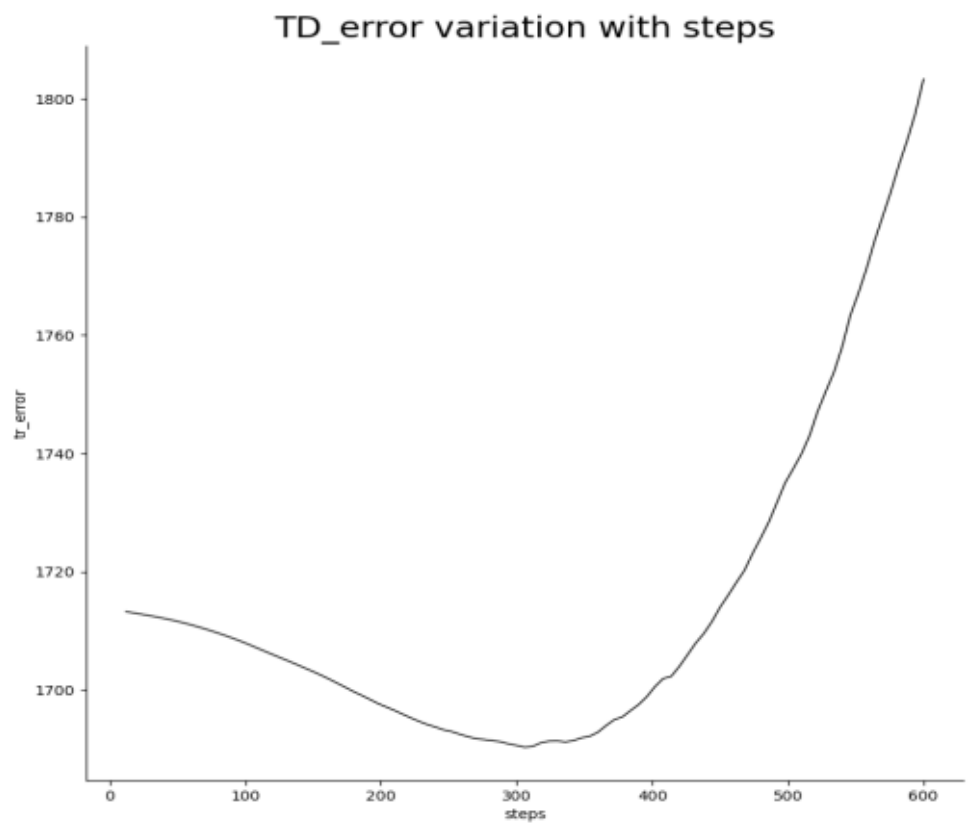

Figure 55: TD_error variation with the steps

According to the figure 28 and 29, it is seen that the minimum TD error has been achieved at around 50 epoch. So, we stopped our training process at the 50th epoch.

### *7.6.2. Value_scale*

This metric returns the average value estimation error. It suggests the scale for estimation of Q functions. If average value estimation is too large, the Q functions will overestimate action-values, which implies the failed training process.

$$\mathbb{E}_{s_t \sim D}\ [max_a Q_\theta(s_t, a)] \tag{59}$$

Following figure 58 and figure 59 explains how the value scale behaves during the training process.

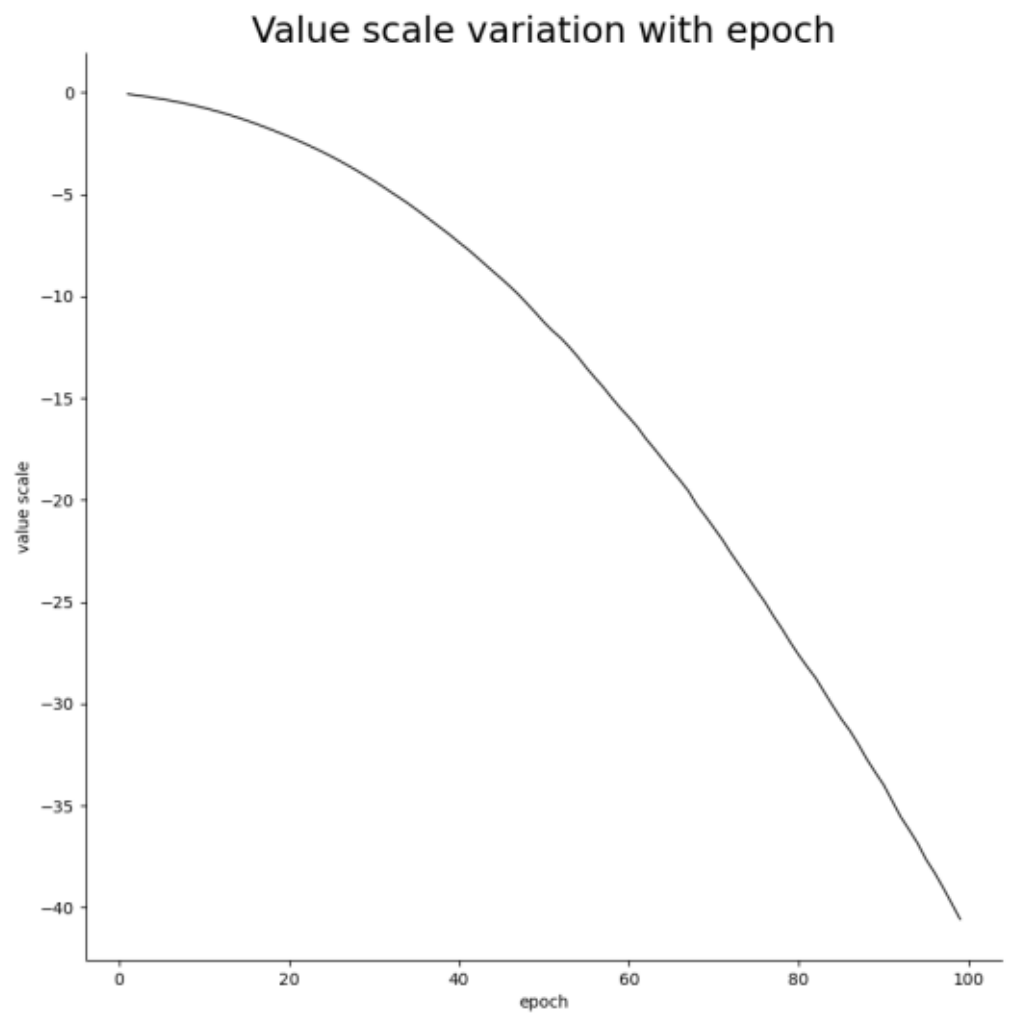

Figure 58: value scale variation with the epochs

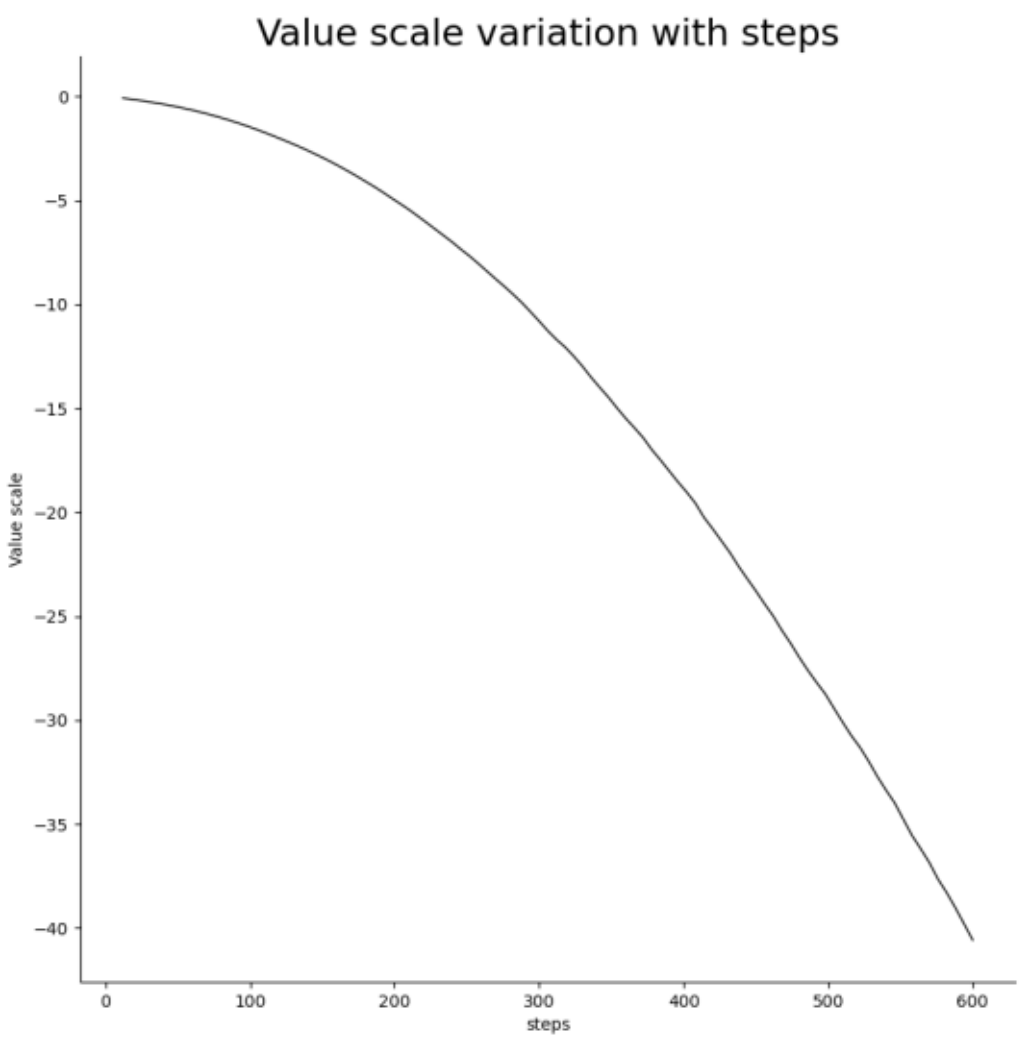

Figure 57: value scale variation with the steps

It is visible that the value estimation is decreasing with the training process.

### *7.6.3. Discounted_advantage*

This metric suggests the average value of the discounted sum of advantage. It suggests how the greedy-policy selects different actions in the action-value space. If the sum of advantage is small, the policy selects actions with a larger estimated action-value.

$$\mathbb{E}_{s_t,\ a_t \sim\ D}\ [\sum_{t'=t} \gamma^{t'-t} \mathrm{A}\ (s_{t'}, a_{t'})] \tag{60}$$

Where, $\mathrm{A}\ (s_{t'}, a_{t'}) = Q_\theta(s_t, a_t) - \ max_a Q_\theta\big((s_t, a)\ \big)$

Following figure 60 and figure 61 explains how the value scale behaves during the training process.

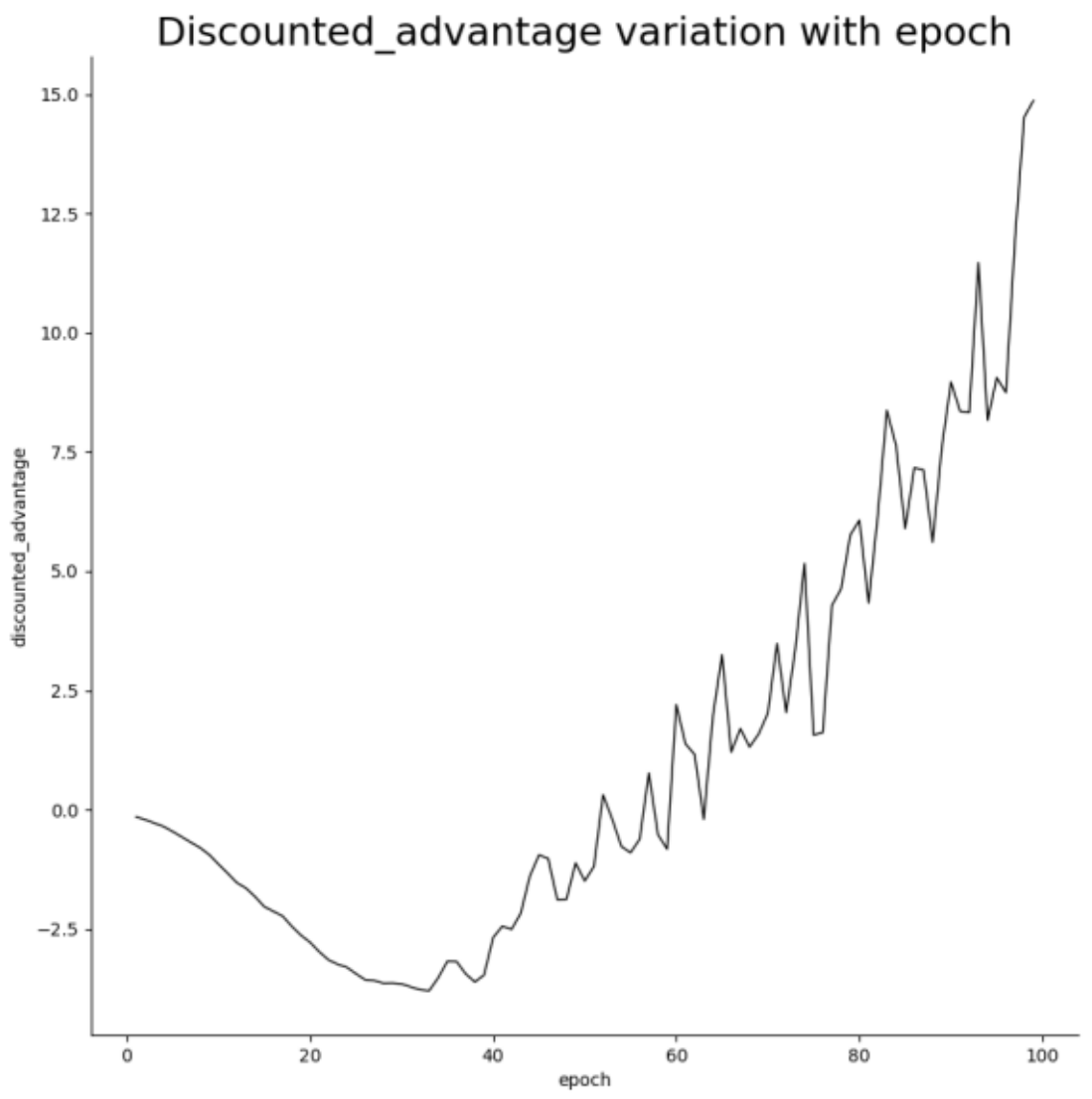

Figure 60: Discounted advantage variation with the epochs

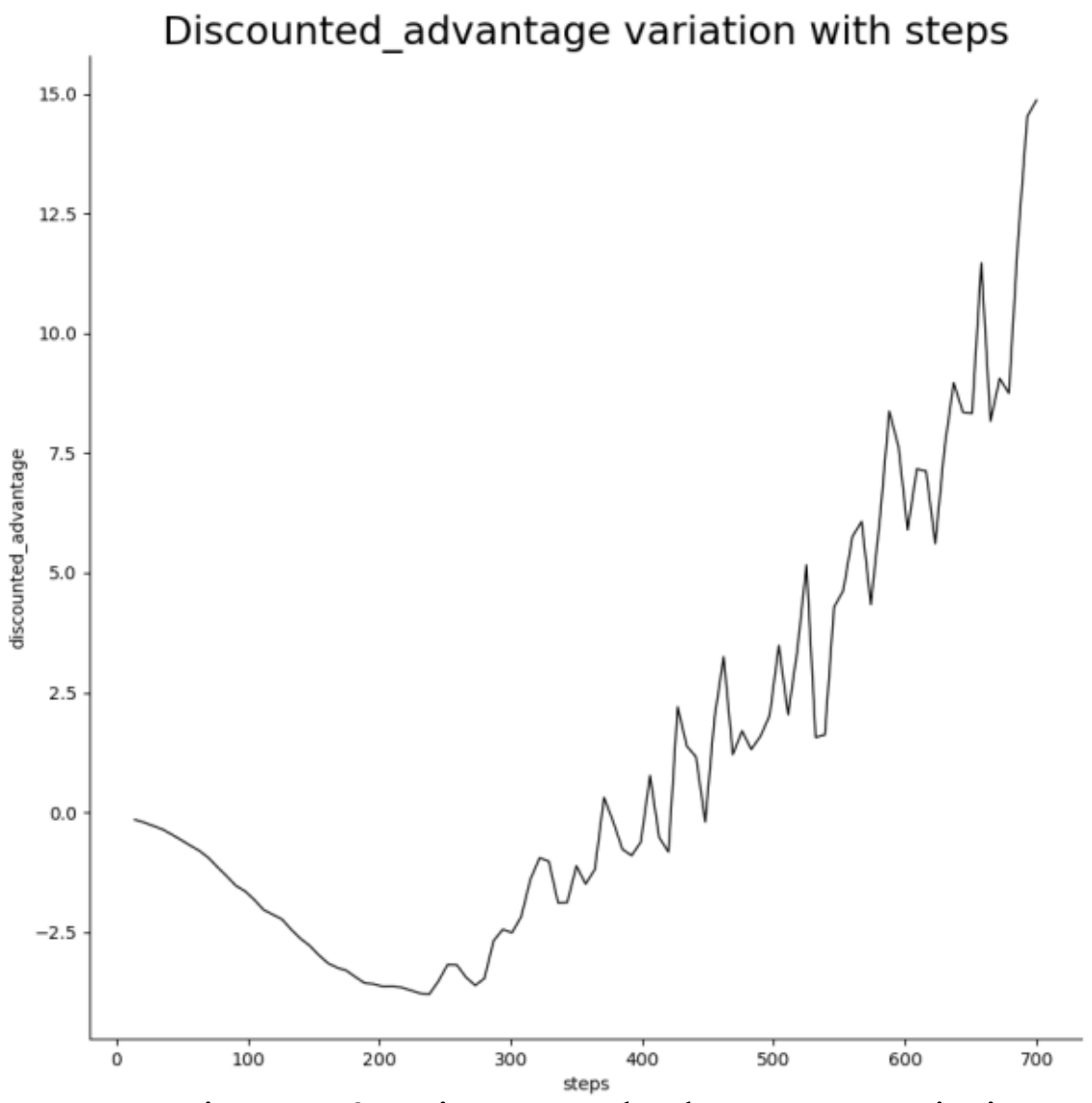

Figure 59: Discounted advantage variation with the steps

### *7.6.4. Initial_state_value_estimation_scorer*

This metric suggests the mean estimated action-values at the initial states. It suggests how much return the trained policy would receive by deploying the policy to the states. If the values are large, the trained policy is expected to get higher returns.

$$\mathbb{E}_{s_0 \sim\ D}\ [Q(s_0, \pi(s_0))] \tag{61}$$

Following figure 62 and figure 63 explains how the initial state value estimation scorer behaves during the training process.

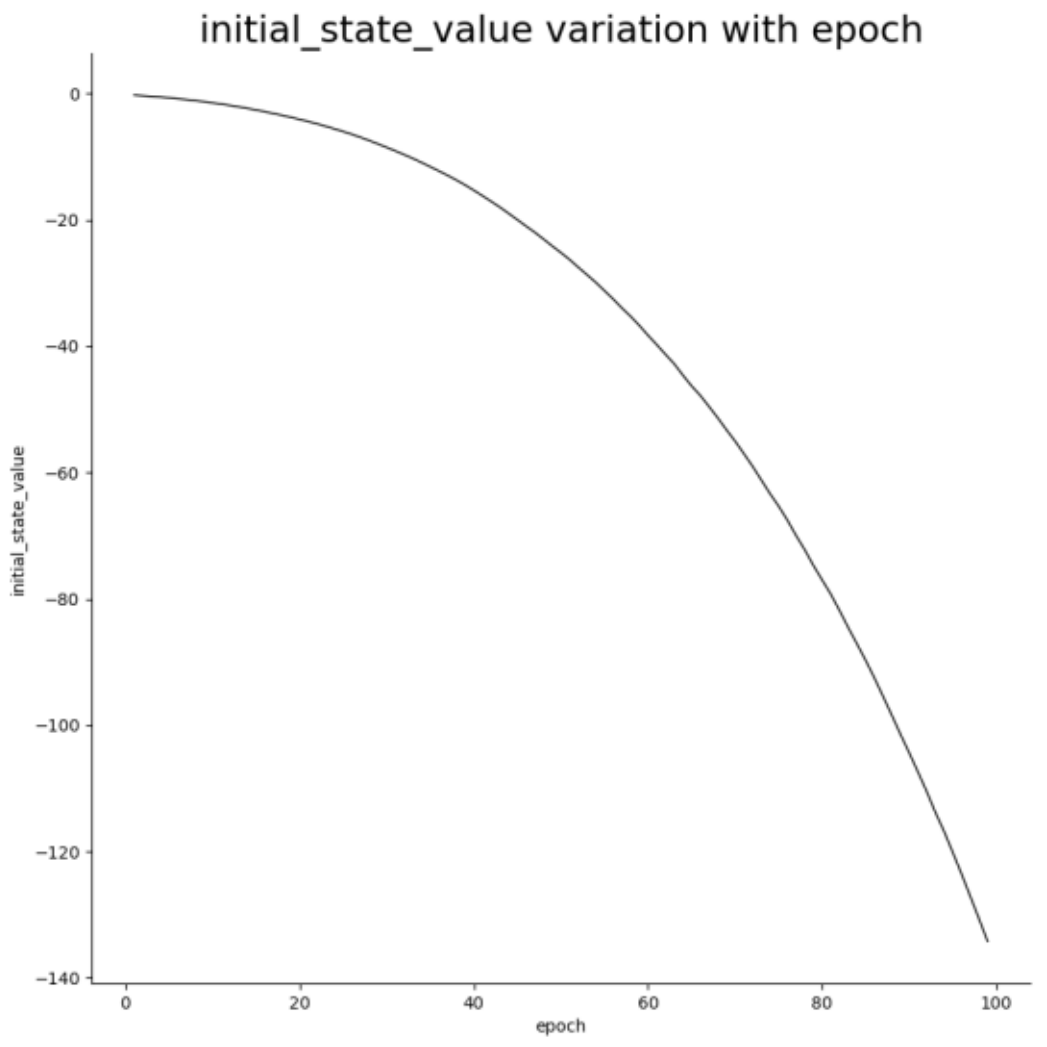

Figure 61: Initial state value estimation scorer variation with the epochs

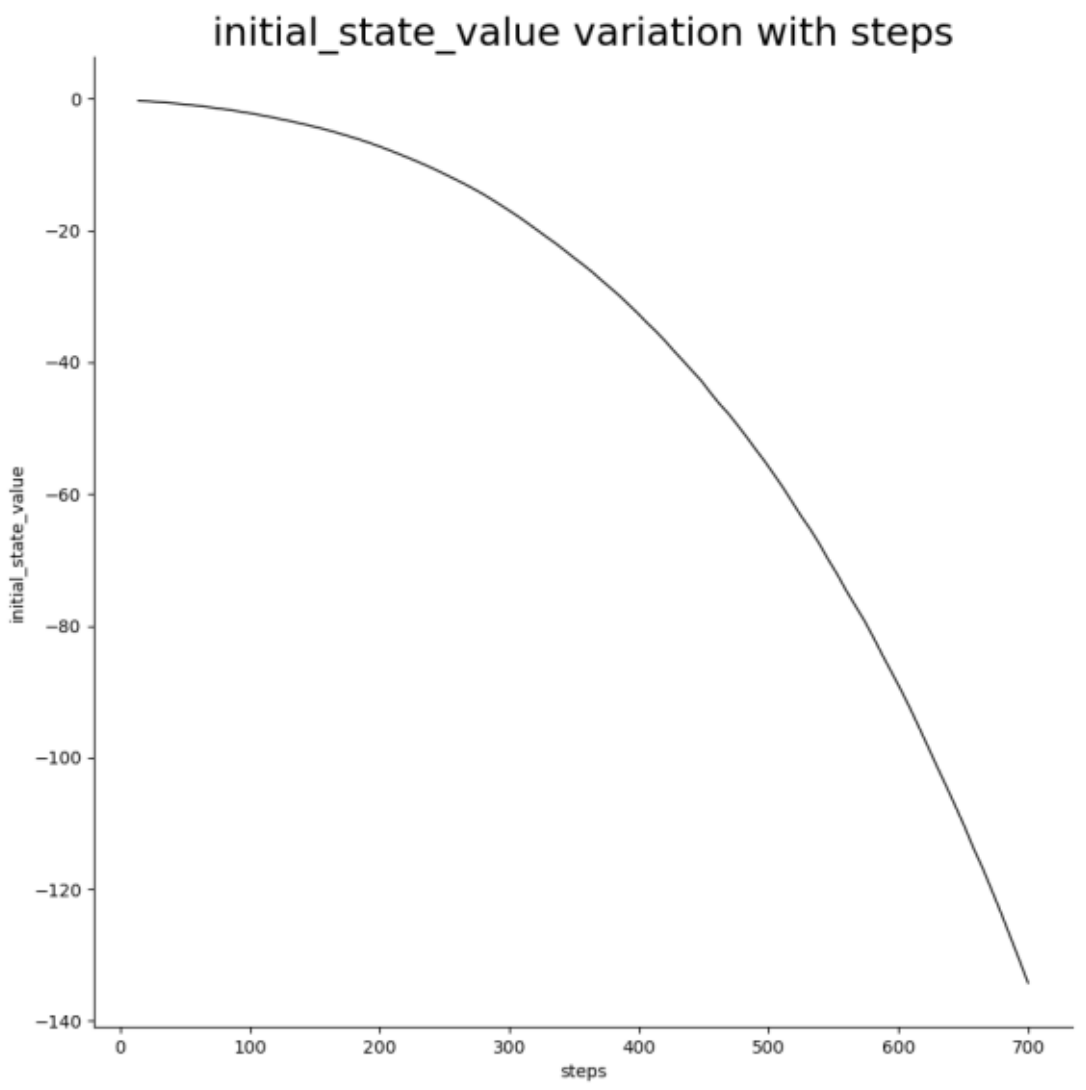

Figure 62: Initial state value estimation scorer variation with the steps

It is seen that the initial state value estimation scorer is decreasing with the training time.

### *7.6.5. MOPO_loss*

This metric shows the loss observed during the training process.

Following figure 64 and figure 65 explains how the initial state value estimation scorer behaves during the training process.

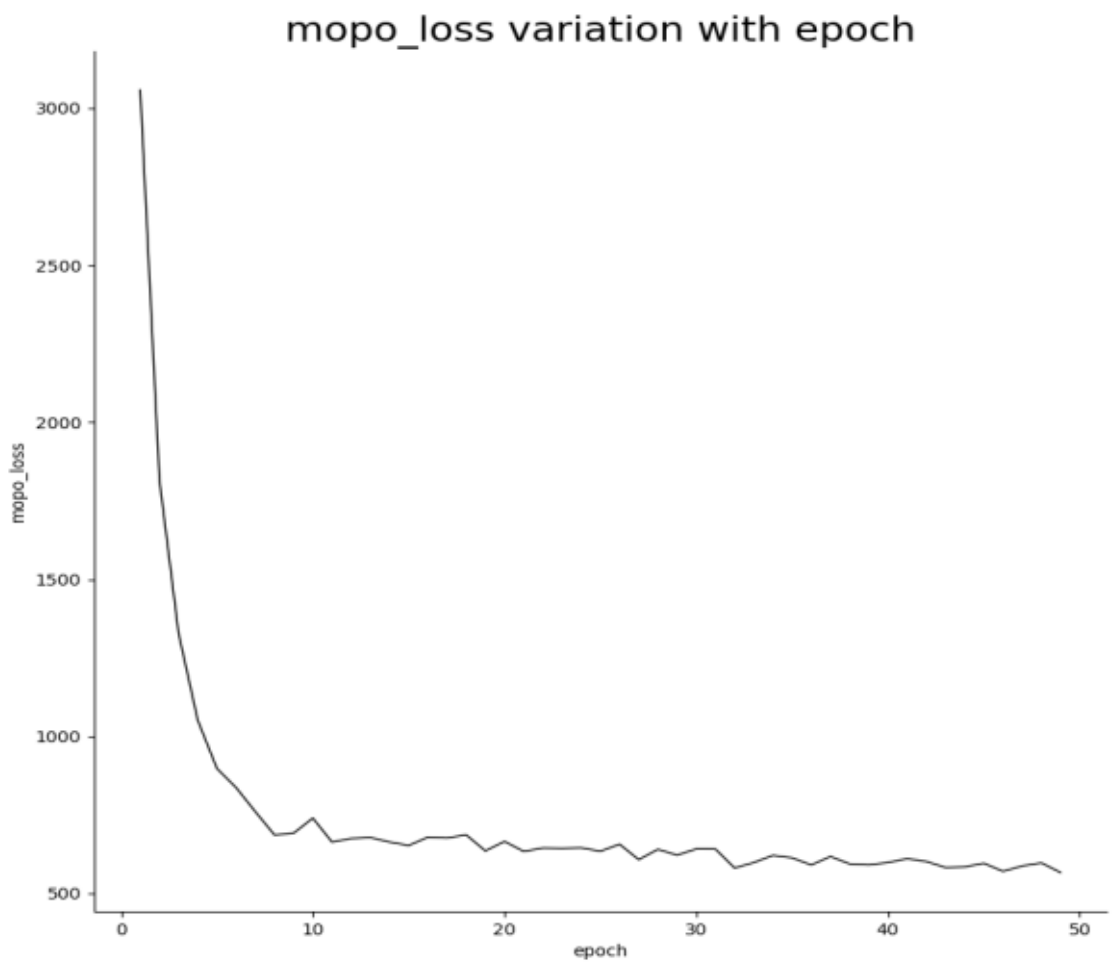

Figure 63: Mopo loss variation with the epochs

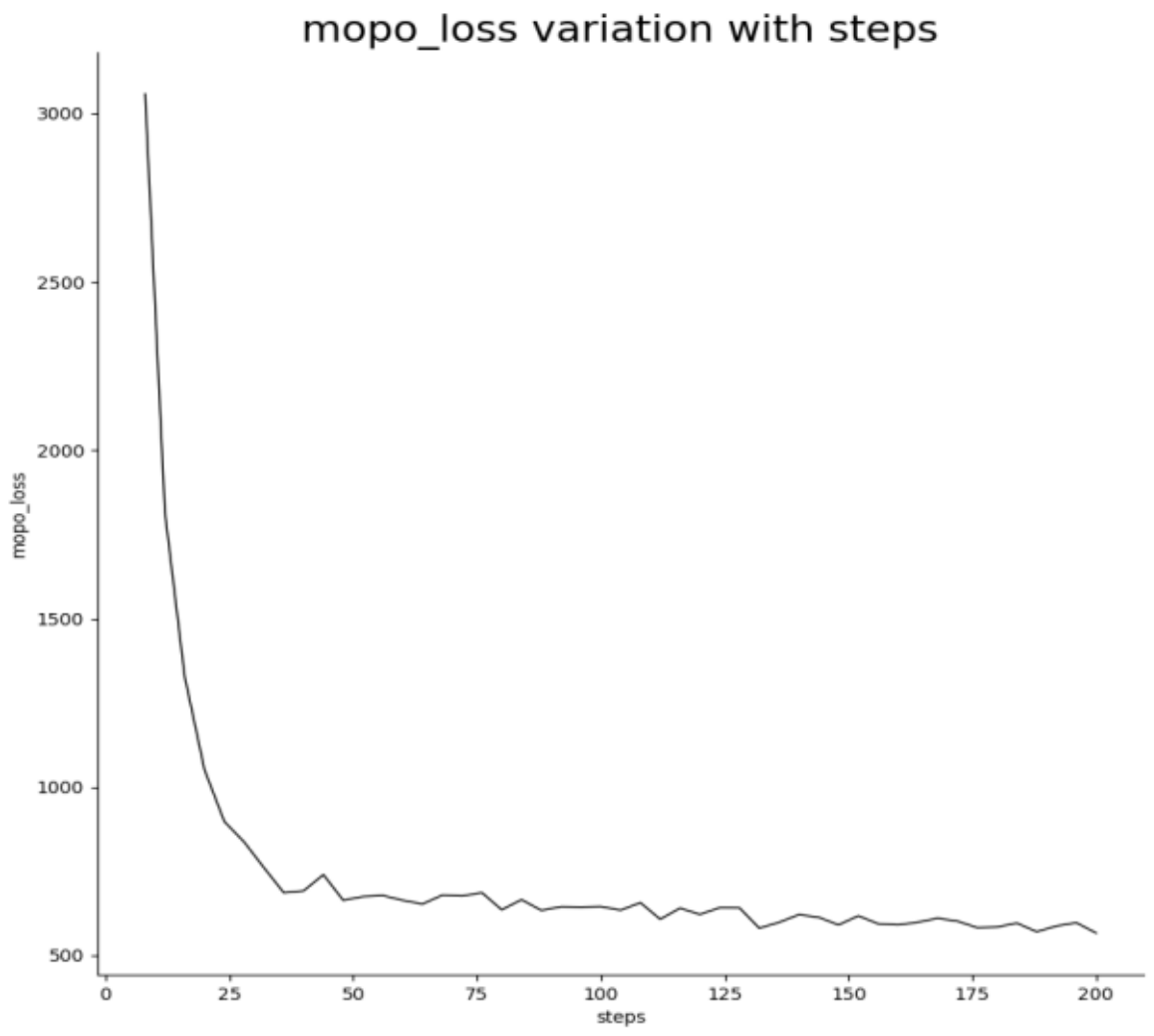

Figure 64: Mopo loss variation with the steps

### *7.6.6. Observational error*

This metric suggests the MSE of observation prediction. It is the difference between actual observations and the observations that is predicted to be received. Alternatively, this metric suggests how dynamics model is generalized to test sets. If the MSE is large, the dynamics model is overfitting.

$$\mathbb{E}_{s_t,a_t,s_{t+1}\sim D}\left[\ (s_{t+1}-s')^2\right] \tag{62}$$

Where $s' \sim T(s_t, a_t)$

Following figure 66 and figure 67 explains how the initial state value estimation scorer behaves during the training process.

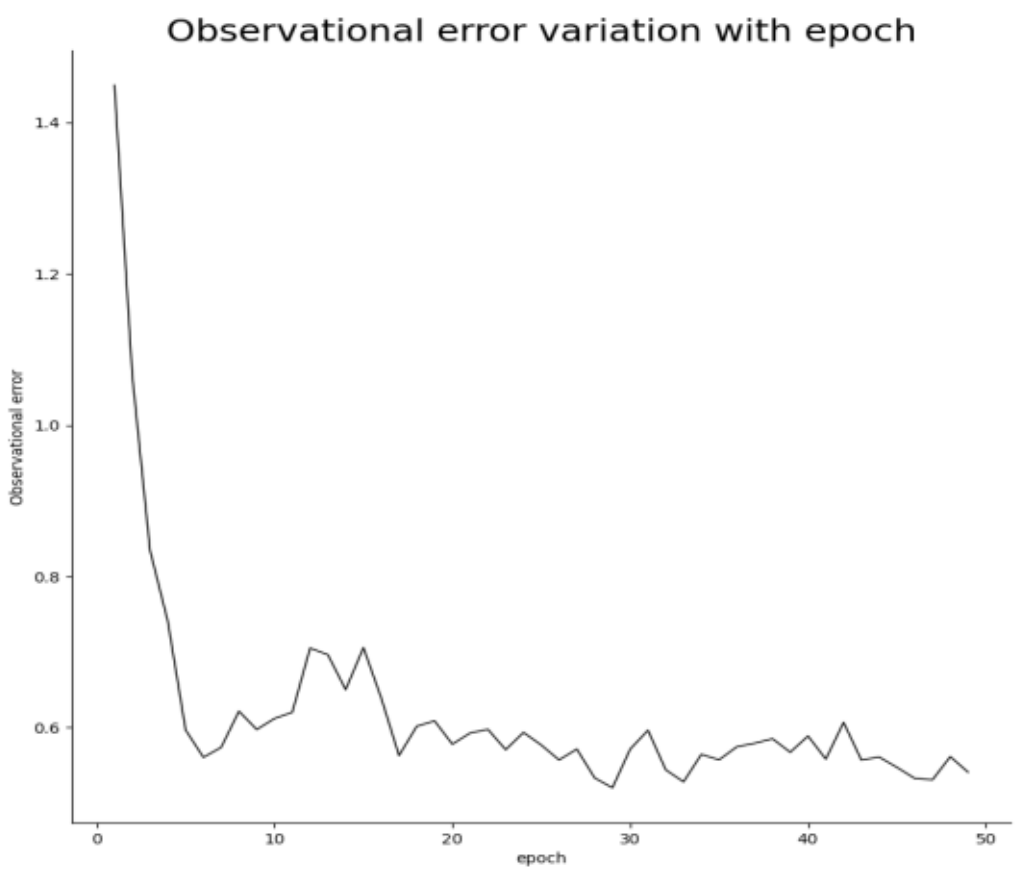

Figure 65: Observational error variation with the epochs

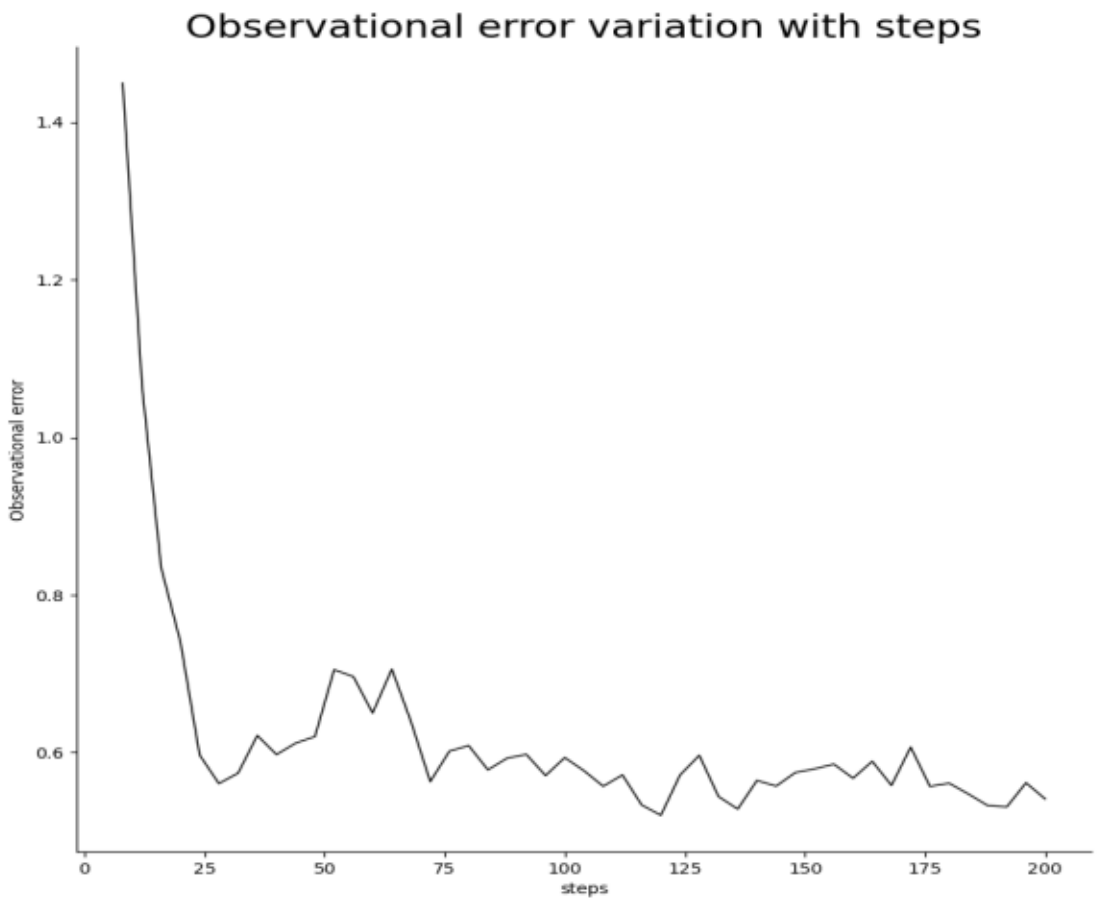

Figure 66: Observational error variation with the steps

### *7.6.7. Reward error*

This metric suggests the MSE of reward prediction. It is the difference between the actual reward and the reward that is predicted to be received. Alternatively, this metric suggests how dynamics model is generalized to test sets. If the MSE is large, the dynamics model is overfitting.

$$\mathbb{E}_{s_t,a_t,r_{t+1}\sim D}\left[\,(r_{t+1}-r')^2\right] \tag{63}$$

Where $r' \sim T(s_t, a_t)$

Following figure 68 and figure 69 explains how the initial state value estimation scorer behaves during the training process.

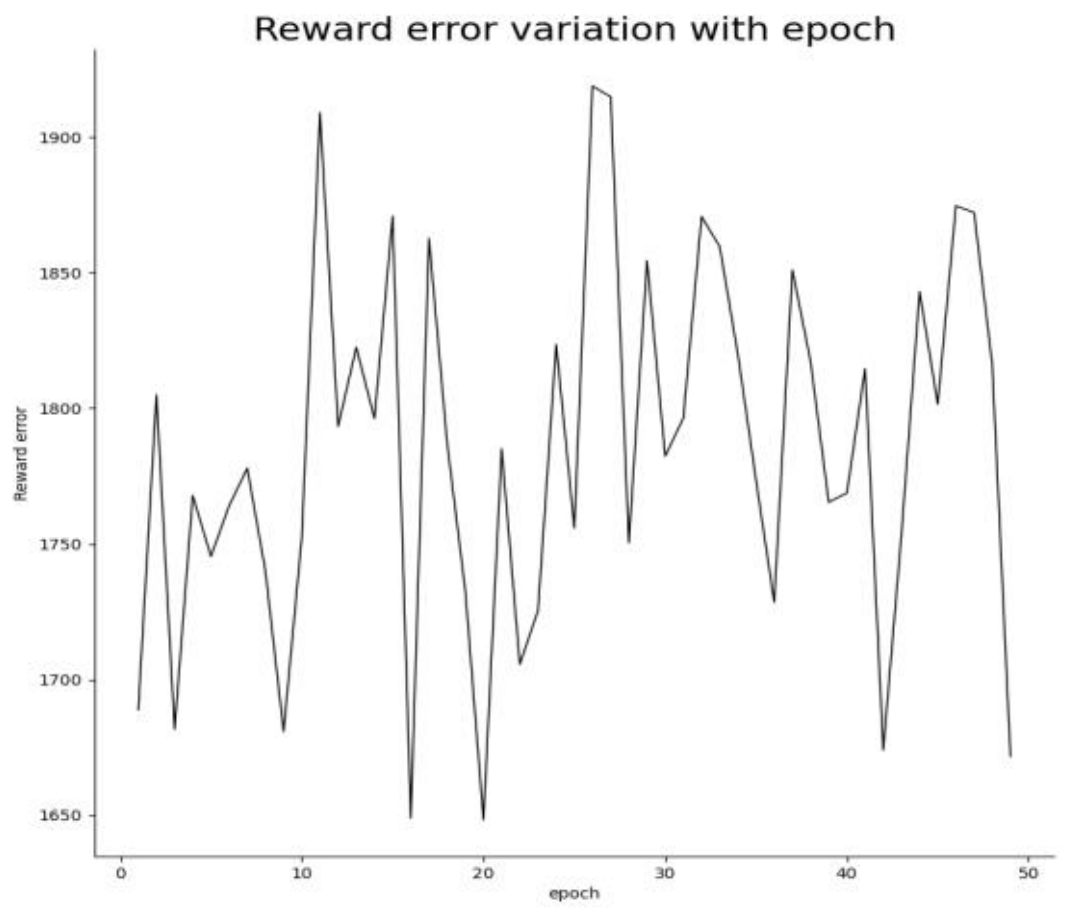

Figure 68: Reward error variation with the epochs

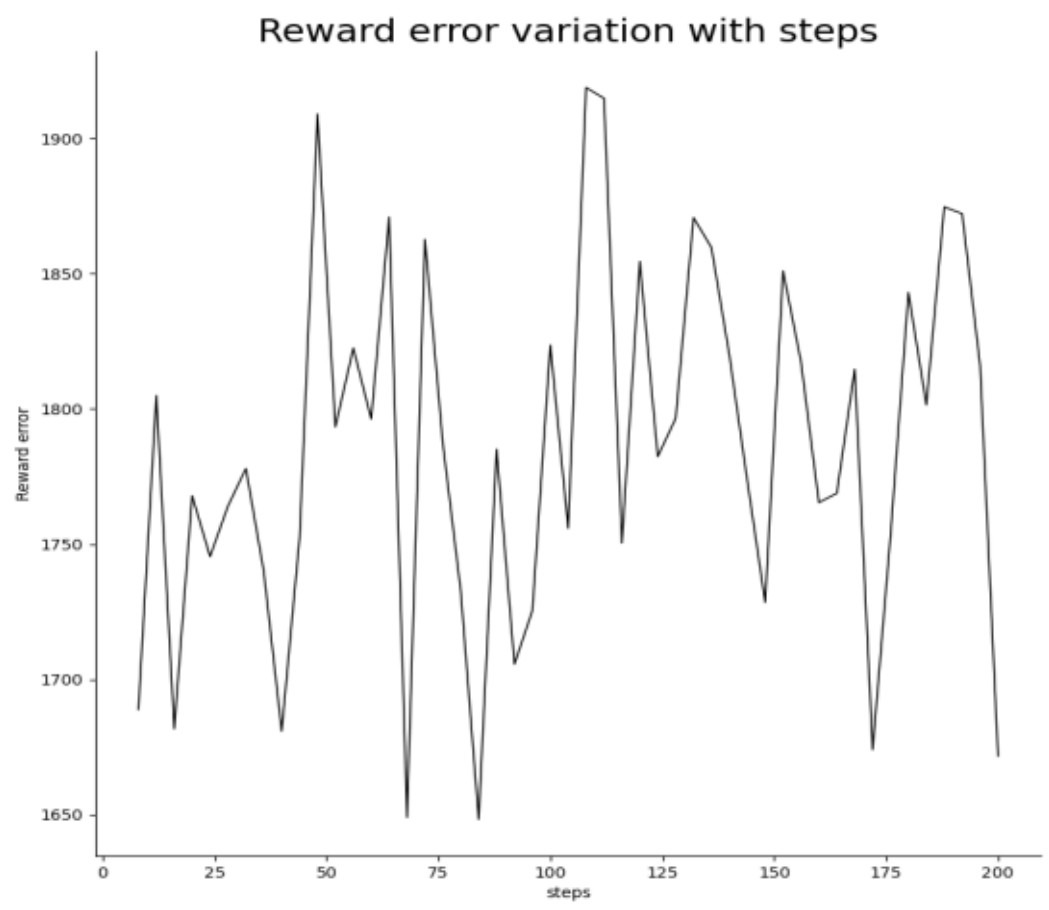

Figure 67: Reward error variation with the steps

It is seen that there an exact pattern of the reward error is not available with the training time.

### *7.6.8. Variance*

This metric shows the prediction variance of ensemble dynamics. It suggests how dynamics model is confident of test sets. If the variance is large, the dynamics model has large uncertainty.

Following figure 70 and figure 71 explains how the initial state value estimation scorer behaves during the training process.

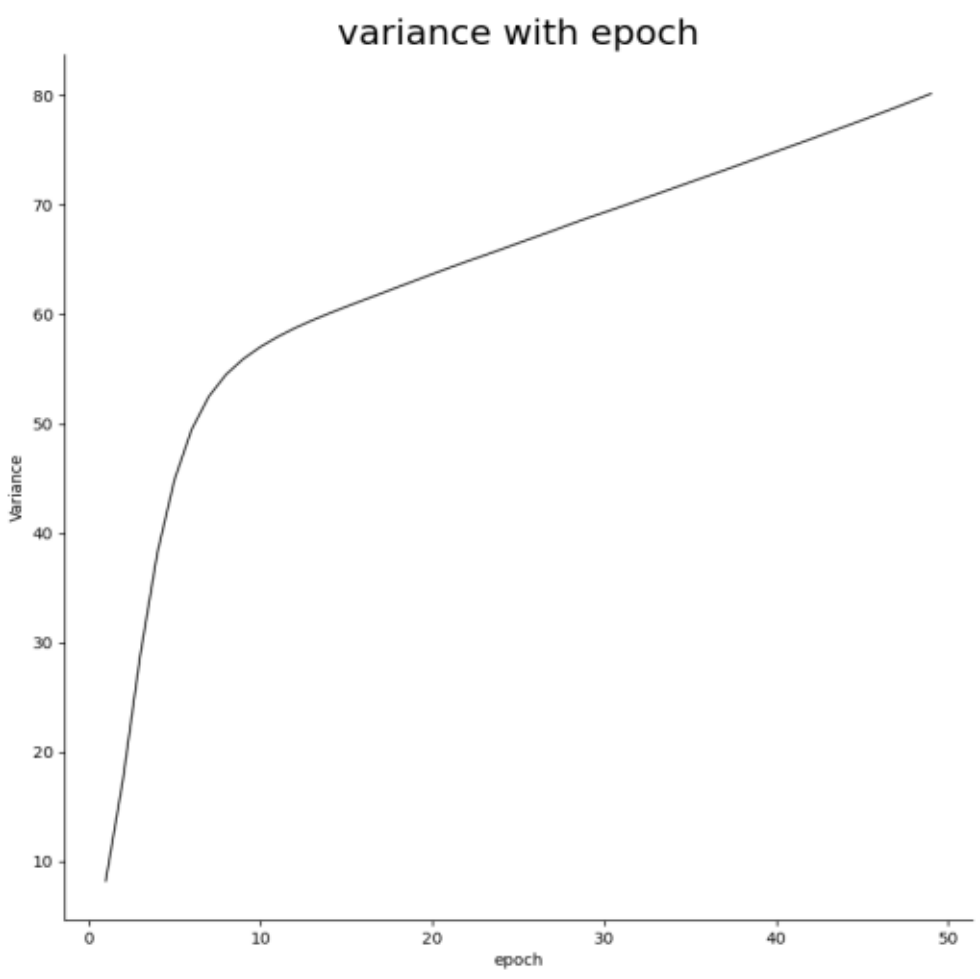

Figure 70: Variance variation with the epochs

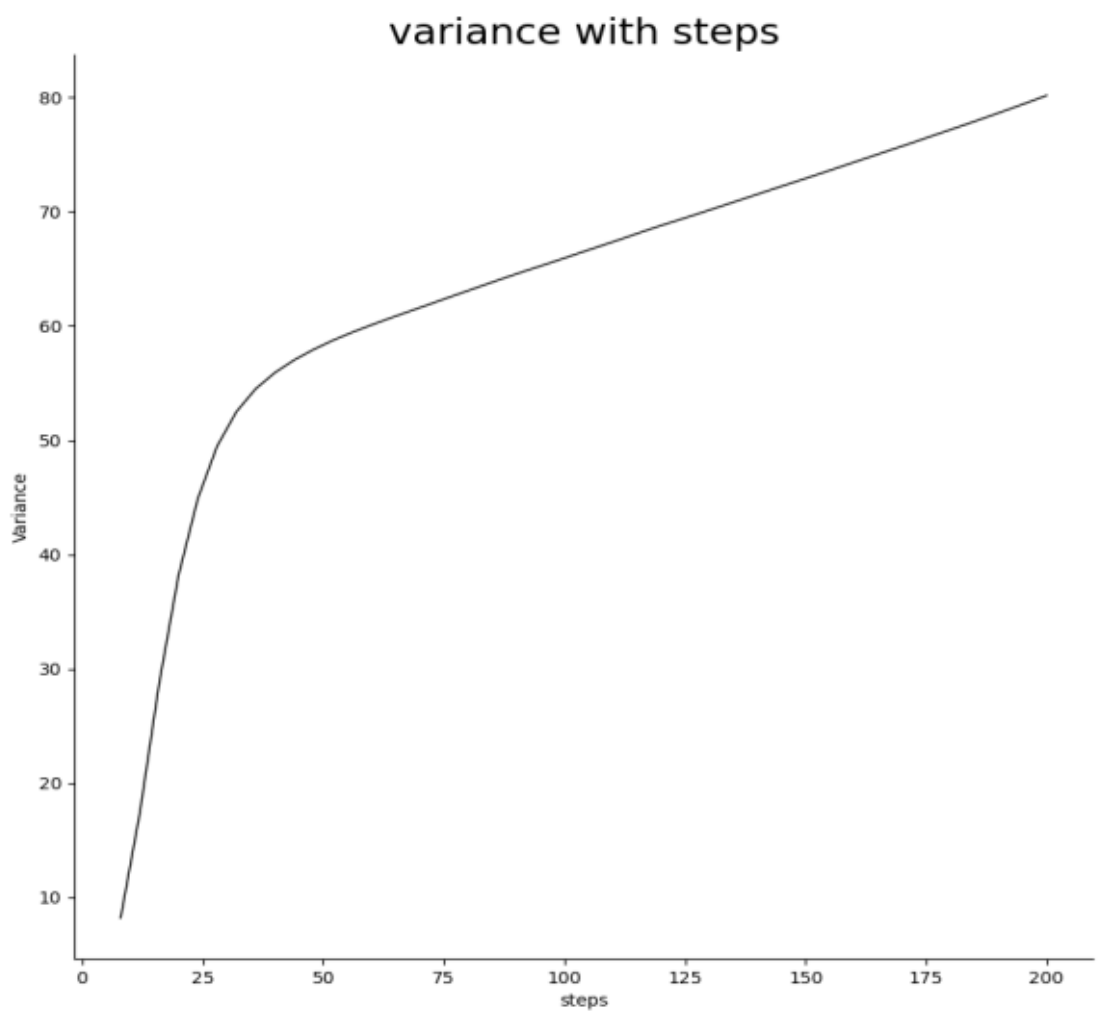

Figure 69: Variance variation with the steps

It is seen that the variance is increasing with the training time.

## 7.7. Statistical Data Analysis

This section includes the analysis results of the clinical data we collected from the hospital. The target patients were level 4-colon cancer patients, who are receiving line 1 chemotherapy treatments within the time of 2017-2020. 40 patients' data was collected from the hospital database with the careful attention and comments by the oncologist. The collected data were as the patients' demographic features of age, weight, height, gender, patients' clinical diagnosis features of tumor size, side effects index, CCI, ECOG PS measure, drug response, decision after the therapy and the patients' given combination chemotherapy drug dose amounts.

In order to in-depth understanding of the collected data, we analyzed them as shown in the table 23. The ranges of the variables, average values and frequent values were computed by using python libraries.

TABLE 33. DATA ANALYSIS RESULTS

| Variable | Range | Average value | Frequent value |
|---|---|---|---|
| ***Patient demographics ($x_i$)*** | | | |
| Age ( Years ) | 39 - 84 | 62.96 | 61.0 |
| Weight (kg) | 38.8 - 93 | 62.20 | - |
| Height (cm) | 143.6 –185.9 | 160.53 | - |
| Gender | - | - | Female ( 58.8% ) |
| ***Clinical diagnosis ($x_i$)*** | | | |
| Tumor size ( $mm^3$ ) | 0 - 4062.63 | 74.92 | - |
| Side effects index | - | 3.67 | 0 |
| CCI | - | 8.5 | 8.0 ( 24.5% ) |
| ECOG PS | - | 0.49 | 0 ( 56.9% ) |
| Drug Response | - | - | Stable Disease ( 51.7 % ) |
| Decision after the therapy | - | - | Maintenance therapy ( 93 % ) |
| Real patient status | 0 - 10 | - | 8 ( 24.0 % ) |
| ***Combination Therapy drugs*** | | | |
| Cetuximab ( DDI ) | 0 - 1.0692 | 0.1738 | - |
| Bevacizumab ( DDI ) | 0 - 1.5243 | 0.6716 | - |
| Irinotecan ( DDI ) | 0 - 1.3315 | 0.8460 | - |
| 5FU ( DDI ) | 0 - 1.0871 | 0.7821 | - |
| Xeloda ( DDI ) | 0 - 0.7067 | 0.8460 | - |
| Oxaliplatin ( DDI ) | 0 - 0.8778 | 0.7822 | - |

Next in order to get an idea of the distribution of the observations, actions and rewards we plotted the clinical diagnosis measures of Tumor size, weight, side effects index, ECOG PS measure, Reward, Cumulative status and the given drug dose presented in the dataset. Plotting these helped in getting an understanding of clinical data distribution under the chemotherapy drug dose changes. Following figures shows the visualizations for 2 patients.

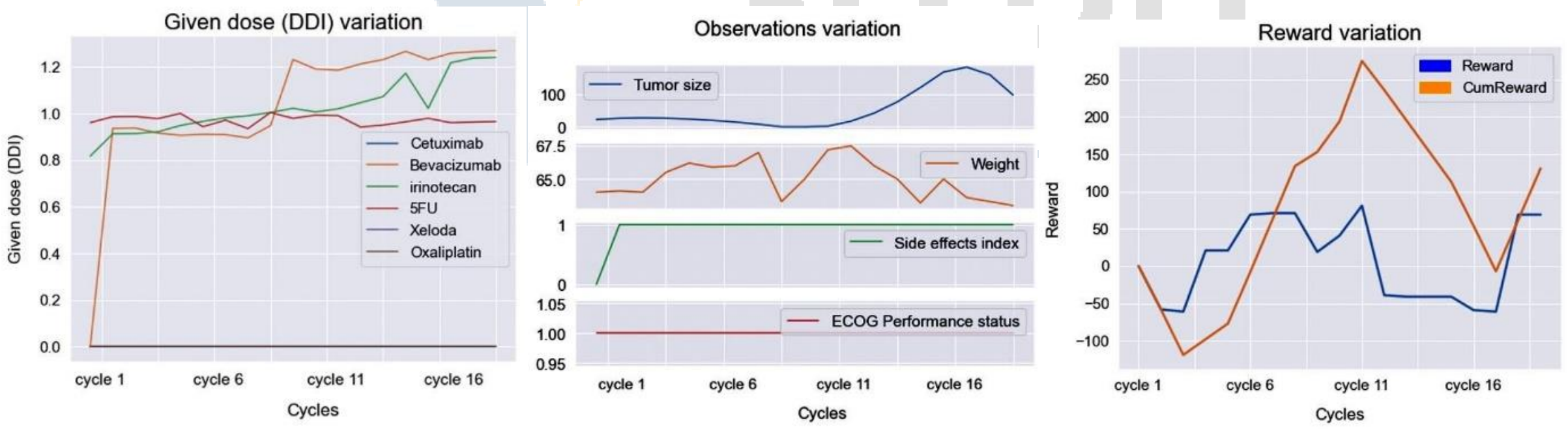

Figure 71: Variations of observations, rewards according to the given dose (DDI): patient No.3

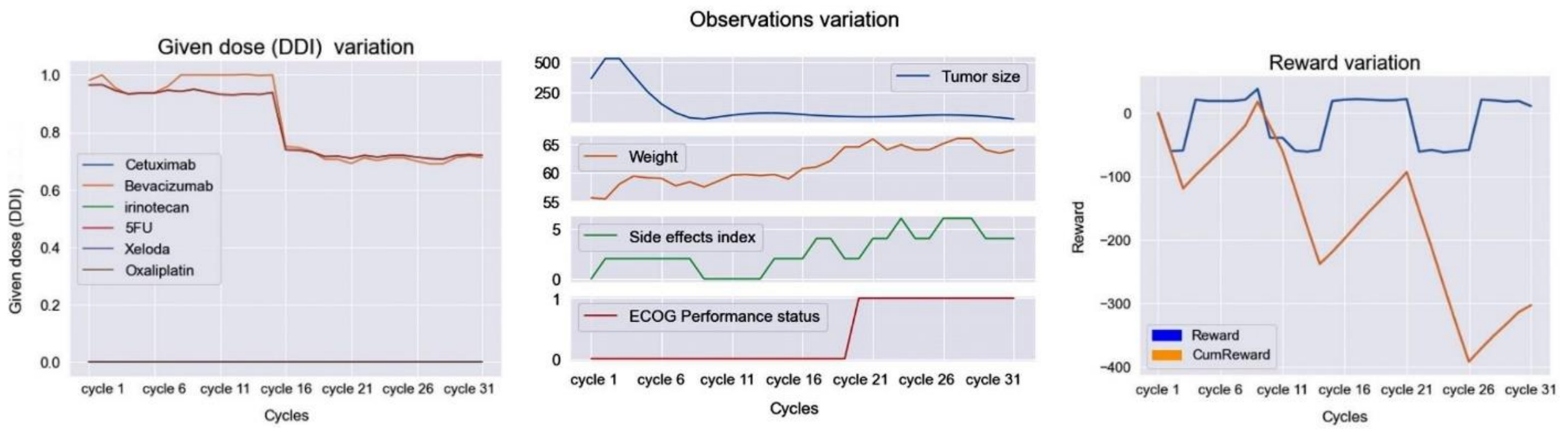

Figure 72: Variations of observations, rewards according to the given dose (DDI): patient No.30

## 7.8. Summary

This chapter presented the implementation process of the developed system. Here, the details on implementation of the modules and the tools and technologies adopted have been presented in detailed. The modular designs identified in the Design chapter were implemented by means of introducing software architectures and this chapter presented the implementation details in detailed. In the next chapter, discuss about the Evaluation of the developed system.

CHAPTER 8

# 8. CONCLUSION & FURTHER WORKS

## 8.1. Introduction

The previous chapter discussed on the evaluation phase conducted on the developed system. This chapter gives an overall conclusion to the whole research project. Project results overview in terms of the targeted objectives, limitations encountered and further research opportunities are considered under this chapter.

## 8.2. Conclusion

At the beginning of the research project, it was hypothesized that offline reinforcement learning can be adapted to solve the problem of cancer chemotherapy treatment planning. As stated in the hypothesis, the inspiration behind this hypothesis was how the online reinforcement learning have been adopted in finding optimal treatment schemes by using the mouse models which are problematic to approach in the real human treatment setting. In order to prove the hypothesis, a complete model based offline reinforcement learning model was constructed on the basis of real colon cancer chemotherapy clinical data.

Firstly, the literature on the area of reinforcement learning in medicine, dynamic treatment regimen planning, reinforcement learning on chemotherapy was comparatively reviewed. Thereby, the limitations in using the mathematical models based online reinforcement learning were identified. Considerable amount of time and effort was undertaken to understand the concepts of reinforcement learning and filling the knowledge gap by reviewing research papers, reference books and knowledge obtained from e-learning courses. The primary development process was started with the design of a basic model with tradition Q Learning, following the validated mathematical models in the online RL setting. However, as the model is not accepted in the real clinical practice, the study was laid towards the offline reinforcement learning setting with the real clinical dataset collected from the hospital. The development process of the new approach was started with the design of the modules and interactions with each module. Implementation was done for the modules identified.

Finally, the evaluation process and the results are presented in the chapter 7. Evaluation process was undertaken using the evaluation criteria setup with the corporation of the oncologist in order to access the real clinical practices in to the designed Chemo RL model. According to the evaluation results, performance evaluation and the oncologist's positive comments it is clear that the objective was achieved.

In order to reach the aim, major objectives were also defined at the initial stage of the research project. The achievement of the objectives with respect to the project results have been summarized in the next sub section.

### 8.3. Project results overview

This session emphasizes the status and the sections related for the project outcomes against the project objectives.

TABLE 34. PROJECT RESULTS OVERVIEW

| Objectives | Status | Sections related |
|---|---|---|
| 1. Comprehensively review the literature review on Reinforcement learning in medicine, Dynamic treatment regimen planning, Reinforcement learning in chemotherapy and offline reinforcement learning | Completed | Chapter 02 |
| 2. Design and develop the Online RL based approach for chemotherapy | Completed | Chapter 04, Chapter 05 |
| 3. Implement the Online RL based approach for chemotherapy using the proposed approach | Completed | Chapter 03, Chapter 06 |
| 4. Evaluation and validation of the developed Online RL based system | Completed | Chapter 07 |
| 5. Design and develop the | Completed | |
| Offline RL based approach for chemotherapy using the real clinical data, practices and knowledge | | Chapter 04, Chapter 05 |
| 6. Implement the Offline RL based approach for chemotherapy using the Proposed approach. | Completed | Chapter 03, Chapter 06 |
| 7. Evaluation and validation of the developed Offline RL based system. | Completed | Chapter 07 |

Thus, all the above objectives were executed in the research as anticipated, as a concluding note it is clear that overall system development and evaluation process was succeeded.

### 8.4. Limitations and Further works

### 8.5. Summary

This chapter presented the implementation process of the developed system. Here, the details on implementation of the modules and the tools and technologies adopted have been presented in detailed. The modular designs identified in the Design chapter were implemented by means of introducing software architectures and this chapter presented the implementation details in detailed. In the next chapter discuss about the Evaluation of the developed system.